\documentclass[letterpaper]{article} 
\usepackage[preprint]{aaai23}
\usepackage{times}  
\usepackage{helvet}  
\usepackage{courier}  
\usepackage[hyphens]{url}  
\usepackage{graphicx} 
\def\UrlFont{\rm}  
\usepackage{natbib}  
\usepackage{caption} 
\usepackage{algorithm}
\usepackage{algorithmic}

\usepackage{newfloat}
\usepackage{listings}
\DeclareCaptionStyle{ruled}{labelfont=normalfont,labelsep=colon,strut=off} 
\floatstyle{ruled}
\newfloat{listing}{tb}{lst}{}
\floatname{listing}{Listing}

\usepackage{subcaption}
\usepackage{booktabs}
\usepackage{multirow}
\usepackage{enumerate}
\usepackage{amsmath}
\usepackage{amssymb}

\title{How to Fine-tune Models with Few Samples:\\Update, Data Augmentation, and Test-time Augmentation}

\author{
    Yujin Kim\textsuperscript{\rm 1}\equalcontrib, Jaehoon Oh\textsuperscript{\rm 2}\equalcontrib, Sungnyun Kim\textsuperscript{\rm 1}, Se-Young Yun\textsuperscript{\rm 1}
}
\affiliations{
    \textsuperscript{\rm 1} Kim Jaechul Graduate School of Artificial Intelligence, KAIST\\
    \textsuperscript{\rm 2} Graduate School of Data Science, KAIST\\
    \{yujin399, jhoon.oh, ksn4397, yunseyoung\}@kaist.ac.kr
}

\begin{document}

\maketitle

\begin{abstract}
Most of the recent few-shot learning (FSL) algorithms are based on transfer learning, where a model is pre-trained using a large amount of source data, and the pre-trained model is fine-tuned using a small amount of target data.
In transfer learning-based FSL, sophisticated pre-training methods have been widely studied for universal representation.
Therefore, it has become more important to utilize the universal representation for downstream tasks, but there are few studies on fine-tuning in FSL.
In this paper, we focus on how to transfer pre-trained models to few-shot downstream tasks from the three perspectives: update, data augmentation, and test-time augmentation.
First, we compare the two popular update methods, full fine-tuning (i.e., updating the entire network, FT) and linear probing (i.e., updating only a linear classifier, LP).
We find that LP is better than FT with extremely few samples, whereas FT outperforms LP as training samples increase.
Next, we show that data augmentation cannot guarantee few-shot performance improvement and investigate the effectiveness of data augmentation based on the intensity of augmentation.
Finally, we adopt augmentation to both a support set for update (i.e., data augmentation) as well as a query set for prediction (i.e., test-time augmentation), considering support-query distribution shifts, and improve few-shot performance.
The code is available at \url{https://github.com/kimyuji/updating_FSL}.
\end{abstract}

\section{Introduction}\label{sec:intro}
To address the issue of expensive labeling processes, the few-shot learning (FSL) paradigm was proposed\,\cite{fink2004object, fei2006one}.
%
%
FSL aims to recognize \emph{novel} classes using few training samples in the target data. Novel classes are disjoint from \emph{base} classes in the source data but are similar to base classes.
However, in real-world scenarios, obtaining labeled target data from certain domains such as X-ray images is difficult, which inevitably leads to the use of dissimilar labeled source data for pre-training\,\cite{guo2020broader, oh2022understanding}.
%
Cross-domain few-shot learning\,(CD-FSL), which has recently been studied to address this practicality issue, is more challenging than traditional FSL owing to the distributional gap between the source and target data.

There are two main approaches for (CD-)FSL: meta-learning and transfer learning.
Meta-learning approaches optimize a generalized model that adapts to new tasks quickly\,\cite{koch2015siamese, vinyals2016matching, snell2017prototypical, finn2017model}.
In contrast, transfer learning approaches aim to transfer knowledge from general source tasks to few-shot target tasks, based on the capacity of deep networks\,\cite{chen2019closer, dhillon2019baseline, wang2019simpleshot}.
%
Recently, \citet{guo2020broader} proposed a new benchmark for CD-FSL, Broader Study of CD-FSL (BSCD-FSL), and showed that transfer learning approaches outshine meta-learning approaches.
Based on these observations, CD-FSL algorithms have focused on the pre-training phase to extract improved and universal representation by exploiting unlabeled target data under a transfer learning framework\,\cite{phoo2020self, islam2021dynamic}.

\begin{figure}[t]
    \centering
    \includegraphics[width=\linewidth]{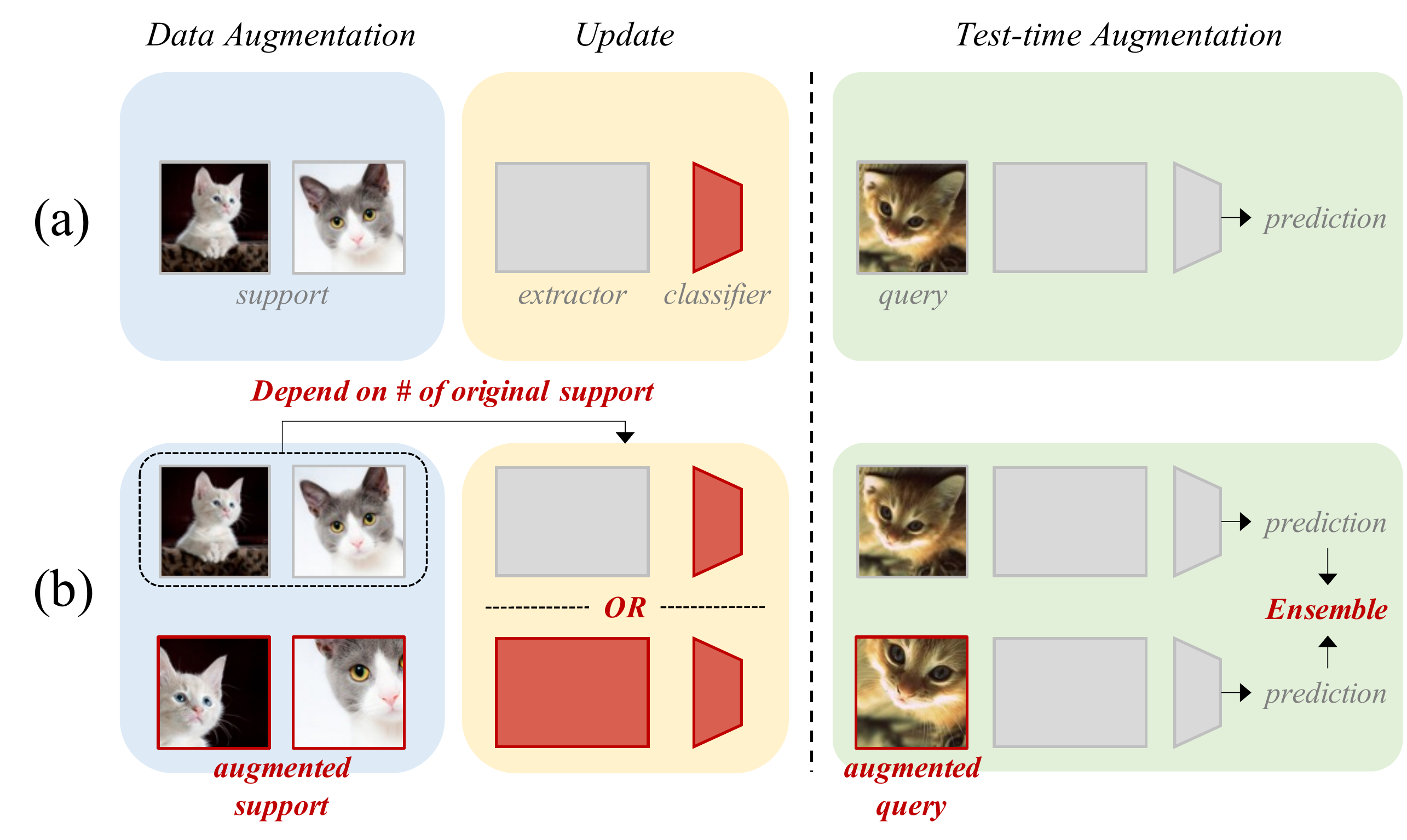}
    \caption{Comparison between strategies for FSL from the three perspectives: update, data augmentation (DA), and test-time augmentation. (a) The conventional strategy updates only a new classifier without DA. Then, prediction is performed using only the original test image. (b) The proposed strategy updates only a new classifier or the entire network, depending on the number of training samples, with DA. Then, prediction is performed using the original test image and augmented test images simultaneously.}
    \label{fig:comparison}
\end{figure}

However, despite the recent attention on (CD-)FSL \cite{phoo2020self, liang2021boosting, hu2022switch}, most of the prior works have followed a conventional fine-tuning strategy. Specifically, only a new classifier is fine-tuned using few training samples\,(a \emph{support} set) without data augmentation, and the fine-tuned model is evaluated using only given samples\,(a \emph{query} set), which is illustrated in Figure \ref{fig:comparison}(a).

Meanwhile, few studies have addressed how to fine-tune and evaluate models for FSL.
For example, \citet{guo2020broader} provided few-shot performance according to the updating parts, but they did not include extreme few-shot cases.
\citet{ni2021data} highlighted that flipping few target samples could improve the performance, but they did not consider augmentation methods other than flipping and cropping.
\citet{yamada2022one} adopted augmentation for evaluation, called test-time augmentation, but they manually determined significantly weak augmentation policies.

In this paper, we focus on how to fine-tune pre-trained models with few samples from three perspectives, as depicted in Figure \ref{fig:comparison}:

\begin{enumerate}[ {(}1{)} ]
    \item \textbf{Update.} We first investigate which update method is better in FSL and why, between \emph{full fine-tuning} (i.e., updating the entire network, FT) and \emph{linear probing} (i.e., updating only a classifier, LP). We show that the number of training samples is a significantly important factor in deciding the update method. For extremely few samples, LP is better than FT because a fine-tuned feature extractor can distort pre-trained representations. However, as the number of training samples increases, the performance gap between LP and FT decreases, and FT outperforms LP.
    \item \textbf{Data Augmentation (DA).} We define \emph{intensity of augmentation} based on the distance between the representations of the original images and augmented images, and demonstrate that the change in performance due to DA depends on the intensity of DA. It is observed that DA with strong intensity, such as MixUp\,\citep{zhang2017mixup} and CutMix\,\citep{yun2019cutmix}, degrades the FSL performance. We further schedule DA to alleviate performance degradation caused by the train-test (i.e., support-query) distribution shift when DA is applied.
    \item \textbf{Test-time Augmentation (TTA).} We propose a strategy that applies augmentation both during training time (i.e., data augmentation, DA) and test time (i.e., test-time augmentation, TTA)\footnote{We distinguish between the terms data augmentation (DA) and test-time augmentation (TTA) in this paper, which is explained in detail in the problem setup section.}, which is another alternative for reducing support-query distribution shifts. We show that the proposed strategy (using both DA and TTA, Figure \ref{fig:comparison}(b)) results in significantly better performance than the conventional strategy (neither DA nor TTA, Figure \ref{fig:comparison}(a)) and other variants (either DA or TTA) for FSL. Furthermore, we demonstrate that \emph{matching the powerful intensity} of augmentation for DA and TTA is essential, and that the performance gain according to the number of augmented images for TTA is significant when augmentation with strong intensity is used.
\end{enumerate}

\section{Preliminaries}\label{sec:preliminaries}

\subsection{Problem Setup}
We explain the training (pre-training and fine-tuning) and evaluation of transfer-based FSL.
Formally, a model $h_{\sf 1} \circ f$ is trained on labeled source data $\mathcal{D}_{B}$ in the pre-training phase, where $f$ is a feature extractor and $h_{\sf 1}$ is a linear classifier for pre-training.
The pre-trained classifier $h_{\sf 1}$ is then replaced with a randomly initialized linear classifier $h_{\sf 2}$ to recognize novel classes $\mathcal{C}_{N}$.
Finally, (1) $h_{\sf 2} \circ f$ is updated (i.e., FT) or (2) only $h_{\sf 2}$ is updated (i.e., LP) on few labeled target data $\mathcal{D}_{S}$ by cross-entropy loss and evaluated on $\mathcal{D}_{Q}$.

Here, $(\mathcal{D}_{S}, \mathcal{D}_{Q})$ is sampled from labeled target data $\mathcal{D}_{N}$ in an episodic manner.
Namely, $n$ classes are sampled from novel classes $\mathcal{C}_{N}$, while $k$ and $k_q$ examples per class are sampled to form $\mathcal{D}_{S}$ and $\mathcal{D}_{Q}$, respectively.
Thus, $\mathcal{D}_{S}$ and $\mathcal{D}_{Q}$ are defined as $\{(x_i^s,y_i^s)\}_{i=1}^{n \times k}$ and $\{(x_i^q,y_i^q)\}_{i=1}^{n \times k_q}$, and each is called a support set and a query set, respectively.
This problem setting is referred to as the $n$-way $k$-shot classification problem, where $n=5$ and $k \in \{1, 5, 20\}$ in general.
The reported few-shot performance is averaged over 600 episodes.

Regarding augmentation, DA refers to when $\mathcal{D}_{S}$ is augmented for the update, and TTA refers to when $\mathcal{D}_{Q}$ is augmented for evaluation.

\subsection{Implementation Details}\label{subsec:impl}
We pre-train ResNet10 on miniImageNet (miniIN) and ResNet18 on tieredImageNet (tieredIN).\footnote{We provide the results of ResNet10 pre-trained on miniIN in the main paper and those of ResNet18 pre-trained on tieredIN in Appendix \ref{appx:FT_LP_tiered} and \ref{appx:datta_tiered}.}
Pre-trained models are fine-tuned on 10 target datasets: miniIN, tieredIN, four non-BSCD-FSL (Places, Plantae, Cars, and CUB), and four BSCD-FSL (CropDiseases, EuroSAT, ISIC, and ChestX). Note that non-BSCD-FSL datasets are more similar to the source data than BSCD-FSL datasets\,\citep{ericsson2021well}.
The detailed setup of datasets is from prior works, which is described in Appendix \ref{appx:setup}.

We follow \citet{guo2020broader} and \citet{oh2022understanding} for the pre-training and fine-tuning setup.
For pre-training, an SGD optimizer with an initial learning rate of 1e-1, the momentum of 0.9, and weight decay of 1e-4 is used. We train for 1000 epochs with a batch size of 64 on miniIN, and learning rate is decayed by 1e-1 at epochs (400, 600, 800). For tieredIN, We train for 90 epochs with a batch size of 256.
For fine-tuning, an SGD optimizer with a learning rate of 1e-2, the momentum of 0.9, and weight decay of 1e-3 is used. We update for 100 epochs with a batch size of 4 for $\{1,5\}$-shot settings and 16 for $20$-shot settings.

For augmentation, we exploit the two types of data augmentation techniques: \emph{single augmentation}, which transforms one image, and \emph{mixing augmentation}, which mixes two images.
Single augmentation includes horizontal flipping (HFlip), random resized cropping (RCrop), color jittering (CJitter), and base augmentation (Base Aug). Base Aug applies HFlip, RCrop, and CJitter simultaneously.
Mixing augmentation includes MixUp\,\citep{zhang2017mixup} and CutMix\,\citep{yun2019cutmix}.
The details of augmentation techniques are given in Appendix \ref{appx:DA}.

\section{Update for (CD-)FSL}\label{sec:analysis1}

\begin{figure}[t]
    \centering
    \includegraphics[width=\linewidth]{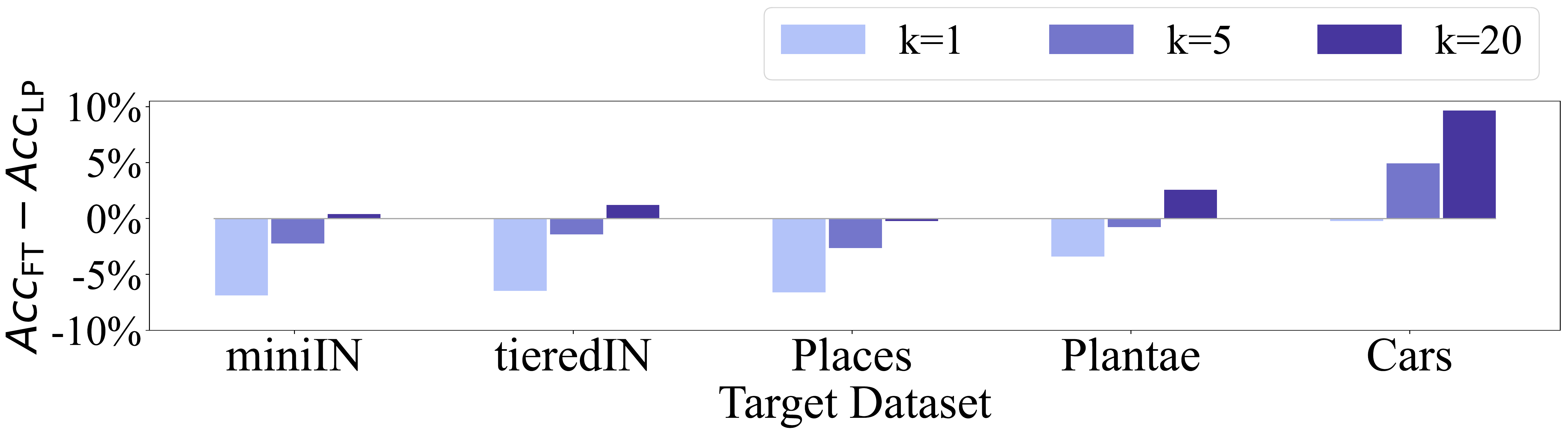}
    \includegraphics[width=\linewidth]{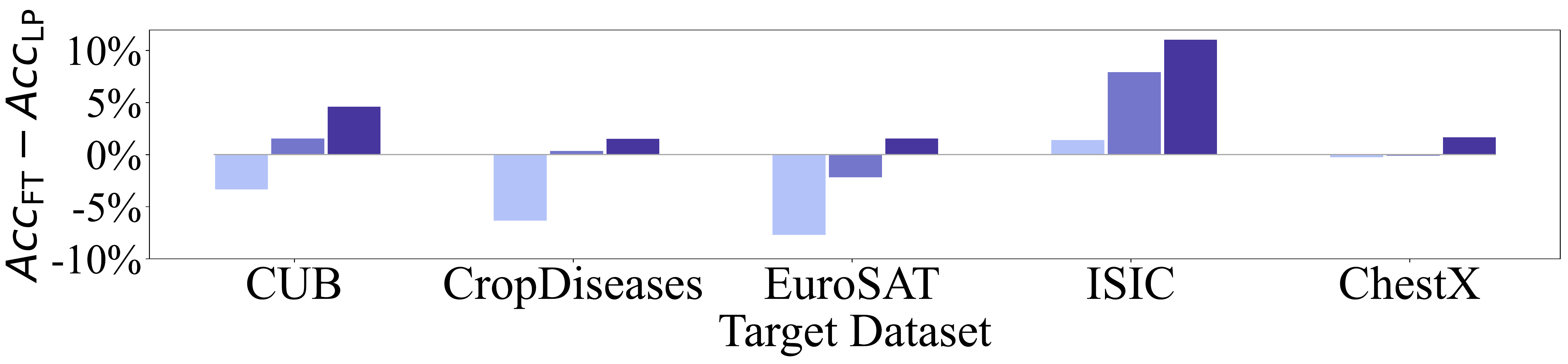}
    \caption{5-way $k$-shot performance difference of FT with LP on various target datasets.}
    \label{fig:why_full}
\end{figure}

We compare popular updating methods, FT\,(i.e., full fine-tuning) and LP\,(i.e., linear probing), under different distributional gaps between the source and target data.
Figure \ref{fig:why_full} describes the performance difference of FT and LP (i.e., $Acc_{\sf FT}-Acc_{\sf LP}$) on various target datasets, according to the number of training samples.
It is observed that shot is a more critical factor than a distributional difference in deciding how to update a pre-trained model.
In line with the observation that FT is a better strategy than LP\,\cite{kornblith2019better, zhai2019large, kumar2022finetuning}, our results support that it even holds for FSL.
The results of ResNet18 pre-trained on tieredIN show the same trend, as reported in Appendix \ref{appx:FT_LP_tiered}.

\begin{figure}[t]
    \centering
    
    
    \begin{subfigure}[h]{0.36\linewidth}
    \includegraphics[width=\linewidth]{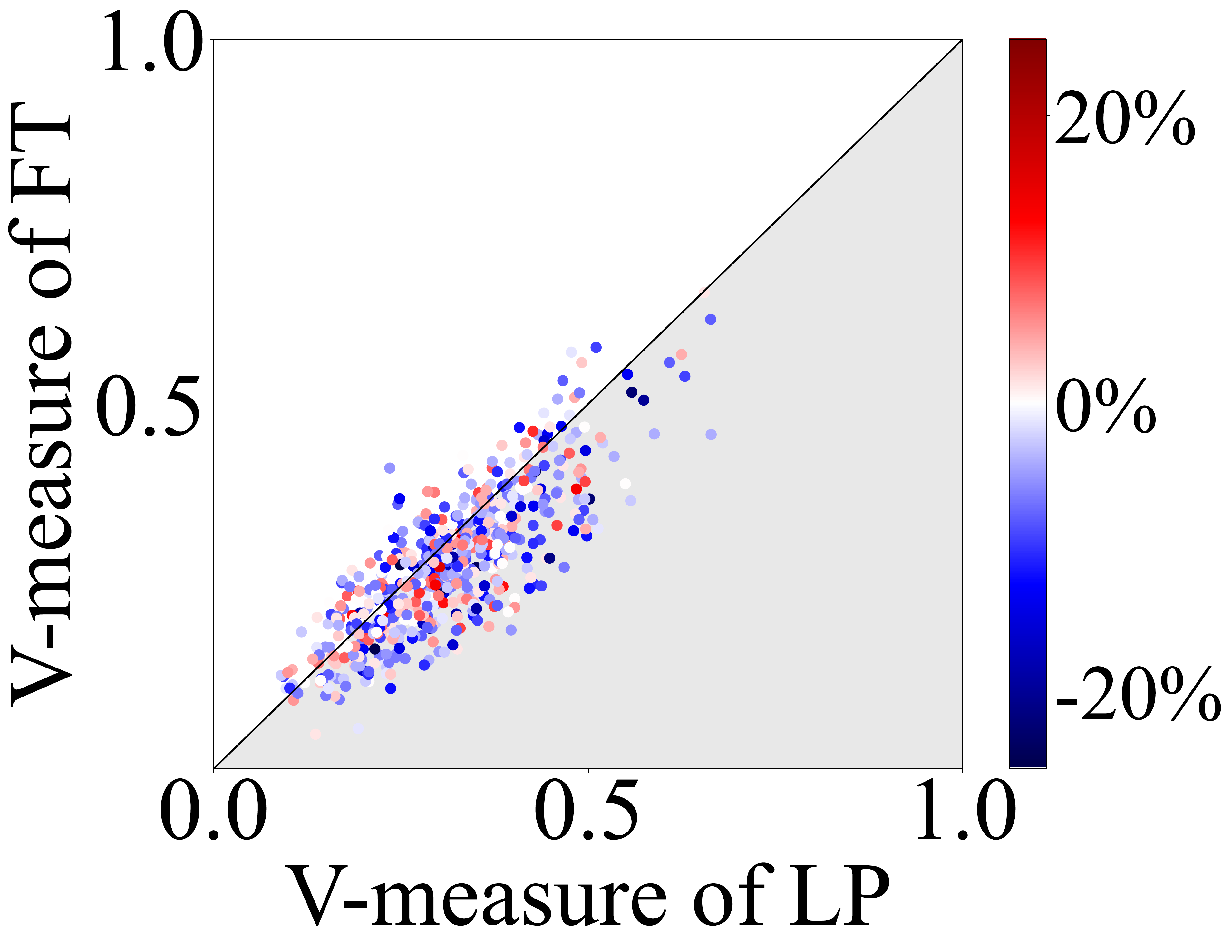}
        \caption{CUB ($k=1$)}
    \end{subfigure}
    \begin{subfigure}[h]{0.31\linewidth}
    \includegraphics[width=\linewidth]{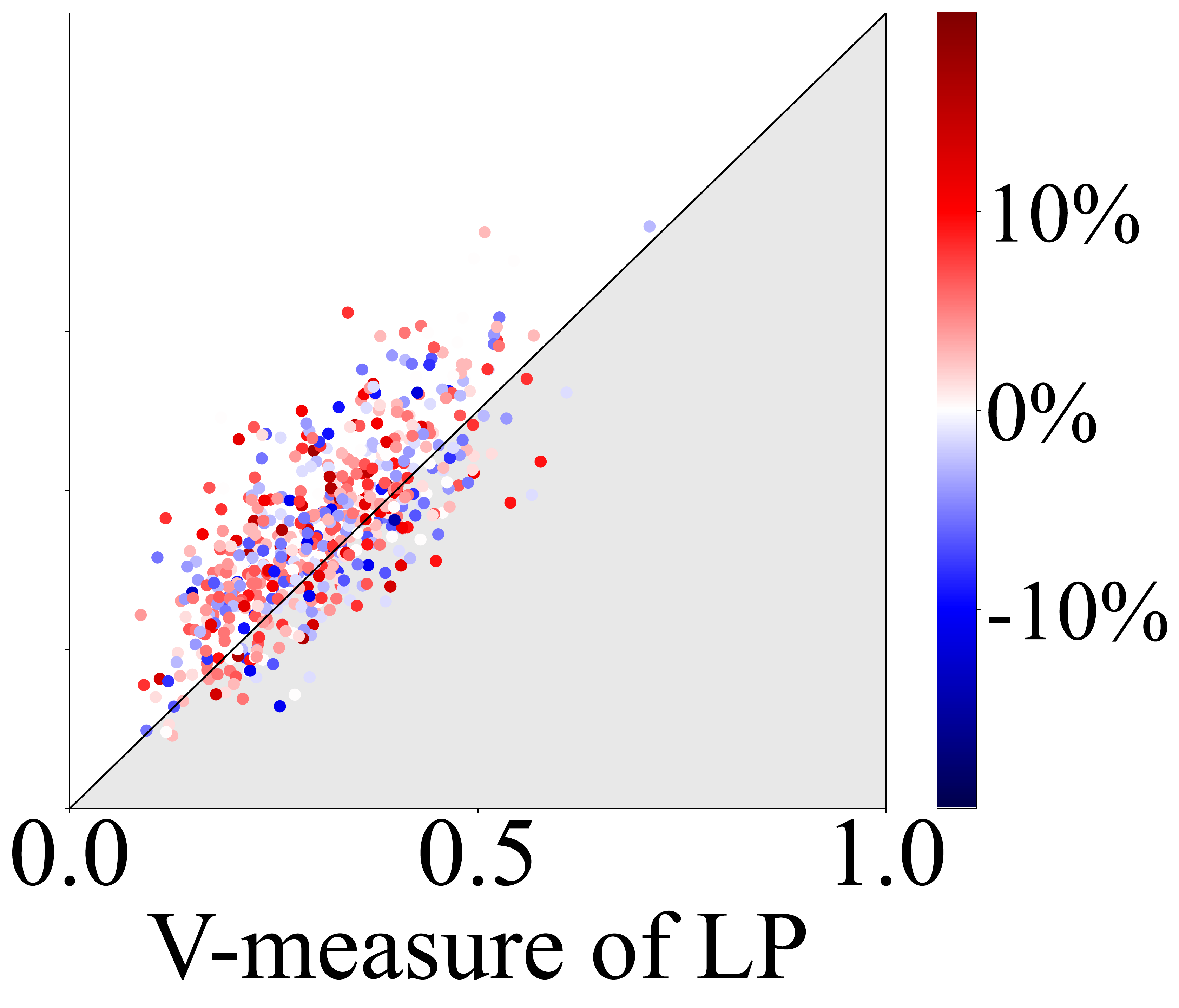}
        \caption{CUB ($k=5$)}
    \end{subfigure}
    \begin{subfigure}[h]{0.31\linewidth}
    \includegraphics[width=\linewidth]{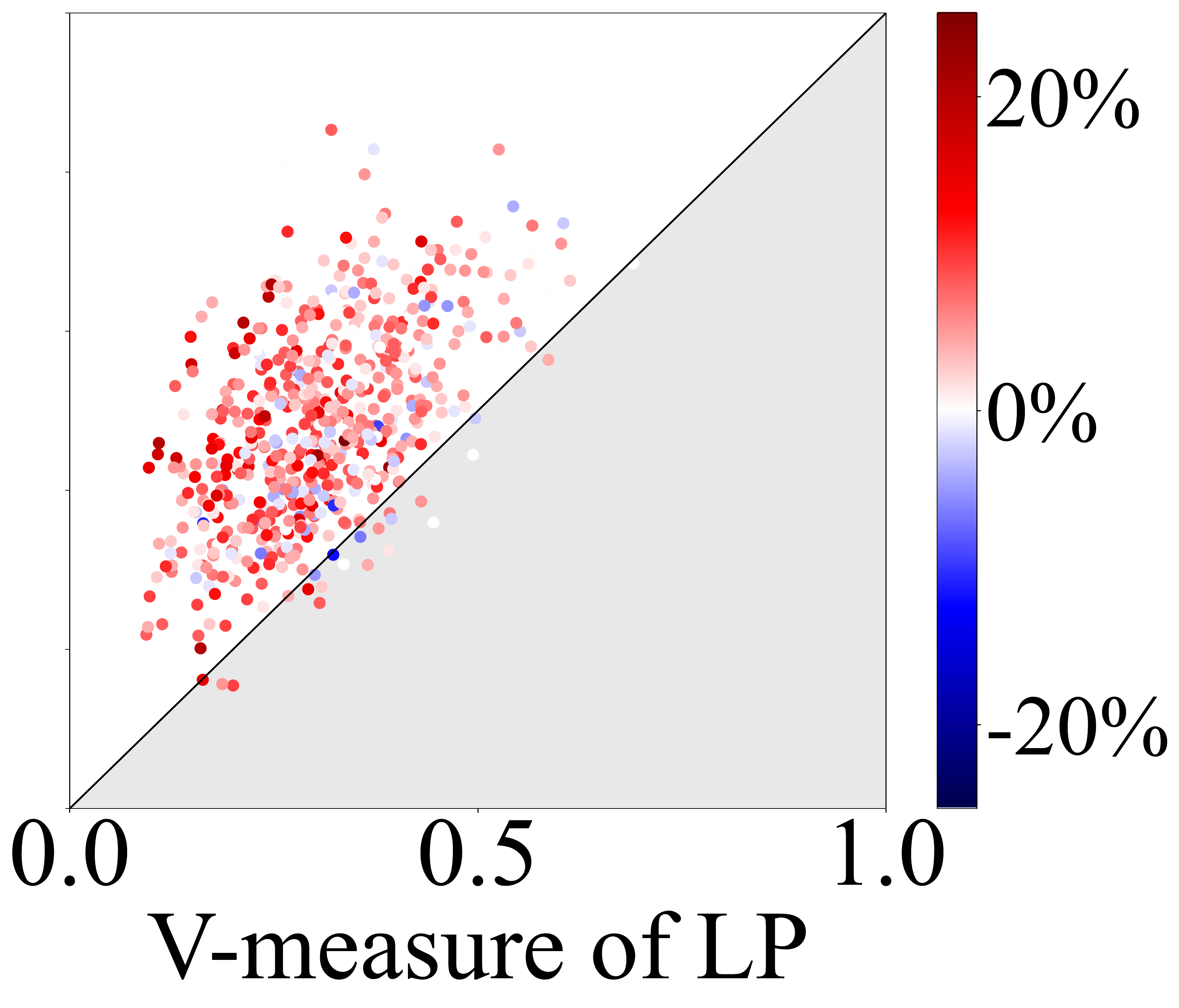}
        \caption{CUB ($k=20$)}
    \end{subfigure}
    \caption{Feature extractor's clustering ability by V-measure and performance differences between LP and FT. Each dot indicates an episode. The color denotes $Acc_{\sf FT}-Acc_{\sf LP}$. Red indicates that FT outperforms LP, whereas blue means that LP outperforms FT.}
    \label{fig:v_measure_main}
\end{figure}

    

However, LP is better than FT when there is only one sample for each class (i.e., $k=1$) in most target datasets.
This is because FT with extremely few samples can distort the representation of a query set used in the evaluation, weakening the clustering ability of the feature extractor.
To understand the effectiveness of FT for learning representation, we analyze the V-measurement cluster score\,\cite{rosenberg2007v} before and after FT. 
Figure \ref{fig:v_measure_main} depicts the relationship between changes in performance and V-measure over 600 episodes.
Each dot indicates a query set of an episode, and the x-axis and y-axis indicate the V-measure of the updated model through LP and FT, respectively. 
Note that LP keeps the same extractor of the pre-trained model, while FT updates the pre-trained extractor using few target data.
The red color indicates that FT outperforms LP, whereas the blue color means that LP outperforms FT.
V-measure is defined as 1) K-Means clustering on representations of a query set using a pre-trained extractor\,(i.e., LP) and an updated extractor\,(i.e., FT), and then 2) calculating the agreement between the ground-truth and predicted K-Means clusters. 
The higher the V-measure score, the better the clustering ability of the feature extractor.
The values of 0 and 1 indicate no agreement and maximum agreement, respectively. 

Figure \ref{fig:v_measure_main}(a), \ref{fig:v_measure_main}(b), and \ref{fig:v_measure_main}(c) describe $1$-, $5$-, and $20$-shot cases on CUB, respectively.
For the 1-shot case, most episodes are located below $y=x$ (the gray region in Figure \ref{fig:v_measure_main}), which means that the V-measure decreases by FT.
This implies that a fine-tuned extractor with one sample per class distorts the pre-trained representation of a query set for clustering.
In this context, most episodes have a color close to blue because the performance of LP is higher than that of FT.
On the contrary, for the 5-shot case, most episodes are located above $y=x$ (the white region in Figure \ref{fig:v_measure_main}) and have a color close to red. 
For the 20-shot case, the increase in V-measure and performance is more significant, indicating that FT forms even better clusters than LP as the number of shots increases.
Therefore, it is concluded that choosing between FT and LP depends on \emph{the number of training samples}.
The results on other datasets are given in Appendix \ref{appx:v_measure_other}.

\section{Data Augmentation for (CD-)FSL}\label{sec:analysis2}
To generate more samples during fine-tuning, we can use DA, which is known to prevent overfitting caused by the lack of variation of data\,\cite{perez2017effectiveness}.
In this section, we study DA techniques for (CD-)FSL when updating a pre-trained model. For consistency in this section, we provide experimental results of FT with DA. The results of LP with DA are provided in Appendix \ref{appx:LP_DA}.
%


\subsection{FT with Data Augmentation}\label{subsec:ft_da}

\begin{figure}[t]
    \centering
    \includegraphics[width=\linewidth]{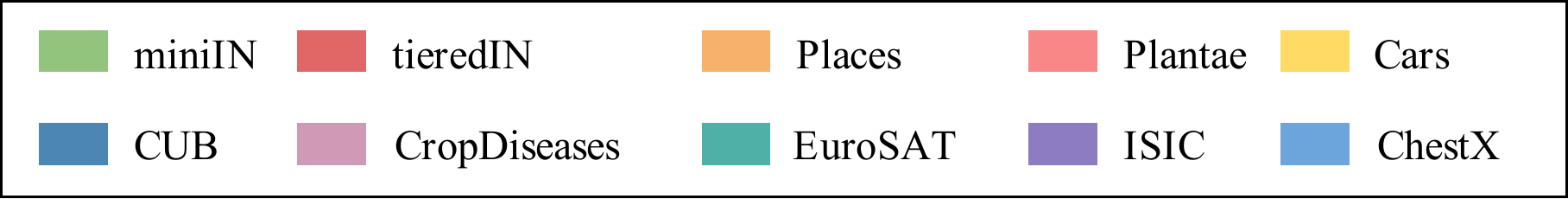}
    
    \begin{subfigure}[h]{0.355\linewidth}
    \includegraphics[width=\linewidth]{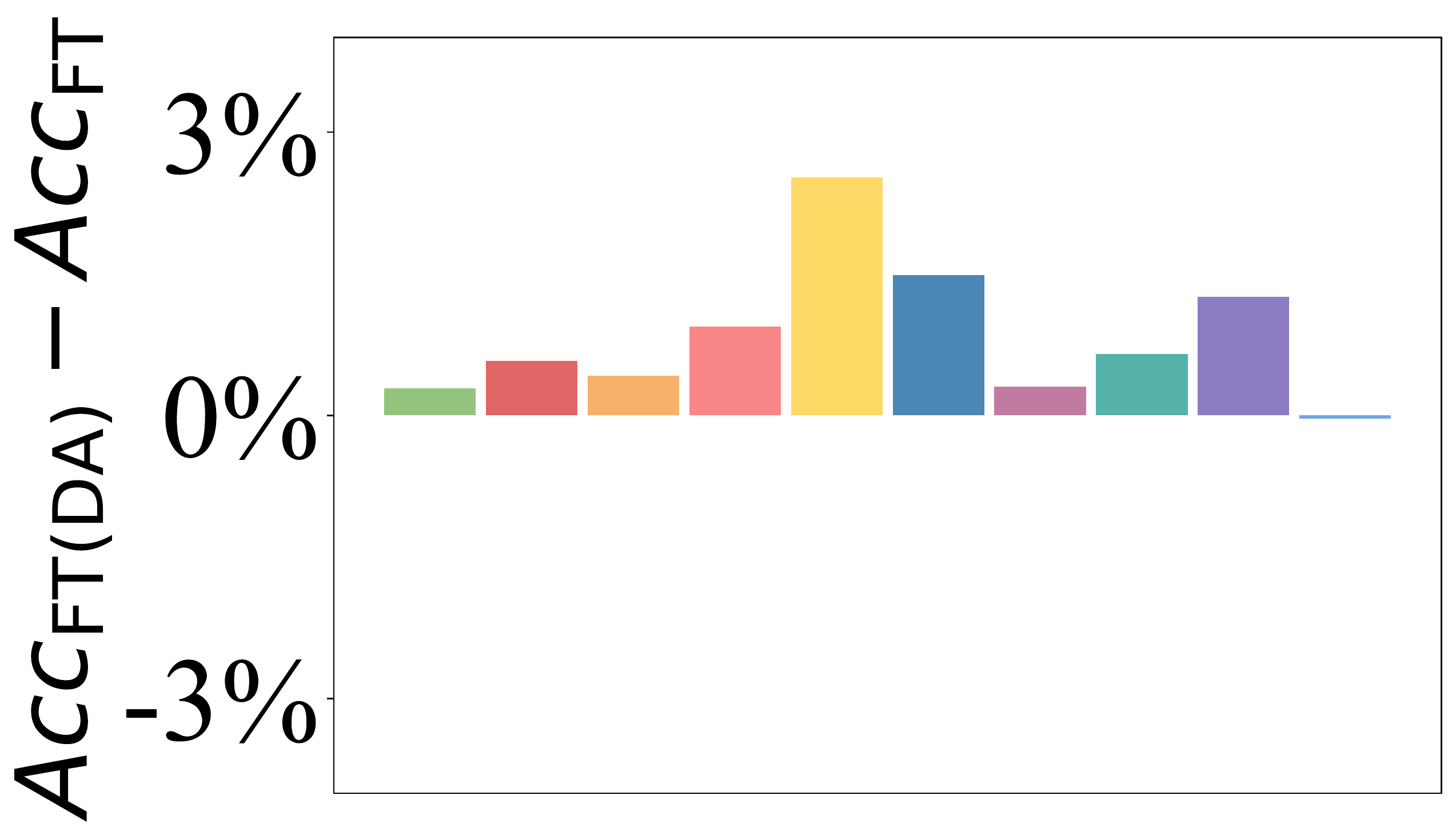}
        \caption{HFlip}
    \end{subfigure} \hfill
    \begin{subfigure}[h]{0.312\linewidth}
    \includegraphics[width=\linewidth]{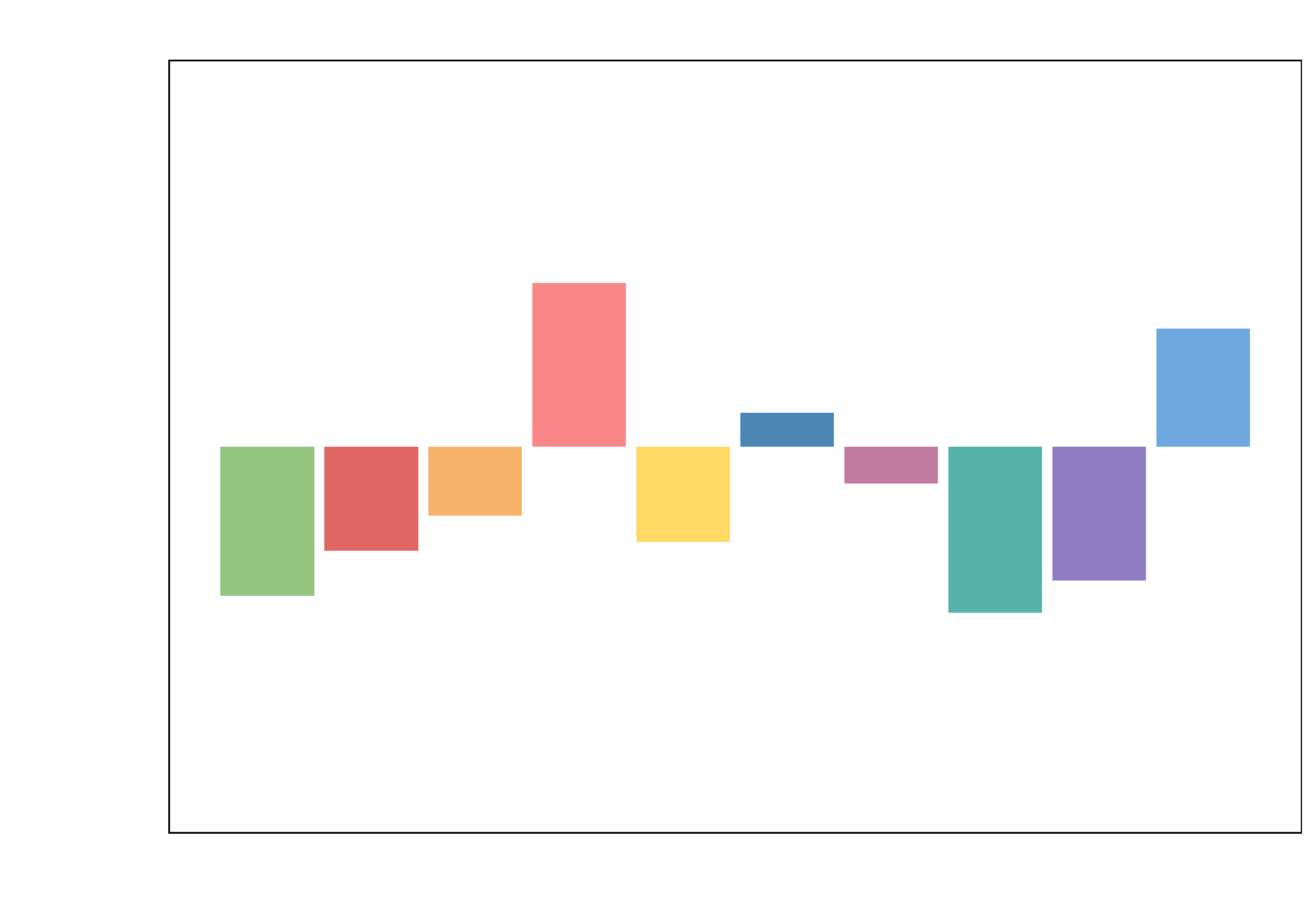}
        \caption{RCrop}
    \end{subfigure} \hfill
    \begin{subfigure}[h]{0.312\linewidth}
    \includegraphics[width=\linewidth]{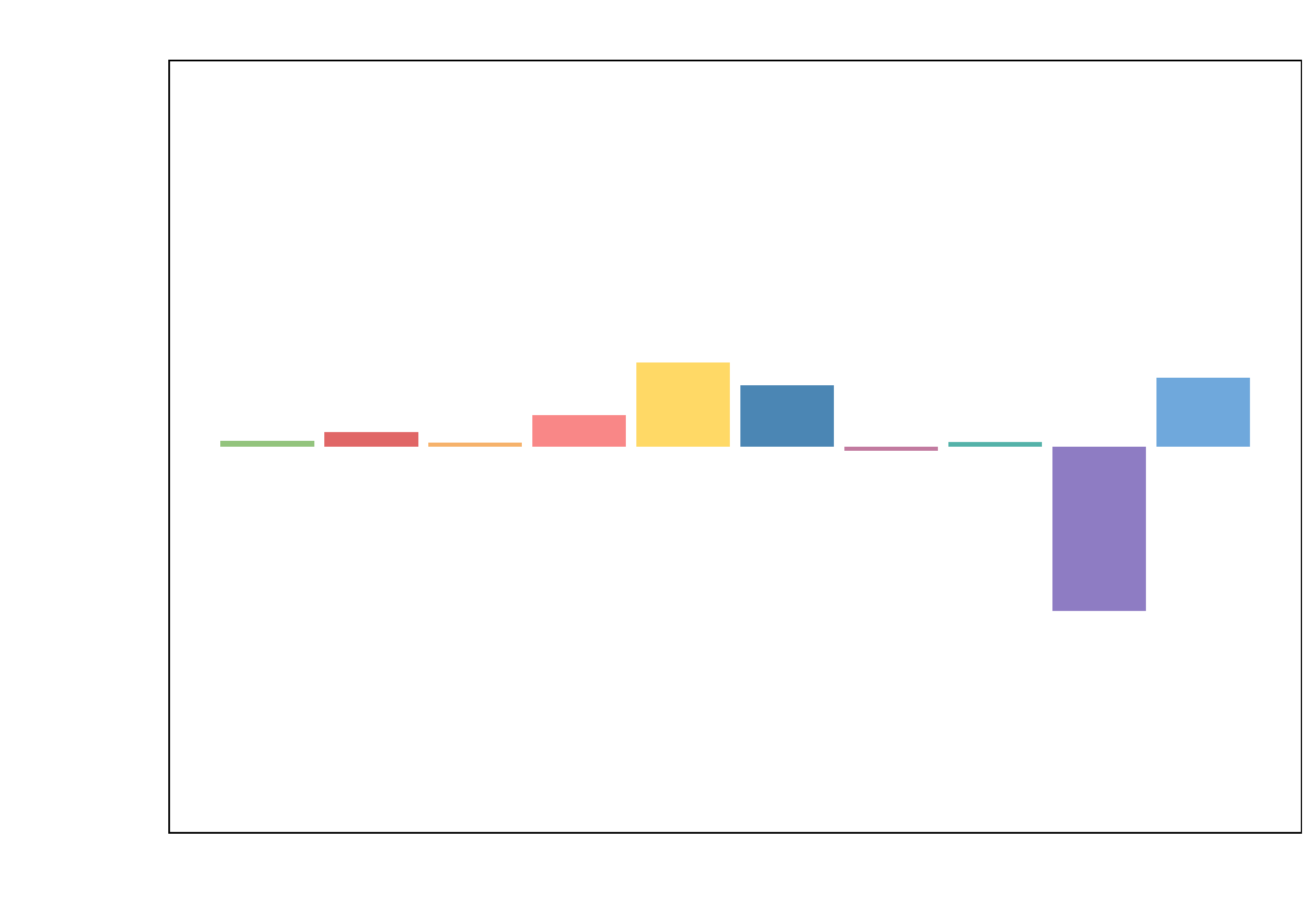}
        \caption{CJitter}
    \end{subfigure}
    
    \begin{subfigure}[h]{0.355\linewidth}
    \includegraphics[width=\linewidth]{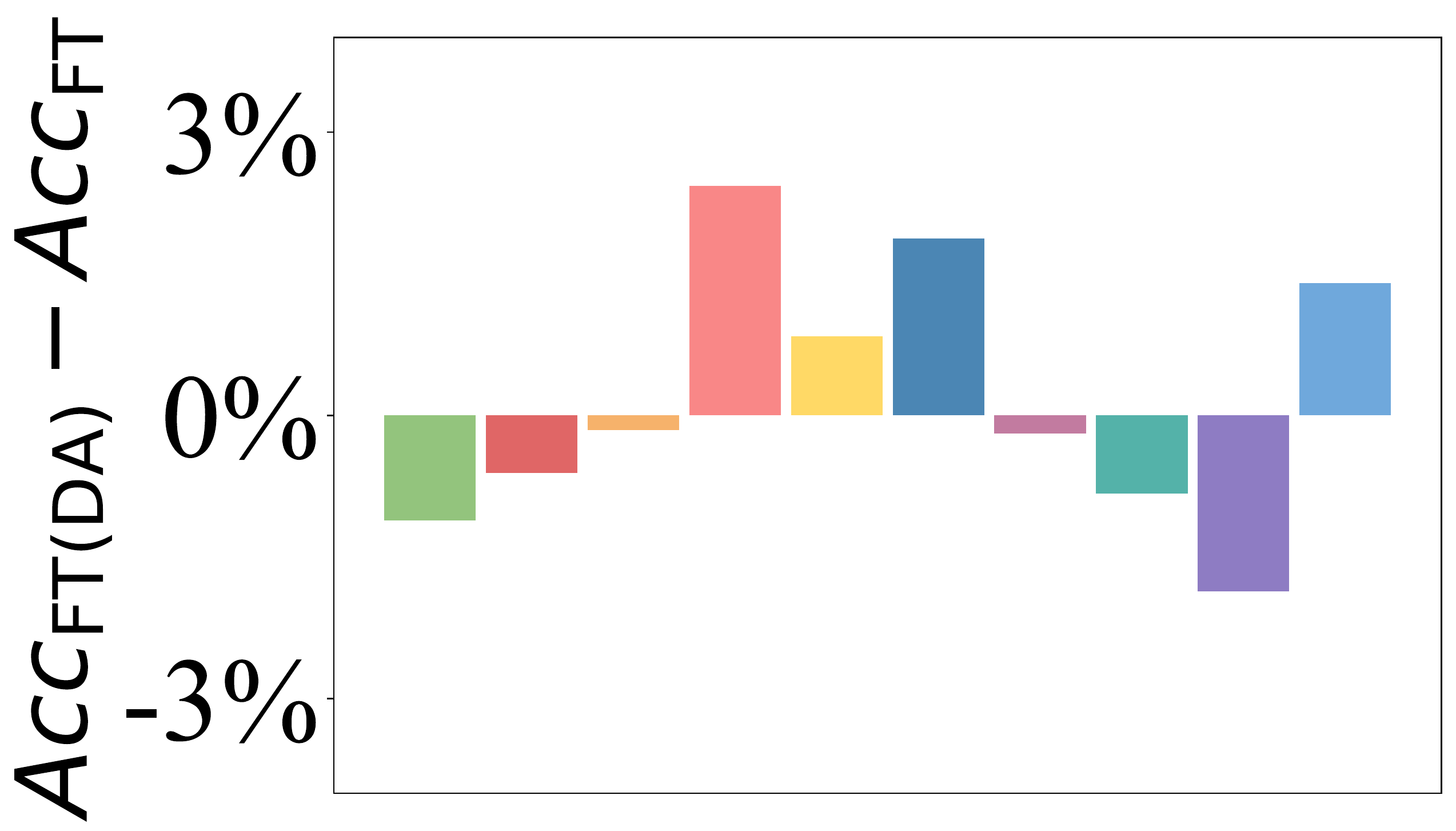}
        \caption{Base Aug}
    \end{subfigure} \hfill
    \begin{subfigure}[h]{0.312\linewidth}
    \includegraphics[width=\linewidth]{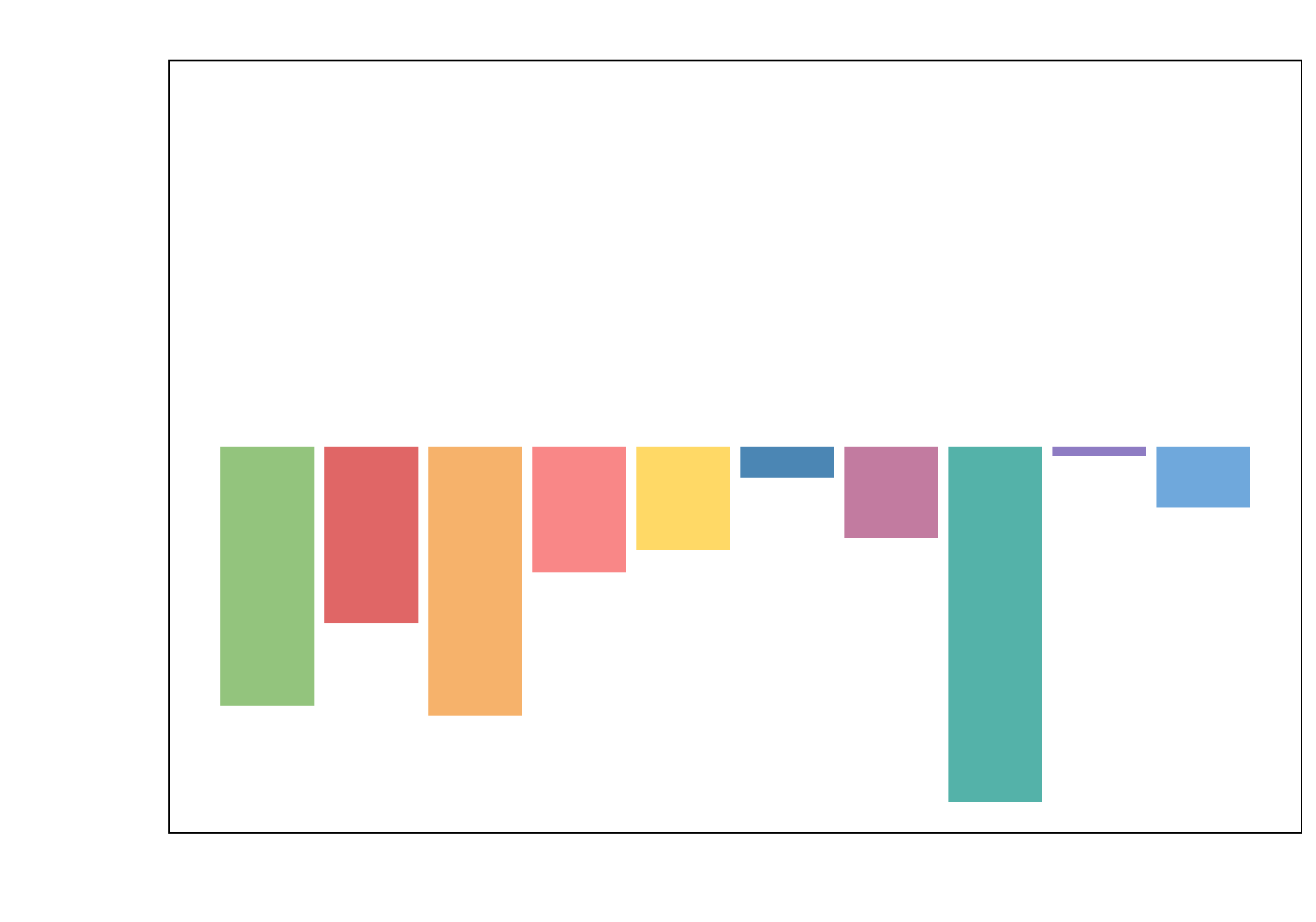}
        \caption{MixUp}
    \end{subfigure} \hfill
    \begin{subfigure}[h]{0.312\linewidth}
    \includegraphics[width=\linewidth]{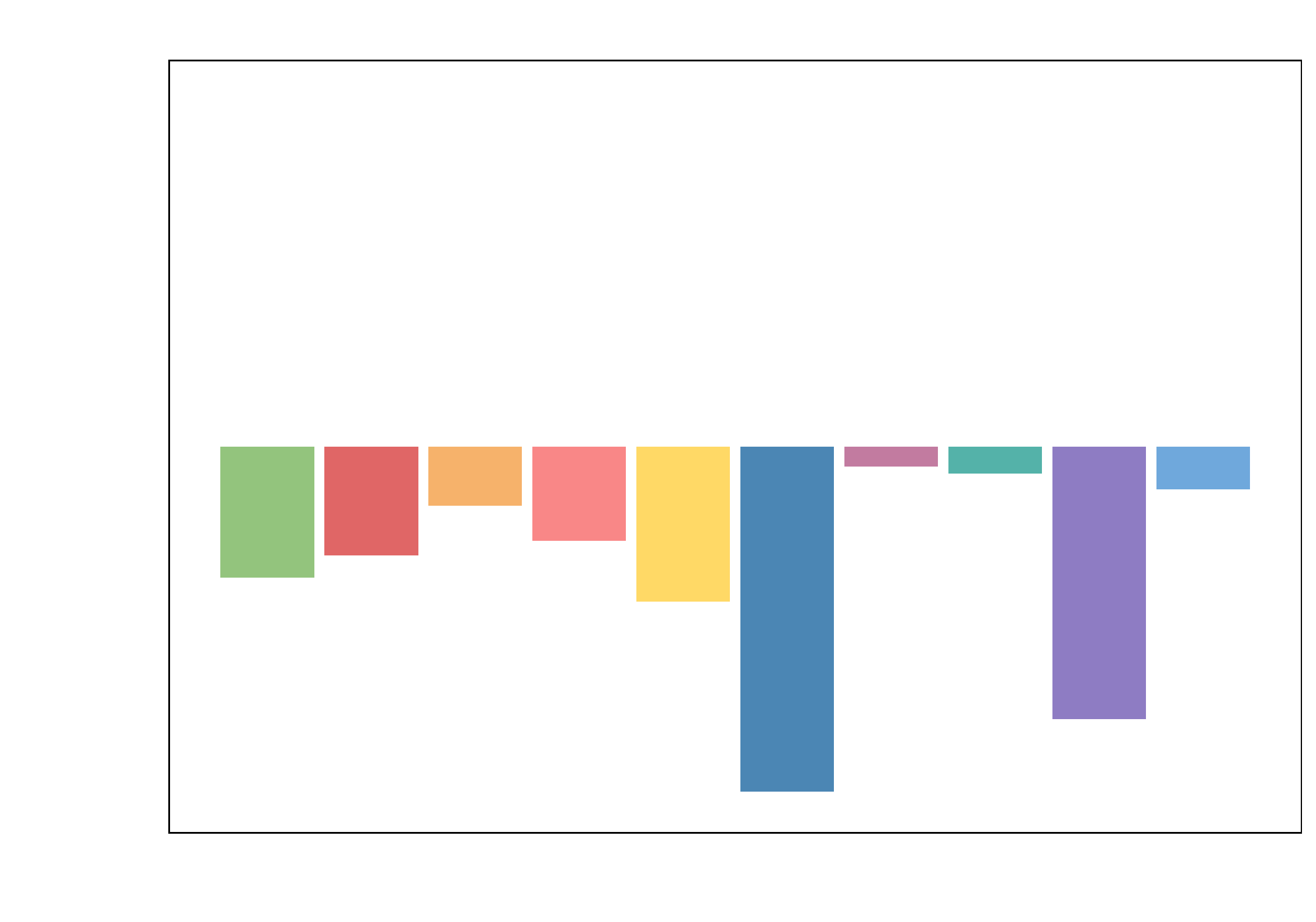}
        \caption{CutMix}
    \end{subfigure}
    \caption{5-way 20-shot performance difference of FT(DA) with FT according to the augmentation techniques. Base augmentation includes horizontal flipping (HFlip), random resized cropping (RCrop), and color jittering (CJitter).}
    \label{fig:miniIN_20_shot_da}
\end{figure}

Although DA generally leads to performance improvement, no significant performance gain occurs overall; rather, it has a negative effect under (CD-)FSL.
Figure \ref{fig:miniIN_20_shot_da} describes the 5-way 20-shot performance gap between FT with DA and FT without DA (i.e., $Acc_{\sf FT(DA)}-Acc_{\sf FT}$). The results include six augmentation techniques, explained in Section \ref{subsec:impl}.
Among the augmentation techniques, HFlip and CJitter improve the few-shot performance on most target datasets.
Meanwhile, other single augmentation techniques such as RCrop and Base Aug increase the FSL performance only for a few target datasets and decrease it for the rest. Furthermore, mixing augmentation techniques significantly decrease the FSL performance for most target datasets. The results of other shots are reported in Appendix \ref{appx:few_shot_da}.

\begin{table}[t]
    \centering
    \resizebox{\linewidth}{!}{
    \begin{tabular}{cccccccc}
    \toprule
    Measurement & $f$ & HFlip & RCrop & CJitter & Base Aug & MixUp & CutMix \\ \midrule
    \multirow{4}{*}{Intensity} & ResNet10 & 1.692 & 7.227 & 1.736 & 7.454 & 9.127 & 9.103 \\
    & ResNet50 & 1.854 & 9.217 & 2.538 & 9.530 & 15.894 & 15.583 \\
    & DesNet121 & 3.447 & 18.177 & 4.927 & 18.623 & 26.418 & 25.895 \\
    & EfficietNet-b0 & 1.483 & 8.749 & 2.720 & 9.117 & 13.840 & 14.447 \\ \midrule
    Diversity & ResNet10 & 0.226 & 0.479 & 0.225 & 0.530 & 0.828 & 0.860 \\
    \bottomrule
    \end{tabular}}
    \caption{Intensity (ours) and diversity \cite{gontijo2020affinity} of augmentation techniques. For $f$, ResNet10 pre-trained on miniIN and others provided by \texttt{timm} open-source library \cite{rw2019timm} are used. For diversity, ResNet10 is only used, because it is defined by the final training loss.}\label{tab:magnitude}
\end{table}

\subsection{Intensity of Augmentation}
We hypothesize that the FSL performance highly depends on the intensity of augmentation.
For instance, the weaker intensity like HFlip\,(i.e., only an original or a flipped image) can increase the performance, whereas the stronger intensity like MixUp can decrease it.
This is because DA itself makes the train-test\,(i.e., support-query) distribution shift, and a stronger intensity can make this shift larger.
In other words, the FSL performance is expected to decrease as the intensity increases, because a model is updated with an augmented support set $\mathcal{D}_{S}$ but evaluated on an original (i.e., non-augmented) query set $\mathcal{D}_{Q}$.

Formally, we define $\mathcal{I}_{\sf Aug}$, which is \emph{intensity of augmentation}, for single augmentation (Eq. (\ref{eq:single})) and mixing augmentation (Eq. (\ref{eq:mixing})) as follows:

\begin{equation}\label{eq:single}
    \frac{1}{|S|} \sum_{x \in S} d(f(x), f({\sf Aug}(x)))
\end{equation}

\begin{equation}\label{eq:mixing}
    \frac{1}{|S|^2} \sum_{(x_1, x_2) \in S \times S} \frac{d(f(x_1), f(\tilde{x})) + d(f(x_2), f(\tilde{x}))}{2} 
\end{equation}

\noindent where $S$ is a subset of a popular dataset such as ImageNet to decrease computation costs, $f$ is a large pre-trained extractor following \citet{cui2018large} and \citet{Li2020Rethinking}, and $d$ is the Euclidean distance. In Eq. (\ref{eq:mixing}), $\tilde{x}$ is an augmented image by mixing $x_1$ and $x_2$ using MixUp or CutMix.
We use a popular dataset and a large pre-trained extractor to focus on \emph{augmentation}, minimizing the bias or dependency on datasets and feature extractors.
Intensity intuitively describes the distance between the representations of the original and augmented images.
Therefore, the larger the value, the stronger the intensity of augmentation.

Similar to intensity, \citet{gontijo2020affinity} proposed \emph{diversity} to measure the complexity of augmentation, defined as the final training loss when training with augmentation.
Their measurement (diversity) requires training costs, which our measurement (intensity) does not require.

It is shown that the intensity or diversity of augmentation is highly correlated with performance degradation for FSL from Figure~\ref{fig:miniIN_20_shot_da} and Table~\ref{tab:magnitude}.
Table \ref{tab:magnitude} describes the intensity and diversity of single and mixing augmentation techniques, and detailed hyperparameters of these techniques are in Appendix \ref{appx:DA}.
To calculate the intensity, miniImageNet and various $f$ such as ResNet \cite{he2016deep}, DenseNet \cite{huang2017densely}, and EfficientNet \cite{tan2019efficientnet} are used.
For diversity, the final training losses over all target datasets are averaged.
The ranking of intensity is almost the same if different feature extractors are used. Therefore, ResNet50 is used to calculate the intensity in the rest of the paper.
Furthermore, the rankings by intensity and diversity are almost identical ({\slshape HFlip\,$\approx$\,CJitter\,$<$\,RCrop\,$\approx$\,Base Aug\,$<$\,MixUp\,$\approx$\,CutMix}).
%

\begin{table}[t]
    \centering
    \small
    \begin{tabular}{c|cccc}
    \toprule
    \textsc{default} & \textsc{weaker} & \textsc{weak} & \textsc{strong} & \textsc{stronger} \\ \midrule
    9.530 & 3.654 & 5.868 & 7.993 & 10.994 \\
    \bottomrule
    \end{tabular}
    \caption{Intensity of Base Aug. \textsc{default} is from \citet{oh2022understanding}, and the other four policies are manually designed.}\label{tab:base_magnitude}
\end{table}

\begin{figure}[t]
    \centering
    \includegraphics[width=\linewidth]{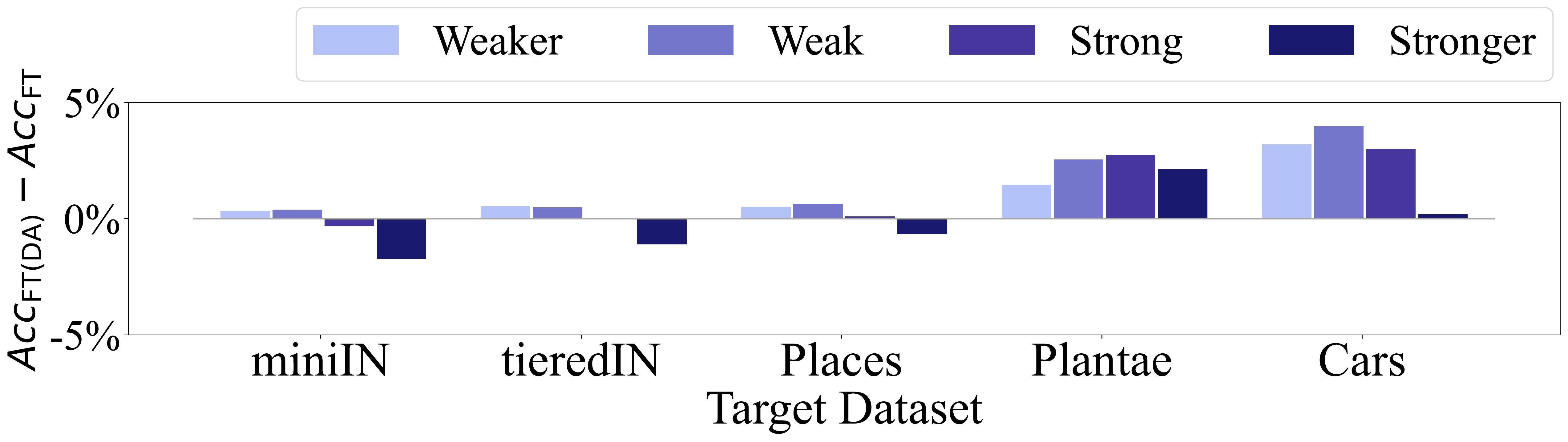}
    \includegraphics[width=\linewidth]{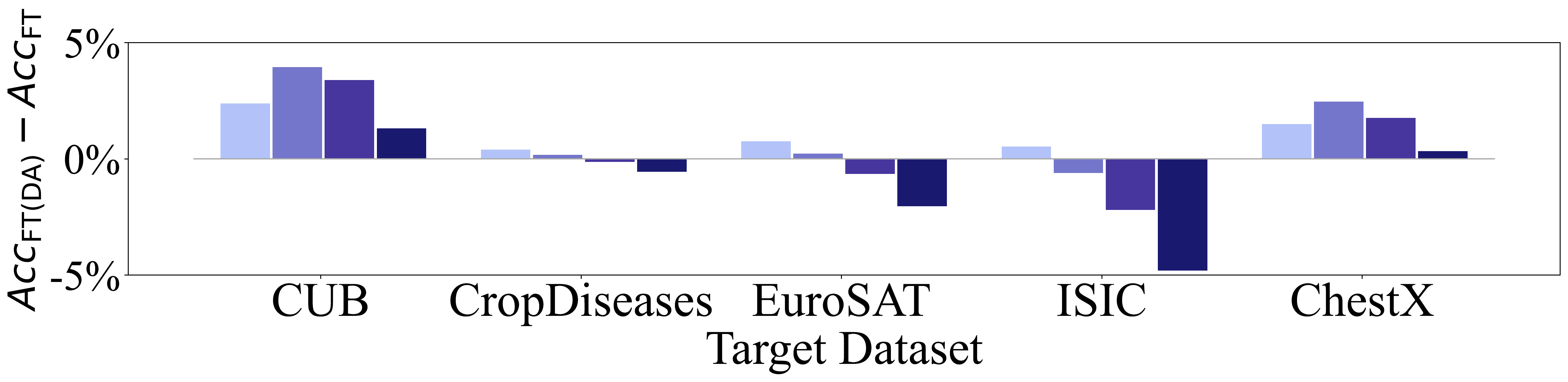}
    \caption{5-way 20-shot performance difference of FT(DA) with FT according to the designed intensity of Base Aug.}
    \label{fig:base_magnitude}
\end{figure}

\subsection{Analysis on Intensity of Single Augmentation}
We analyze the intensity of single augmentation using Base Aug.
We vary the hyperparameters of RCrop (range of crop scale) and CJitter (brightness, contrast, and saturation, which have the same range) to regulate the intensity of Base Aug.
In other words, a stronger intensity can be specified by a lower crop scale ($<1$) and values of brightness, contrast, and saturation far from 1. For analysis, four combinations of (crop scale [min, max], values of CJitter [min, max]) are designed: \textsc{weaker} ([0.9, 1.0], [0.8, 1.2]) $\subset$ \textsc{weak} ([0.6, 1.0], [0.6, 1.4]) $\subset$ \textsc{strong} ([0.3, 1.0], [0.4, 1.6]) $\subset$ \textsc{stronger} ([0.01, 1.0], [0.2, 1.8]). These settings are similar to the \textsc{default} ([0.08, 1.0], [0.6, 1.4]).
Table \ref{tab:base_magnitude} provides the intensity of augmentation using Eq. (\ref{eq:single}), which implies that intensity well-reflects the changes according to crop scale and values of brightness, contrast, and saturation.

Figure \ref{fig:base_magnitude} depicts the 5-way 20-shot performance difference of FT with DA and FT according to the designed intensity.
Each dataset has the best intensity among four designed intensities, in which the best performance is achieved. For example, \textsc{weak} is the best intensity on CUB, while \textsc{strong} is the best intensity on Plantae.
Furthermore, the performance increases before the best intensity, then decreases after the best intensity on each dataset.
%
%
%
Namely, monotonic decreasing cases (e.g., EuroSAT and ISIC) are expected to have the optimal intensity less than \textsc{weaker}.
%
%
%
%
The results of other shots are reported in Appendix \ref{appx:single_aug}.
%
%

\begin{table}[t]
    \centering
    \resizebox{\linewidth}{!}{
    \begin{tabular}{cccccc}
    \toprule
    MixUp & MixUp(W) & MixUp(B) & CutMix & CutMix(W) & CutMix(B) \\ \midrule
    15.894 & 9.002 & 16.020 & 15.583 & 8.737 & 15.716 \\
    \bottomrule
    \end{tabular}}
    \caption{Decoupled intensity of mixing augmentation.}\label{tab:decomp_magnitude}
\end{table}

\begin{table}[t]
\centering
\resizebox{\linewidth}{!}{
\begin{tabular}{ccccccc}
    \toprule
    $k$ & MixUp & miniIN & \!\!\!\!CropDiseases\!\!\!\! & EuroSAT & ISIC & ChestX \\
    \midrule
    \multirow{2}{*}{1} & -  & 49.34{\scriptsize$\pm$.77} & 66.82{\scriptsize$\pm$.89} & 57.25{\scriptsize$\pm$.83} & 31.20{\scriptsize$\pm$.55} & 22.06{\scriptsize$\pm$.38} \\
    & B & 46.30{\scriptsize$\pm$.73} & 63.67{\scriptsize$\pm$.88} & 50.53{\scriptsize$\pm$.87} & 31.22{\scriptsize$\pm$.54} & 22.15{\scriptsize$\pm$.37} \\
    \midrule
    \multirow{4}{*}{5} & -   & 75.55{\scriptsize$\pm$.60} & 91.64{\scriptsize$\pm$.49} & 82.32{\scriptsize$\pm$.57} & 48.82{\scriptsize$\pm$.62} & 25.97{\scriptsize$\pm$.40} \\
    & W+B & 71.72{\scriptsize$\pm$.65} & 90.12{\scriptsize$\pm$.52} & 75.23{\scriptsize$\pm$.66} & 48.65{\scriptsize$\pm$.61} & 25.90{\scriptsize$\pm$.42} \\
    & W   & 74.11{\scriptsize$\pm$.64} & 91.48{\scriptsize$\pm$.50} & 81.96{\scriptsize$\pm$.62} & 49.90{\scriptsize$\pm$.63} & 26.30{\scriptsize$\pm$.43} \\
    & B   & 70.74{\scriptsize$\pm$.66} & 89.47{\scriptsize$\pm$.56} & 73.47{\scriptsize$\pm$.69} & 48.21{\scriptsize$\pm$.60} & 26.00{\scriptsize$\pm$.43} \\
    \midrule
    \multirow{4}{*}{20} & -   & 87.25{\scriptsize$\pm$.42} & 97.69{\scriptsize$\pm$.22} & 92.62{\scriptsize$\pm$.34} & 62.38{\scriptsize$\pm$.59} & 32.69{\scriptsize$\pm$.45} \\
    & W+B & 84.26{\scriptsize$\pm$.48} & 96.77{\scriptsize$\pm$.27} & 88.86{\scriptsize$\pm$.43} & 62.27{\scriptsize$\pm$.56} & 32.02{\scriptsize$\pm$.45} \\
    & W   & 85.51{\scriptsize$\pm$.46} & 97.33{\scriptsize$\pm$.24} & 91.19{\scriptsize$\pm$.39} & 62.48{\scriptsize$\pm$.58} & 32.85{\scriptsize$\pm$.46} \\
    & B   & 83.80{\scriptsize$\pm$.50} & 96.83{\scriptsize$\pm$.28} & 87.96{\scriptsize$\pm$.45} & 61.45{\scriptsize$\pm$.56} & 32.02{\scriptsize$\pm$.45} \\
    \midrule
    $k$ & CutMix & miniIN & \!\!\!\!CropDiseases\!\!\!\! & EuroSAT & ISIC & ChestX \\
    \midrule
    \multirow{2}{*}{1} & -  & 49.34{\scriptsize$\pm$.77} & 66.82{\scriptsize$\pm$.89} & 57.25{\scriptsize$\pm$.83} & 31.20{\scriptsize$\pm$.55} & 22.06{\scriptsize$\pm$.38} \\
    & B & 48.98{\scriptsize$\pm$.77} & 67.38{\scriptsize$\pm$.91} & 59.61{\scriptsize$\pm$.86} & 31.34{\scriptsize$\pm$.56} & 21.81{\scriptsize$\pm$.39} \\
    \midrule
    \multirow{4}{*}{5} & -   & 75.55{\scriptsize$\pm$.60} & 91.64{\scriptsize$\pm$.49} & 82.32{\scriptsize$\pm$.57} & 48.82{\scriptsize$\pm$.62} & 25.97{\scriptsize$\pm$.40} \\
    & W+B & 74.04{\scriptsize$\pm$.63} & 91.84{\scriptsize$\pm$.49} & 82.93{\scriptsize$\pm$.56} & 46.39{\scriptsize$\pm$.60} & 25.90{\scriptsize$\pm$.42} \\
    & W   & 74.97{\scriptsize$\pm$.63} & 92.00{\scriptsize$\pm$.48} & 82.51{\scriptsize$\pm$.59} & 49.21{\scriptsize$\pm$.61} & 25.98{\scriptsize$\pm$.42} \\
    & B   & 73.82{\scriptsize$\pm$.63} & 91.73{\scriptsize$\pm$.49} & 83.24{\scriptsize$\pm$.58} & 45.58{\scriptsize$\pm$.60} & 25.92{\scriptsize$\pm$.40} \\
    \midrule
    \multirow{4}{*}{20} & -   & 87.25{\scriptsize$\pm$.42} & 97.69{\scriptsize$\pm$.22} & 92.62{\scriptsize$\pm$.34} & 62.38{\scriptsize$\pm$.59} & 32.69{\scriptsize$\pm$.45} \\
    & W+B & 85.58{\scriptsize$\pm$.46} & 97.52{\scriptsize$\pm$.23} & 92.26{\scriptsize$\pm$.35} & 59.54{\scriptsize$\pm$.56} & 32.20{\scriptsize$\pm$.46} \\
    & W   & 86.21{\scriptsize$\pm$.46} & 97.74{\scriptsize$\pm$.21} & 92.28{\scriptsize$\pm$.35} & 61.98{\scriptsize$\pm$.59} & 32.47{\scriptsize$\pm$.46} \\
    & B   & 85.17{\scriptsize$\pm$.47} & 97.48{\scriptsize$\pm$.25} & 92.20{\scriptsize$\pm$.36} & 58.98{\scriptsize$\pm$.56} & 31.76{\scriptsize$\pm$.46} \\
    \bottomrule
\end{tabular}}
\caption{5-way $k$-shot performance according to decomposition of MixUp and CutMix. ``-'' indicates no augmentation. W and B indicate mixing within the same class (for $k>1$) and mixing between different classes, respectively.}\label{tab:mixup_decomp}
\end{table}

\subsection{Analysis on Intensity of Mixing Augmentation}\label{subsec:mixup}

We further explore the intensity of mixing augmentation, such as MixUp and CutMix.
%
Unlike single augmentation, mixing augmentation uses two different images.
Therefore, we hypothesize that the intensity of mixing augmentation highly depends on whether the two images are sampled from the same class or from different classes.
To validate our hypothesis, we decompose mixing augmentation into mixing within the same class (W in Tables \ref{tab:decomp_magnitude} and \ref{tab:mixup_decomp}) and mixing between different classes (B in Tables \ref{tab:decomp_magnitude} and \ref{tab:mixup_decomp}).

Table \ref{tab:decomp_magnitude} provides the decoupled intensity of mixing augmentation using Eq. (\ref{eq:mixing}). The intensity of mixing augmentation (i.e., MixUp and CutMix) is close to that of mixing augmentation from different classes (i.e., MixUp(B) and CutMix(B)), because mixed images from the same class are rare among $|S|^2$ mixed images.
The intensity of mixing augmentation is larger when the two images are sampled from different classes than from the same class.

This tendency is highly correlated with the few-shot performance according to the decomposition of MixUp and CutMix, as given in Table \ref{tab:mixup_decomp}.
Mixing images from different classes (B in Table \ref{tab:mixup_decomp}) reduces performance more than mixing images from the same class (W in Table \ref{tab:mixup_decomp}), although the latter also reduces the performance.
In the original MixUp paper\,\cite{zhang2017mixup}, MixUp between different classes is the key factor in improving performance.
However, under the few-shot setting, this factor causes deterioration of performance, which contradicts the belief in the standard setting.
It implies that the benefit of interpolating different classes arises based on the large number of training samples.

\begin{table}[t]
    \centering
    \resizebox{\linewidth}{!}{
    \begin{tabular}{ccccccc}
    \toprule
    $k$ & Aug. epochs & miniIN & \!\!\!CropDiseases\!\!\! & EuroSAT & ISIC & ChestX \\ \midrule
    \multirow{5}{*}{1} & \!\!- & 49.34{\scriptsize$\pm$.77} & 66.82{\scriptsize$\pm$.89} & 57.25{\scriptsize$\pm$.83} & 31.20{\scriptsize$\pm$.55} & 22.06{\scriptsize$\pm$.38} \\
    & \,\,\,\,1-100  & 46.30{\scriptsize$\pm$.73} & 63.67{\scriptsize$\pm$.88} & 50.53{\scriptsize$\pm$.87} & 31.22{\scriptsize$\pm$.54} & 22.15{\scriptsize$\pm$.37} \\
    & \,31-100 & 47.06{\scriptsize$\pm$.77} & 64.62{\scriptsize$\pm$.90} & 50.06{\scriptsize$\pm$.84} & 30.71{\scriptsize$\pm$.54} & 22.05{\scriptsize$\pm$.38} \\
    & \,1-70   & 47.20{\scriptsize$\pm$.75} & 64.22{\scriptsize$\pm$.92} & 55.25{\scriptsize$\pm$.86} & 31.33{\scriptsize$\pm$.54} & 21.63{\scriptsize$\pm$.38} \\
    & \,31-70\,\,\,  & 48.43{\scriptsize$\pm$.76} & 64.91{\scriptsize$\pm$.89} & 55.97{\scriptsize$\pm$.87} & 30.80{\scriptsize$\pm$.54} & 21.78{\scriptsize$\pm$.36} \\
    \midrule
    \multirow{5}{*}{5} & \!\!- & 75.55{\scriptsize$\pm$.60} & 91.64{\scriptsize$\pm$.49} & 82.32{\scriptsize$\pm$.57} & 48.82{\scriptsize$\pm$.62} & 25.97{\scriptsize$\pm$.40} \\
    & \,\,\,\,1-100  & 71.72{\scriptsize$\pm$.65} & 90.12{\scriptsize$\pm$.52} & 75.23{\scriptsize$\pm$.66} & 48.65{\scriptsize$\pm$.61} & 25.90{\scriptsize$\pm$.42} \\
    & \,31-100 & 71.70{\scriptsize$\pm$.64} & 89.88{\scriptsize$\pm$.54} & 74.47{\scriptsize$\pm$.68} & 47.80{\scriptsize$\pm$.61} & 26.00{\scriptsize$\pm$.42} \\
    & \,1-70   & 74.17{\scriptsize$\pm$.64} & 91.47{\scriptsize$\pm$.51} & 81.14{\scriptsize$\pm$.59} & 49.31{\scriptsize$\pm$.61} & 25.82{\scriptsize$\pm$.42} \\
    & \,31-70\,\,\,  & 74.98{\scriptsize$\pm$.63} & 91.35{\scriptsize$\pm$.49} & 81.68{\scriptsize$\pm$.59} & 49.06{\scriptsize$\pm$.61} & 25.66{\scriptsize$\pm$.41} \\
    \midrule
    \multirow{5}{*}{20} & \!\!- & 87.25{\scriptsize$\pm$.42} & 97.69{\scriptsize$\pm$.22} & 92.62{\scriptsize$\pm$.34} & 62.38{\scriptsize$\pm$.59} & 32.69{\scriptsize$\pm$.45} \\
    & \,\,\,\,1-100  & 84.26{\scriptsize$\pm$.48} & 96.77{\scriptsize$\pm$.27} & 88.86{\scriptsize$\pm$.43} & 62.27{\scriptsize$\pm$.56} & 32.02{\scriptsize$\pm$.45} \\
    & \,31-100 & 84.11{\scriptsize$\pm$.49} & 96.90{\scriptsize$\pm$.25} & 88.40{\scriptsize$\pm$.44} & 61.27{\scriptsize$\pm$.59} & 32.06{\scriptsize$\pm$.46} \\
    & \,1-70   & 86.40{\scriptsize$\pm$.45} & 97.60{\scriptsize$\pm$.23} & 92.03{\scriptsize$\pm$.36} & 63.16{\scriptsize$\pm$.58} & 32.77{\scriptsize$\pm$.46} \\
    & \,31-70\,\,\,  & 86.81{\scriptsize$\pm$.43} & 97.74{\scriptsize$\pm$.22} & 92.34{\scriptsize$\pm$.36} & 62.65{\scriptsize$\pm$.58} & 32.38{\scriptsize$\pm$.47} \\
    \bottomrule
    \end{tabular}}
    \caption{5-way $k$-shot performance according to the data augmentation scheduler. ``-'' and ``1-100'' indicate no augmentation and augmentation during all epochs, respectively.}\label{tab:scheduling}
\end{table}

\subsection{Scheduling Data Augmentation}\label{subsec:scheduling}

We investigate whether scheduling DA affects few-shot performance because using DA for fine-tuning causes the support-query distribution shift.
It means that a support set $\mathcal{D}_{S}$ is augmented while updating a pre-trained model, whereas a query set $\mathcal{D}_{Q}$ is not augmented during evaluation.

To alleviate this problem, we propose an augmentation scheduler that applies DA to a certain range of epochs.
Table \ref{tab:scheduling} shows the few-shot performance according to the augmentation scheduler with MixUp augmentation.
Note that a pre-trained model is updated for 100 epochs, as explained in implementation details.
An augmentation scheduler without DA at later epochs\,(i.e., 1-70 and 31-70 in Table \ref{tab:scheduling}) outperforms an augmentation scheduler including DA at later epochs\,(i.e., 1-100 and 31-100 in Table \ref{tab:scheduling}). 
This demonstrates the significance of updating a model with non-augmented data before evaluating a query set to remedy the support-query distribution shift.
Namely, scheduling DA can be explained by moving a support set distribution close to a query set distribution at later epochs.

\begin{table*}[t]
\centering
\resizebox{\linewidth}{!}{
\begin{tabular}{cccccccccccccc}
    \toprule
    $k$ & Update & DA & TTA & miniIN & tieredIN & Places & Plantae & Cars & CUB & \!\!\!\!CropDiseases\!\!\!\! & EuroSAT & ISIC & ChestX \\
    \midrule
    \multirow{8.5}{*}{1} & \multirow{4}{*}{LP} & & & 56.23{\scriptsize$\pm$.77} & 55.43{\scriptsize$\pm$.91} & 52.83{\scriptsize$\pm$.81} & 36.49{\scriptsize$\pm$.69} & 29.79{\scriptsize$\pm$.51} & 40.91{\scriptsize$\pm$.76} & \textbf{73.16}{\scriptsize$\pm$.87} & 64.96{\scriptsize$\pm$.86} & 29.80{\scriptsize$\pm$.55} & 22.32{\scriptsize$\pm$.42} \\ 
    & & \checkmark & & 56.92{\scriptsize$\pm$.79} & 55.50{\scriptsize$\pm$.90} & 52.29{\scriptsize$\pm$.84} & 37.43{\scriptsize$\pm$.68} & 30.29{\scriptsize$\pm$.56} & 42.00{\scriptsize$\pm$.78} & 66.46{\scriptsize$\pm$.93} & 66.13{\scriptsize$\pm$.92} & 29.79{\scriptsize$\pm$.55} & 21.97{\scriptsize$\pm$.37} \\ 
    & & & \checkmark & 57.87{\scriptsize$\pm$.79} & 53.42{\scriptsize$\pm$.93} & 53.04{\scriptsize$\pm$.83} & 36.58{\scriptsize$\pm$.70} & 29.84{\scriptsize$\pm$.52} & 40.75{\scriptsize$\pm$.77} & 69.89{\scriptsize$\pm$.91} & 60.66{\scriptsize$\pm$.90} & 29.76{\scriptsize$\pm$.55} & 21.75{\scriptsize$\pm$.35} \\ 
    & & \checkmark & \checkmark & \textbf{60.73}{\scriptsize$\pm$.80} & \textbf{55.99}{\scriptsize$\pm$.93} & \textbf{53.59}{\scriptsize$\pm$.83} & \textbf{39.56}{\scriptsize$\pm$.72} & \textbf{33.49}{\scriptsize$\pm$.64} & \textbf{44.12}{\scriptsize$\pm$.79} & 73.12{\scriptsize$\pm$.86} & \textbf{67.98}{\scriptsize$\pm$.90} & 31.39{\scriptsize$\pm$.60} & \textbf{22.61}{\scriptsize$\pm$.40} \\ \cmidrule{2-14}
    & \multirow{4}{*}{FT} & & & 49.34{\scriptsize$\pm$.77} & 48.96{\scriptsize$\pm$.86} & 46.24{\scriptsize$\pm$.78} & 33.08{\scriptsize$\pm$.59} & 29.55{\scriptsize$\pm$.52} & 37.57{\scriptsize$\pm$.70} & 66.82{\scriptsize$\pm$.89} & 57.25{\scriptsize$\pm$.83} & 31.20{\scriptsize$\pm$.55} & 22.06{\scriptsize$\pm$.38} \\
    & & \checkmark & & 51.59{\scriptsize$\pm$.79} & 49.01{\scriptsize$\pm$.86} & 47.31{\scriptsize$\pm$.81} & 34.74{\scriptsize$\pm$.61} & 30.12{\scriptsize$\pm$.56} & 39.57{\scriptsize$\pm$.73} & 65.49{\scriptsize$\pm$.91} & 58.30{\scriptsize$\pm$.86} & 32.97{\scriptsize$\pm$.56} & 22.07{\scriptsize$\pm$.37} \\
    & & & \checkmark & 49.76{\scriptsize$\pm$.79} & 46.72{\scriptsize$\pm$.90} & 45.54{\scriptsize$\pm$.78} & 33.44{\scriptsize$\pm$.64} & 28.77{\scriptsize$\pm$.54} & 37.12{\scriptsize$\pm$.71} & 61.73{\scriptsize$\pm$.94} & 48.92{\scriptsize$\pm$.88} & 29.74{\scriptsize$\pm$.57} & 21.31{\scriptsize$\pm$.33} \\
    & & \checkmark & \checkmark & 54.50{\scriptsize$\pm$.83} & 49.57{\scriptsize$\pm$.92} & 47.47{\scriptsize$\pm$.84} & 36.95{\scriptsize$\pm$.68} & 32.50{\scriptsize$\pm$.60} & 41.15{\scriptsize$\pm$.75} & 68.33{\scriptsize$\pm$.89} & 58.10{\scriptsize$\pm$.86} & \textbf{33.97}{\scriptsize$\pm$.60} & 22.34{\scriptsize$\pm$.36} \\
    \midrule
    \multirow{8.5}{*}{5} & \multirow{4}{*}{LP} & & & 77.77{\scriptsize$\pm$.61} & 75.02{\scriptsize$\pm$.71} & 73.26{\scriptsize$\pm$.66} & 54.21{\scriptsize$\pm$.74} & 44.92{\scriptsize$\pm$.67} & 58.38{\scriptsize$\pm$.76} & 91.27{\scriptsize$\pm$.48} & 84.48{\scriptsize$\pm$.57} & 40.88{\scriptsize$\pm$.54} & 26.07{\scriptsize$\pm$.44} \\ 
    & & \checkmark & & 77.51{\scriptsize$\pm$.60} & 74.65{\scriptsize$\pm$.72} & 72.43{\scriptsize$\pm$.67} & 54.93{\scriptsize$\pm$.72} & 45.34{\scriptsize$\pm$.70} & 59.84{\scriptsize$\pm$.75} & 87.29{\scriptsize$\pm$.58} & 83.34{\scriptsize$\pm$.61}  & 41.26{\scriptsize$\pm$.54} & 25.58{\scriptsize$\pm$.43} \\ 
    & & & \checkmark & 79.42{\scriptsize$\pm$.60} & 73.73{\scriptsize$\pm$.77} & 72.93{\scriptsize$\pm$.68} & 54.83{\scriptsize$\pm$.74} & 44.63{\scriptsize$\pm$.70} & 59.24{\scriptsize$\pm$.76} & 86.66{\scriptsize$\pm$.61} & 80.24{\scriptsize$\pm$.66} & 41.33{\scriptsize$\pm$.58} & 24.80{\scriptsize$\pm$.37} \\ 
    & & \checkmark & \checkmark & \textbf{82.34}{\scriptsize$\pm$.54} & \textbf{76.31}{\scriptsize$\pm$.72} & \textbf{73.96}{\scriptsize$\pm$.67} & 59.78{\scriptsize$\pm$.74} & 53.60{\scriptsize$\pm$.77} & 65.28{\scriptsize$\pm$.74} & 92.03{\scriptsize$\pm$.44}& \textbf{84.91}{\scriptsize$\pm$.58} & 45.30{\scriptsize$\pm$.58} & 27.56{\scriptsize$\pm$.45} \\ \cmidrule{2-14}
    & \multirow{4}{*}{FT} & & & 75.55{\scriptsize$\pm$.60} & 73.59{\scriptsize$\pm$.74} & 70.60{\scriptsize$\pm$.68} & 53.45{\scriptsize$\pm$.74} & 49.85{\scriptsize$\pm$.73} & 59.93{\scriptsize$\pm$.79} & 91.64{\scriptsize$\pm$.49} & 82.32{\scriptsize$\pm$.57} & 48.82{\scriptsize$\pm$.62} & 25.97{\scriptsize$\pm$.40} \\
    & & \checkmark & & 74.74{\scriptsize$\pm$.62} & 72.25{\scriptsize$\pm$.74} & 69.79{\scriptsize$\pm$.68} & 56.10{\scriptsize$\pm$.71} & 51.15{\scriptsize$\pm$.73} & 62.43{\scriptsize$\pm$.74} & 91.54{\scriptsize$\pm$.46} & 79.71{\scriptsize$\pm$.61} & 48.42{\scriptsize$\pm$.59} & 27.93{\scriptsize$\pm$.43} \\
    & & & \checkmark & 76.53{\scriptsize$\pm$.62} & 71.82{\scriptsize$\pm$.75} & 68.93{\scriptsize$\pm$.73} & 54.39{\scriptsize$\pm$.75} & 47.72{\scriptsize$\pm$.75} & 60.74{\scriptsize$\pm$.79} & 88.90{\scriptsize$\pm$.56} & 72.01{\scriptsize$\pm$.74} & 46.18{\scriptsize$\pm$.62} & 24.20{\scriptsize$\pm$.37} \\
    & & \checkmark & \checkmark& 79.17{\scriptsize$\pm$.59} & 73.78{\scriptsize$\pm$.73} & 70.00{\scriptsize$\pm$.69} & \textbf{60.89}{\scriptsize$\pm$.73} & \textbf{58.29}{\scriptsize$\pm$.77} & \textbf{68.43}{\scriptsize$\pm$.74} & \textbf{93.07}{\scriptsize$\pm$.42} & 81.38{\scriptsize$\pm$.57} & \textbf{52.10}{\scriptsize$\pm$.62} & \textbf{28.21}{\scriptsize$\pm$.42} \\
    \midrule
    \multirow{8.5}{*}{20} & \multirow{4}{*}{LP} & & & 86.54{\scriptsize$\pm$.44} & 83.26{\scriptsize$\pm$.61} & 81.53{\scriptsize$\pm$.53} & 66.52{\scriptsize$\pm$.69} & 60.95{\scriptsize$\pm$.66} & 71.29{\scriptsize$\pm$.70} & 96.20{\scriptsize$\pm$.29} & 91.00{\scriptsize$\pm$.40} & 51.32{\scriptsize$\pm$.54} & 30.98{\scriptsize$\pm$.45} \\ 
    & & \checkmark & & 85.63{\scriptsize$\pm$.46} & 82.56{\scriptsize$\pm$.60} & 81.05{\scriptsize$\pm$.53} & 66.56{\scriptsize$\pm$.66} & 58.07{\scriptsize$\pm$.63} & 71.92{\scriptsize$\pm$.64} & 93.22{\scriptsize$\pm$.40} & 89.38{\scriptsize$\pm$.46} & 49.71{\scriptsize$\pm$.54} & 29.82{\scriptsize$\pm$.43} \\ 
    & & & \checkmark & 88.09{\scriptsize$\pm$.43} & 82.51{\scriptsize$\pm$.63} & 81.40{\scriptsize$\pm$.54} & 67.84{\scriptsize$\pm$.69} & 60.98{\scriptsize$\pm$.67} & 71.39{\scriptsize$\pm$.69} & 91.70{\scriptsize$\pm$.47} & 91.09{\scriptsize$\pm$.39} & 51.48{\scriptsize$\pm$.55} & 30.83{\scriptsize$\pm$.46} \\ 
    & & \checkmark & \checkmark & \textbf{89.93}{\scriptsize$\pm$.39} & 84.47{\scriptsize$\pm$.57} & \textbf{82.52}{\scriptsize$\pm$.51} & 72.38{\scriptsize$\pm$.65} & 69.56{\scriptsize$\pm$.66} & 78.45{\scriptsize$\pm$.62} & 96.41{\scriptsize$\pm$.29} & 90.49{\scriptsize$\pm$.44} & 55.25{\scriptsize$\pm$.53} & 33.08{\scriptsize$\pm$.46} \\ \cmidrule{2-14}
    & \multirow{4}{*}{FT} & & & 86.94{\scriptsize$\pm$.44} & 84.48{\scriptsize$\pm$.59} & 81.30{\scriptsize$\pm$.53} & 69.08{\scriptsize$\pm$.68} & 70.62{\scriptsize$\pm$.64} & 75.91{\scriptsize$\pm$.67} & 97.72{\scriptsize$\pm$.22} & \textbf{92.54}{\scriptsize$\pm$.36} & 62.36{\scriptsize$\pm$.59} & 32.65{\scriptsize$\pm$.47} \\
    & & \checkmark & & 85.83{\scriptsize$\pm$.47} & 83.87{\scriptsize$\pm$.57} & 81.15{\scriptsize$\pm$.53} & 71.51{\scriptsize$\pm$.65} & 71.46{\scriptsize$\pm$.59} & 77.78{\scriptsize$\pm$.60} & 97.53{\scriptsize$\pm$.22} & 91.71{\scriptsize$\pm$.35} & 60.50{\scriptsize$\pm$.57} & 34.05{\scriptsize$\pm$.44} \\
    & & & \checkmark & 87.22{\scriptsize$\pm$.44} & 83.06{\scriptsize$\pm$.61} & 80.06{\scriptsize$\pm$.56} & 69.77{\scriptsize$\pm$.70} & 66.67{\scriptsize$\pm$.66} & 77.00{\scriptsize$\pm$.68} & 95.79{\scriptsize$\pm$.32} & 83.97{\scriptsize$\pm$.55} & 58.63{\scriptsize$\pm$.66} & 28.31{\scriptsize$\pm$.43} \\
    & & \checkmark & \checkmark & 89.62{\scriptsize$\pm$.38} & \textbf{85.17}{\scriptsize$\pm$.54} & 81.59{\scriptsize$\pm$.52} & \textbf{76.82}{\scriptsize$\pm$.61} & \textbf{80.38}{\scriptsize$\pm$.52} & \textbf{84.19}{\scriptsize$\pm$.53} & \textbf{98.25}{\scriptsize$\pm$.18} & 92.31{\scriptsize$\pm$.34} & \textbf{65.00}{\scriptsize$\pm$.59} & \textbf{36.08}{\scriptsize$\pm$.46} \\
    \bottomrule
    \end{tabular}}
\caption{5-way $k$-shot performance comparison according to update, DA, and TTA. Base Aug is used for both DA and TTA. For TTA, 32 predictions are ensembled. The best performance for each $k$ is boldfaced.} \label{tab:comparison_TTA}
\end{table*}

\section{Test-time Augmentation for (CD-)FSL}

We adopt TTA with DA for FSL, which can be explained by moving a query set distribution close to an augmented support set distribution to overcome the support-query distribution shift, as opposed to scheduling DA.
Namely, the original test image in $\mathcal{D}_{Q}$ is augmented by the policy used for $\mathcal{D}_{S}$, and predictions from the augmented test images are ensembled.
Note that mixing augmentation techniques cannot be appropriate for TTA, because mixed images are not used for evaluation.
%

Table \ref{tab:comparison_TTA} describes few-shot performance according to the three perspectives: update, DA, and TTA. 
Our strategy, updating with DA and predicting with TTA, outperforms the other strategies in most cases.
%
TTA with DA not only overcomes the performance degradation of DA on few samples but also brings out improvement.
In contrast, TTA without DA decreases performance on most datasets, because augmentation is not applied to a support set and a query set simultaneously, which is similar to using only DA analyzed in Section \ref{sec:analysis2}.
Therefore, augmentation should be used for both a support set and a query set to improve (CD-)FSL performance.
%
%

With regard to updating methods, LP+DA+TTA is better than FT+DA+TTA when there are extremely few samples (e.g., $k=1$), while FT+DA+TTA is better than LP+DA+TTA as shot increases (e.g., $k=20$) on most datasets when ResNet10 is fine-tuned.
This trend is almost the same as the comparison of LP and FT, as explained in Section \ref{sec:analysis1}.
%
%
%
%
%
The results using ResNet18 pre-trained on tieredIN are reported in Appendix \ref{appx:datta_tiered}.

\begin{figure}[t]
    \centering
    \begin{subfigure}[h]{0.325\linewidth}
    \includegraphics[width=\linewidth]{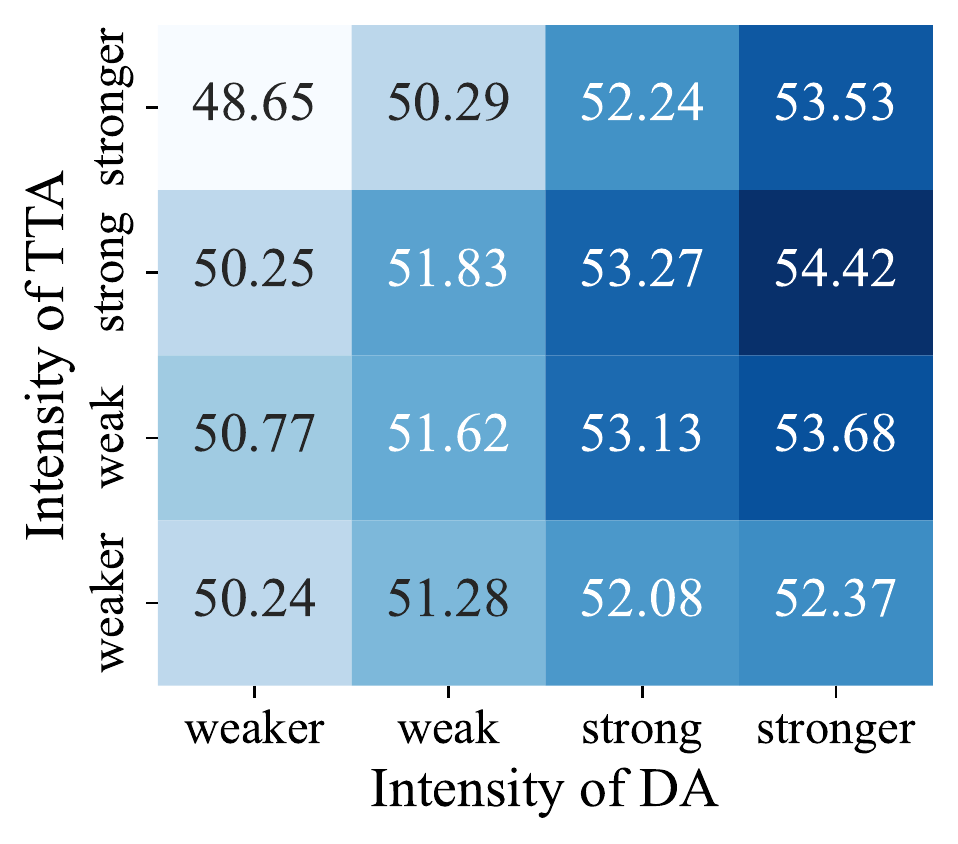}
        \caption{miniIN ($k=1$)}
    \end{subfigure}
    \begin{subfigure}[h]{0.325\linewidth}
    \includegraphics[width=\linewidth]{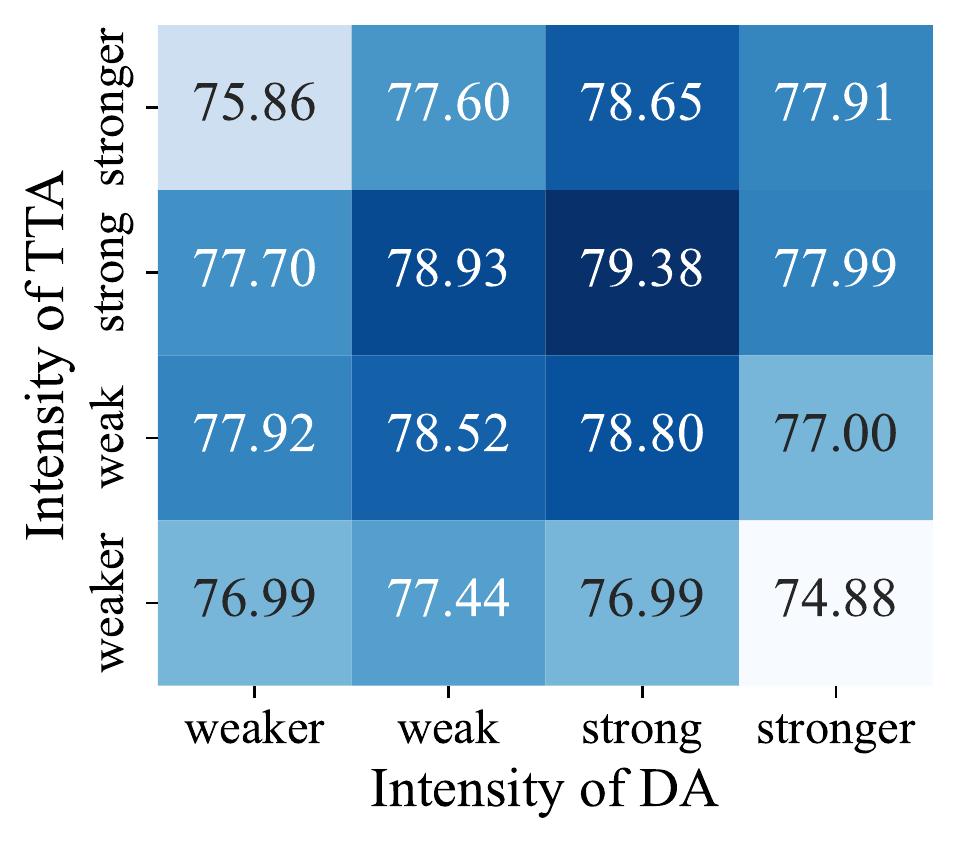}
        \caption{miniIN ($k=5$)}
    \end{subfigure}
    \begin{subfigure}[h]{0.325\linewidth}
    \includegraphics[width=\linewidth]{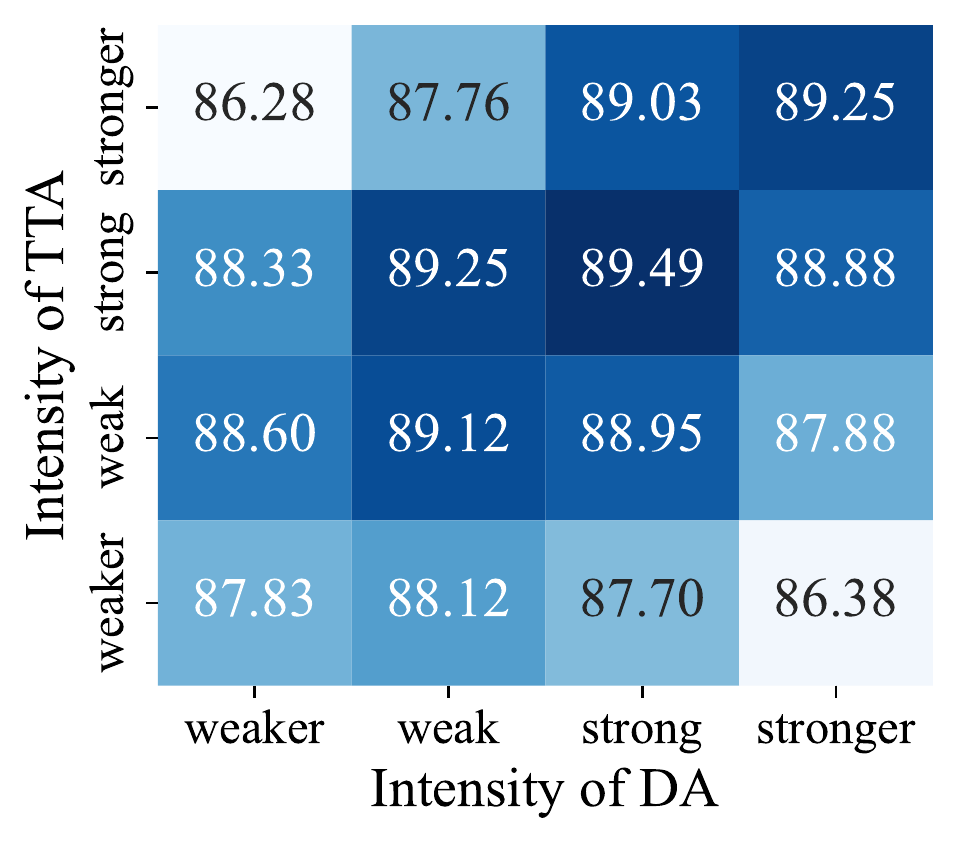}
        \caption{miniIN ($k=20$)}
    \end{subfigure}
    
    \begin{subfigure}[h]{0.325\linewidth}
    \includegraphics[width=\linewidth]{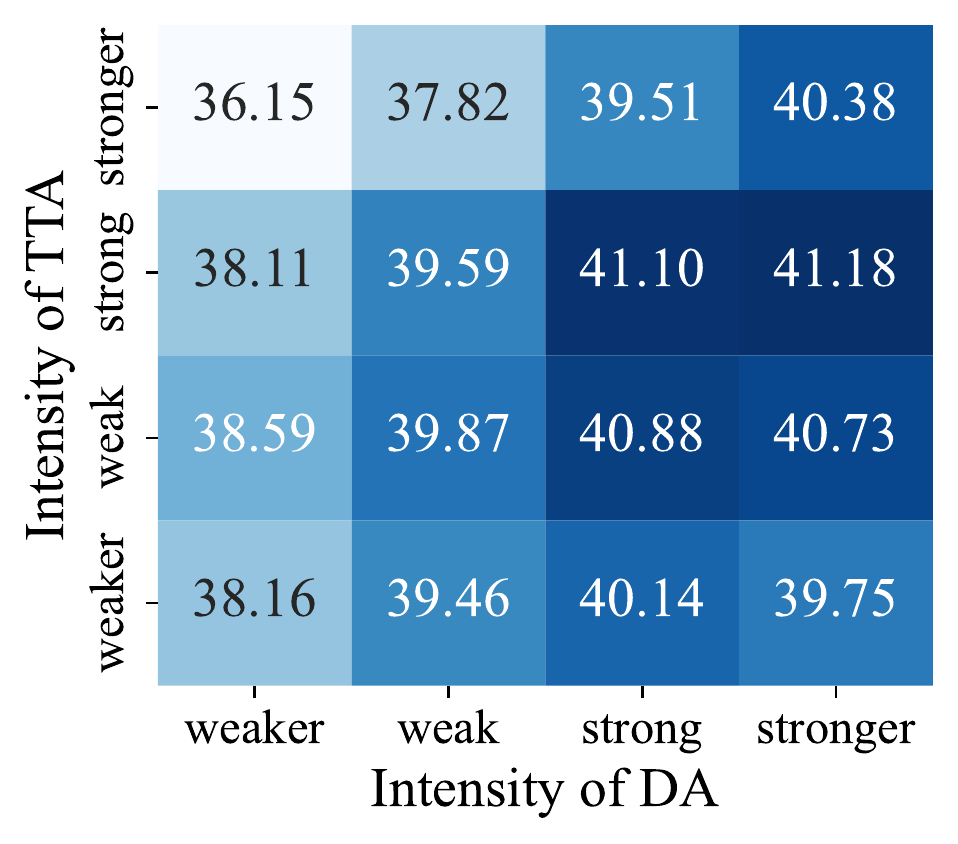}
        \caption{CUB ($k=1$)}
    \end{subfigure}
    \begin{subfigure}[h]{0.325\linewidth}
    \includegraphics[width=\linewidth]{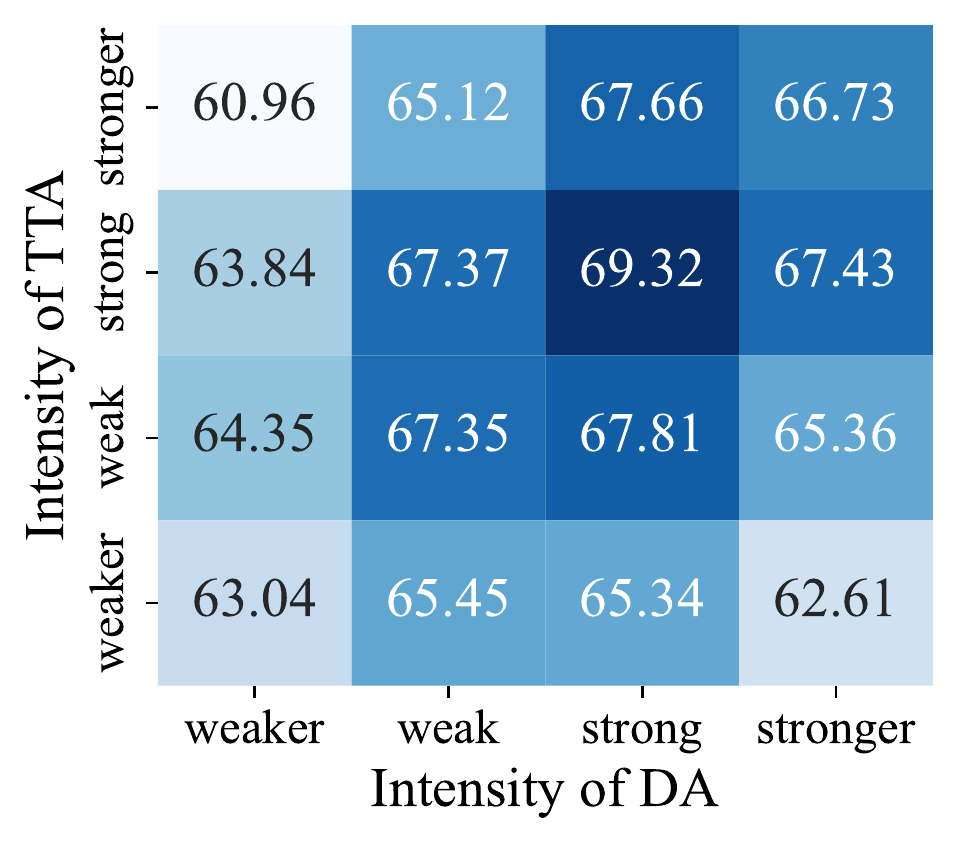}
        \caption{CUB ($k=5$)}
    \end{subfigure}
    \begin{subfigure}[h]{0.325\linewidth}
    \includegraphics[width=\linewidth]{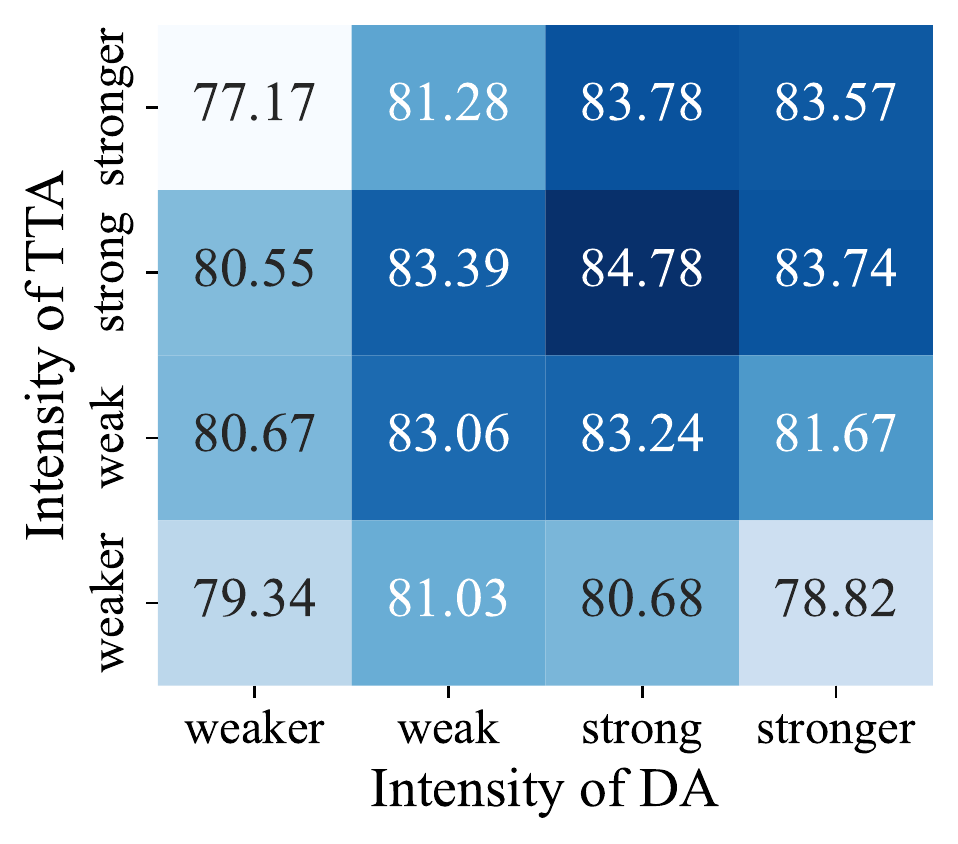}
        \caption{CUB ($k=20$)}
    \end{subfigure}
    \caption{5-way $k$-shot performance according to the designed intensity of DA and TTA. Base Aug is used for both DA and TTA. For TTA, 32 predictions are ensembled.}
    \label{fig:matching_main}
\end{figure}

\subsection{Ablation Study on Combination of Intensity}\label{sec:comb}
We investigate the combination of intensity for DA and TTA, using four designed intensities provided in Table \ref{tab:base_magnitude}.
%
%
Figure \ref{fig:matching_main} describes the few-shot performance according to the combination of intensities, when both DA and TTA are used.
%
X-axis and y-axis denote designed intensity for DA and TTA, respectively.
The upright ($\nearrow$) diagonal elements are the matched combination where the intensity of DA is the same as that of TTA.
%

The best performance is achieved when \textsc{strong(er)} augmentation is used for DA and TTA, although \textsc{strong(er)} is not the best intensity when it is used only for DA, shown in Figure \ref{fig:base_magnitude}.
Furthermore, the performance tends to degrade when the intensity gap between DA and TTA increases.
For example, \textsc{strong}-\textsc{strong} outperforms \textsc{strong}-\textsc{weaker} and \textsc{stronger}-\textsc{stronger} outperforms \textsc{stronger}-\textsc{weaker}.
Therefore, matching the same \emph{powerful} intensity between DA and TTA is essential as a rule of thumb, because optimizing the combination is infeasible.
%
The results of LP are provided in Appendix \ref{appx:abla_comb}.

\begin{figure}[t]
    \centering
    \includegraphics[width=\linewidth]{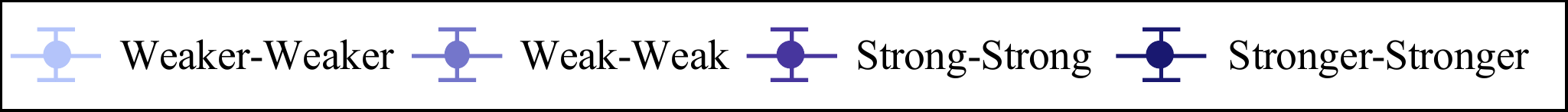}
    
    \begin{subfigure}[h]{0.34\linewidth}
    \includegraphics[width=\linewidth]{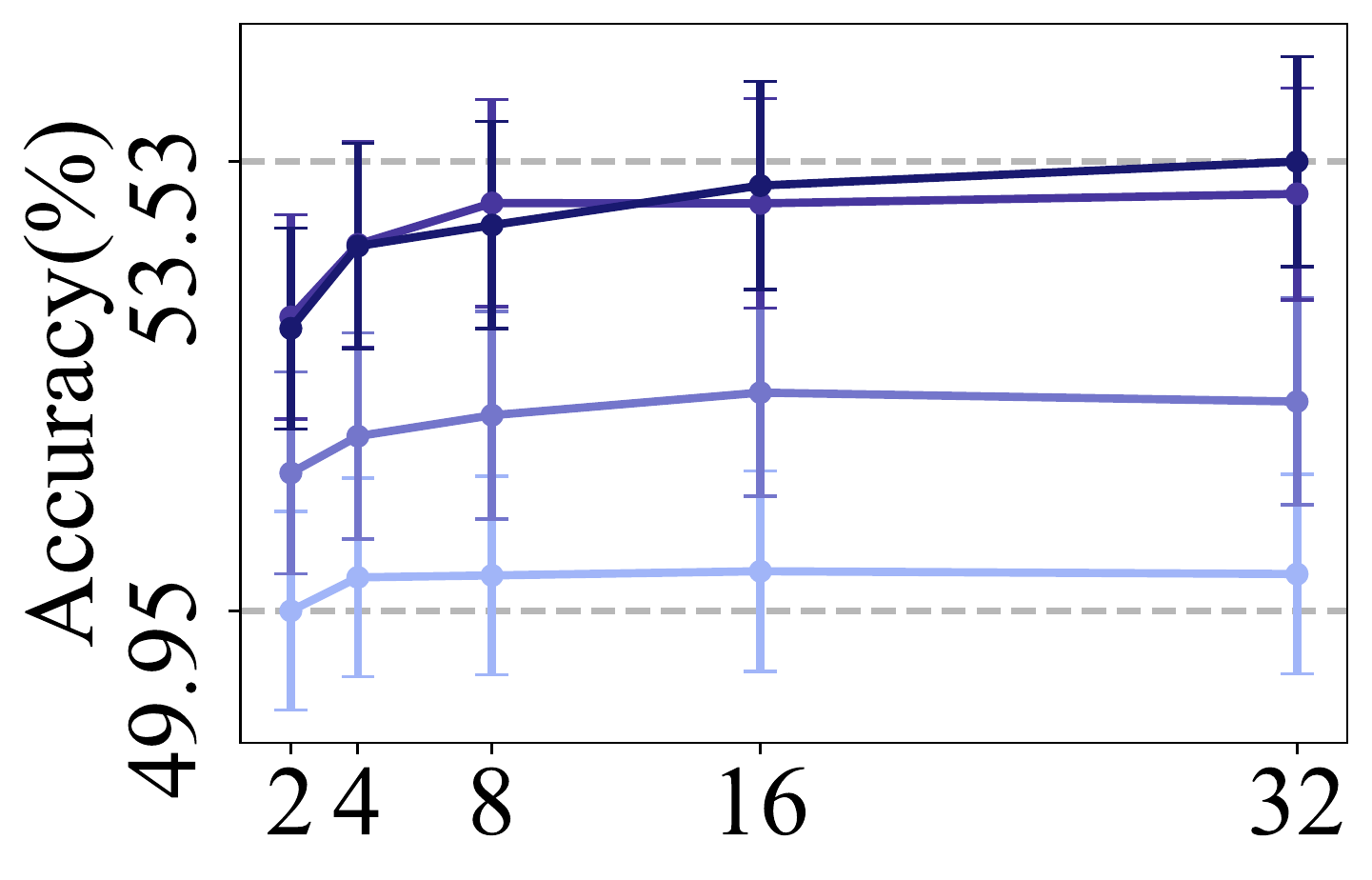}
        \caption{miniIN ($k=1$)}
    \end{subfigure}
    \begin{subfigure}[h]{0.315\linewidth}
    \includegraphics[width=\linewidth]{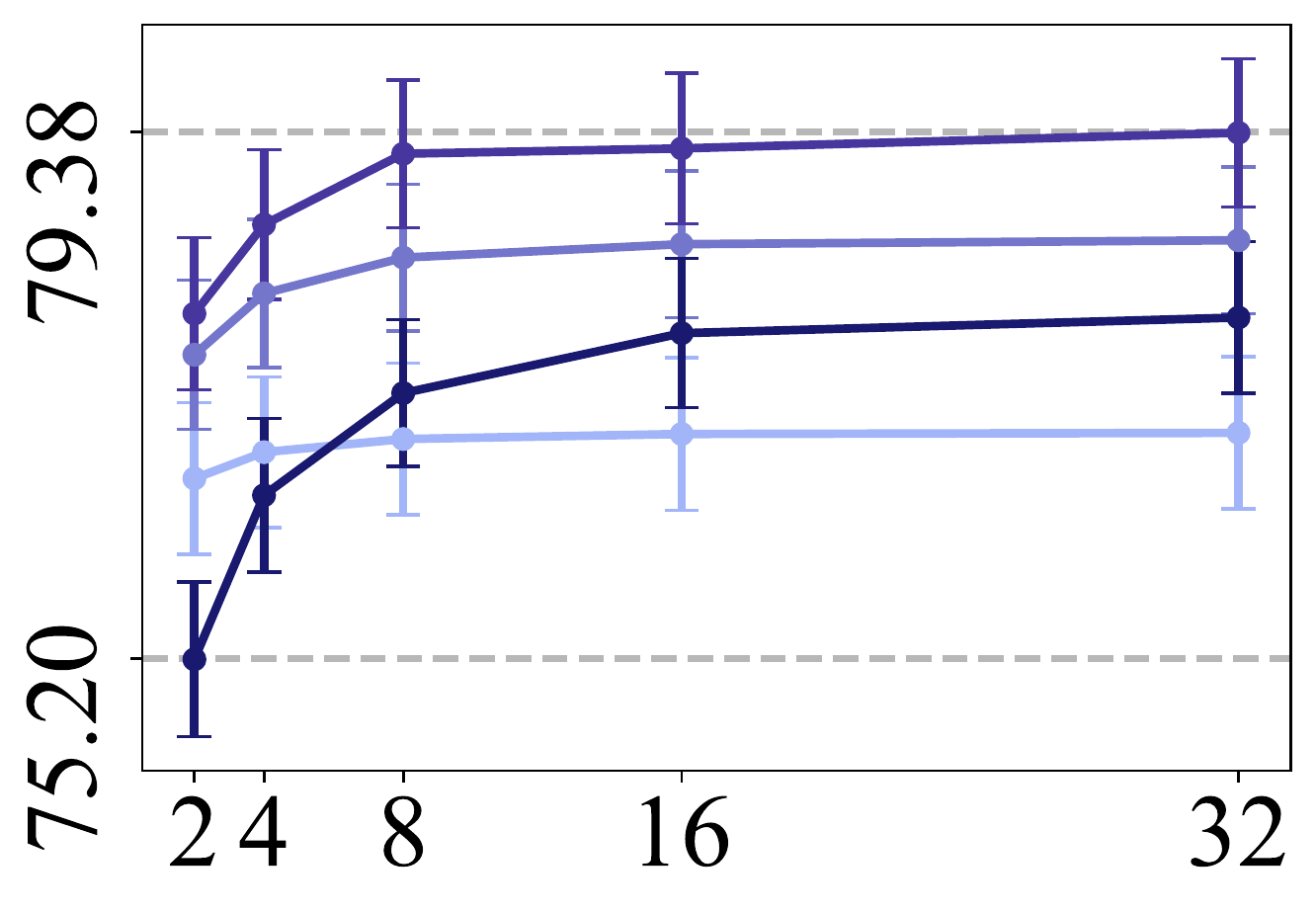}
        \caption{miniIN ($k=5$)}
    \end{subfigure}
    \begin{subfigure}[h]{0.315\linewidth}
    \includegraphics[width=\linewidth]{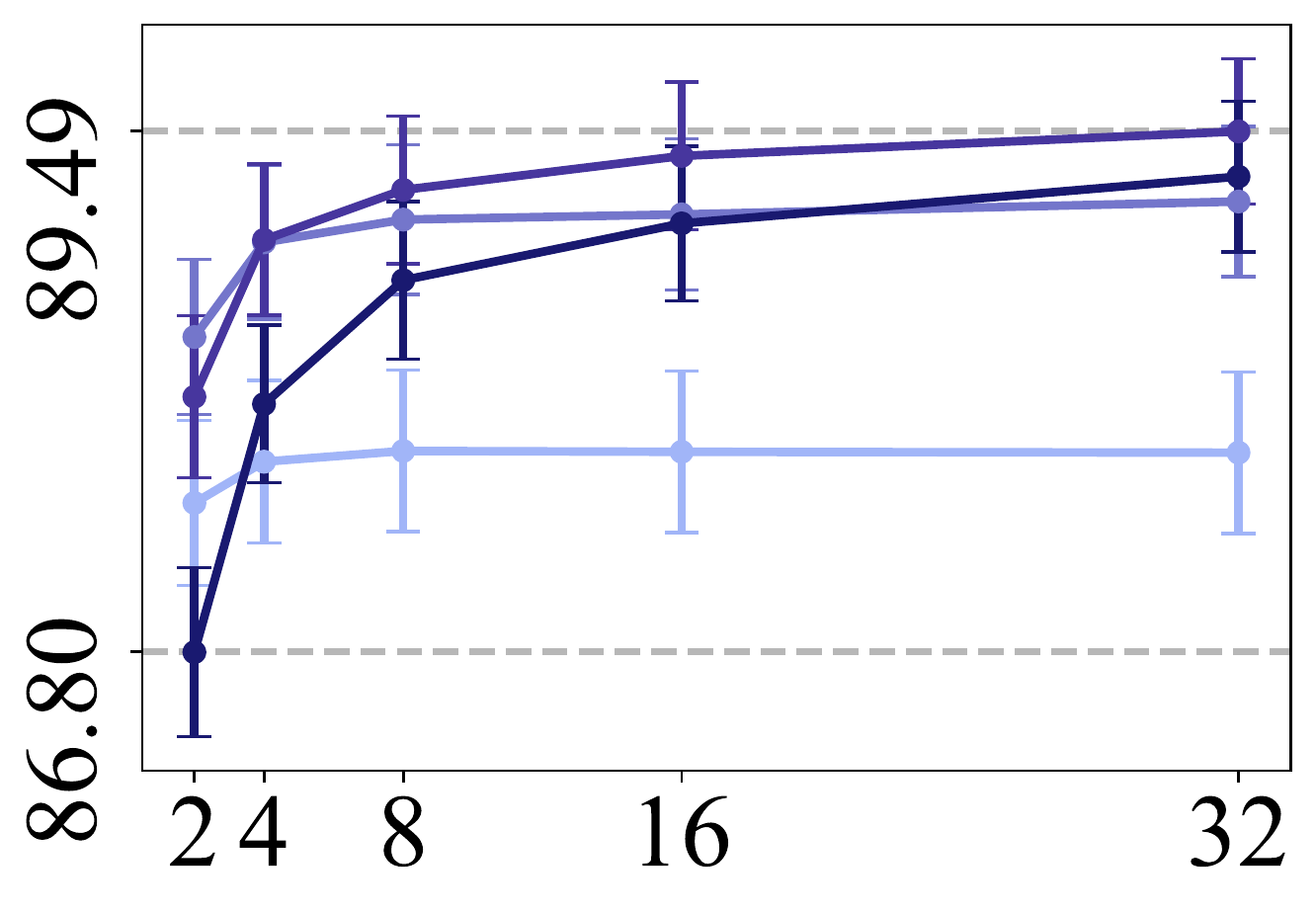}
        \caption{miniIN ($k=20$)}
    \end{subfigure}
    
    \begin{subfigure}[h]{0.34\linewidth}
    \includegraphics[width=\linewidth]{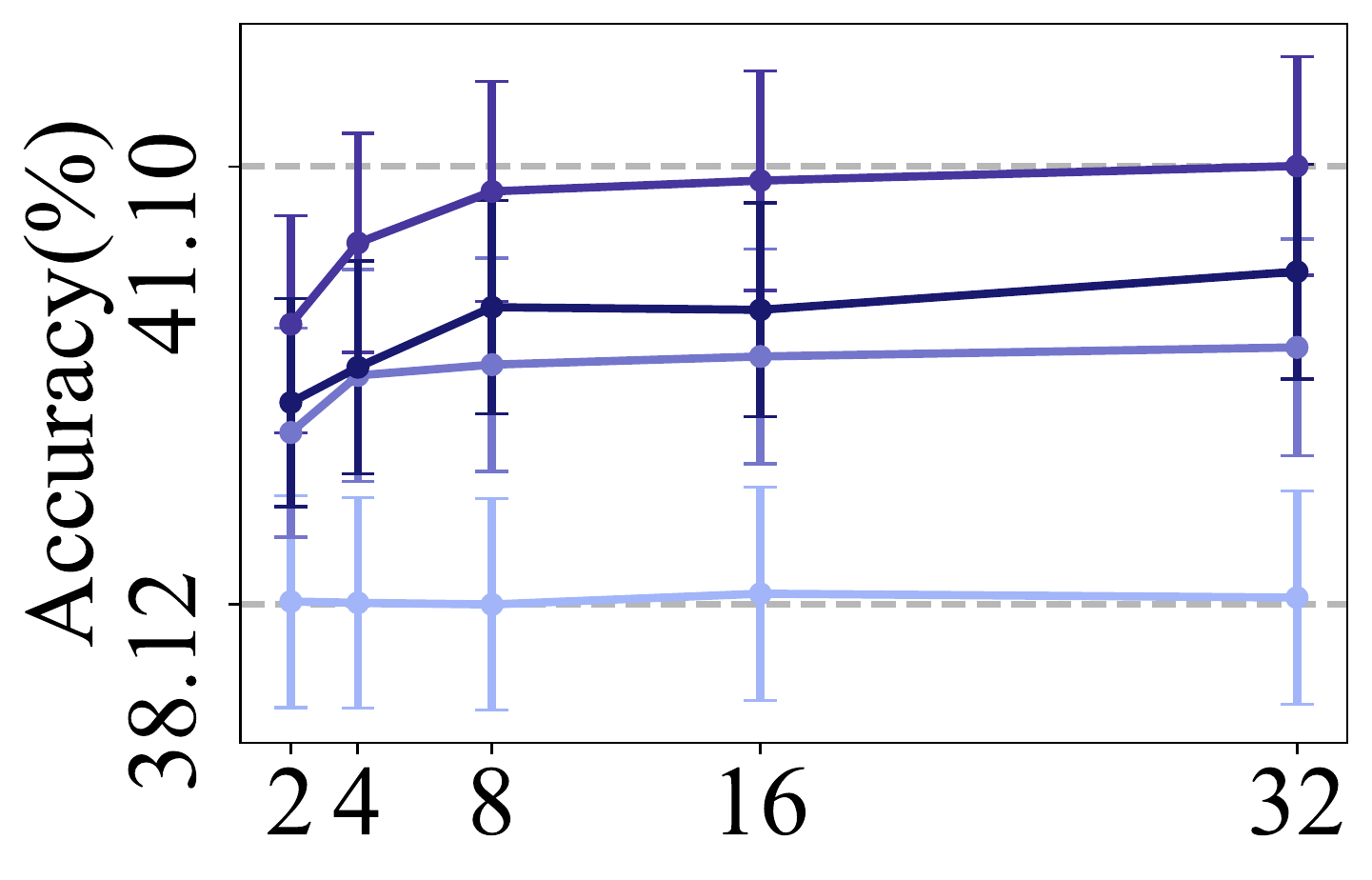}
        \caption{CUB ($k=1$)}
    \end{subfigure}
    \begin{subfigure}[h]{0.315\linewidth}
    \includegraphics[width=\linewidth]{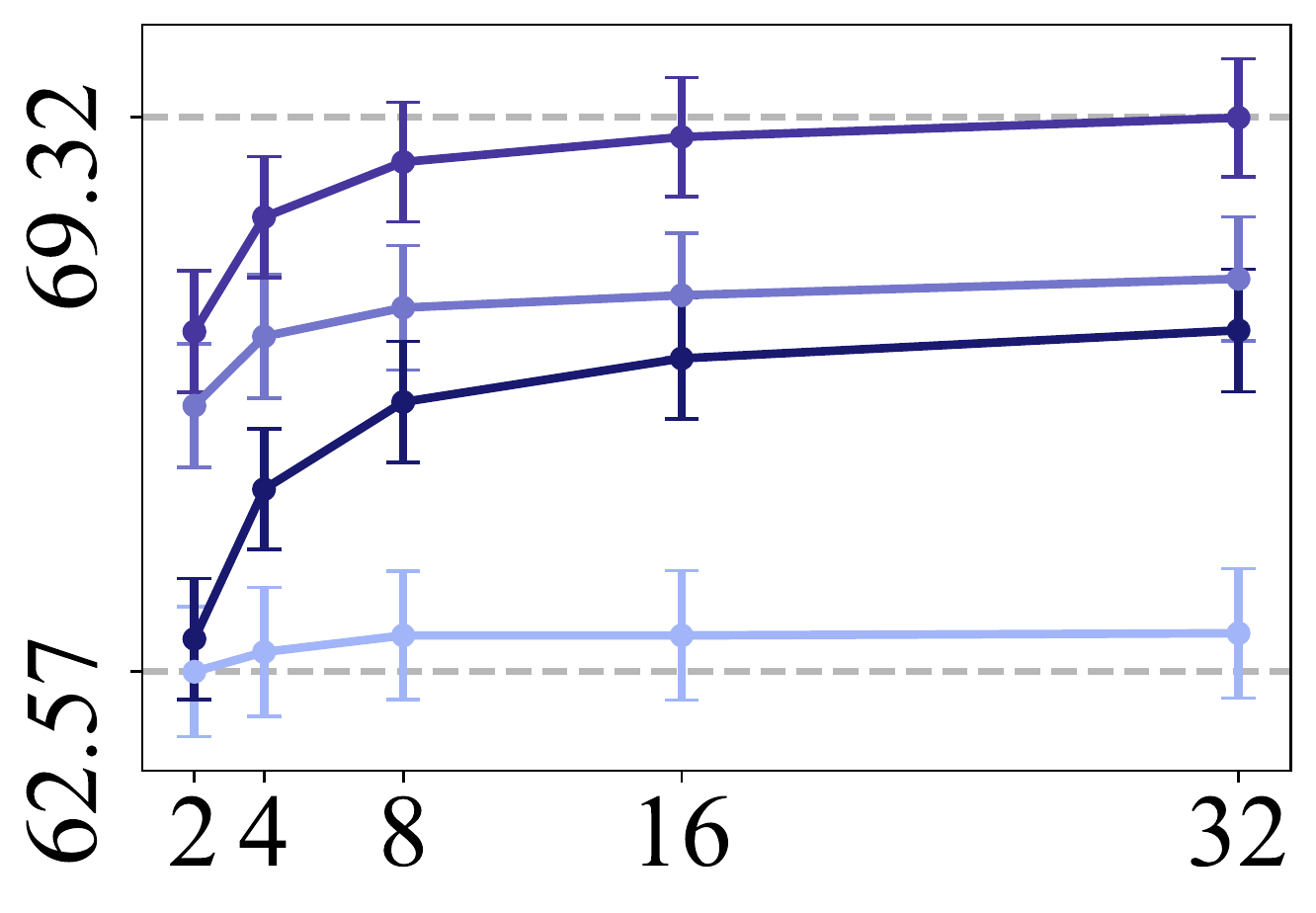}
        \caption{CUB ($k=5$)}
    \end{subfigure}
    \begin{subfigure}[h]{0.315\linewidth}
    \includegraphics[width=\linewidth]{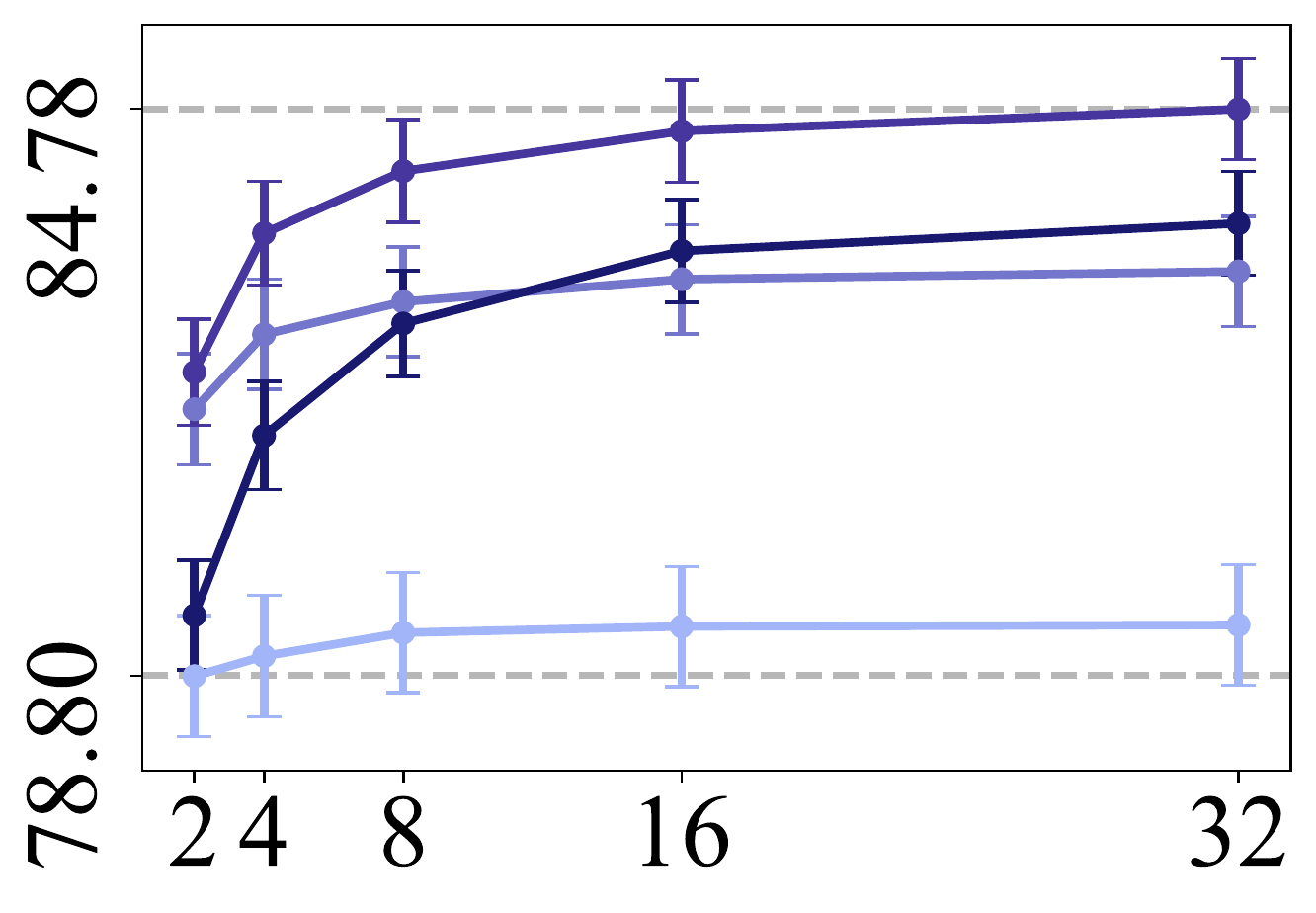}
        \caption{CUB ($k=20$)}
    \end{subfigure}
    \caption{5-way $k$-shot performance according to the number of augmented samples for TTA.}
    \label{fig:tta_abla}
\end{figure}

\subsection{Ablation Study on Number of TTA Samples}
We investigate the effectiveness of increasing the number of samples for TTA according to the intensity of augmentation.
Figure \ref{fig:tta_abla} describes the performance according to the number of augmented samples for TTA using the four matched intensities.
The value $v$ on the x-axis denotes the number of samples used in the prediction, which includes one original test image and ($v-1$) augmented test images.

When \textsc{strong(er)} augmentation is used for DA and TTA, the performance gain is significant as augmented images for TTA increase, enhancing the effectiveness of the ensemble, while performance gain is marginal for \textsc{weak(er)} augmentation.
With the result of Section \ref{sec:comb}, we conclude that a sufficient number of TTA samples is required with matched powerful intensity for better few-shot performance.
The results of LP are provided in Appendix \ref{appx:abla_num}.
 
\section{Related Works}\label{sec:related}

\subsection{Full Fine-tuning (FT) vs. Linear Probing (LP)}
Most of the works on updating pre-trained models have been mainly studied for language tasks\,\cite{dodge2020fine, zhao2021closer, zhou2021closer}.
In general computer vision setting, transfer learning and updating methods gained much attention\,\cite{zhai2019large, kornblith2019better, ericsson2021well, kumar2022finetuning}.
\citet{kornblith2019better} showed that FT outperformed LP, by investigation on many-shot target datasets.
Meanwhile, \citet{kumar2022finetuning} demonstrated that FT could distort pre-trained representation on out-of-distribution tasks, where datasets for updating and evaluation are different. They further showed that LP $\rightarrow$ FT updates could remedy this distortion.

\subsection{Data Augmentation}
DA is a general technique to prevent overfitting and increase generalization performance\,\cite{simard2003best, krizhevsky2012imagenet}.
Single augmentation techniques such as flipping, cropping, and color jittering are the most generic and prevalent practices in computer vision tasks.
 Mixing augmentation techniques include MixUp \cite{zhang2017mixup}, CutMix \cite{yun2019cutmix}, and PuzzleMix \cite{kim2020puzzle}, which mix two images while mixing corresponding class labels.
%
%
In addition, there are deep learning approaches for data augmentation\,\cite{mirza2014conditional,gatys2016image,zhu2017unpaired}.
%

For DA in FSL\footnote{Related works about (CD-)FSL are explained in Appendix \ref{appx:related}.}, few studies have been conducted to remedy the lack of target samples by data generation\,\cite{yang2021free, kumar2022gdc} or augmentation\,\cite{mangla2020charting, ni2021data}.
Generation-based algorithms sample target representation from the calibrated target distribution based on the source distribution.
Among augmentation-based algorithms, S2M2\,\cite{mangla2020charting} utilizes Manifold MixUp loss\,\cite{verma2019manifold} that works as a feature-level augmentation in pre-training.
\citet{ni2021data} analyze the effectiveness of data augmentation in the meta-learning pipeline and propose Meta-MaxUp for pre-training. 
However, for few samples augmentation, only horizontal flipping and random clipping are considered.

\subsection{Test-time Augmentation}
Data augmentation can be used in the evaluation, which is called TTA. Namely, a test sample is augmented and the predictions from each augmented sample are ensembled for improved and robust performance, and uncertainty estimation\,\cite{matsunaga2017image, prakash2018deflecting, smith2018understanding, kim2020learning}.
For TTA in FSL, \citet{yamada2022one} showed that TTA with manually deterministic augmentation policies improves one-shot performances.
Unlike prior works, we investigate the efficacy of various data augmentation for few samples (i.e., support set) and adopt stochastic TTA for evaluation (i.e., query set).

\section{Conclusion}\label{sec:conclusion}

We focused on maximally transferring pre-trained models to few-shot downstream tasks from three perspectives: update, data augmentation, and test-time augmentation.
We showed that the number of training samples is a critical factor in deciding between LP and FT.
In this context, we investigated DA for few samples using the intensity of augmentation and concluded that the few-shot performance depends on the intensity of augmentation.
Finally, we adopted TTA with DA and showed the improved few-shot performance. 
We hope that our observations will inspire how to exploit pre-trained models with few samples for (CD-)FSL.
%
%

\clearpage
\bibliography{aaai23.bib}

\clearpage
\appendix
\section{Dataset Details}\label{appx:setup}
We employ two source datasets and ten target datasets for cross-domain scenario, following the same dataset setting with \citet{oh2022understanding}\footnote{We followed the pre-processing details from \url{https://github.com/sungnyun/CD-FSL}}.

We use miniImageNet\,\citep{vinyals2016matching} and tieredImageNet\,\citep{ren2018meta} as source datasets which are disjoint portions of ImageNet-1k\,\citep{deng2009imagenet}. MiniImageNet includes 64 classes for source data in pre-training, and 20 classes for target data in fine-tuning. Likewise, tieredImageNet has 351 classes and 160 classes for source and target data, respectively. Cases when a model is pre-trained and fine-tuned by one of these datasets\,(e.g., tieredIN$\rightarrow$miniIN), are regarded as single-domain scenarios.

We use BSCD-FSL\,\citep{guo2020broader} and non-BSCD-FSL datasets as target data. BSCD-FSL benchmark\,\citep{guo2020broader} includes CropDiseases, EuroSAT, ISIC, and ChestX\,(arranged in the order of similarity to source domain). CropDiseases\,\citep{mohanty2016using} is a plant disease dataset and EuroSAT\,\citep{helber2019eurosat} is a satellite image dataset. In addition, ISIC\,\citep{codella2019skin} and ChestX\,\citep{wang2017chestx} each include images of skin lesion and X-ray. Similarly, non-BSCD-FSL includes Places\,\citep{zhou2017places}, Plantae\,\citep{van2018inaturalist}, Cars\,\citep{krause20133d} and CUB\,\citep{wah2011caltech}. Each dataset contains images of different scene categories, plant species, car models and bird species, respectively. 

\section{Augmentation Details}\label{appx:DA}
We consider data augmentation methods\footnote{https://pytorch.org/vision/stable/transforms.html} as follows:
\begin{itemize}
    \item Random Horizontal Flip (HFlip) flips an image horizontally with probability of 0.5.
    \item Random Resized Crop (RCrop) crops an image with a random scale and ratio, and resize it to 224$\times$224. The scale and ratio are sampled every epoch from [0.08, 1.0] and [0.75, 1.33], respectively. 
    \item Color Jitter (CJitter) changes properties of an image including brightness, contrast, saturation and hue. The values of brightness, contrast, and saturation are each sampled from [0.6, 1.4] every epoch, while the hue is fixed as the value of 0. 
    \item Base augmentation (Base Aug) applies HFlip, RCrop, and CJitter simultaneously.
    \item MixUp\,\citep{zhang2017mixup} linearly interpolates two images with a ratio of $\lambda \sim \beta(1, 1)$, which is sampled every epoch. Labels are mixed correspondingly.
    \item CutMix\,\citep{yun2019cutmix} crops one image with a ratio of $\lambda \sim \beta(1, 1)$, and pastes the cropped image onto the other image. $\lambda$ is sampled every epoch, and the labels are mixed with the ratio of areas between the cropped image and the original image.
\end{itemize}

\section{Update using ResNet18}\label{appx:FT_LP_tiered}
\begin{figure}[h!]
    \centering
    \includegraphics[width=\linewidth]{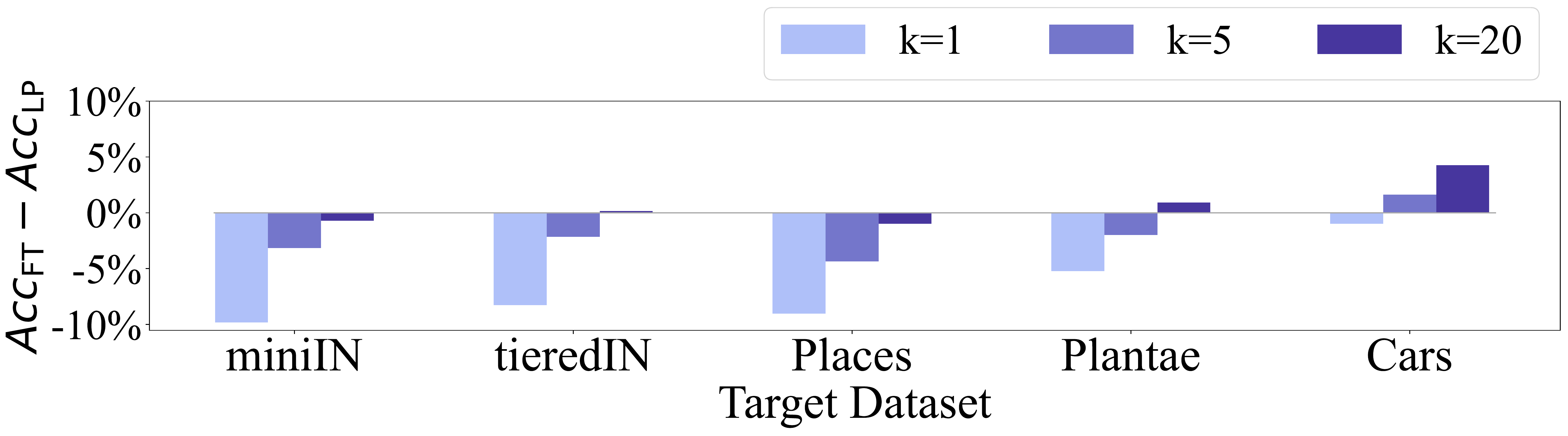}
    \includegraphics[width=\linewidth]{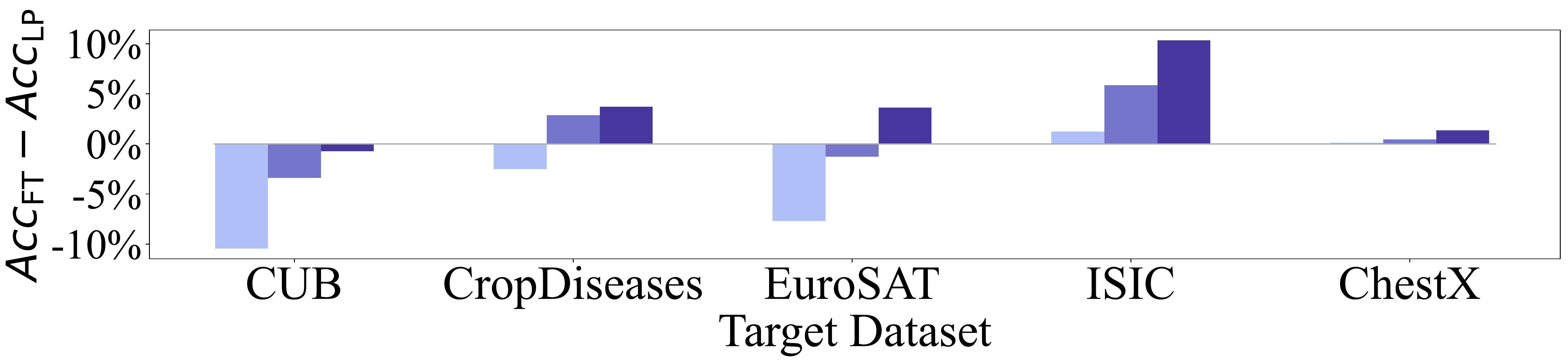}
    \caption{Performance difference of FT with LP on target datasets according to the shot $k$. ResNet18 is pre-trained on tieredIN.}\label{fig:why_full_tiered}
\end{figure}

We provide the performance difference of FT and LP (i.e., $Acc_{\sf FT}-Acc_{\sf LP}$) with ResNet18 pre-trained on tieredIN, shown in Figure \ref{fig:why_full_tiered}.
As we explained, FT becomes better than LP as shot increases in most cases, which is the same trend when ResNet10 is pre-trained on miniIN.

\section{V-measure on Other Datasets}\label{appx:v_measure_other}
\begin{figure}[h!]
\centering
     \begin{subfigure}[h]{0.21\linewidth}
         \includegraphics[width=\linewidth]{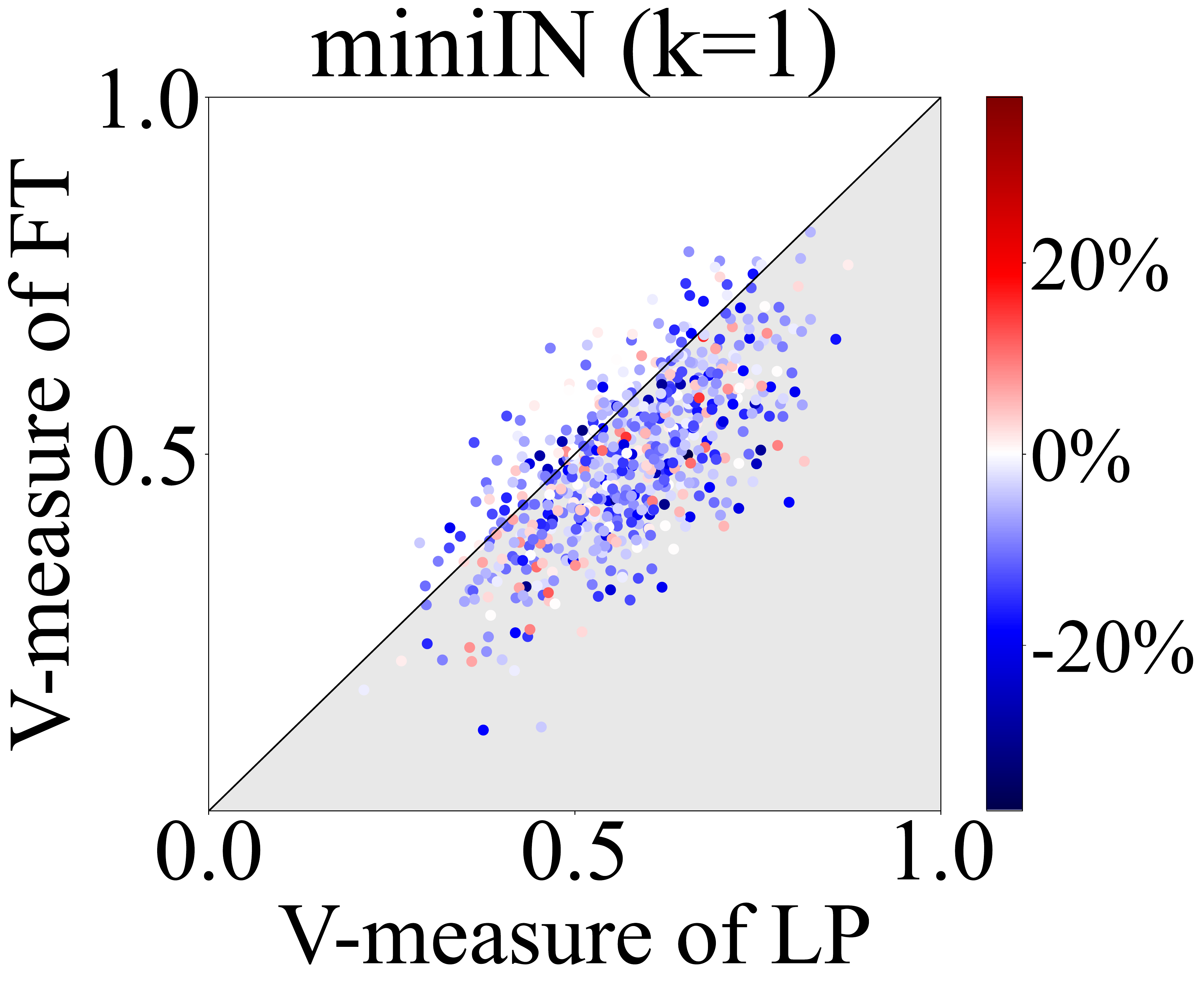}
     \end{subfigure}
     \begin{subfigure}[h]{0.185\linewidth}
         \includegraphics[width=\linewidth]{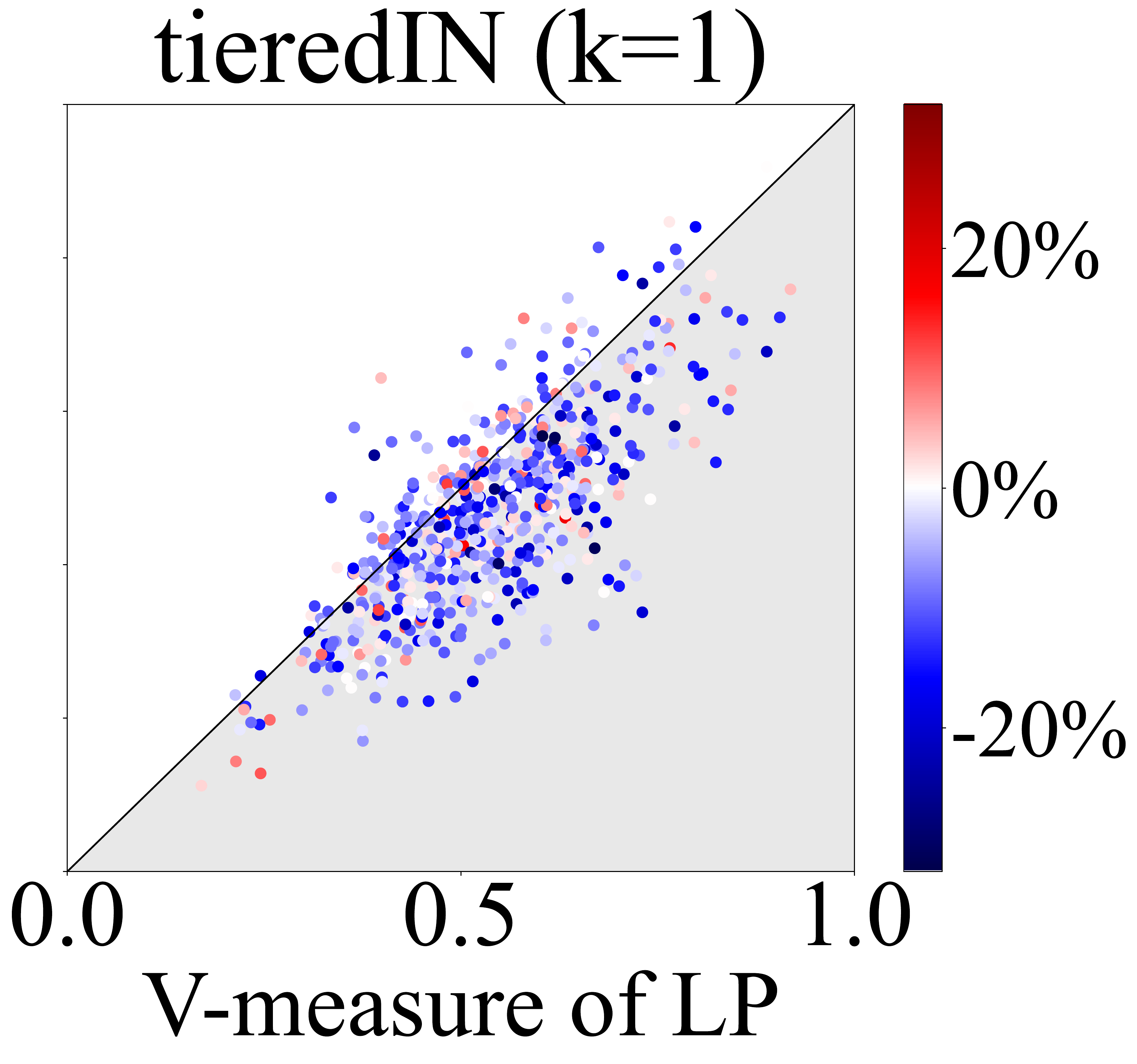}
     \end{subfigure}
     \begin{subfigure}[h]{0.185\linewidth}
         \includegraphics[width=\linewidth]{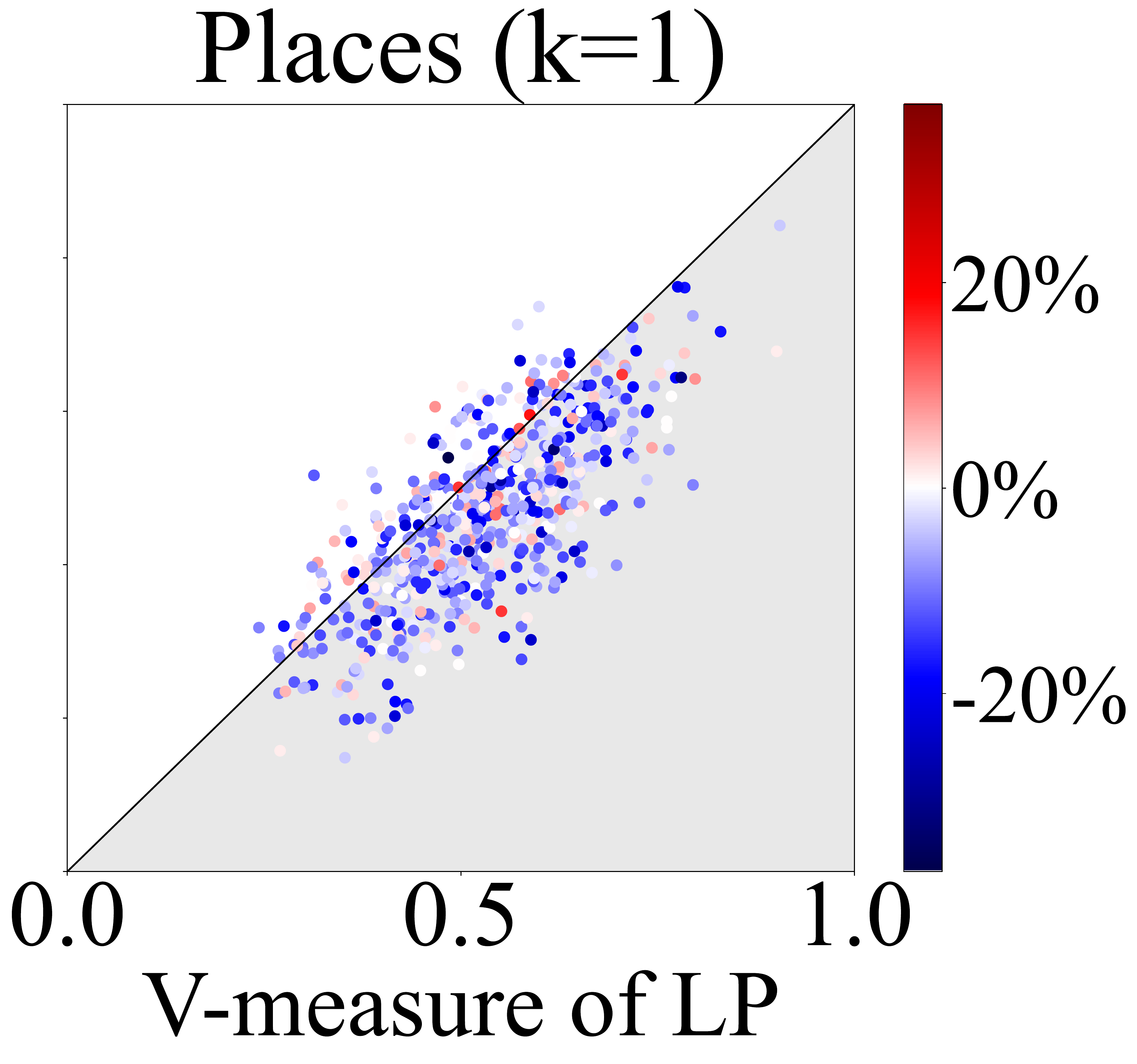}
     \end{subfigure}
     \begin{subfigure}[h]{0.185\linewidth}
         \includegraphics[width=\linewidth]{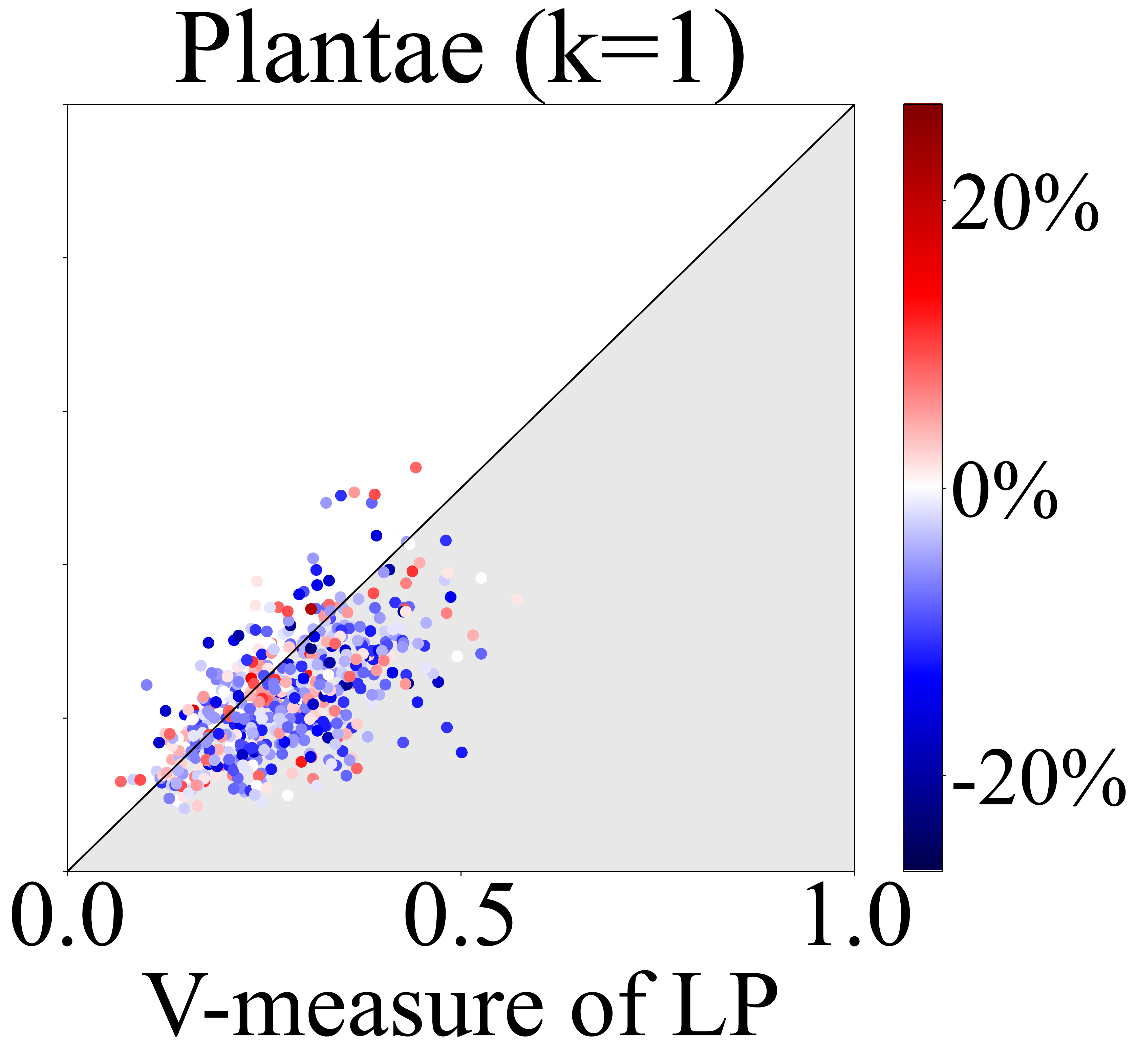}
     \end{subfigure}
     \begin{subfigure}[h]{0.185\linewidth}
         \includegraphics[width=\linewidth]{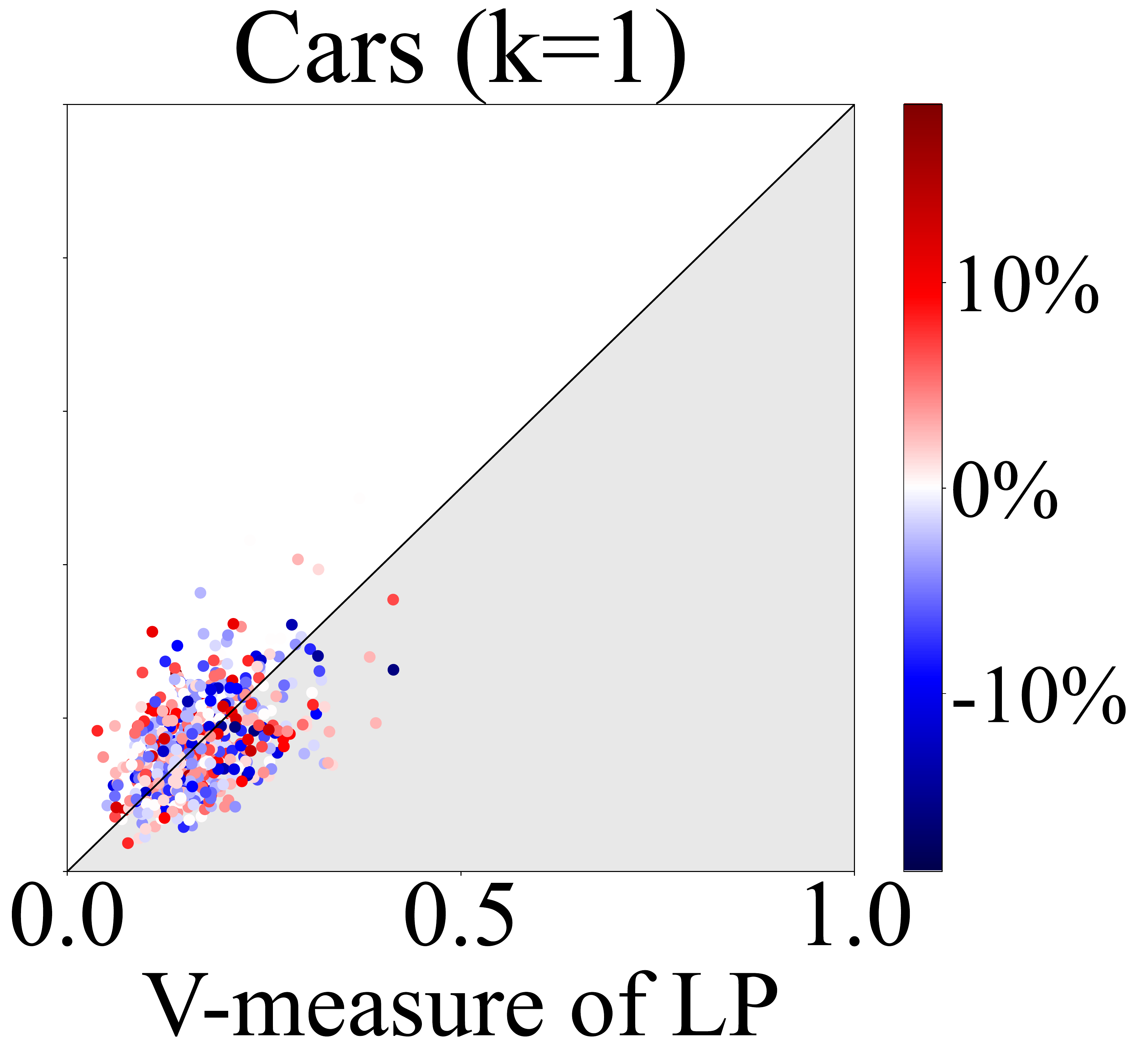}
     \end{subfigure}
     
     \begin{subfigure}[h]{0.21\linewidth}
         \includegraphics[width=\linewidth]{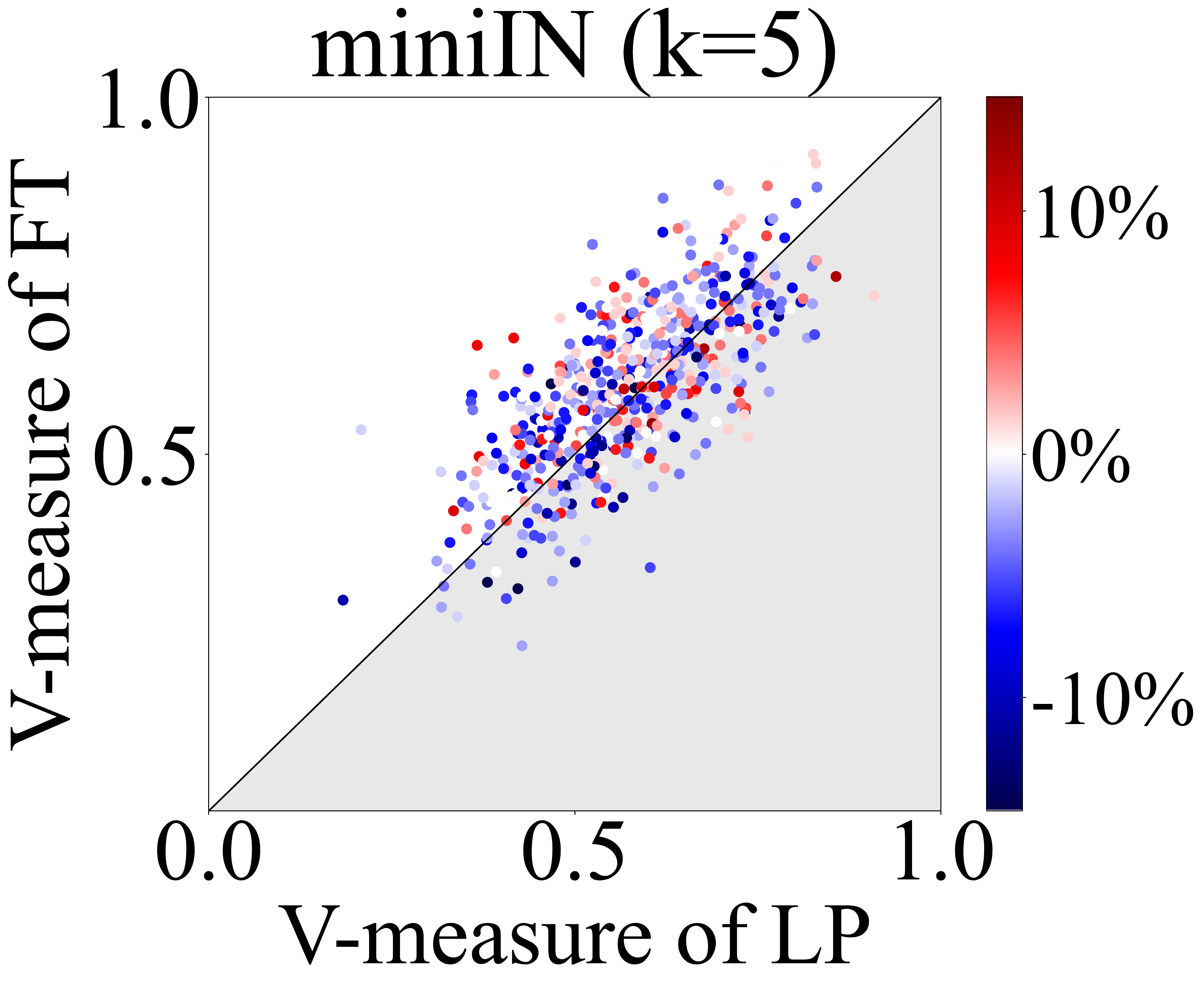}
     \end{subfigure}
     \begin{subfigure}[h]{0.185\linewidth}
         \includegraphics[width=\linewidth]{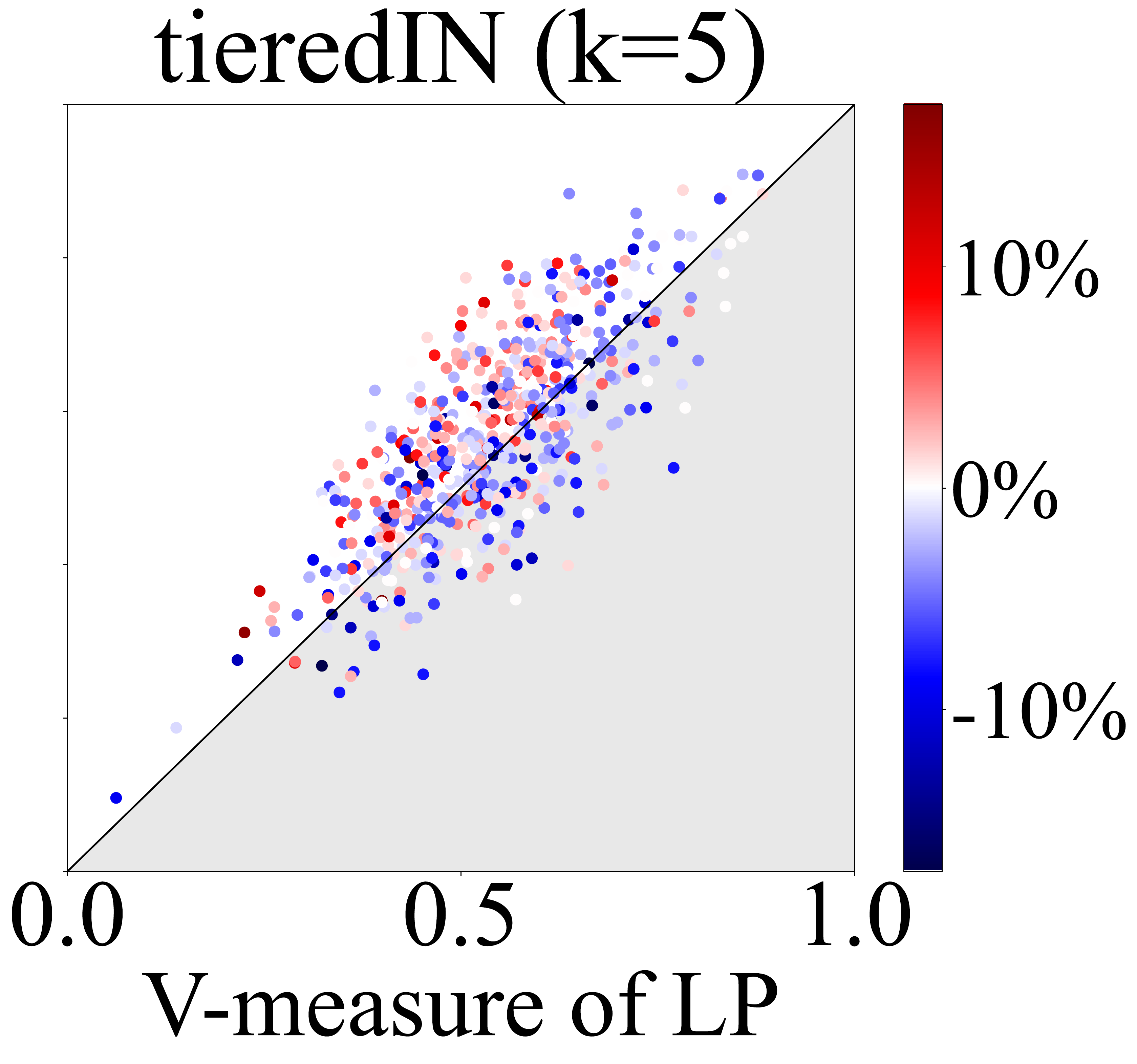}
     \end{subfigure}
     \begin{subfigure}[h]{0.185\linewidth}
         \includegraphics[width=\linewidth]{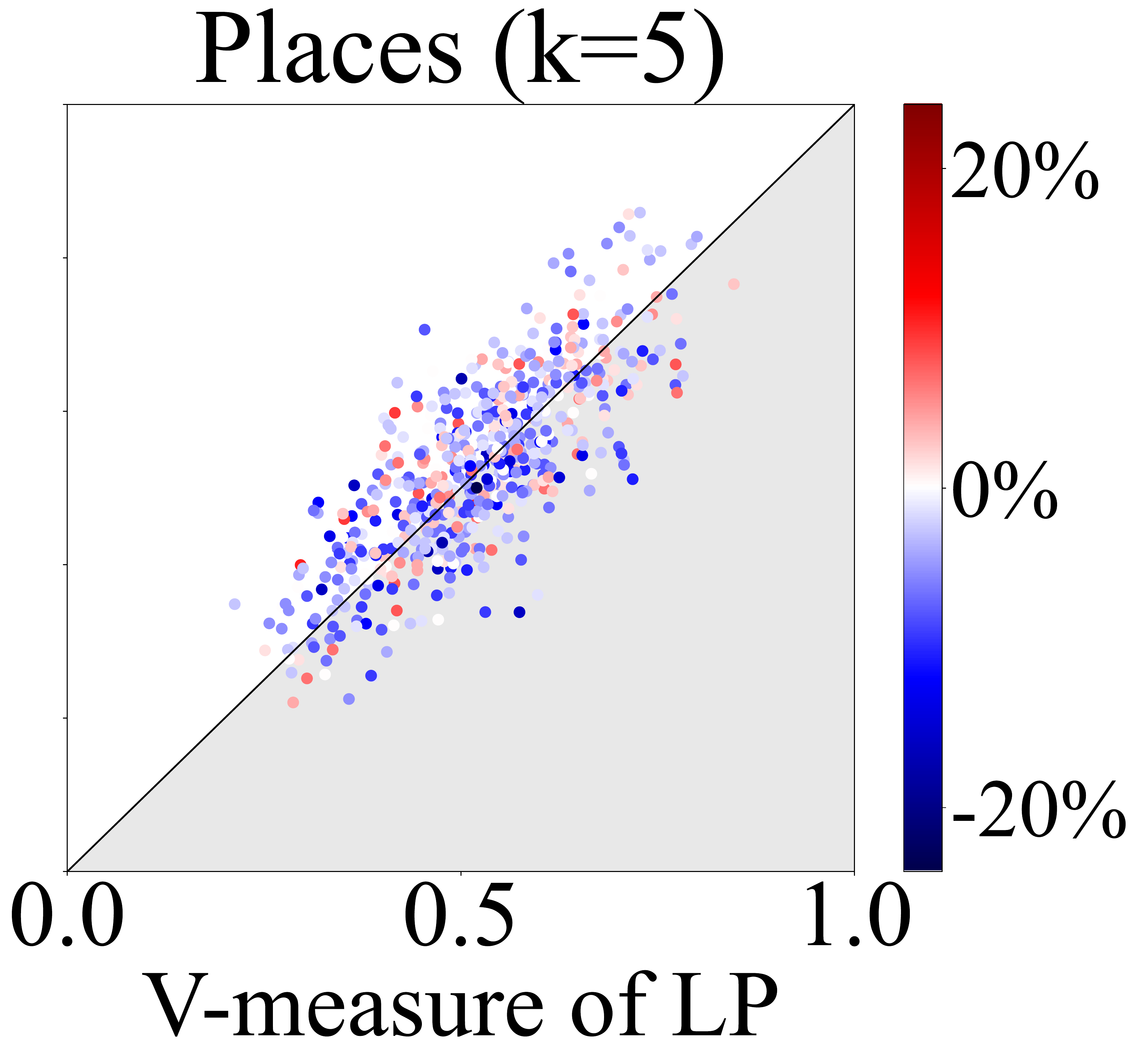}
     \end{subfigure}
     \begin{subfigure}[h]{0.185\linewidth}
         \includegraphics[width=\linewidth]{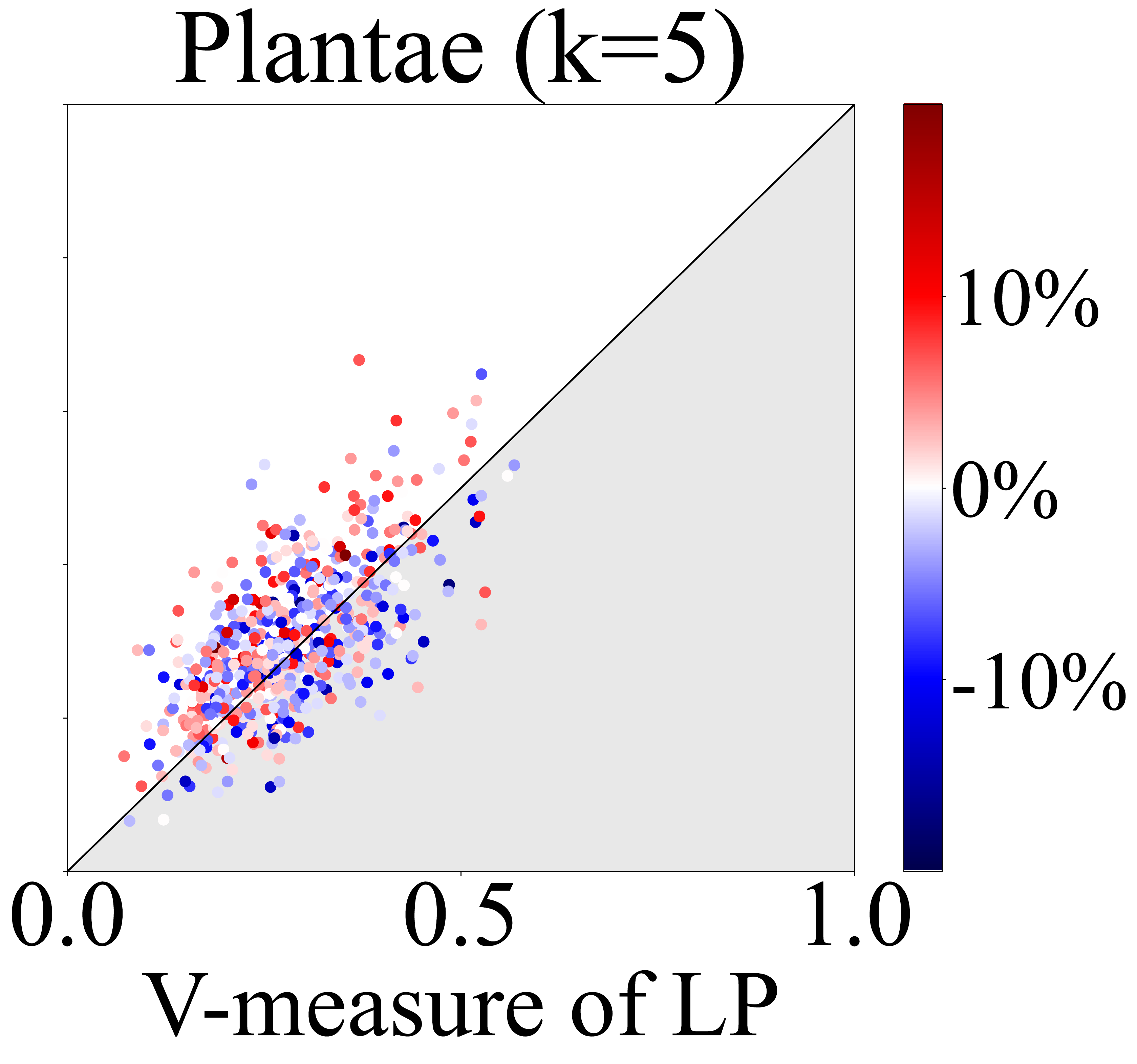}
     \end{subfigure}
     \begin{subfigure}[h]{0.185\linewidth}
         \includegraphics[width=\linewidth]{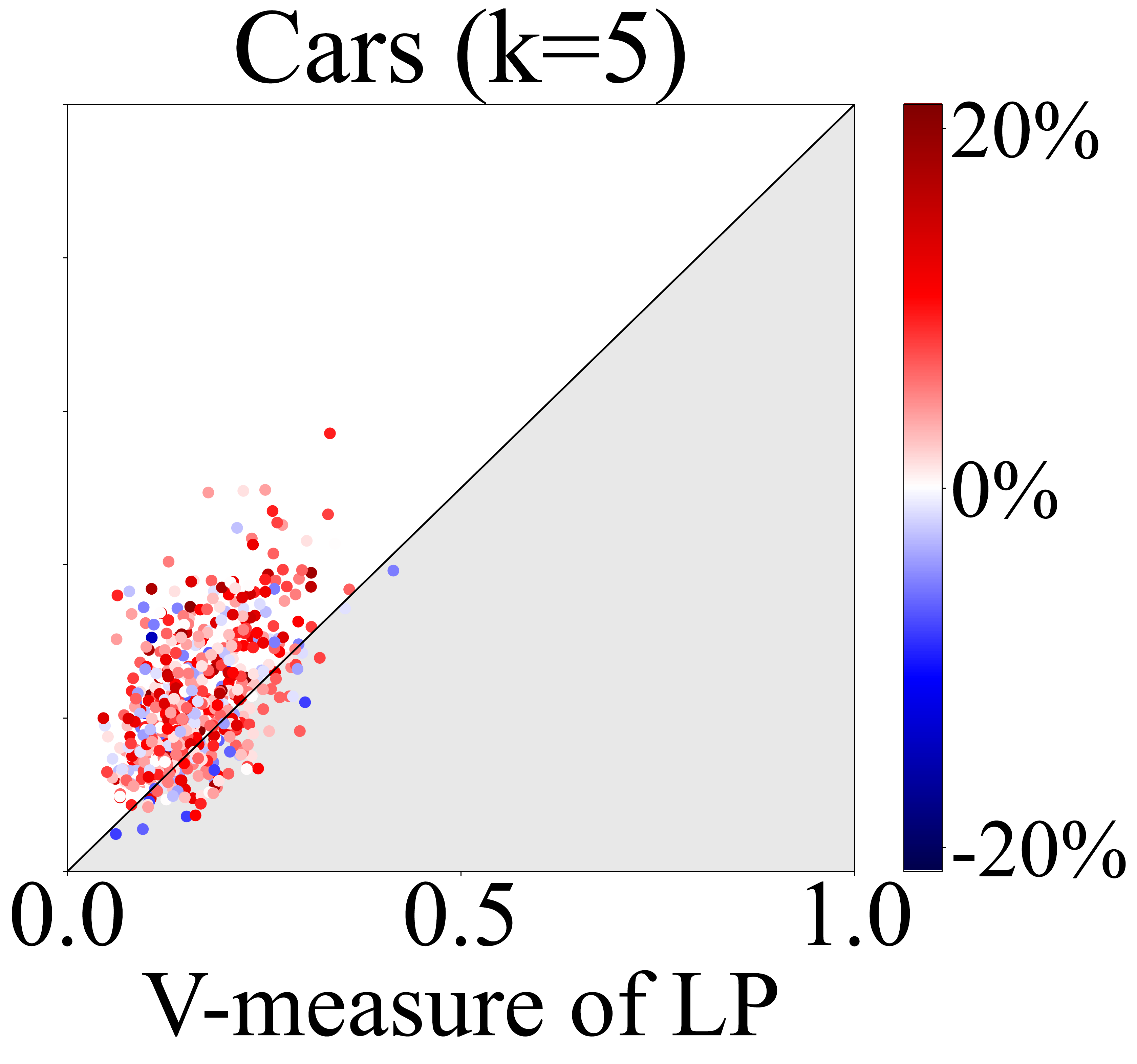}
     \end{subfigure}
     
     \begin{subfigure}[h]{0.21\linewidth}
         \includegraphics[width=\linewidth]{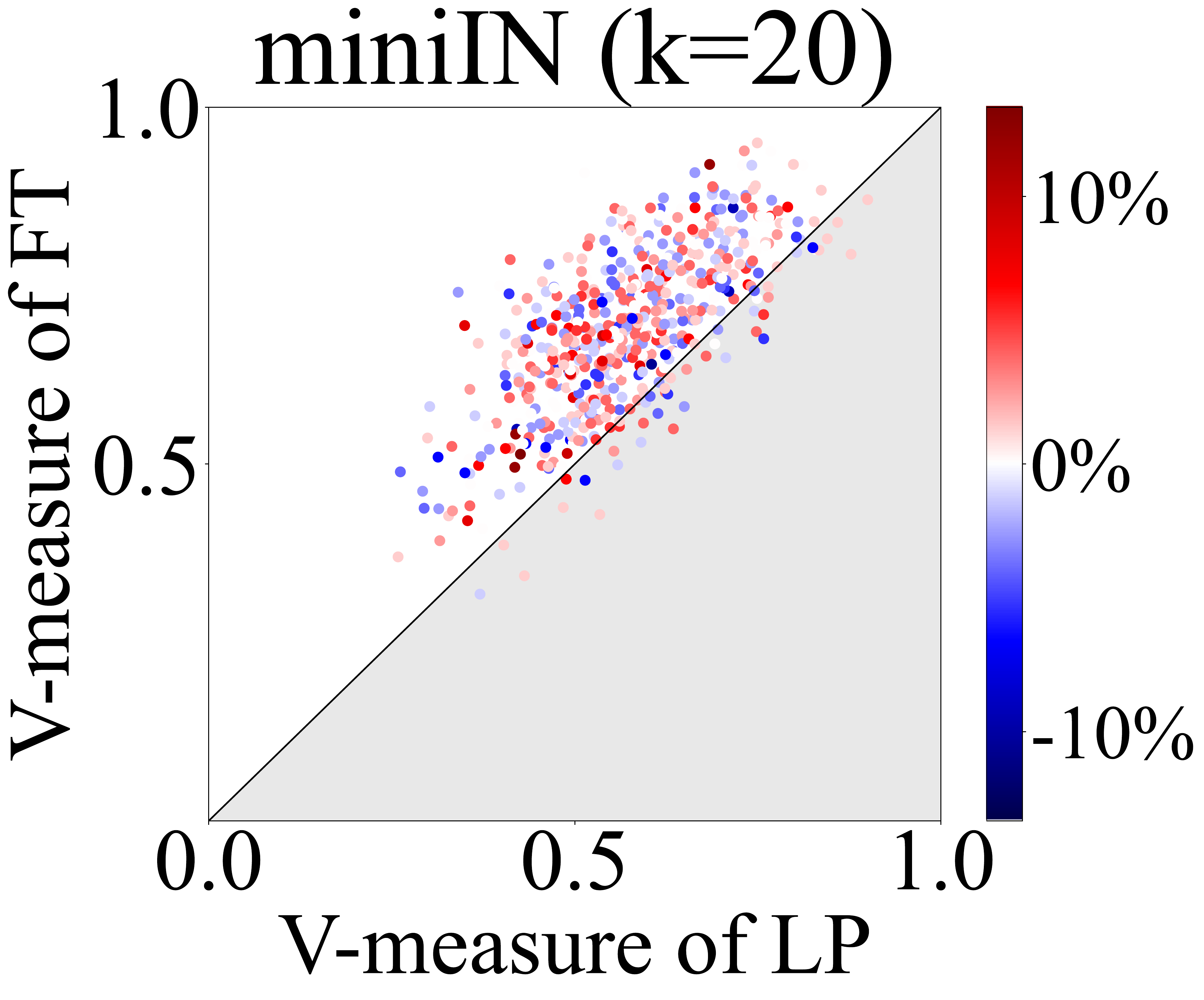}
     \end{subfigure}
     \begin{subfigure}[h]{0.185\linewidth}
         \includegraphics[width=\linewidth]{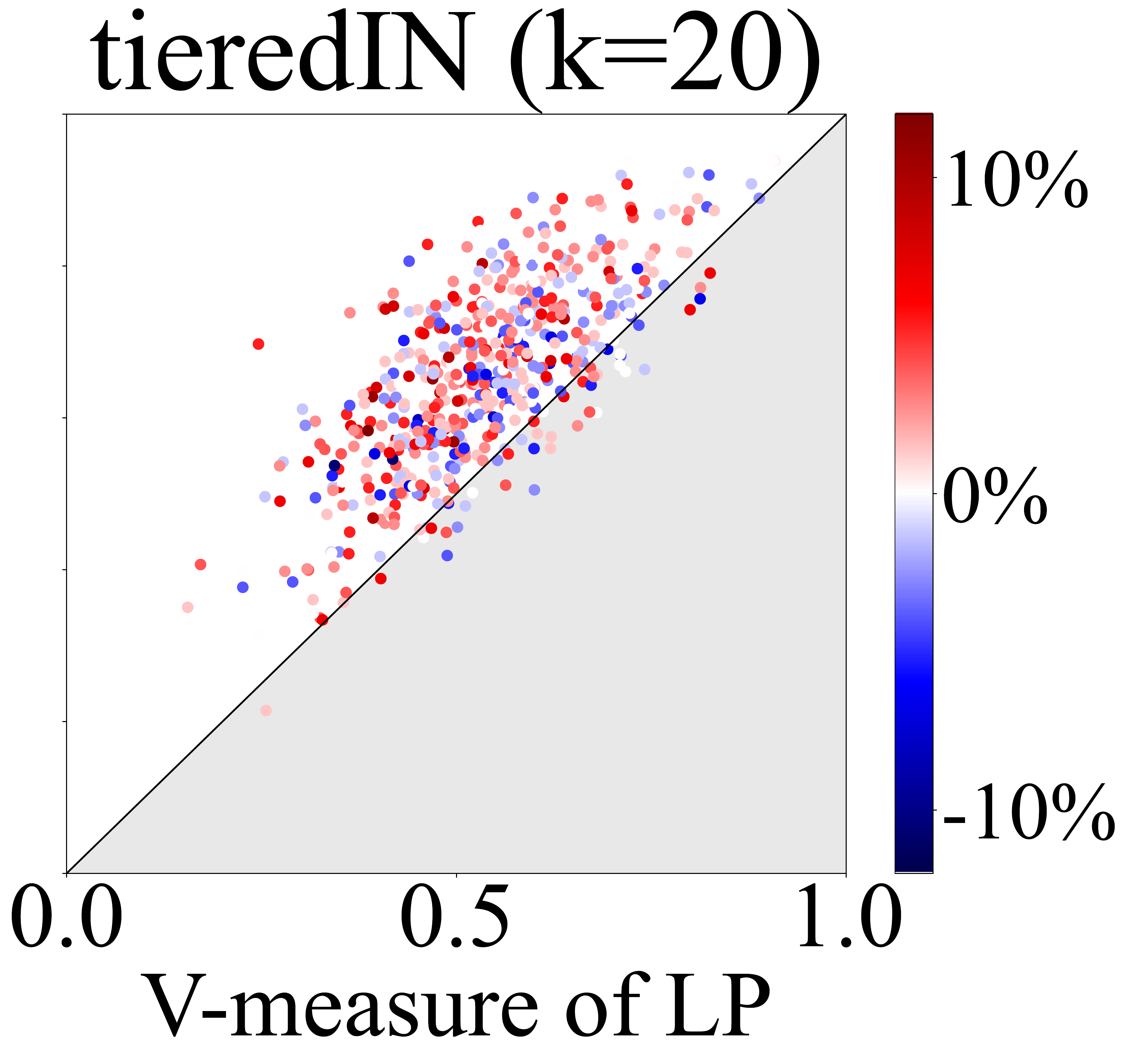}
     \end{subfigure}
     \begin{subfigure}[h]{0.185\linewidth}
         \includegraphics[width=\linewidth]{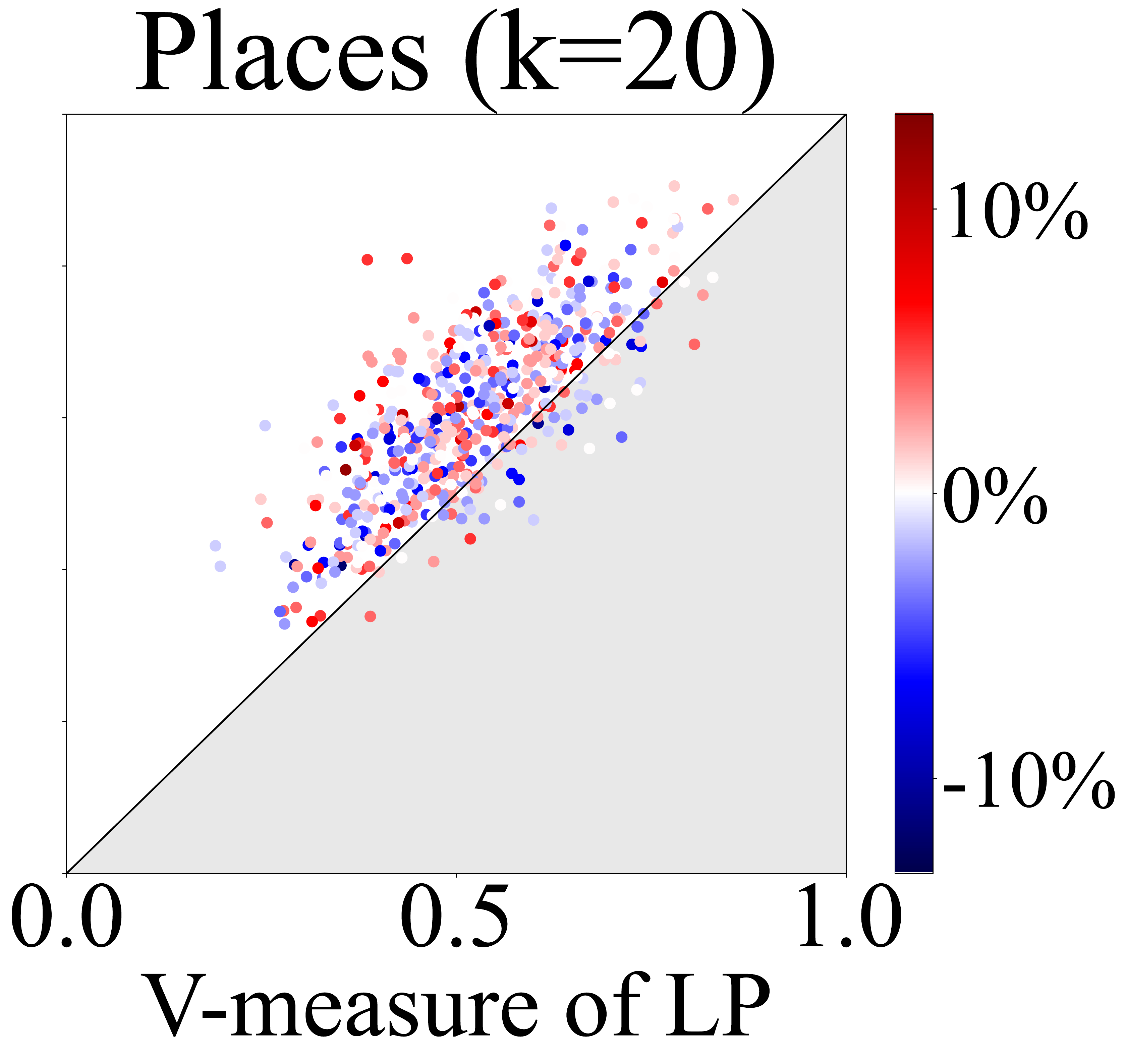}
     \end{subfigure}
     \begin{subfigure}[h]{0.185\linewidth}
         \includegraphics[width=\linewidth]{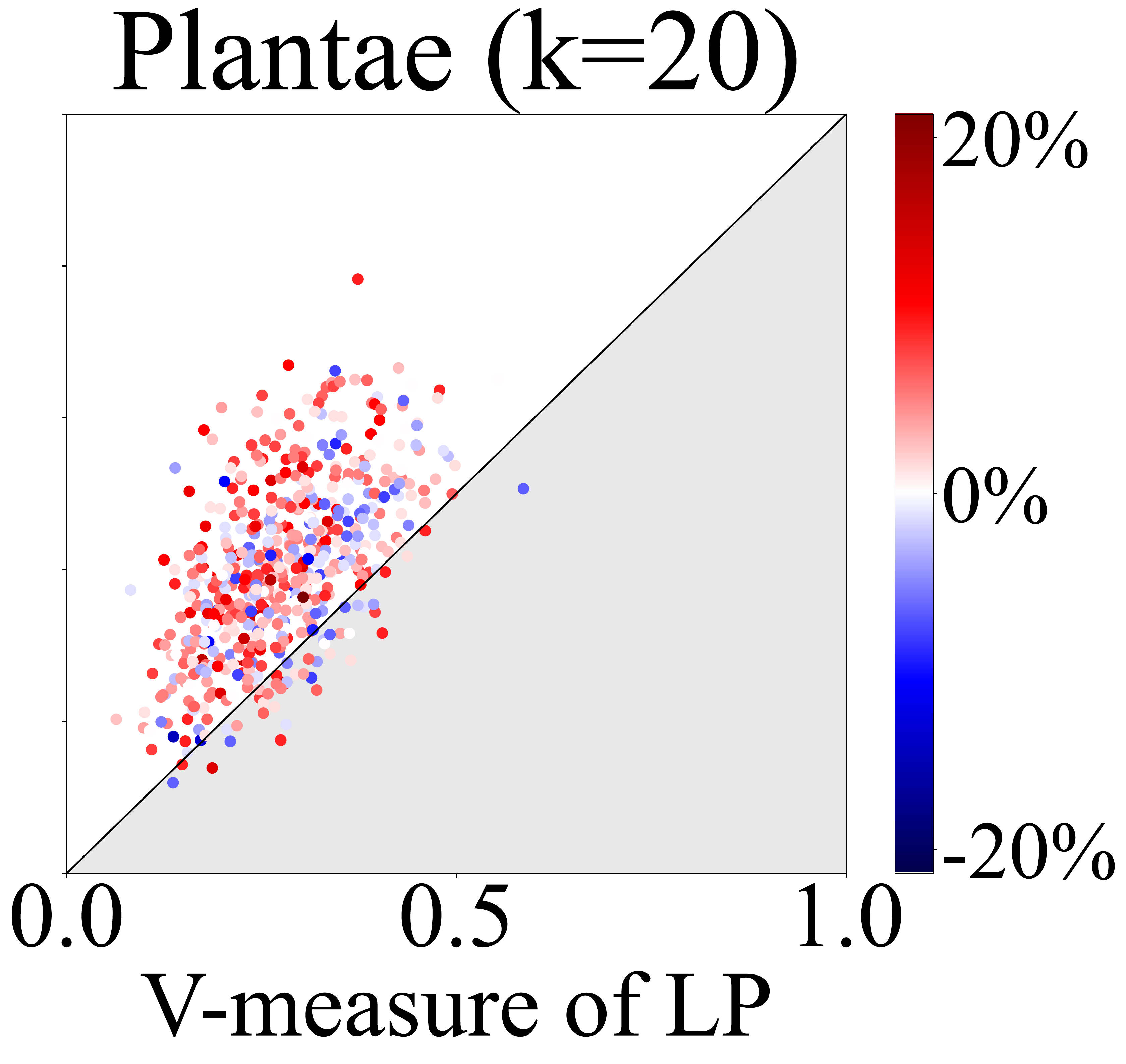}
     \end{subfigure}
     \begin{subfigure}[h]{0.185\linewidth}
         \includegraphics[width=\linewidth]{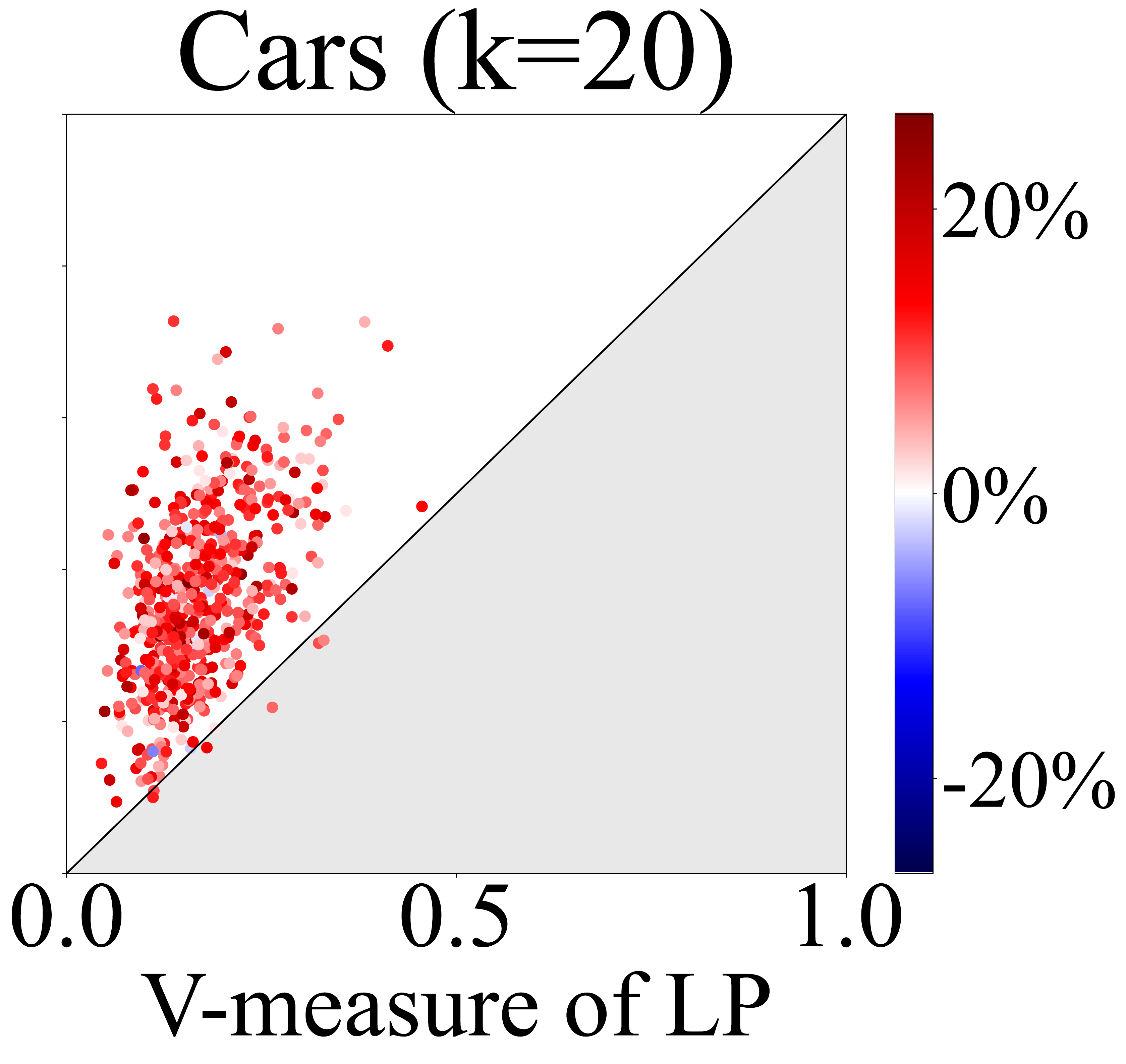}
     \end{subfigure}
     
     \begin{subfigure}[h]{0.21\linewidth}
         \includegraphics[width=\linewidth]{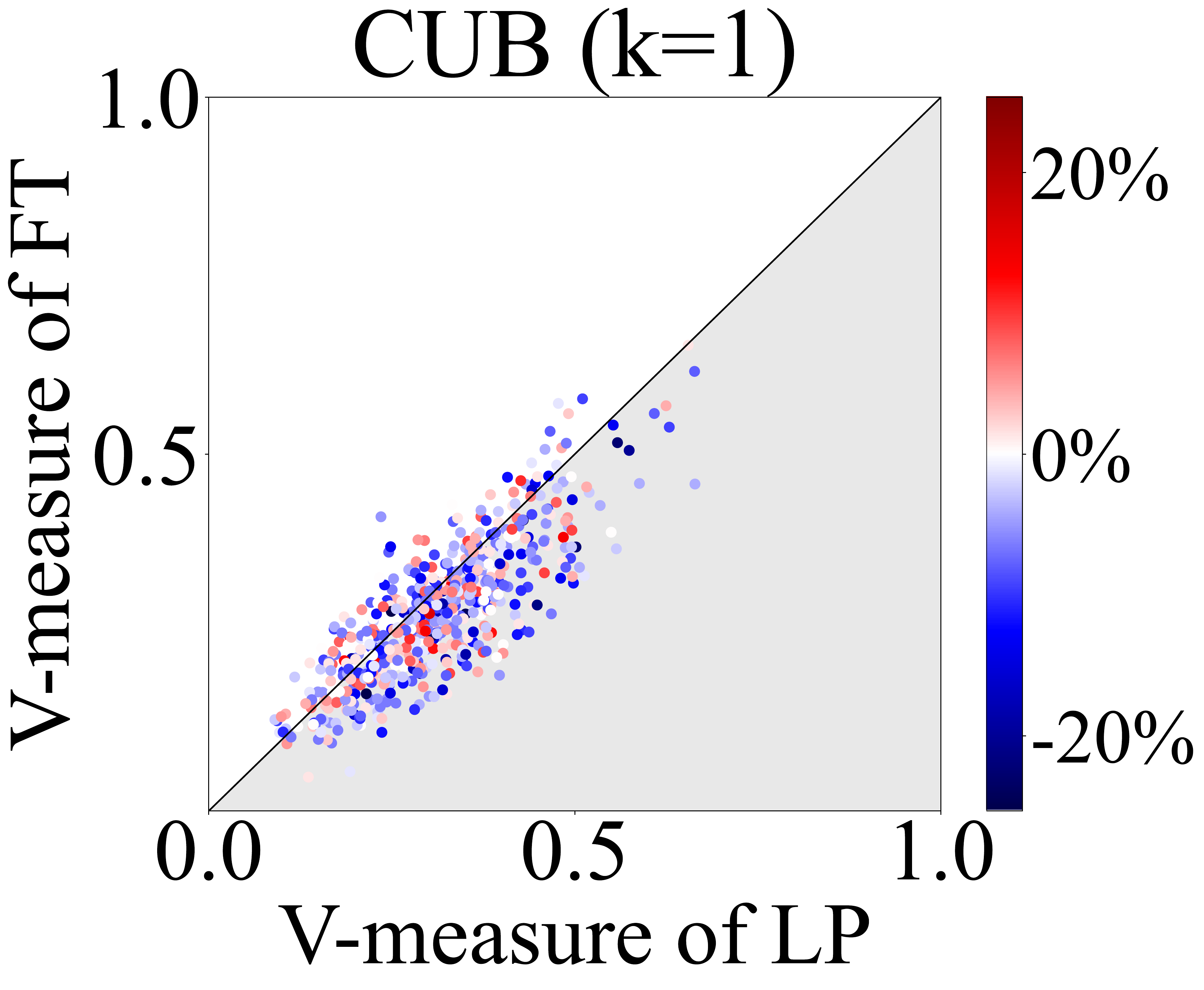}
     \end{subfigure}
     \begin{subfigure}[h]{0.185\linewidth}
         \includegraphics[width=\linewidth]{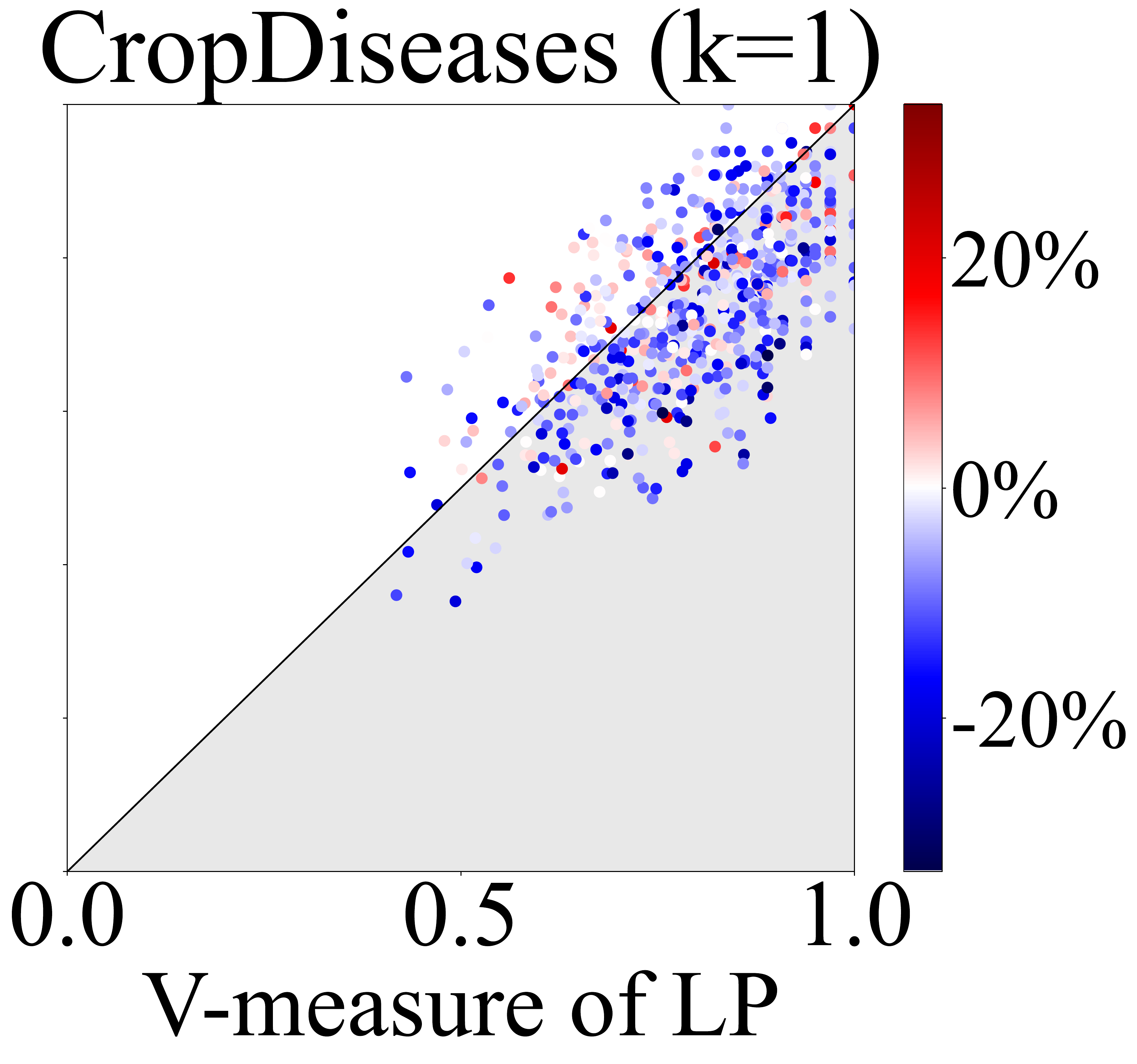}
     \end{subfigure}
     \begin{subfigure}[h]{0.185\linewidth}
         \includegraphics[width=\linewidth]{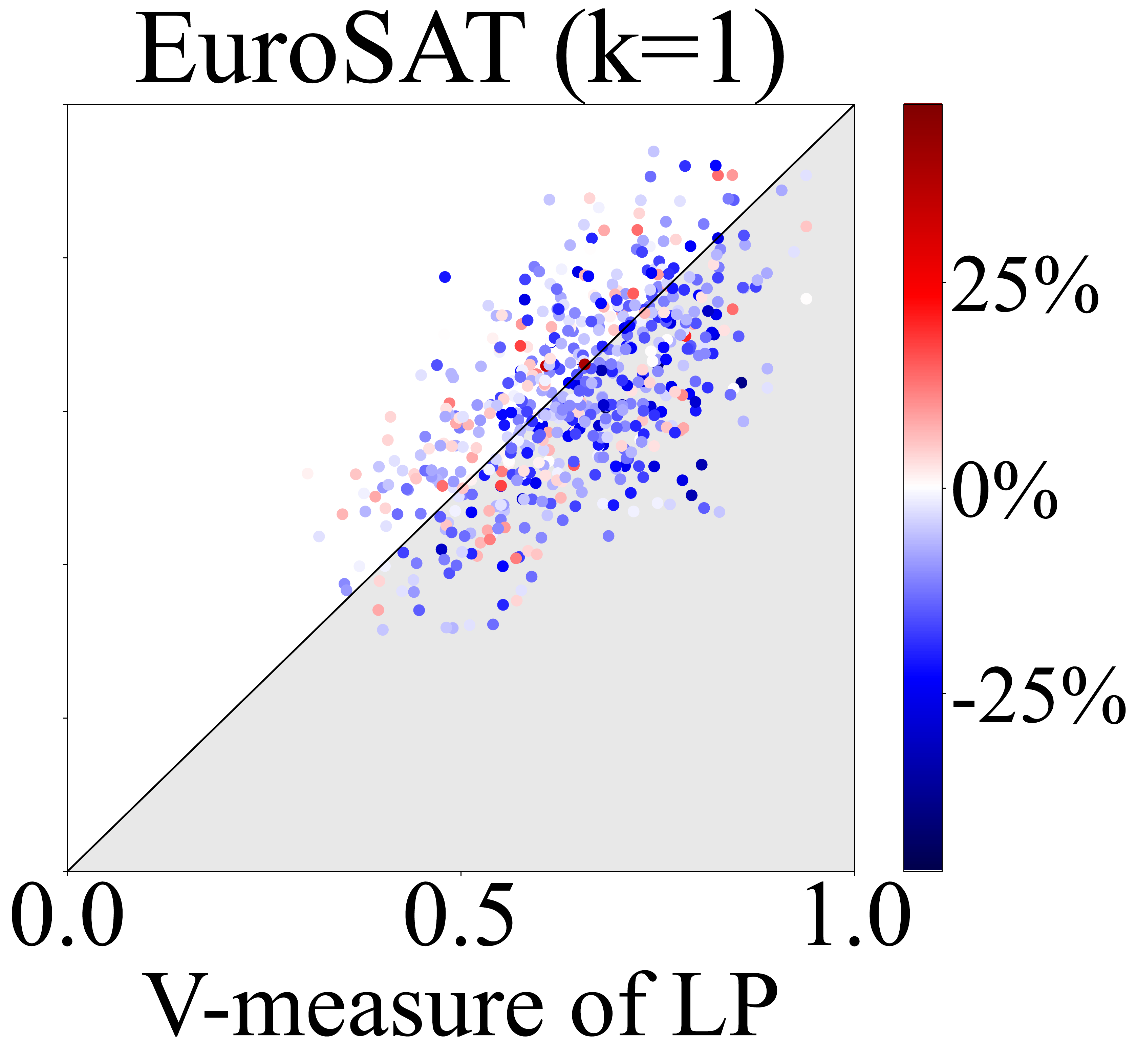}
     \end{subfigure}
     \begin{subfigure}[h]{0.185\linewidth}
         \includegraphics[width=\linewidth]{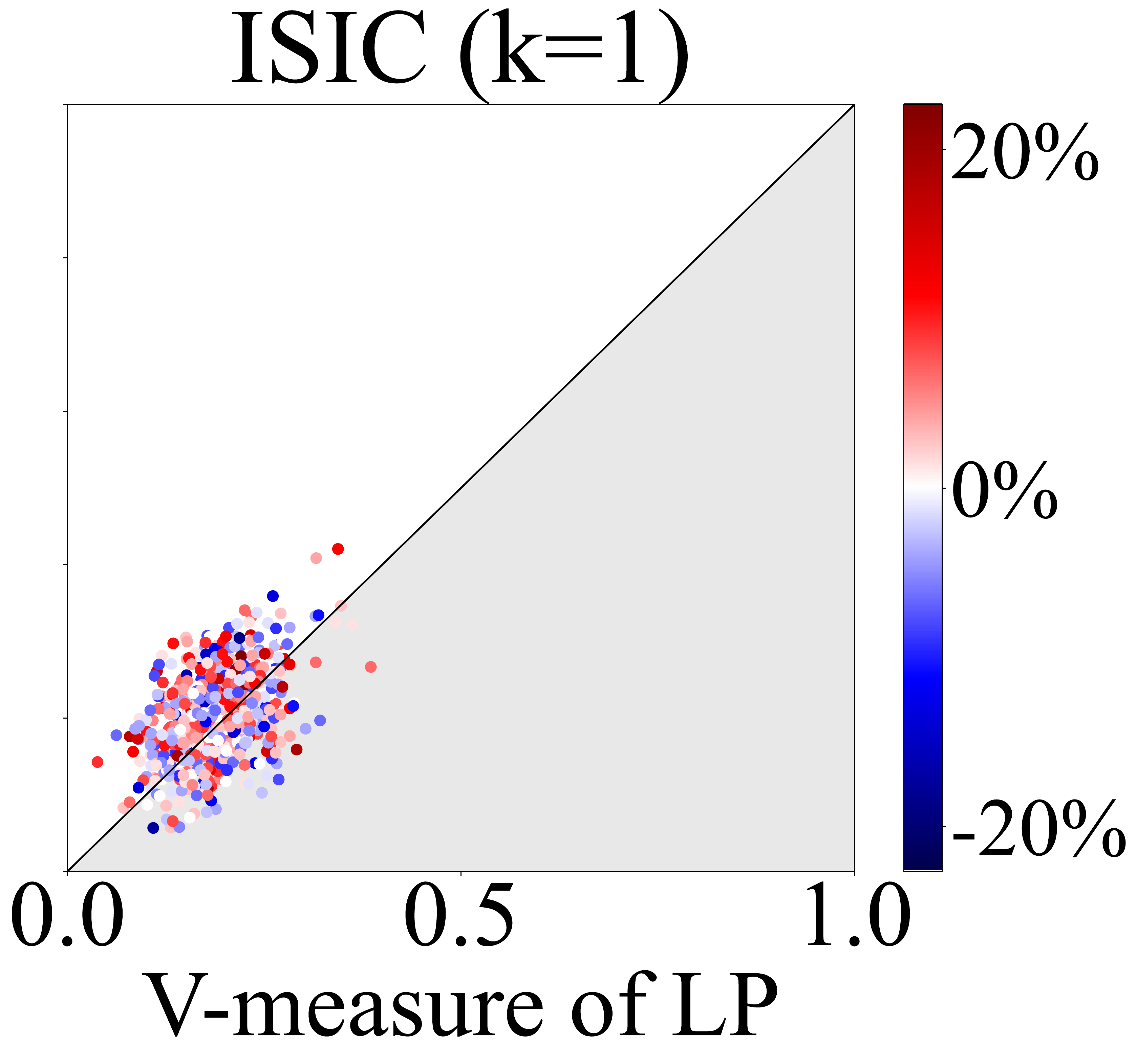}
     \end{subfigure}
     \begin{subfigure}[h]{0.185\linewidth}
         \includegraphics[width=\linewidth]{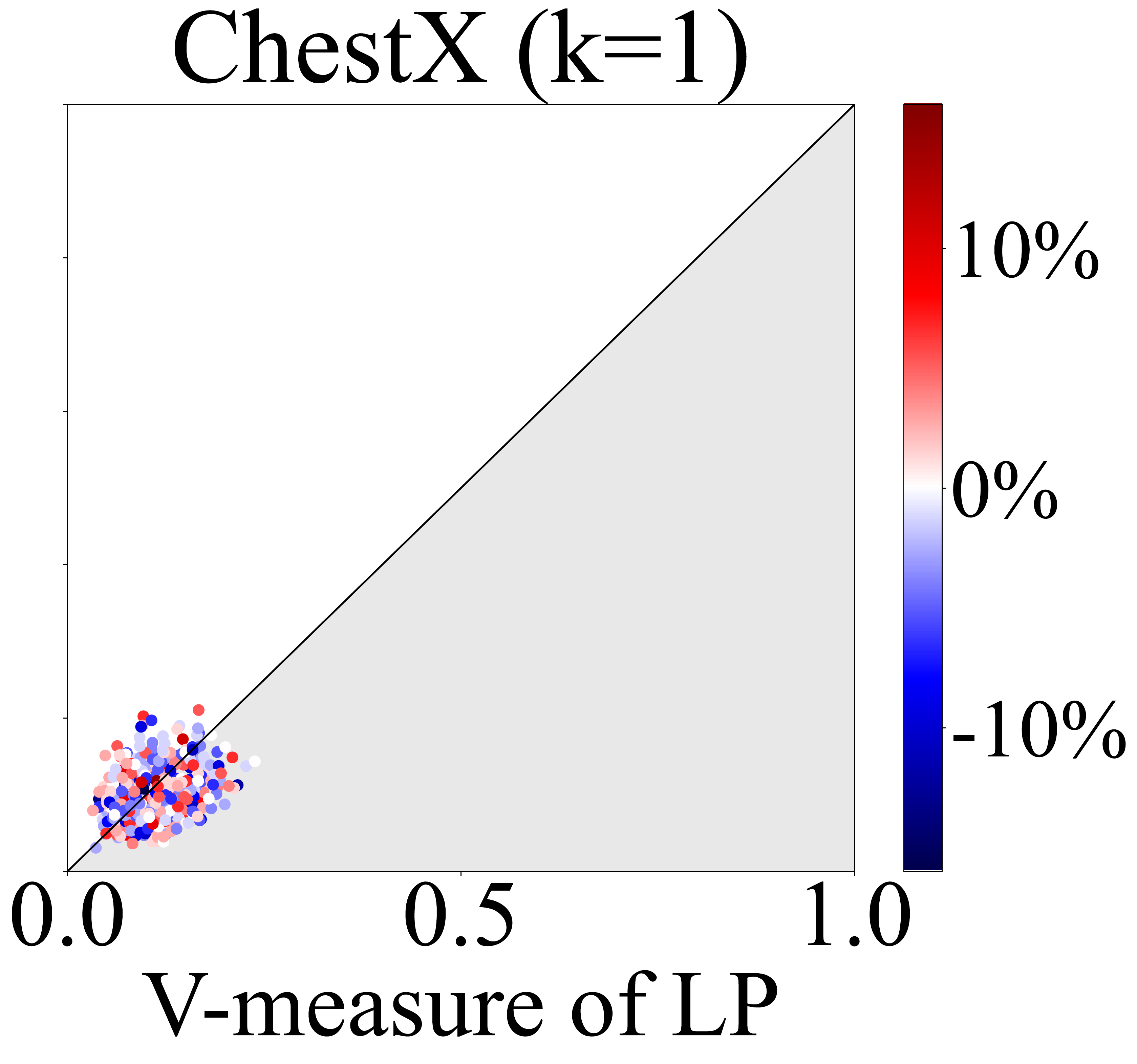}
     \end{subfigure}
     
     \begin{subfigure}[h]{0.21\linewidth}
         \includegraphics[width=\linewidth]{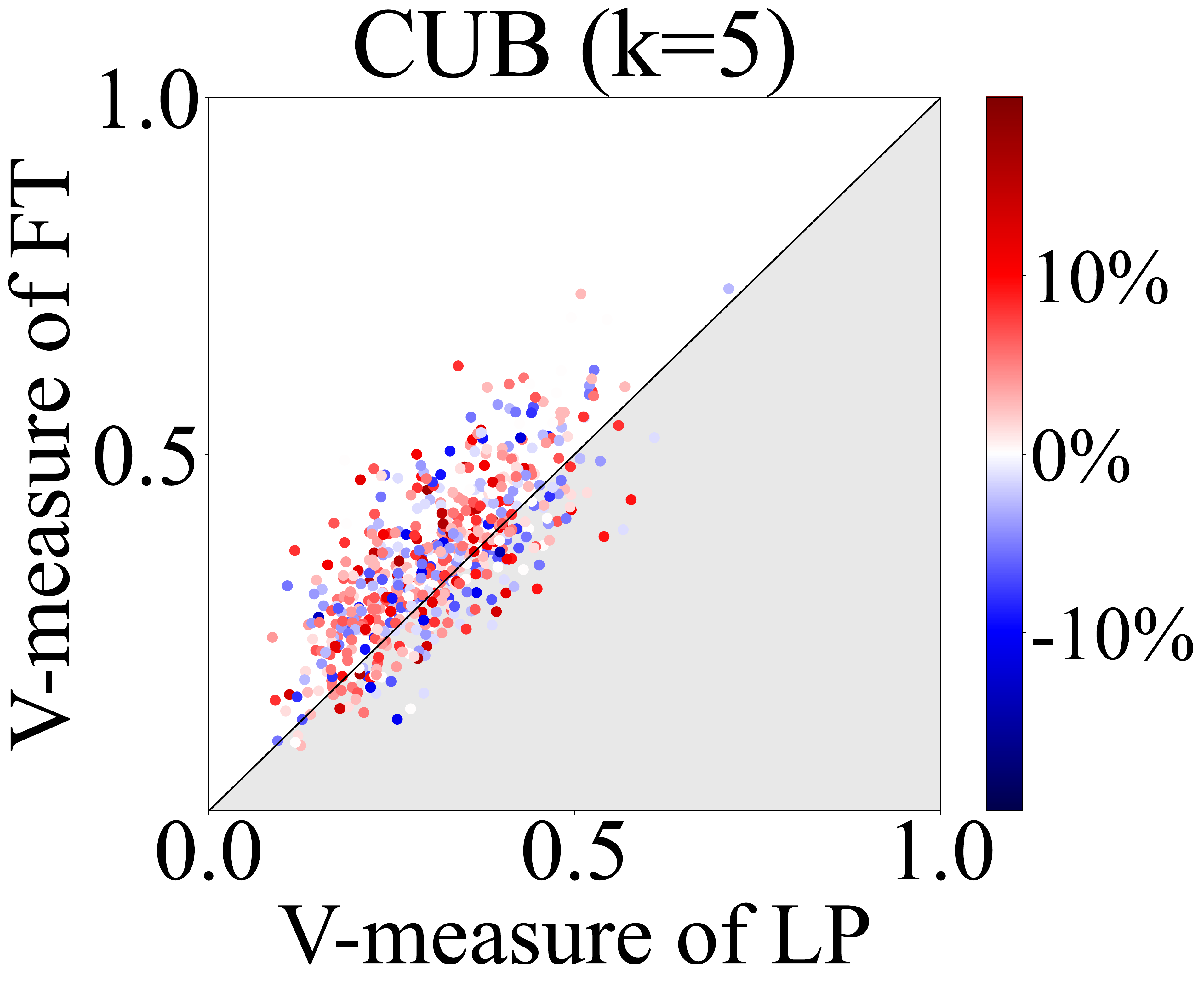}
     \end{subfigure}
     \begin{subfigure}[h]{0.185\linewidth}
         \includegraphics[width=\linewidth]{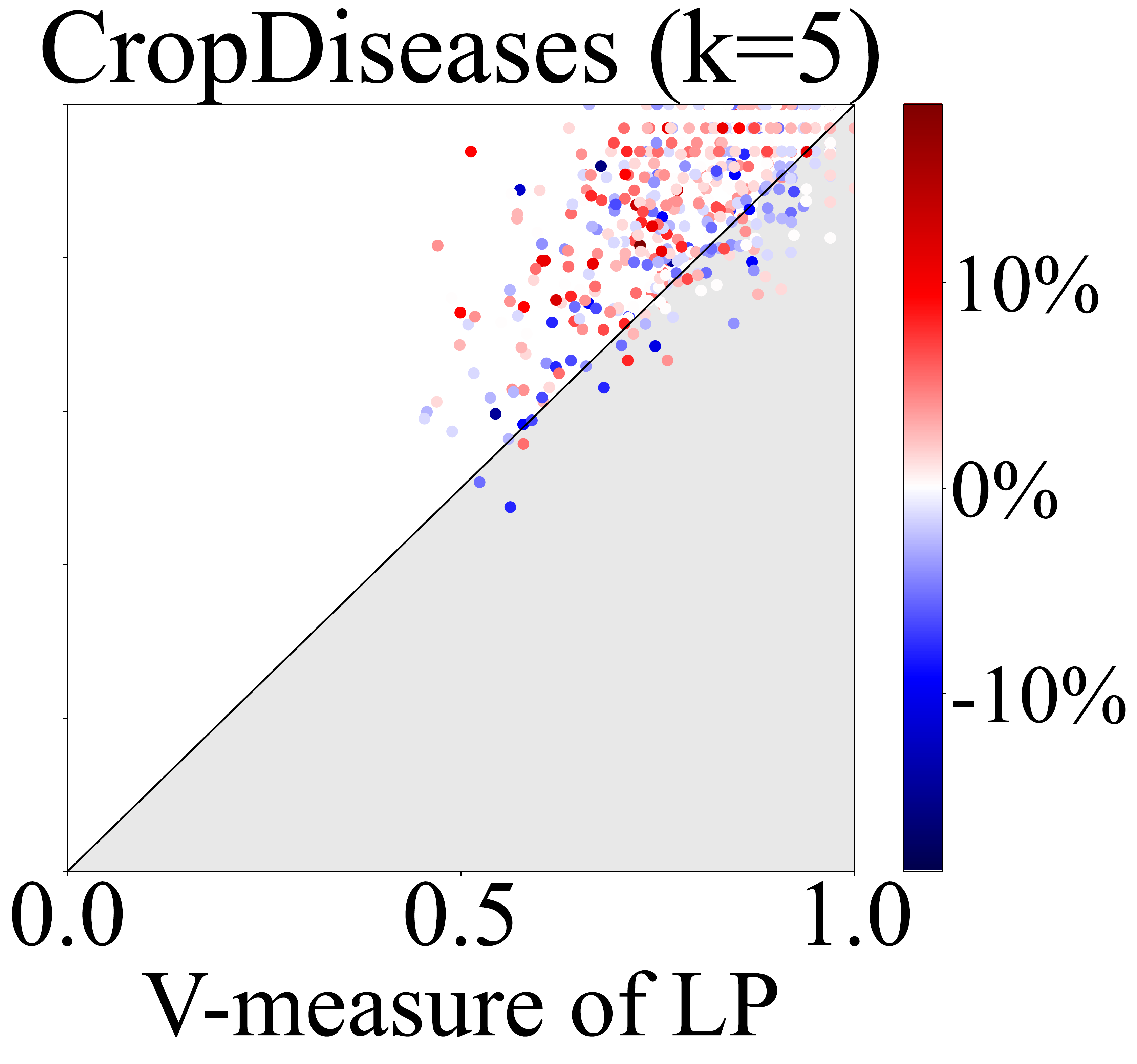}
     \end{subfigure}
     \begin{subfigure}[h]{0.185\linewidth}
         \includegraphics[width=\linewidth]{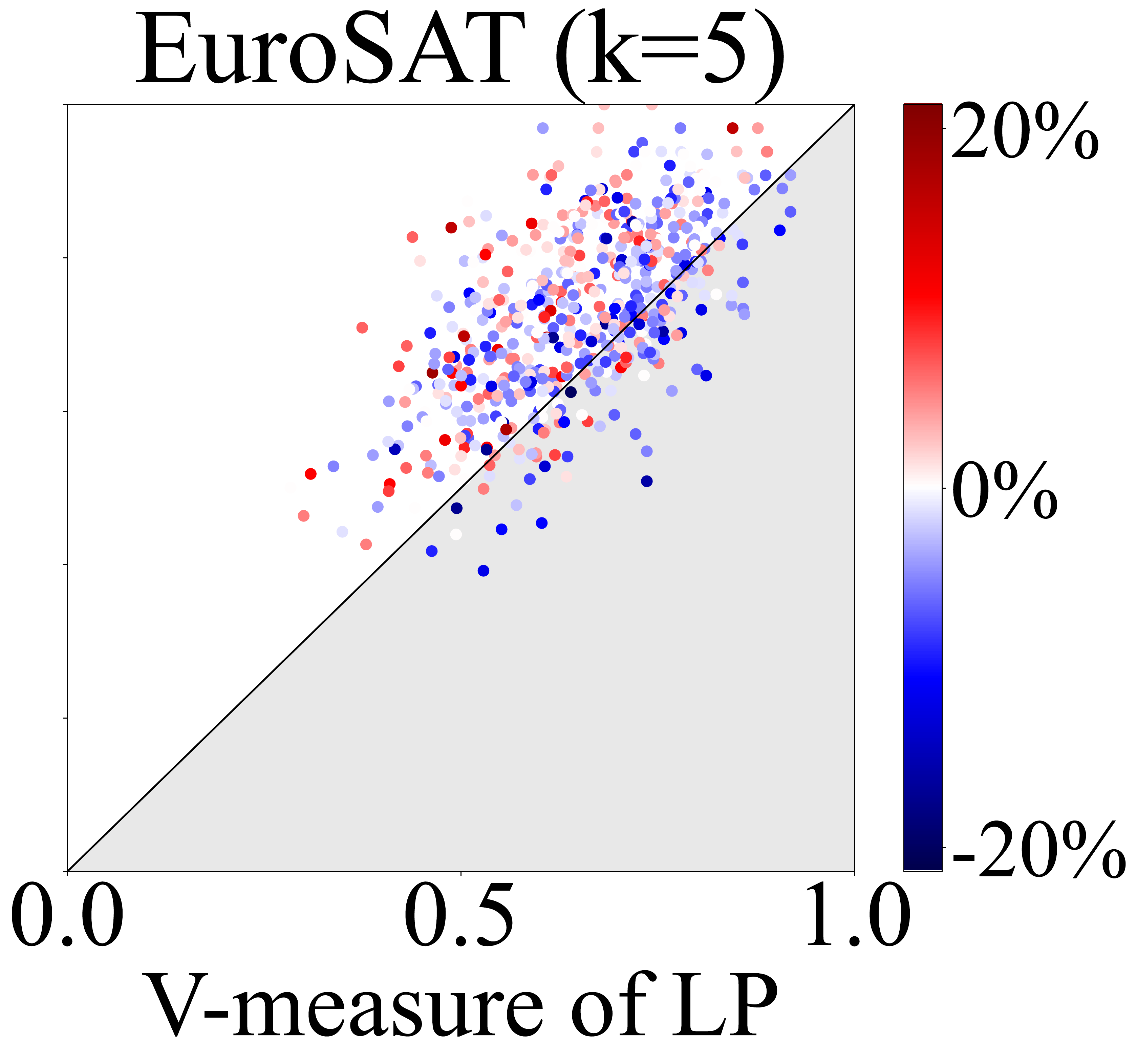}
     \end{subfigure}
     \begin{subfigure}[h]{0.185\linewidth}
         \includegraphics[width=\linewidth]{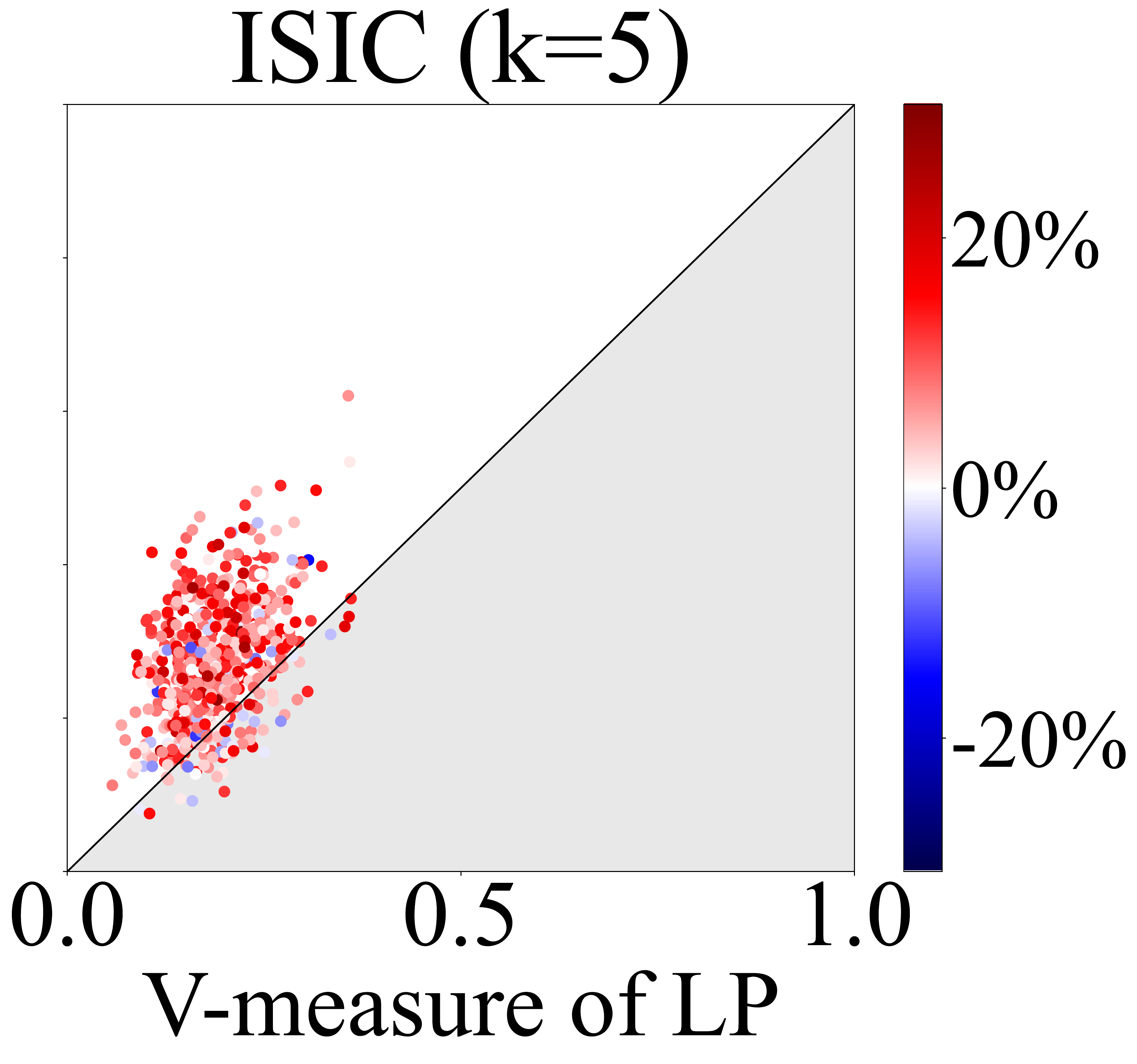}
     \end{subfigure}
     \begin{subfigure}[h]{0.185\linewidth}
         \includegraphics[width=\linewidth]{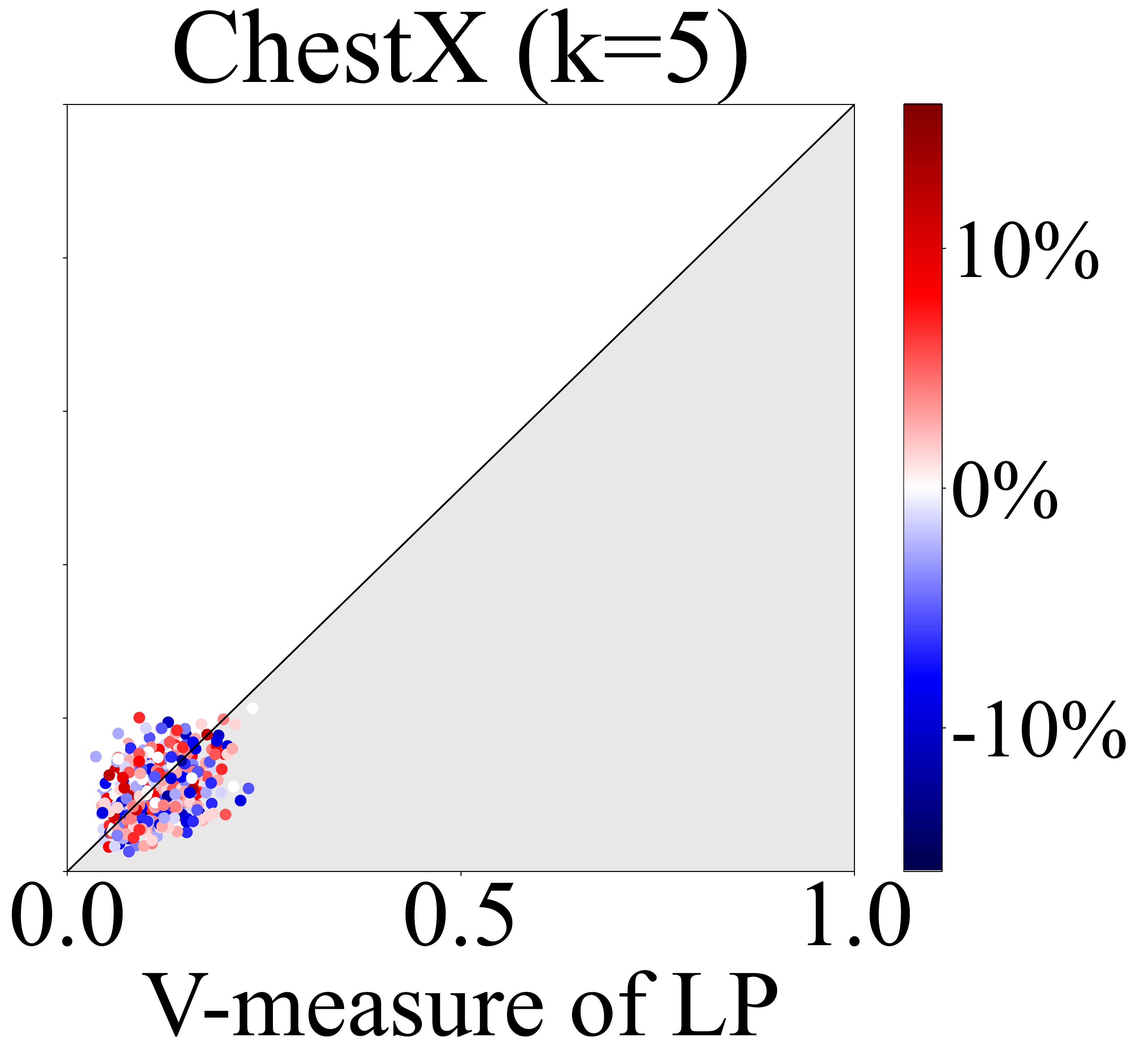}
     \end{subfigure}
     
     \begin{subfigure}[h]{0.21\linewidth}
         \includegraphics[width=\linewidth]{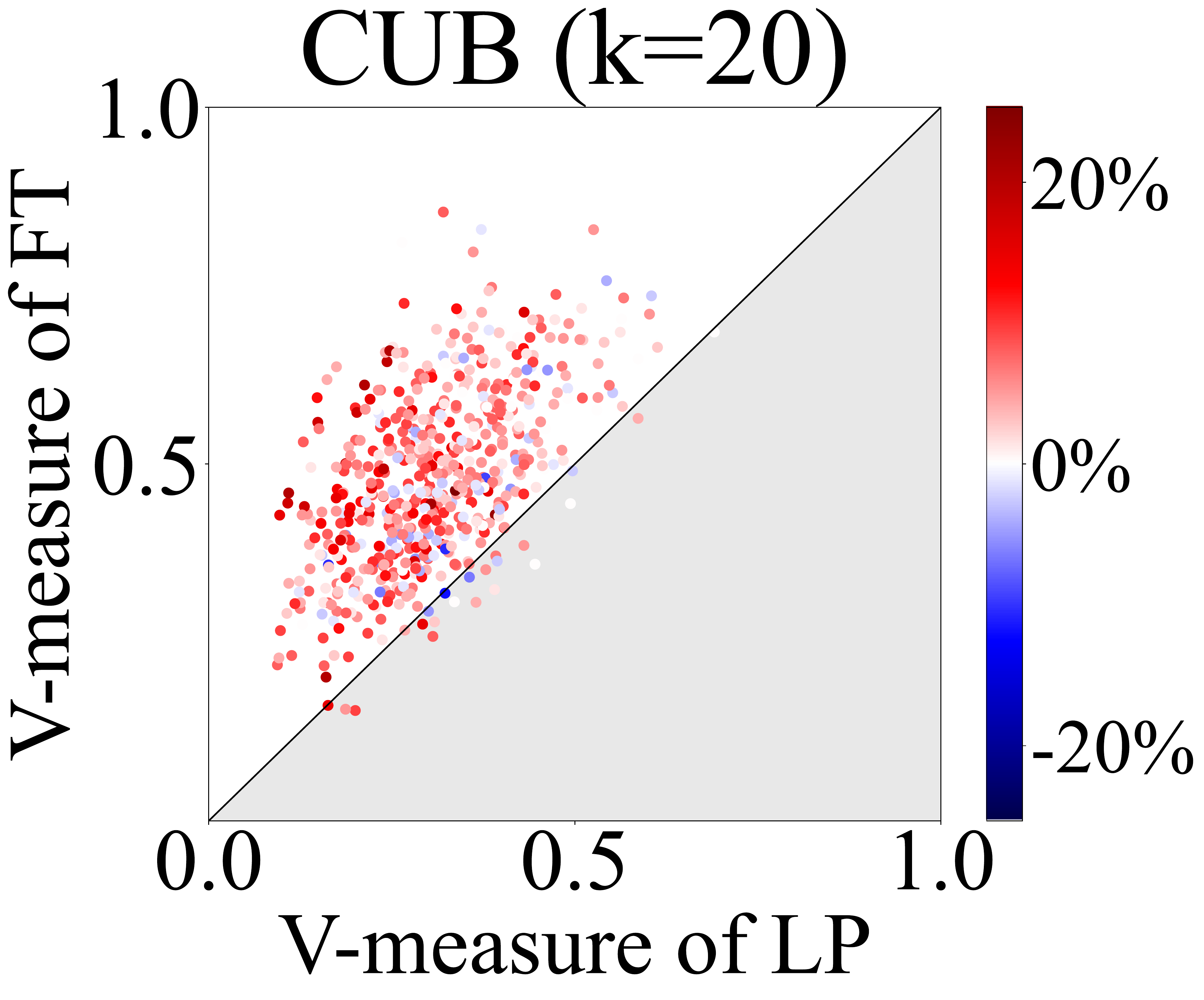}
     \end{subfigure}
     \begin{subfigure}[h]{0.185\linewidth}
         \includegraphics[width=\linewidth]{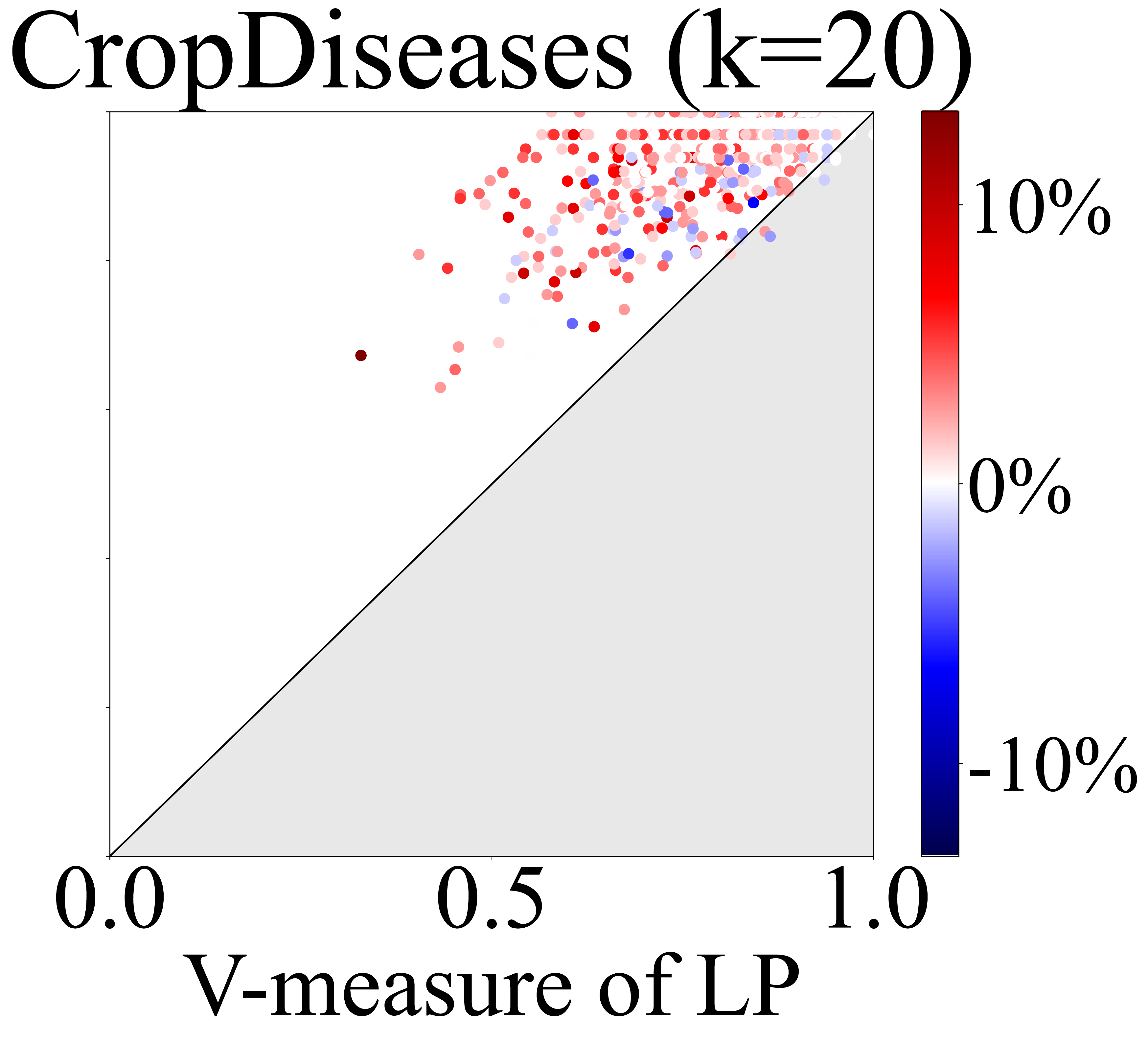}
     \end{subfigure}
     \begin{subfigure}[h]{0.185\linewidth}
         \includegraphics[width=\linewidth]{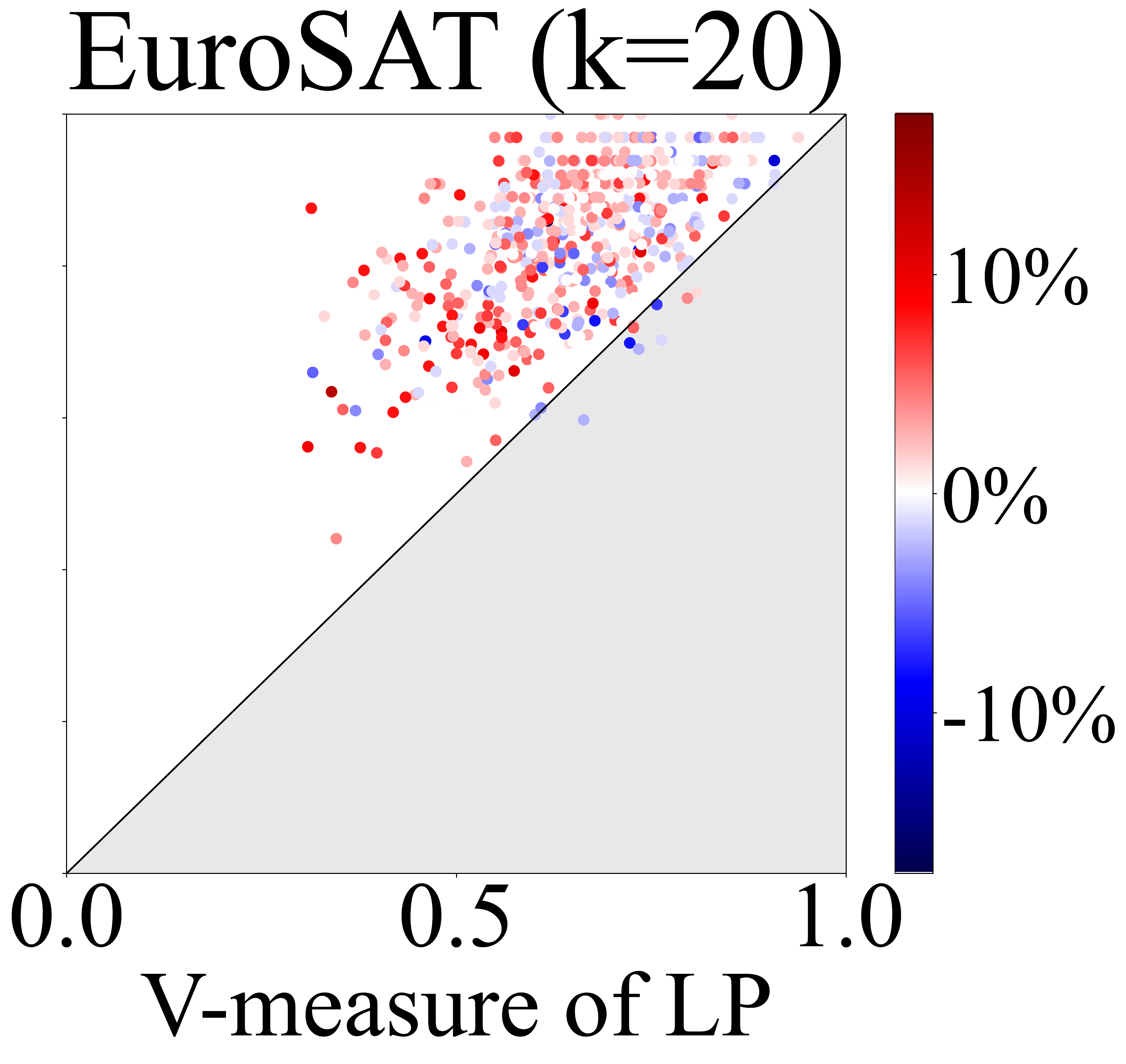}
     \end{subfigure}
     \begin{subfigure}[h]{0.185\linewidth}
         \includegraphics[width=\linewidth]{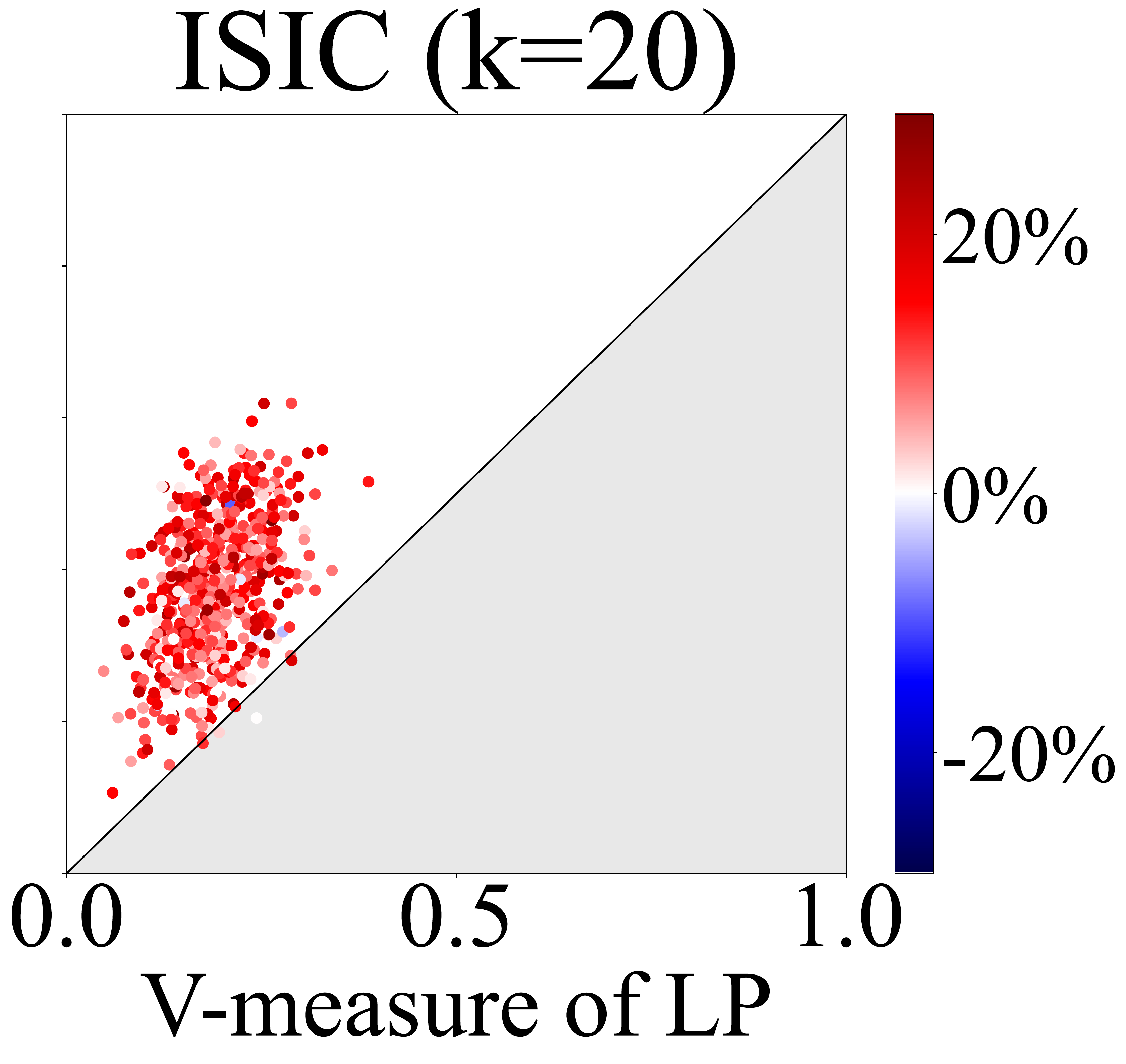}
     \end{subfigure}
     \begin{subfigure}[h]{0.185\linewidth}
         \includegraphics[width=\linewidth]{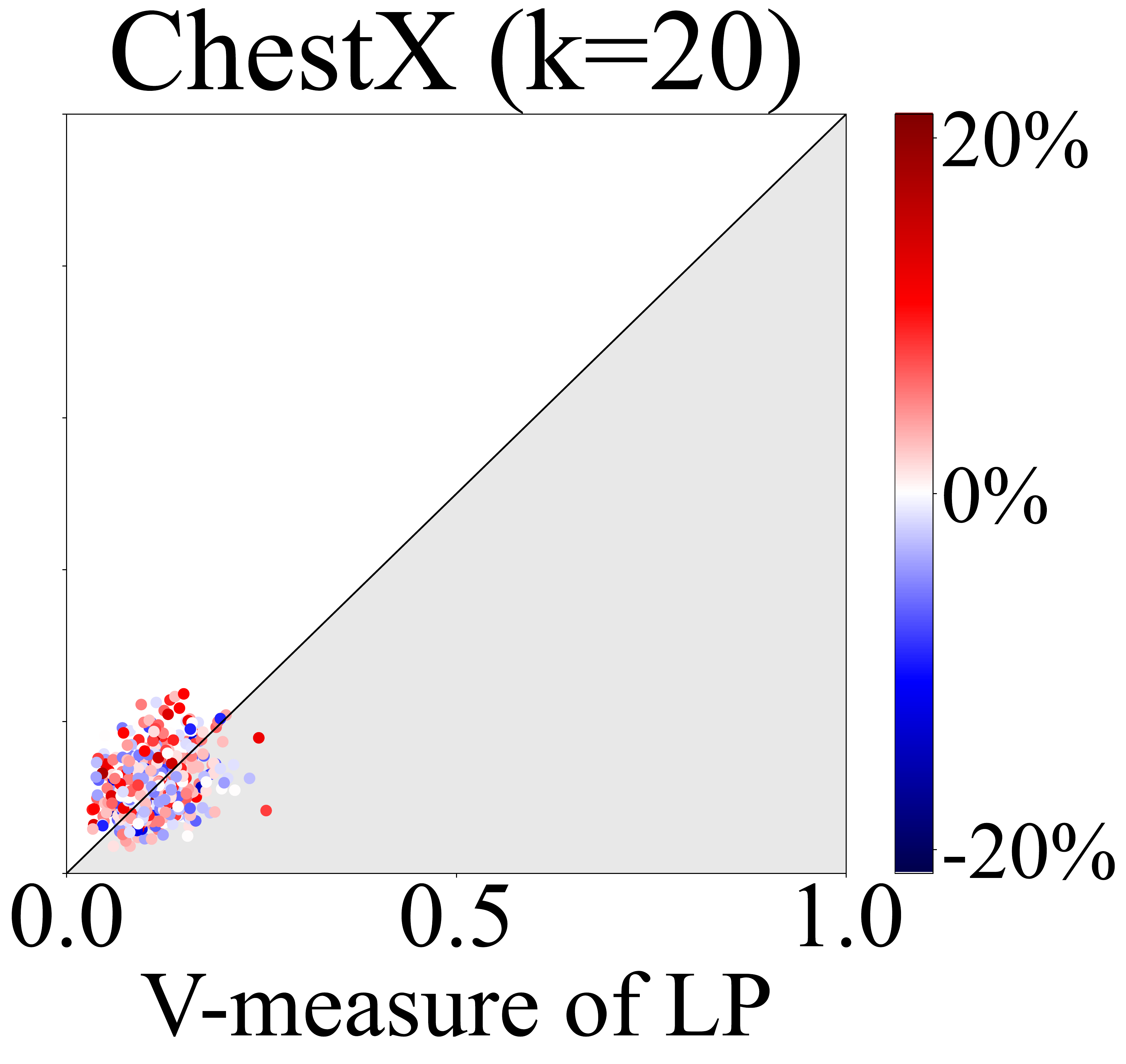}
     \end{subfigure}
     \caption{Feature extractor’s clustering ability by V-measure and performance differences between LP and FT. The pre-trained model is ResNet10 on miniIN.} \label{fig:v_measure_miniIN}
\end{figure}

Figure \ref{fig:v_measure_miniIN} describes the relationship between V-measure and performance when ResNet10 pre-trained on miniIN is updated. The trend is the same for all target datasets.

\section{FT with Data Augmentation}\label{appx:few_shot_da}

\begin{figure}[h]
    \centering

    \begin{subfigure}[h]{0.355\linewidth}
    \includegraphics[width=\linewidth]{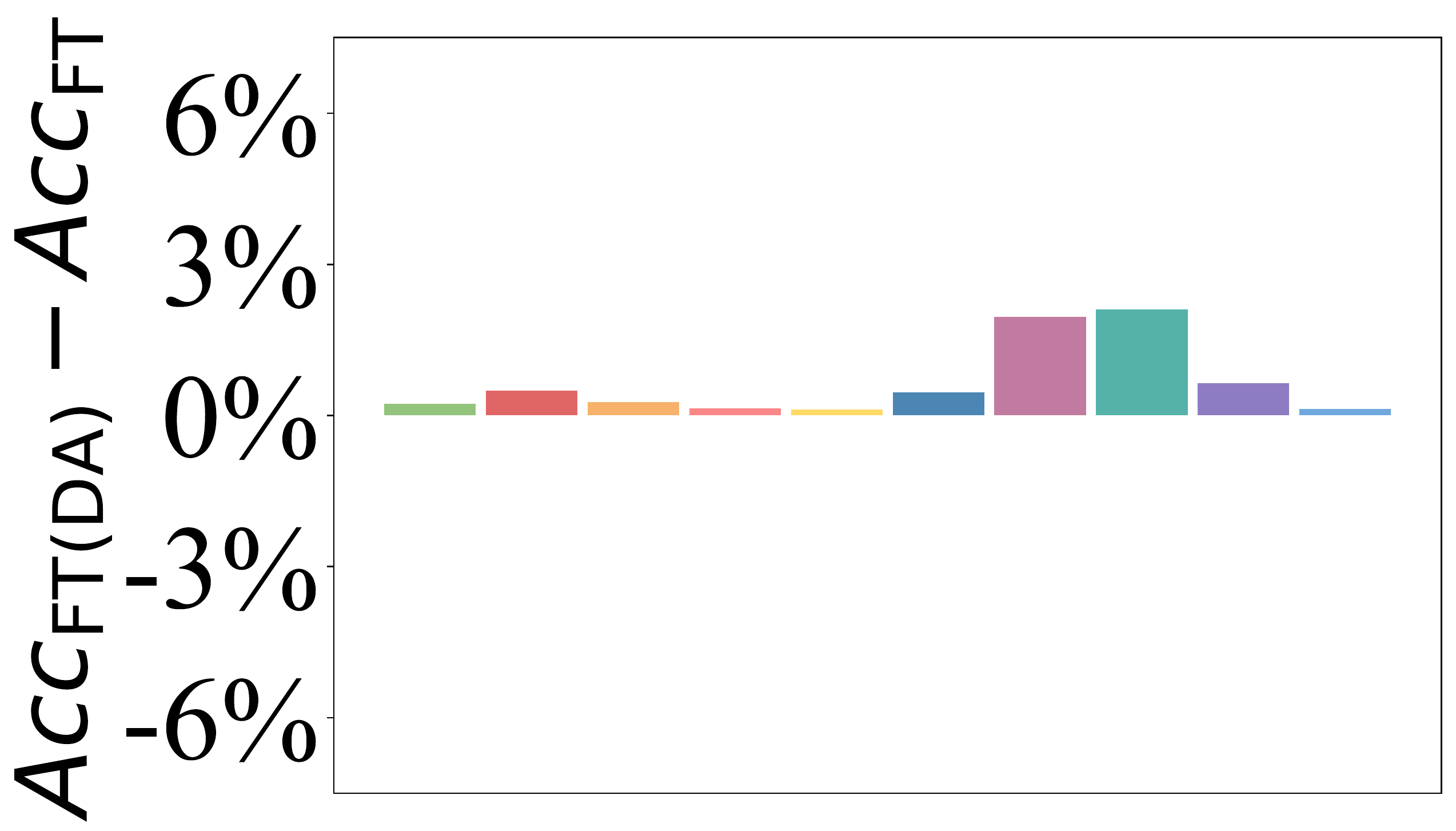}
        \caption{HFlip\,($k$ = 1)}
    \end{subfigure} \hfill
    \begin{subfigure}[h]{0.312\linewidth}
    \includegraphics[width=\linewidth]{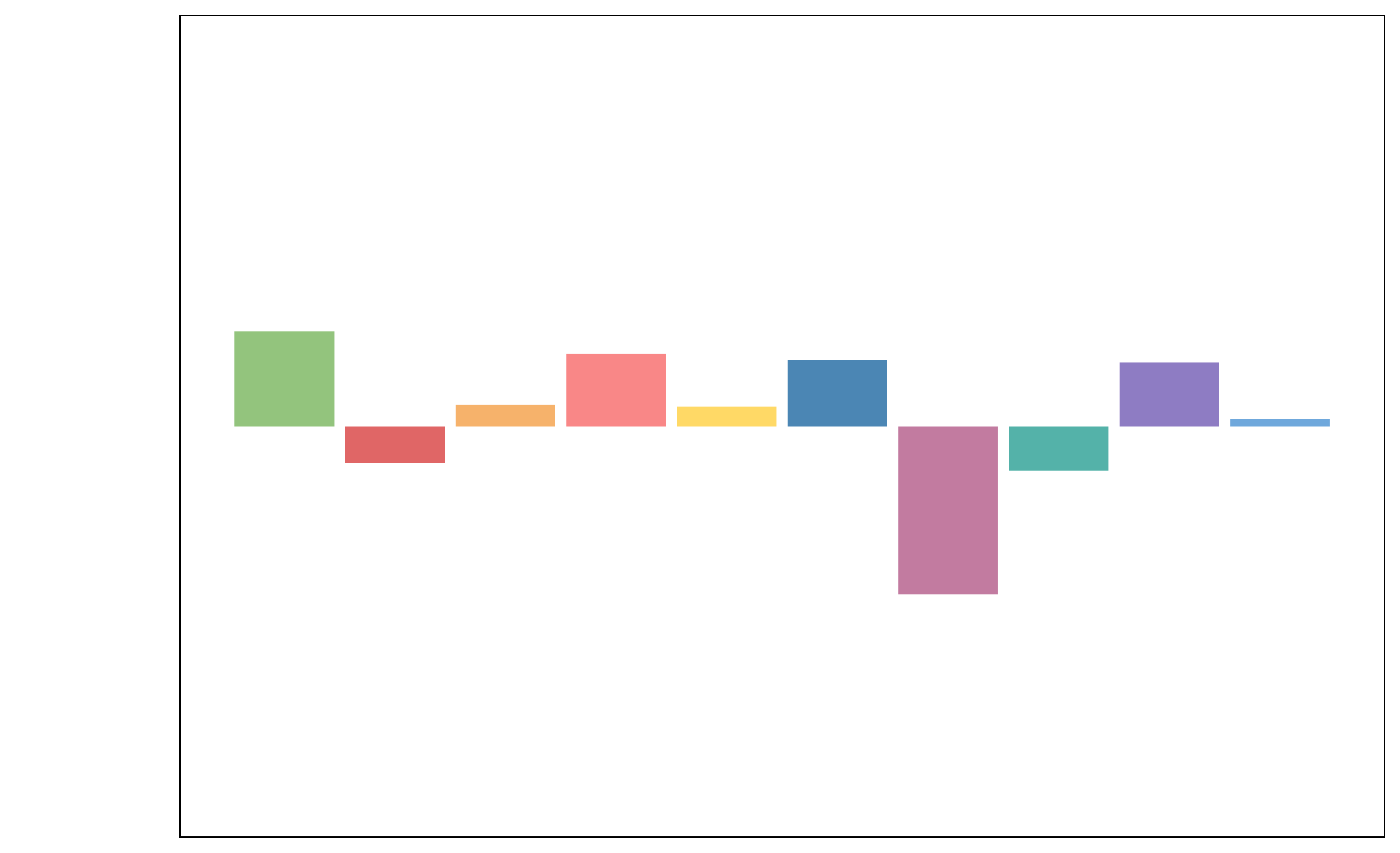}
        \caption{RCrop\,($k$ = 1)}
    \end{subfigure} \hfill
    \begin{subfigure}[h]{0.312\linewidth}
    \includegraphics[width=\linewidth]{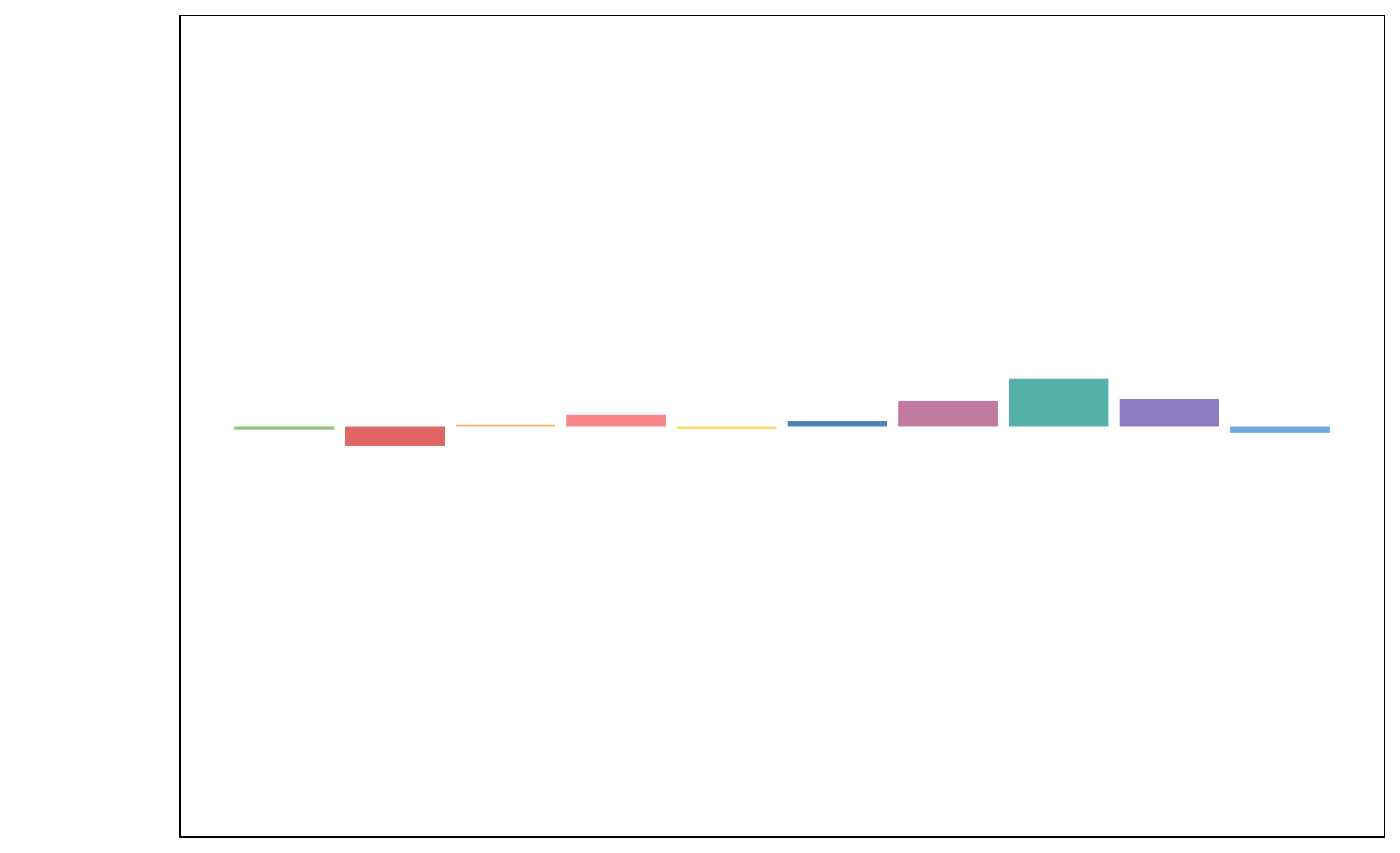}
        \caption{CJitter\,($k$ = 1)}
    \end{subfigure}
    
    \begin{subfigure}[h]{0.355\linewidth}
    \includegraphics[width=\linewidth]{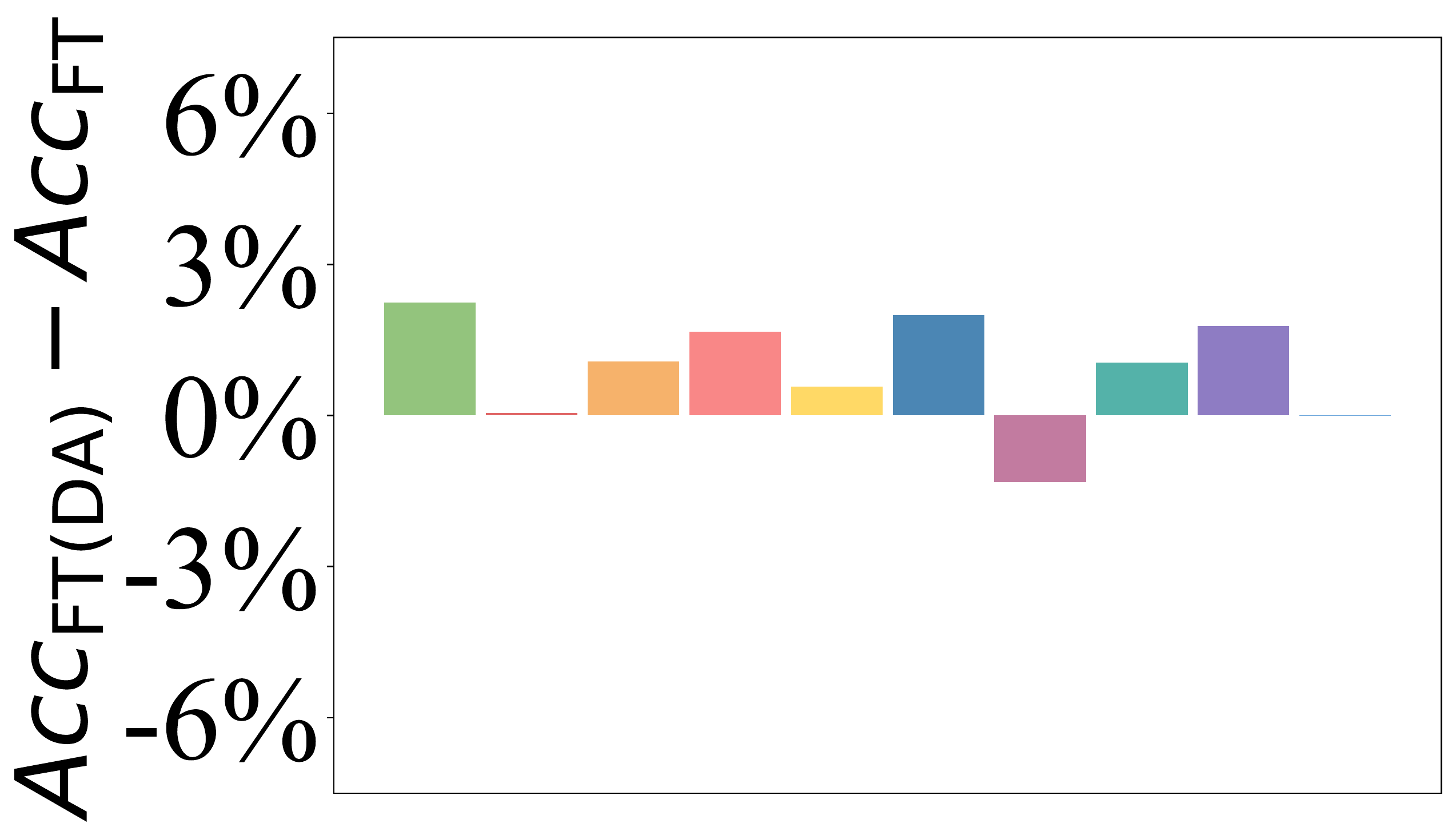}
        \caption{Base Aug\,($k$ = 1)}
    \end{subfigure} \hfill
    \begin{subfigure}[h]{0.312\linewidth}
    \includegraphics[width=\linewidth]{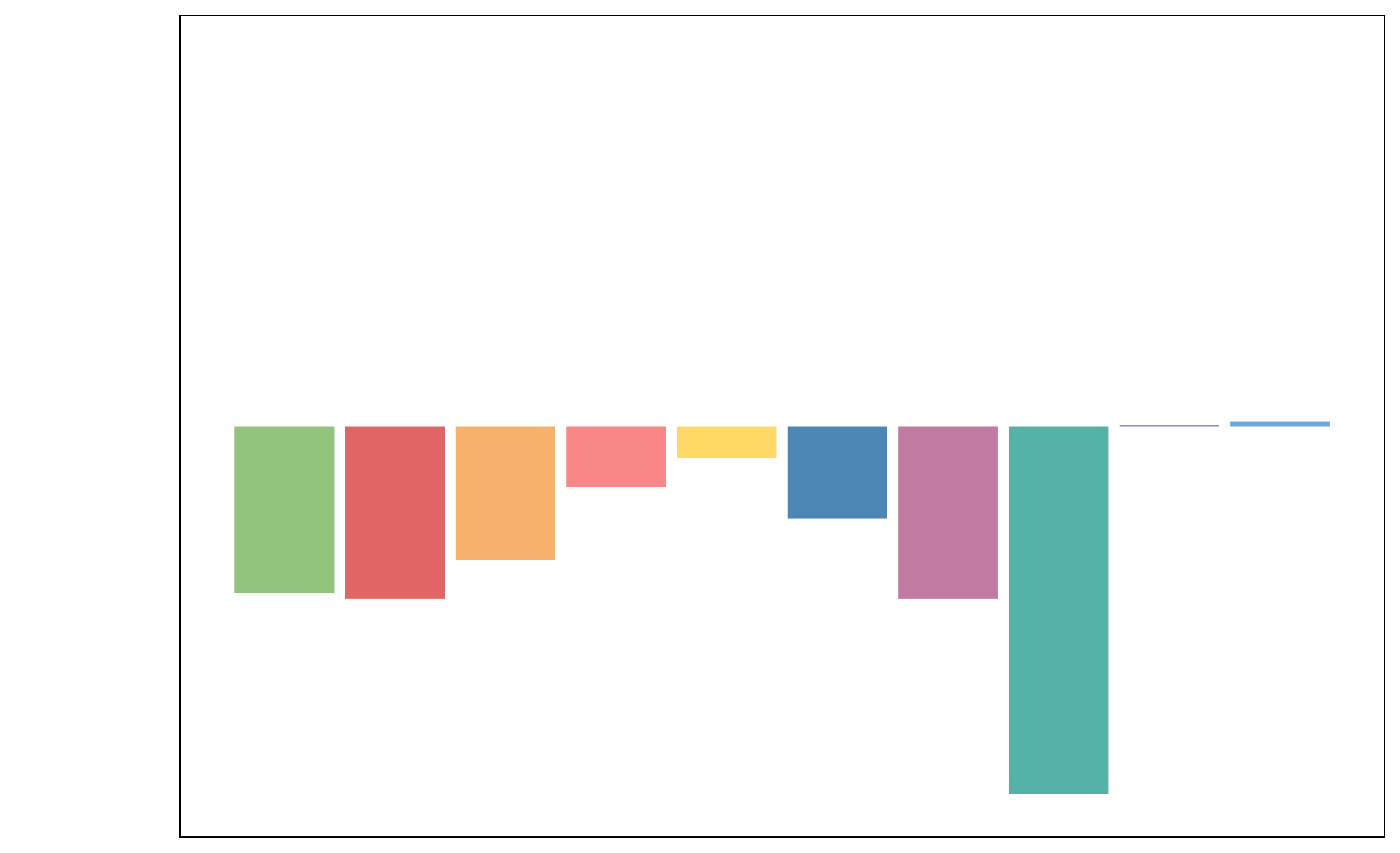}
        \caption{MixUp\,($k$ = 1)}
    \end{subfigure} \hfill
    \begin{subfigure}[h]{0.312\linewidth}
    \includegraphics[width=\linewidth]{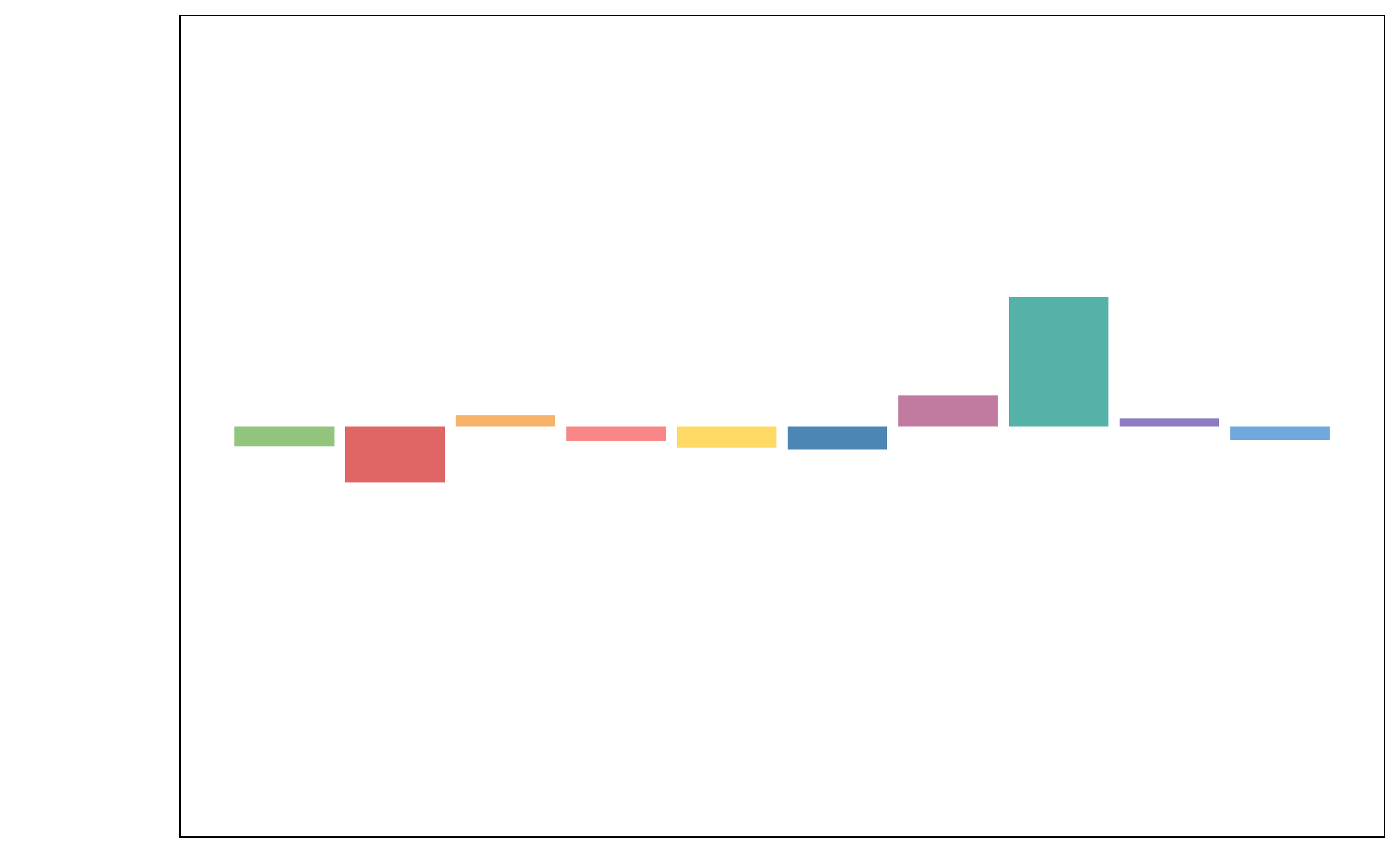}
        \caption{CutMix\,($k$ = 1)}
    \end{subfigure}
    
    \begin{subfigure}[h]{0.355\linewidth}
    \includegraphics[width=\linewidth]{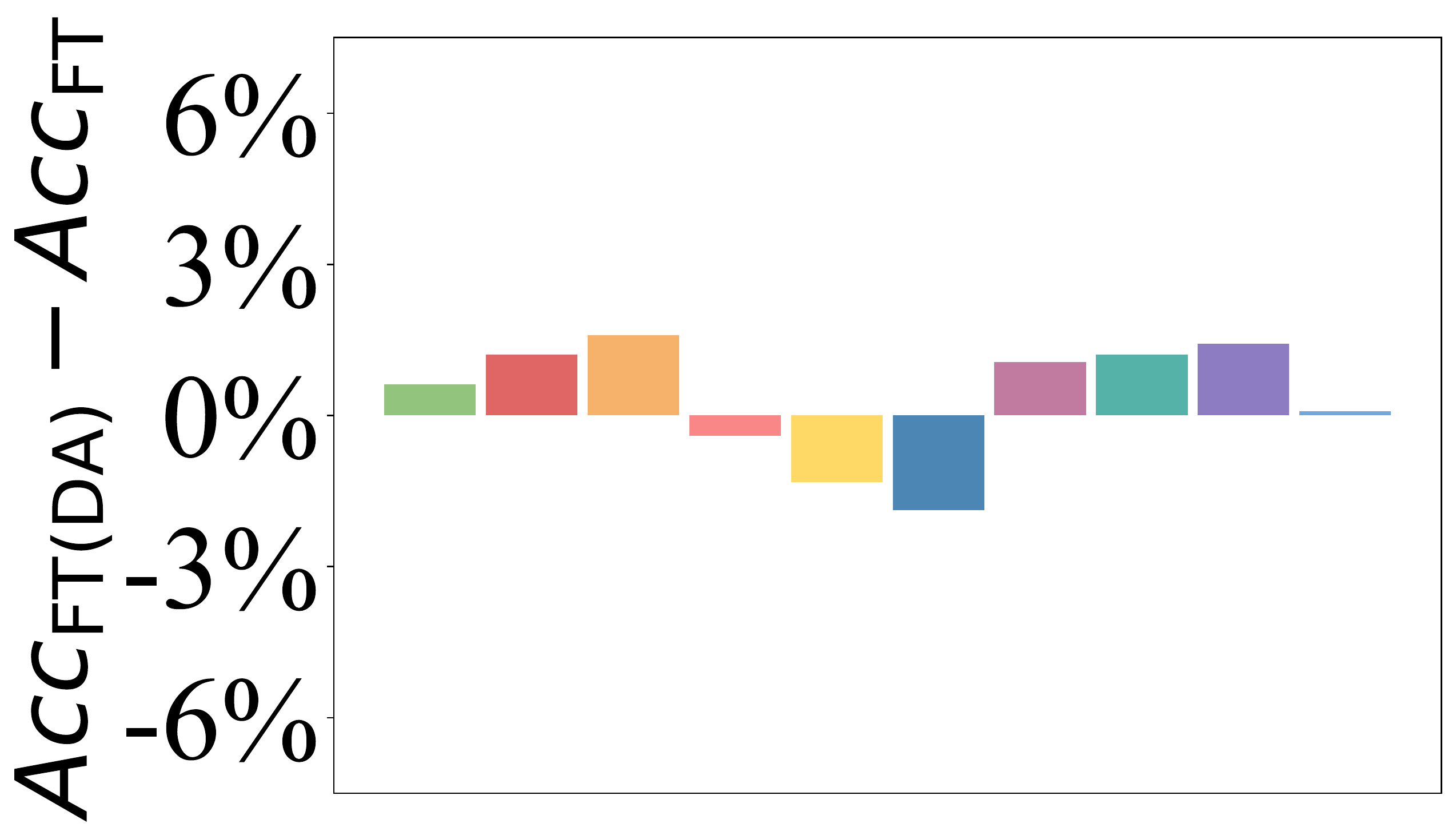}
        \caption{HFlip\,($k$ = 5)}
    \end{subfigure} \hfill
    \begin{subfigure}[h]{0.312\linewidth}
    \includegraphics[width=\linewidth]{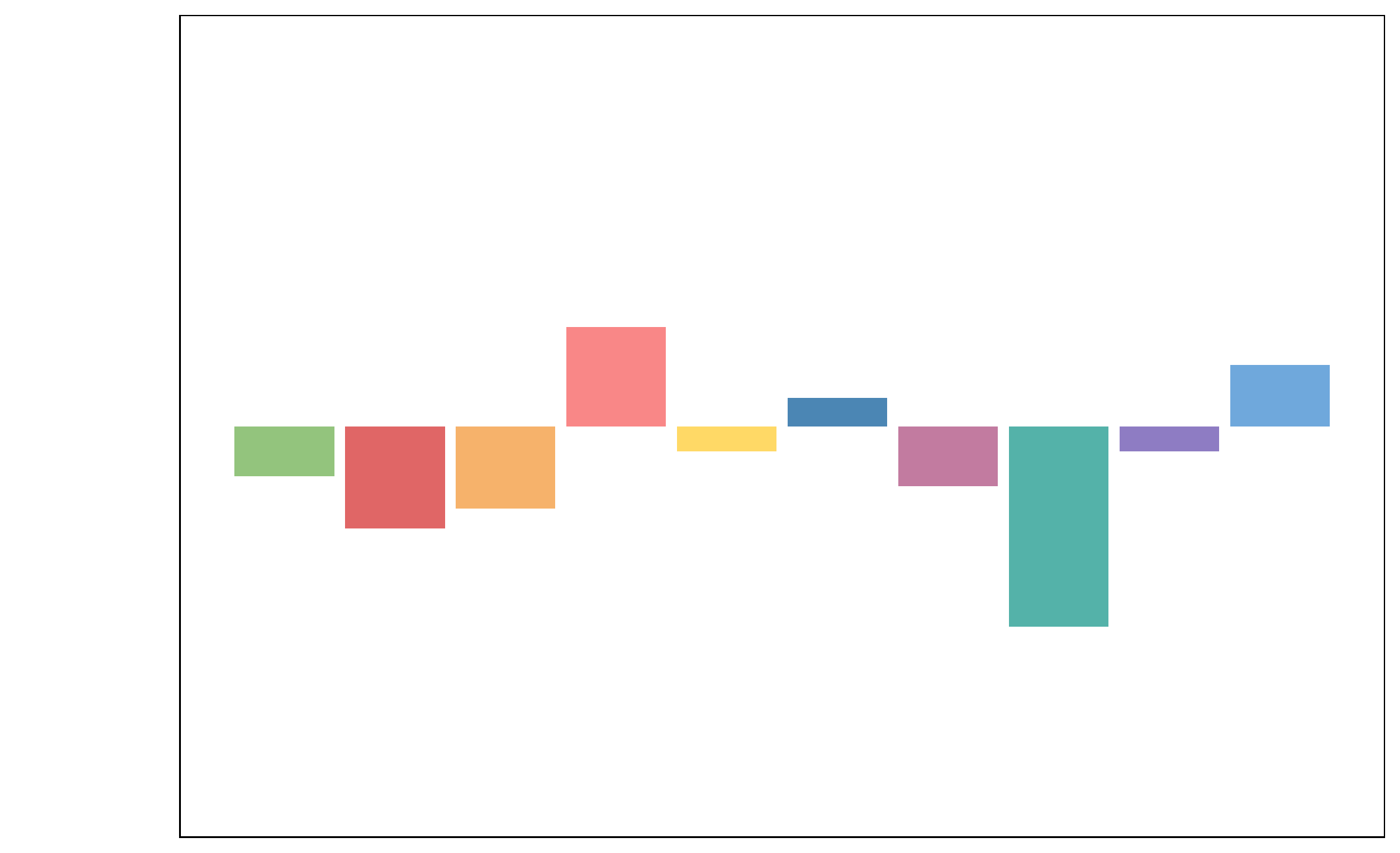}
        \caption{RCrop\,($k$ = 5)}
    \end{subfigure} \hfill
    \begin{subfigure}[h]{0.312\linewidth}
    \includegraphics[width=\linewidth]{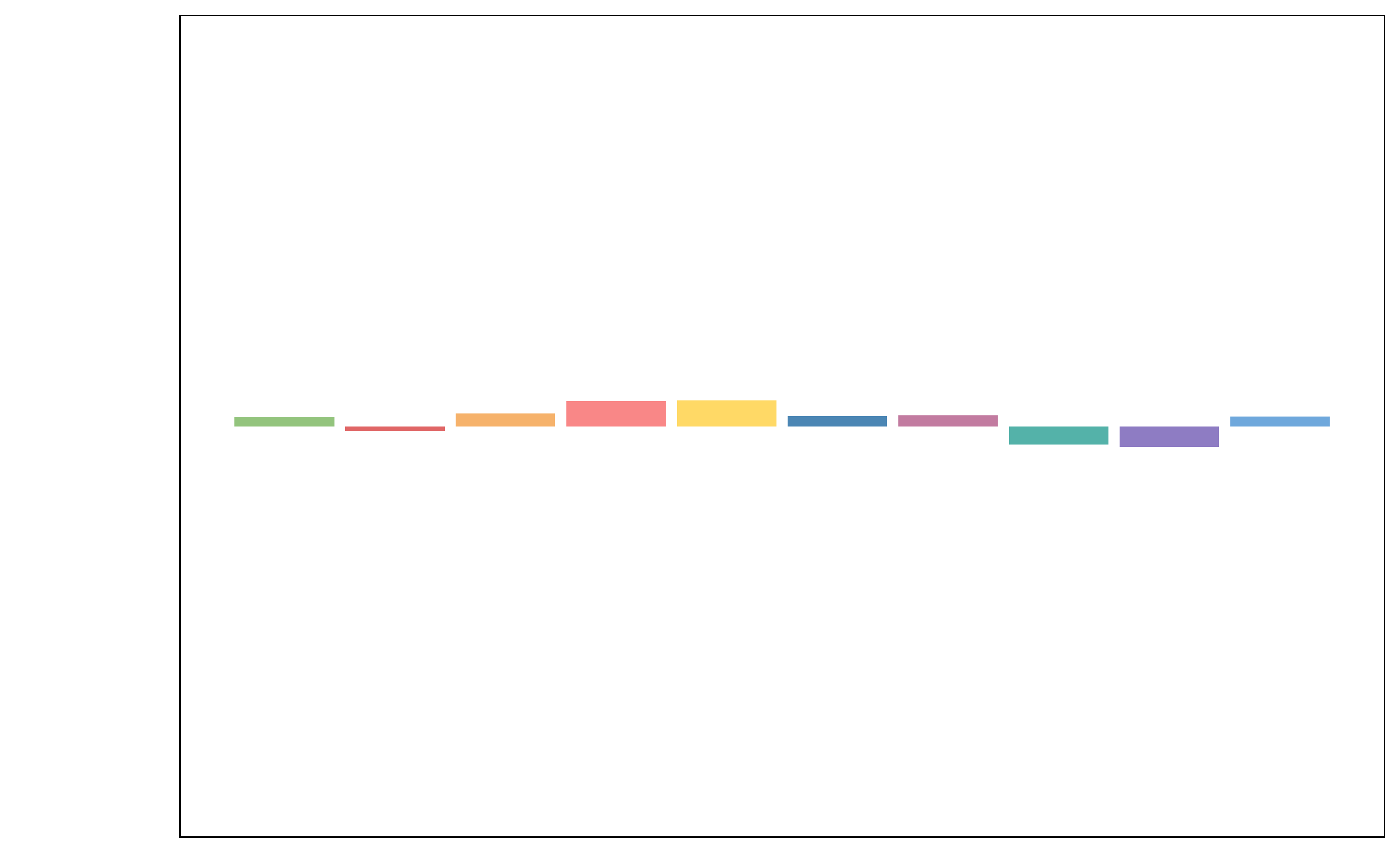}
        \caption{CJitter\,($k$ = 5)}
    \end{subfigure}
    
    \begin{subfigure}[h]{0.355\linewidth}
    \includegraphics[width=\linewidth]{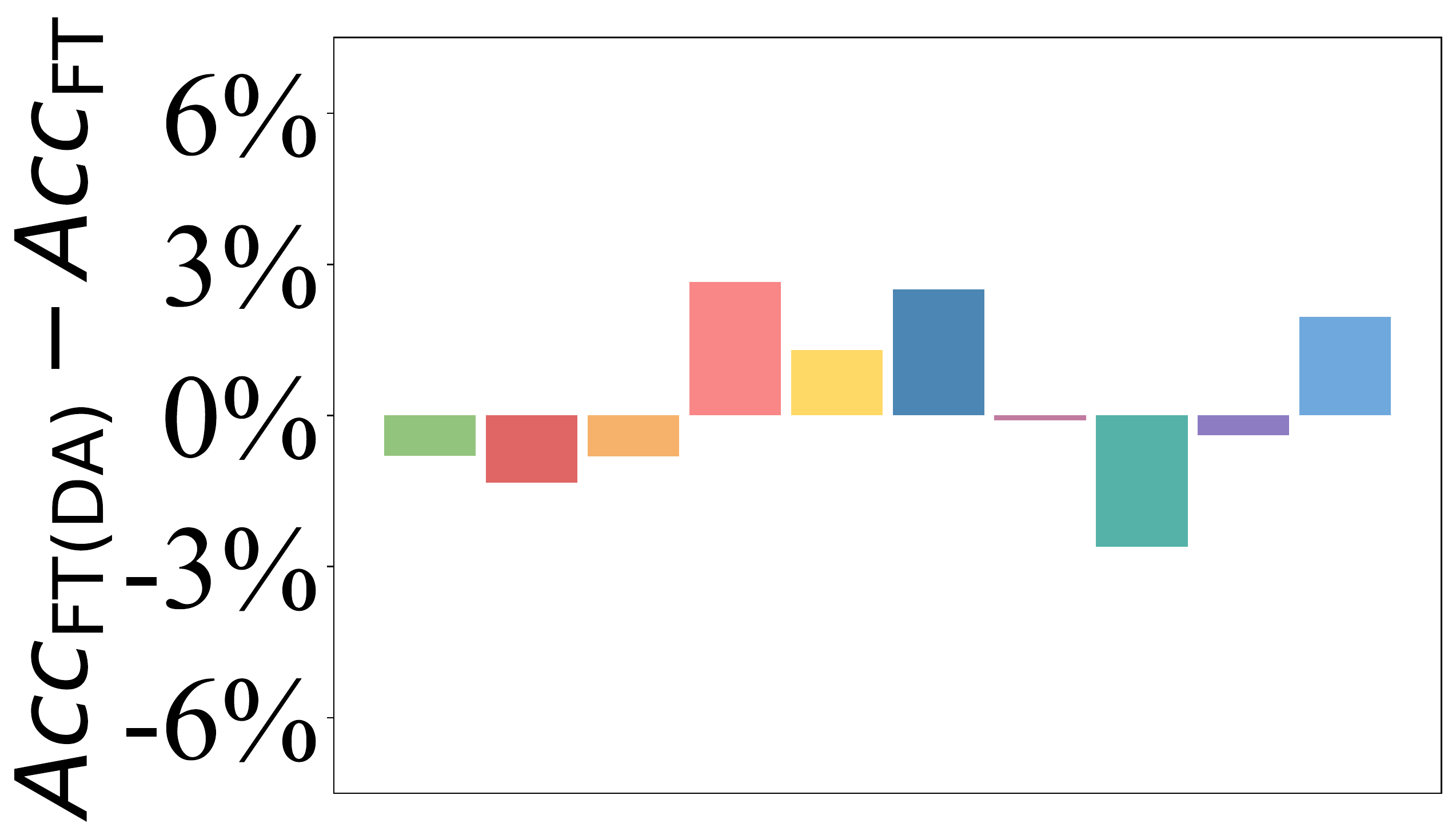}
        \caption{Base Aug\,($k$ = 5)}
    \end{subfigure} \hfill
    \begin{subfigure}[h]{0.312\linewidth}
    \includegraphics[width=\linewidth]{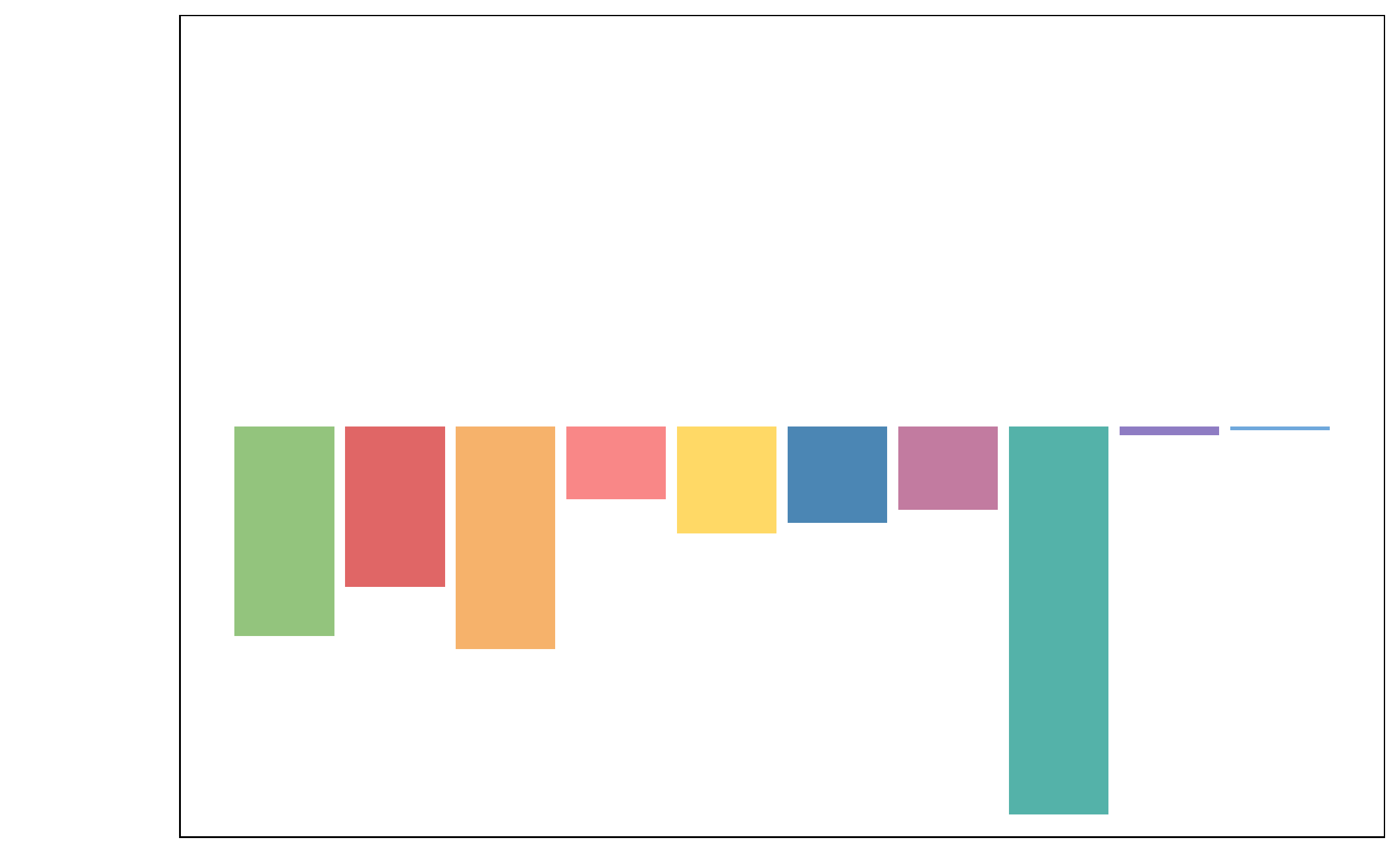}
        \caption{MixUp\,($k$ = 5)}
    \end{subfigure} \hfill
    \begin{subfigure}[h]{0.312\linewidth}
    \includegraphics[width=\linewidth]{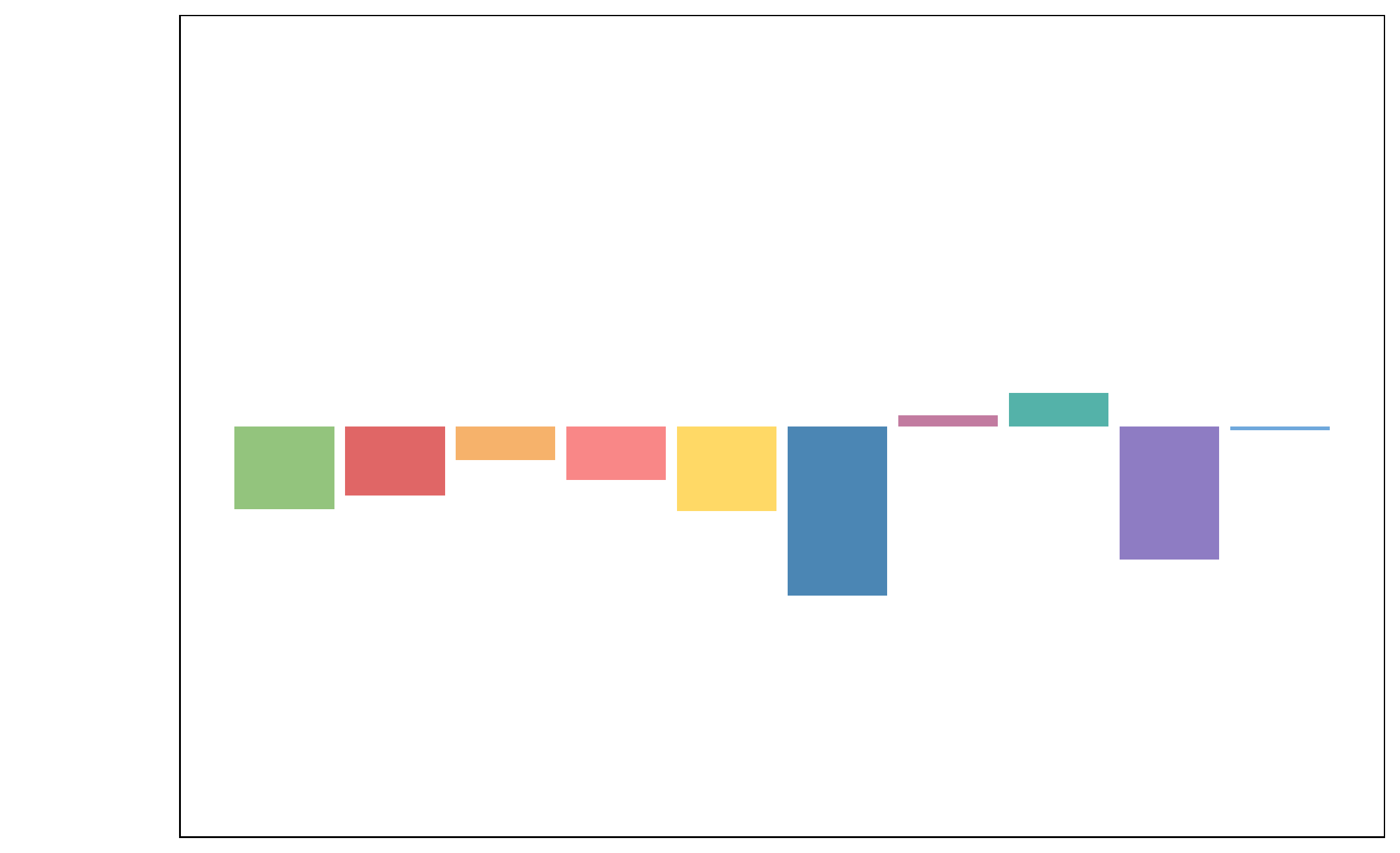}
        \caption{CutMix\,($k$ = 5)}
    \end{subfigure}
    
    \caption{5-way $k$-shot performance difference of FT(DA) with FT according to the augmentation techniques.}
    \label{fig:miniIN_da}
\end{figure}

Figure \ref{fig:miniIN_da} is visualized using the difference between FT(DA) and FT according to different augmentation techniques for $k=1$ and $k=5$.

\section{Intensity of Single Augmentation}\label{appx:single_aug}

\begin{figure}[h!]
    \centering
    \includegraphics[width=\linewidth]{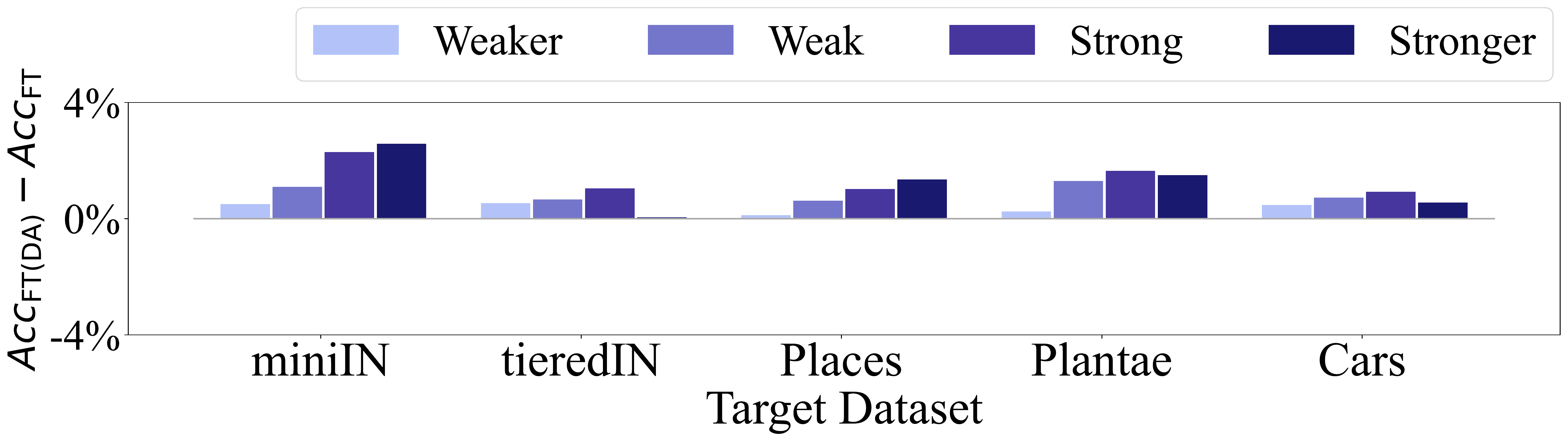}
    \includegraphics[width=\linewidth]{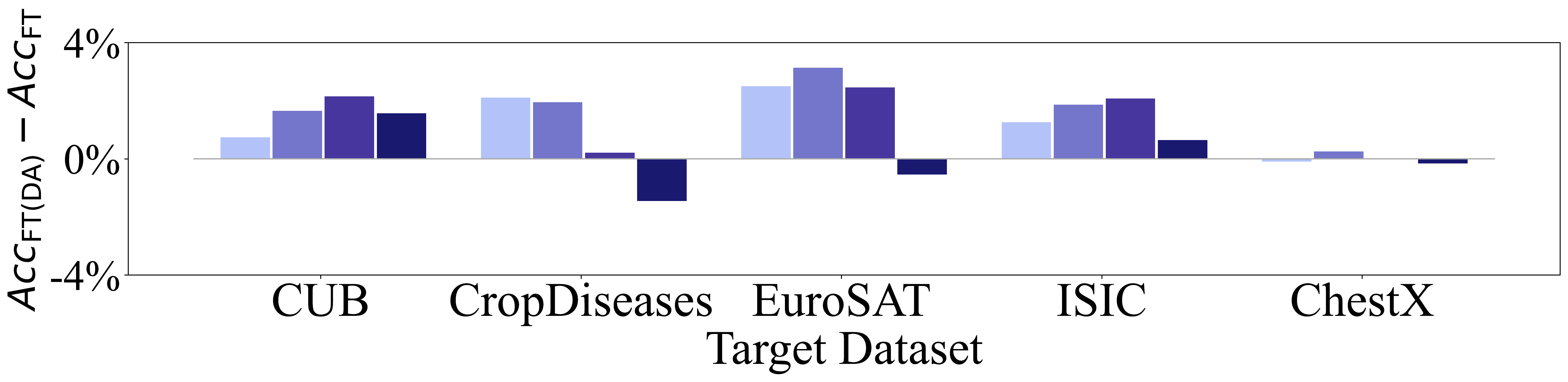}
    \caption{5-way 1-shot performance difference of FT(DA) with FT according to the designed intensity of Base Aug.}
    \label{fig:base_magnitude_1shot}
\end{figure}

\begin{figure}[h!]
    \centering
    \includegraphics[width=\linewidth]{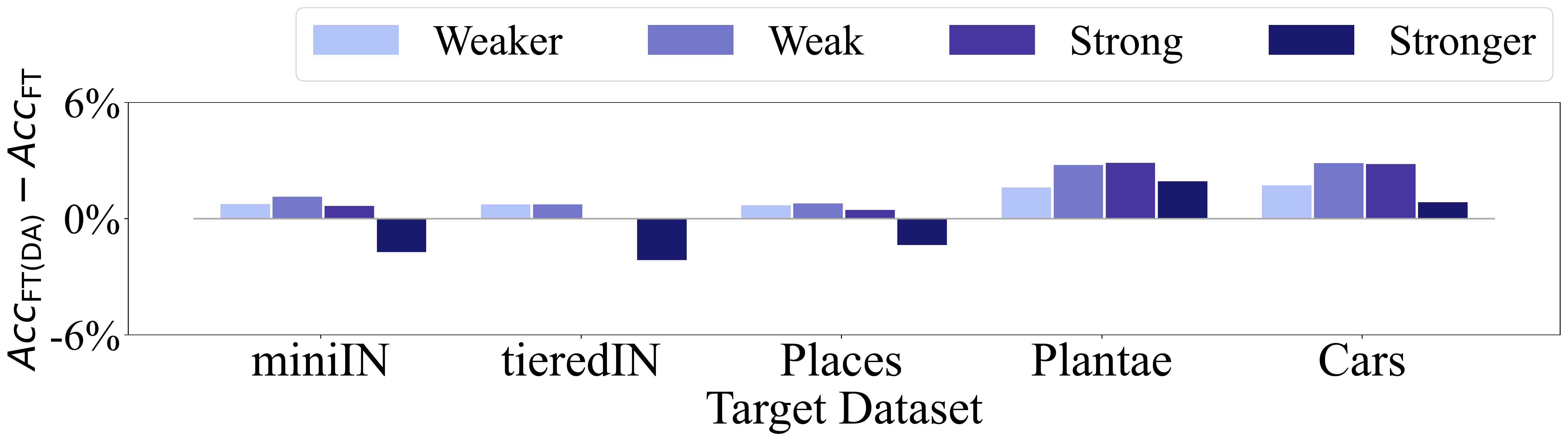}
    \includegraphics[width=\linewidth]{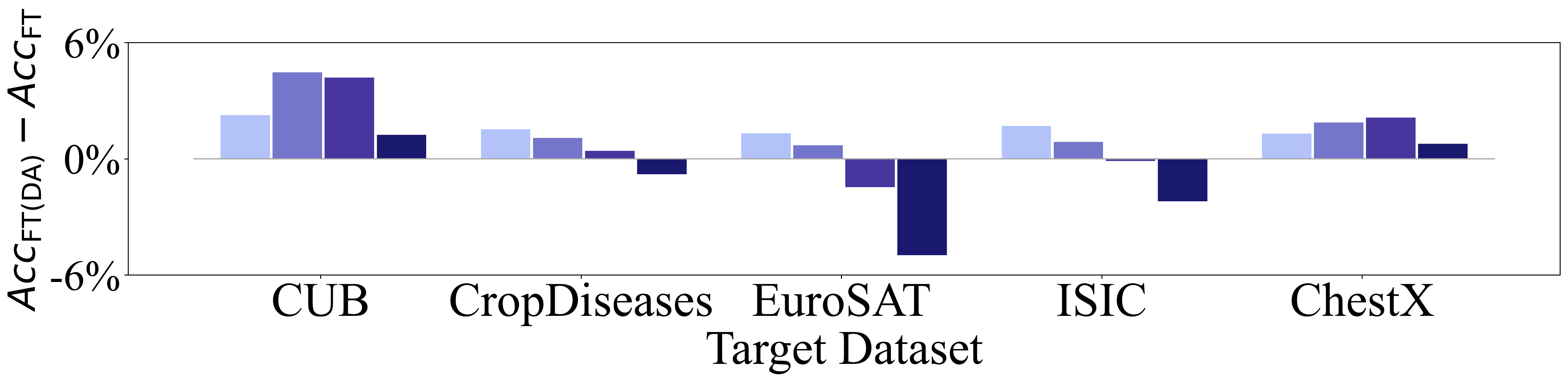}
    \caption{5-way 5-shot performance difference of FT(DA) with FT according to the designed intensity of Base Aug.}
    \label{fig:base_magnitude_5shot}
\end{figure}

Figure \ref{fig:base_magnitude_1shot} and \ref{fig:base_magnitude_5shot} are visualized using the difference between FT(DA) and FT according to the designed intensities for $k$=1 and $k$=5, respectively. It is still observed that the performance increases then decreases based on the best intensity. We further show that the best intensity can vary depending on shot. For instance, miniIN has \textsc{stronger} as the best intensity for $k=1$, while it is the worst for $k=5$.

\section{LP with Data Augmentation}\label{appx:LP_DA}
We provide the results of LP on the analysis of Section \ref{sec:analysis2}. The following are the results of BSCD-FSL datasets as  target data. 

\subsection{Performance Degradation}

\begin{figure}[h!]
    \centering
    \includegraphics[width=\linewidth]{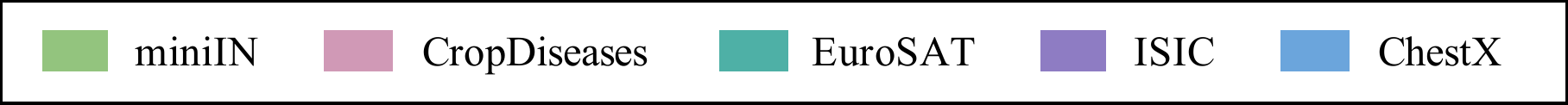}
    
    \begin{subfigure}[h]{0.355\linewidth}
    \includegraphics[width=\linewidth]{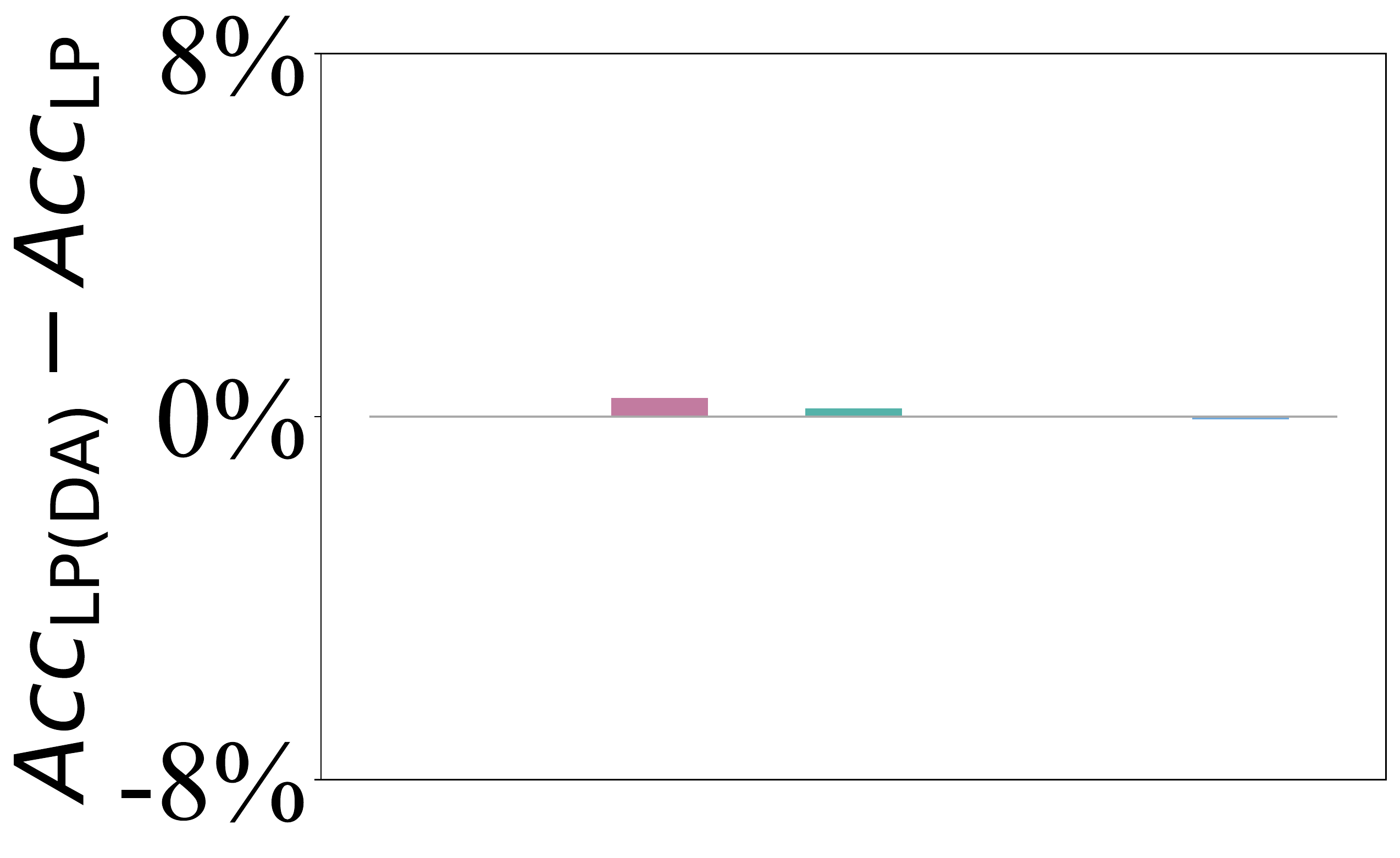}
        \caption{HFlip}
    \end{subfigure} \hfill
    \begin{subfigure}[h]{0.312\linewidth}
    \includegraphics[width=\linewidth]{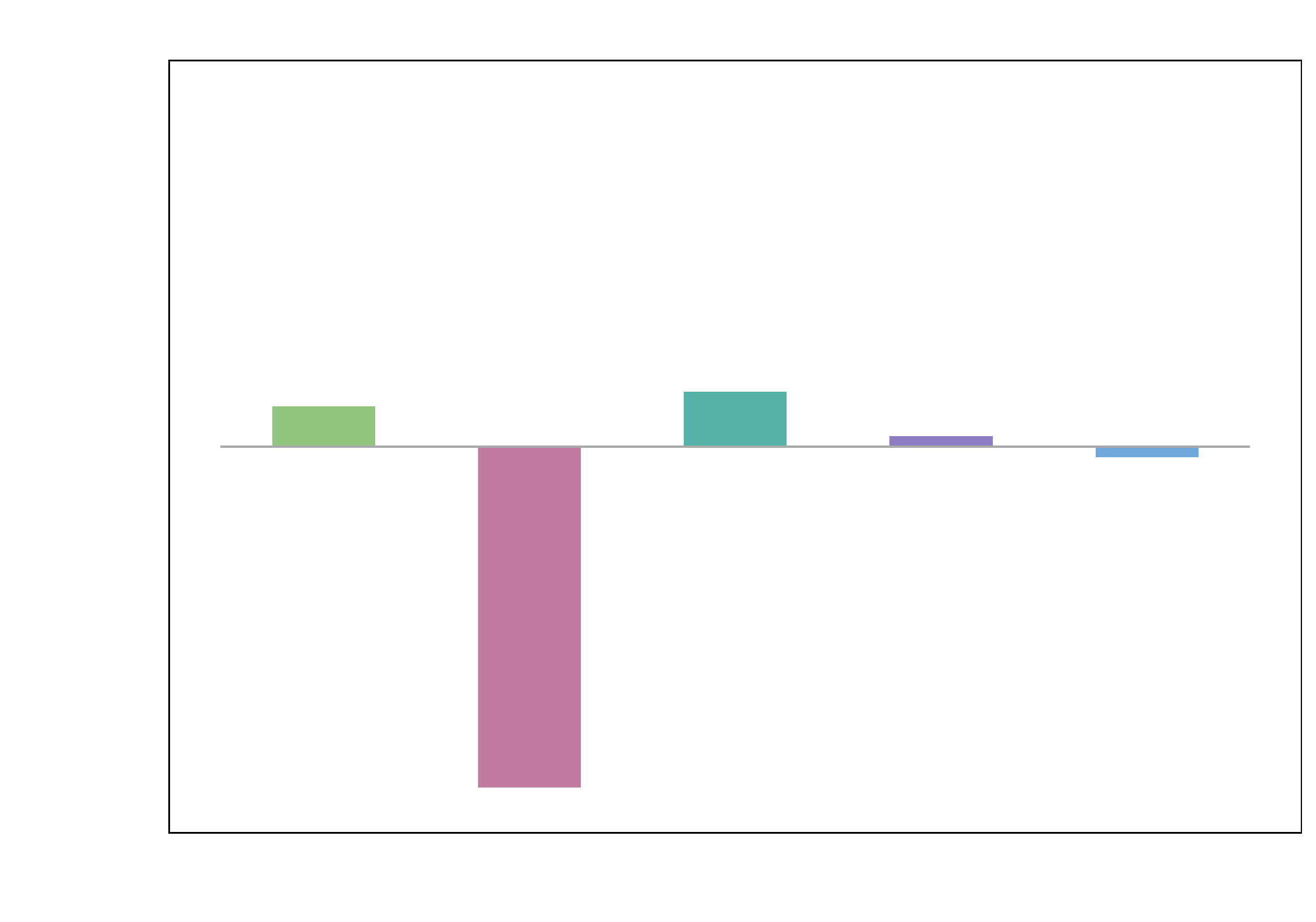}
        \caption{RCrop}
    \end{subfigure} \hfill
    \begin{subfigure}[h]{0.312\linewidth}
    \includegraphics[width=\linewidth]{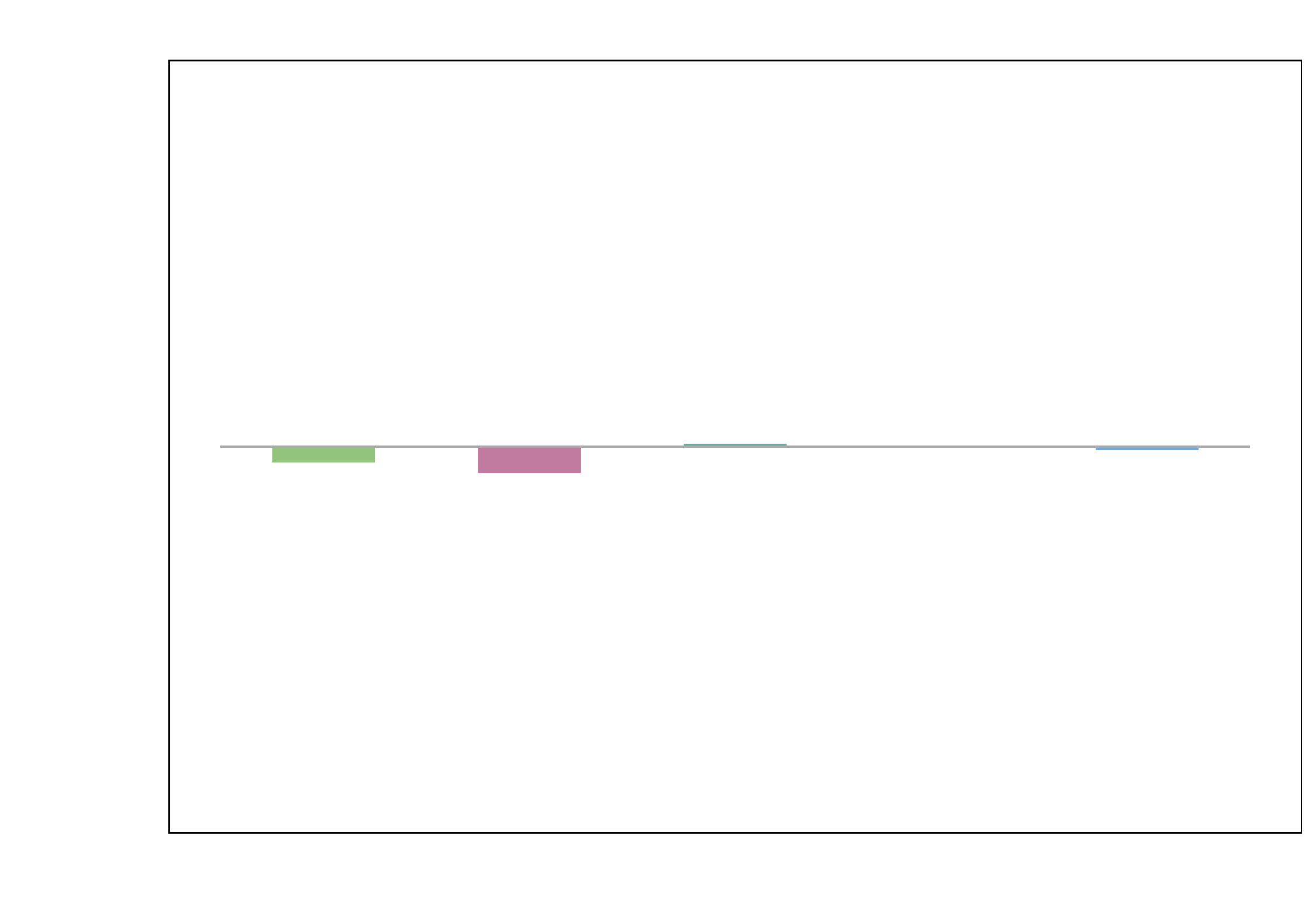}
        \caption{CJitter}
    \end{subfigure}
    
    \begin{subfigure}[h]{0.355\linewidth}
    \includegraphics[width=\linewidth]{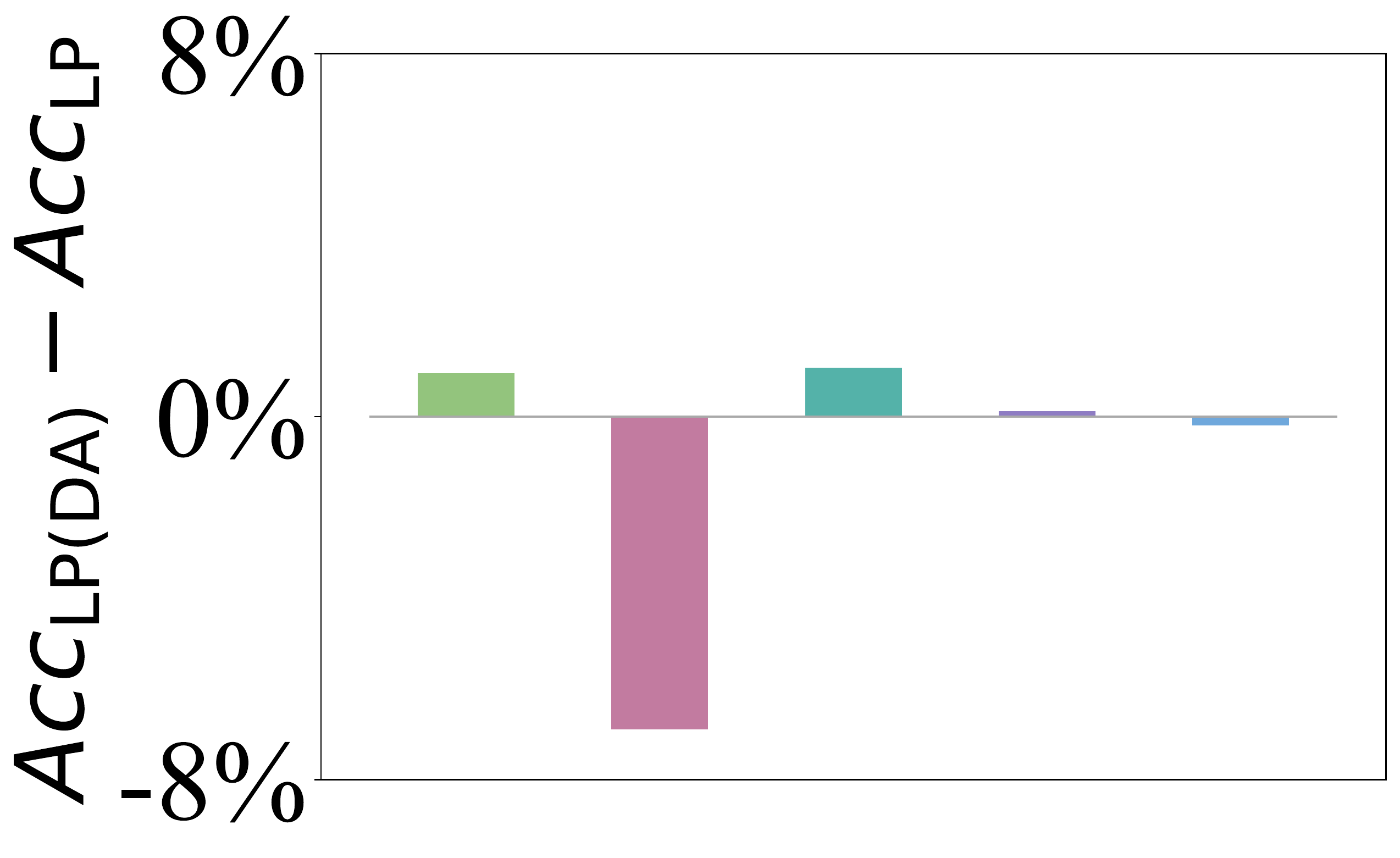}
        \caption{Base Aug}
    \end{subfigure} \hfill
    \begin{subfigure}[h]{0.312\linewidth}
    \includegraphics[width=\linewidth]{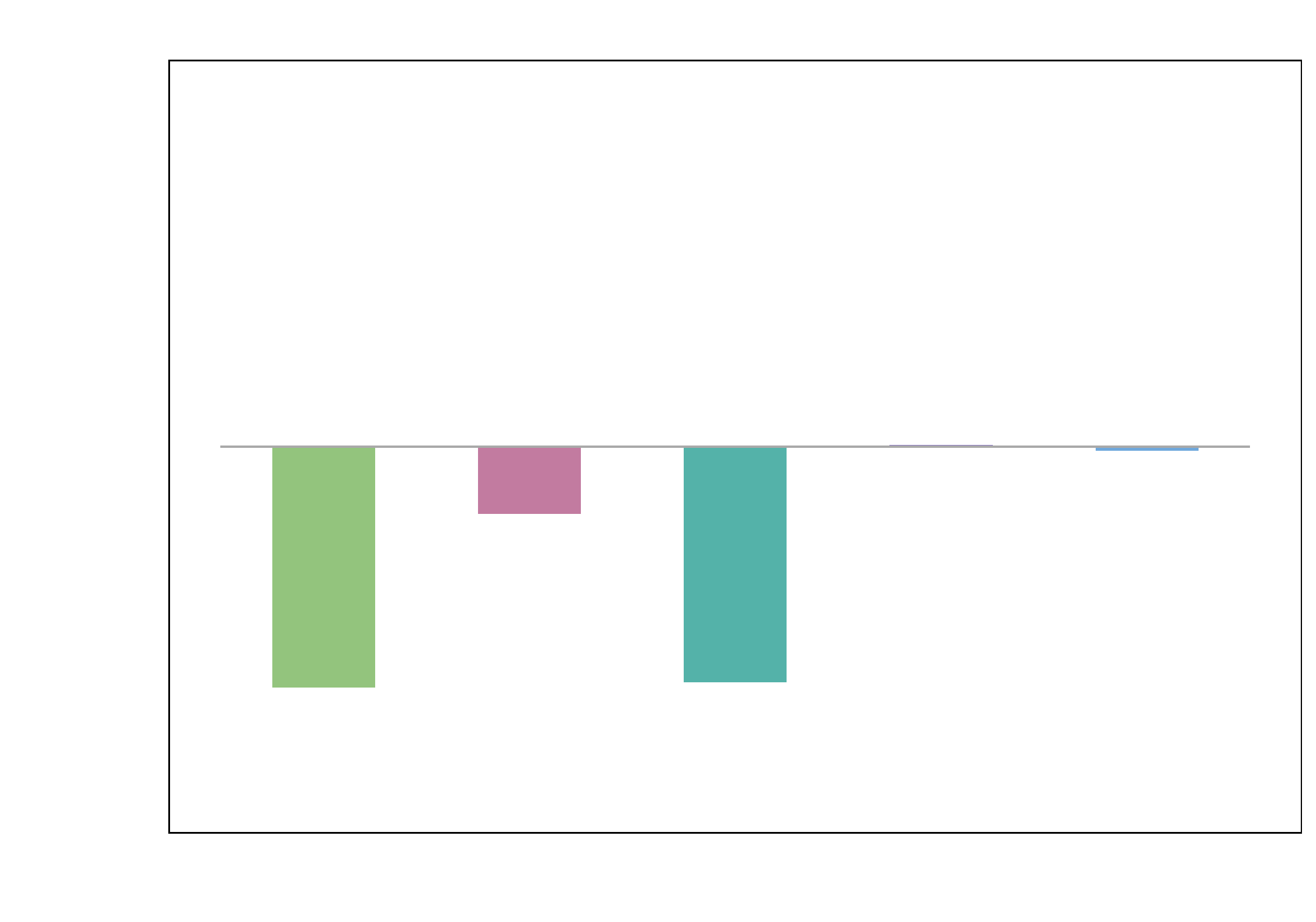}
        \caption{MixUp}
    \end{subfigure} \hfill
    \begin{subfigure}[h]{0.312\linewidth}
    \includegraphics[width=\linewidth]{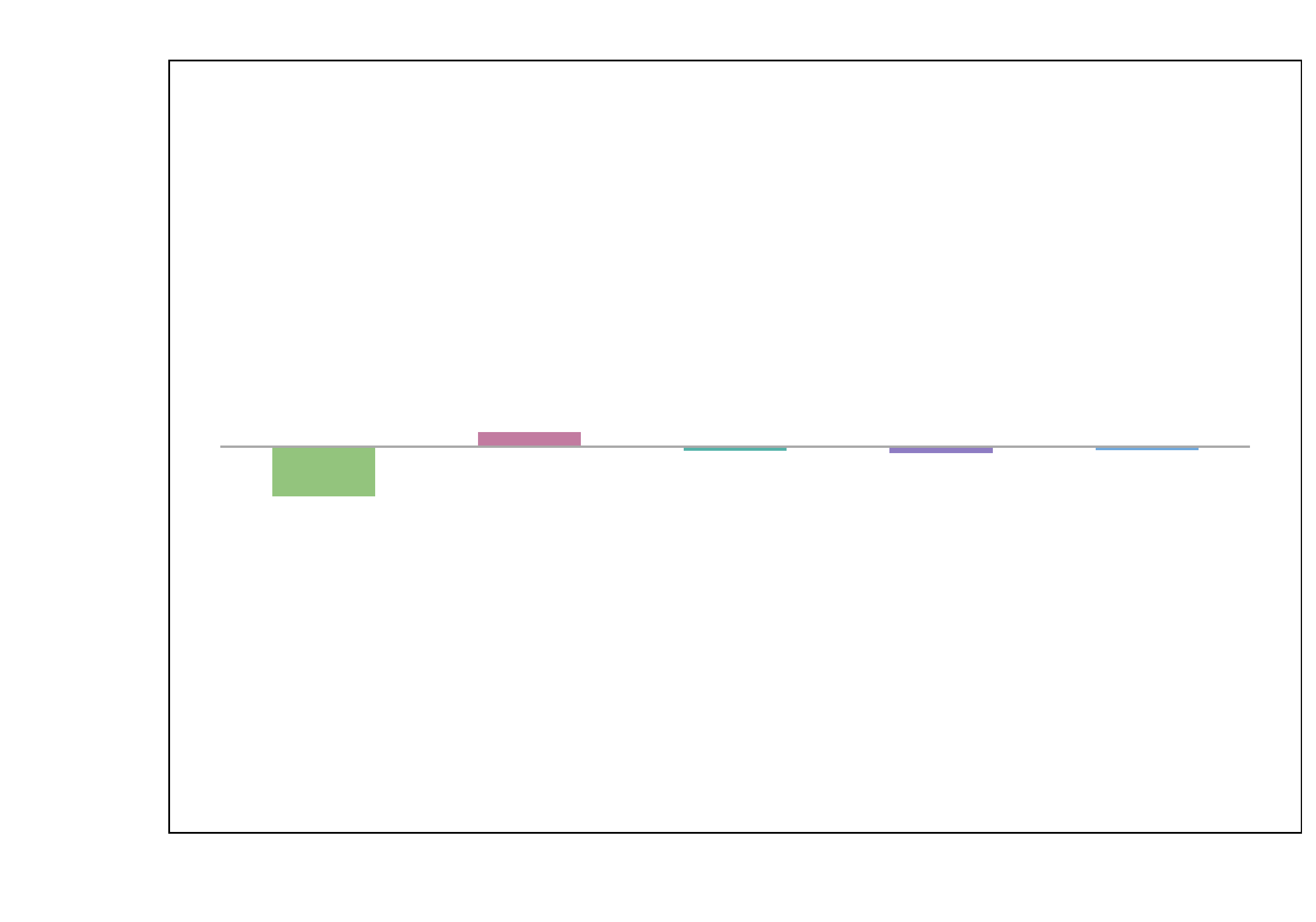}
        \caption{CutMix}
    \end{subfigure}
    \caption{5-way 1-shot performance difference of LP(Aug) with LP according to the augmentation techniques.}
    \label{fig:miniIN_da_lp_1shot}
\end{figure}

\begin{figure}[h!]
    \centering   
    \begin{subfigure}[h]{0.355\linewidth}
    \includegraphics[width=\linewidth]{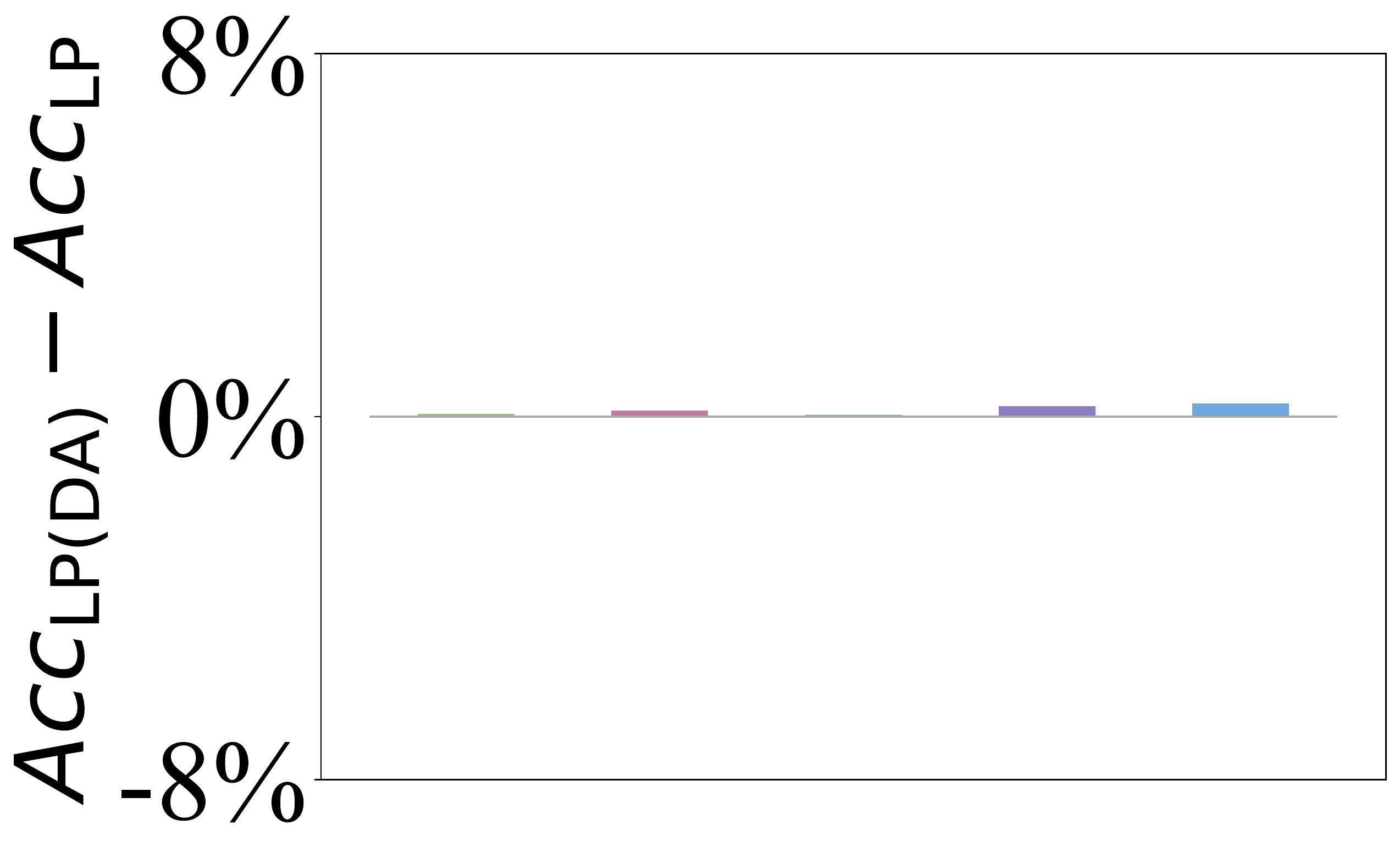}
        \caption{HFlip}
    \end{subfigure} \hfill
    \begin{subfigure}[h]{0.312\linewidth}
    \includegraphics[width=\linewidth]{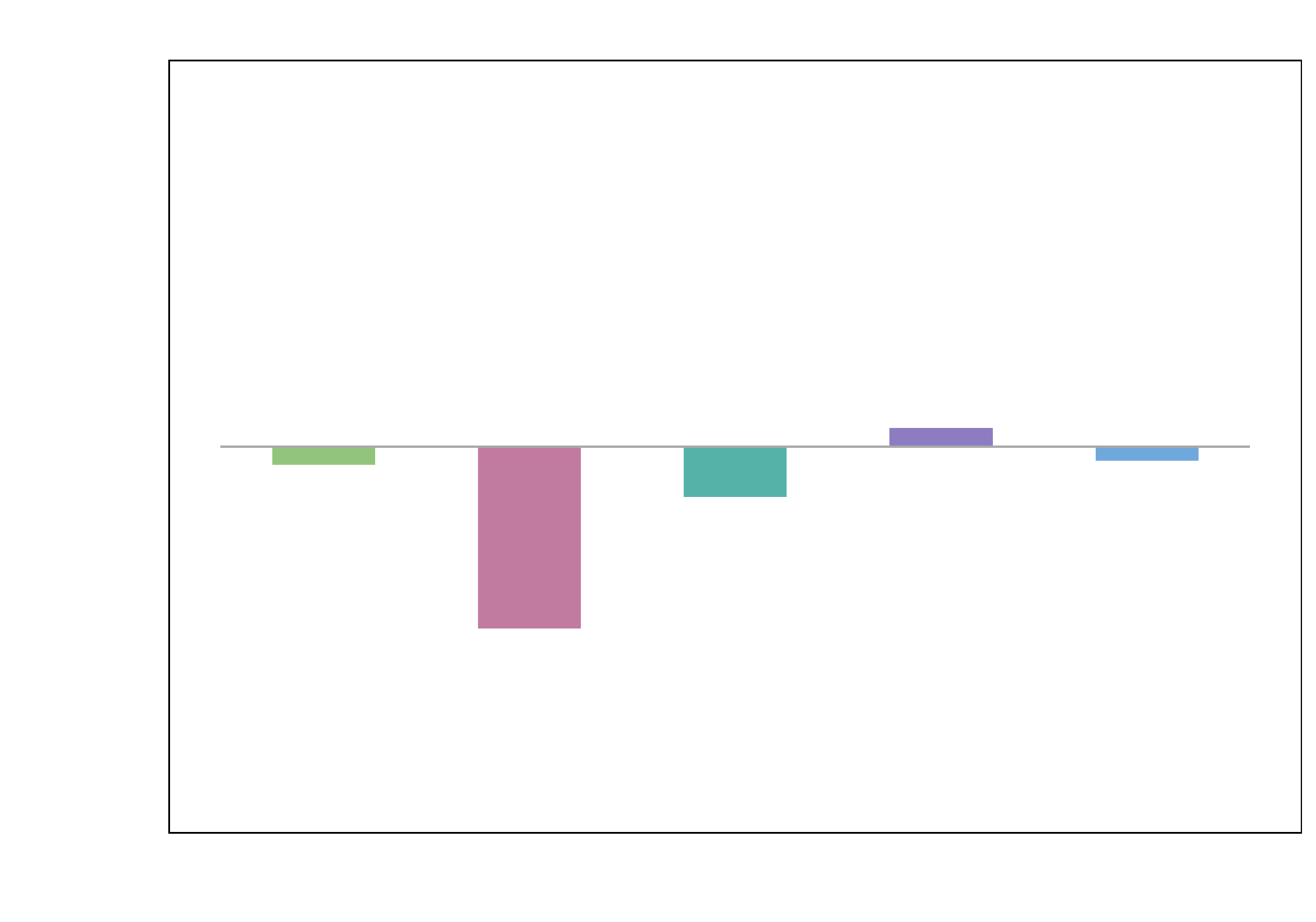}
        \caption{RCrop}
    \end{subfigure} \hfill
    \begin{subfigure}[h]{0.312\linewidth}
    \includegraphics[width=\linewidth]{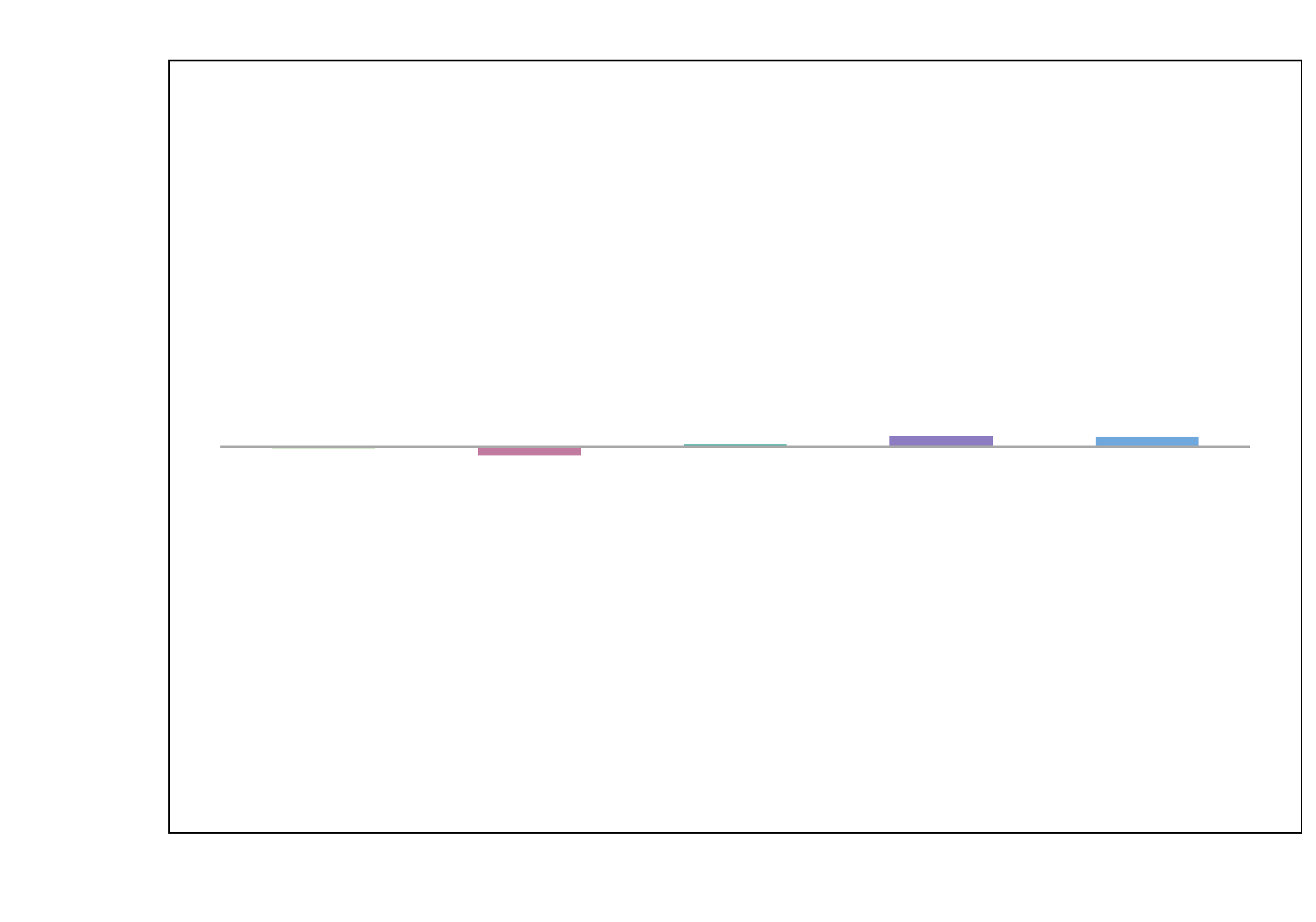}
        \caption{CJitter}
    \end{subfigure}
    
    \begin{subfigure}[h]{0.355\linewidth}
    \includegraphics[width=\linewidth]{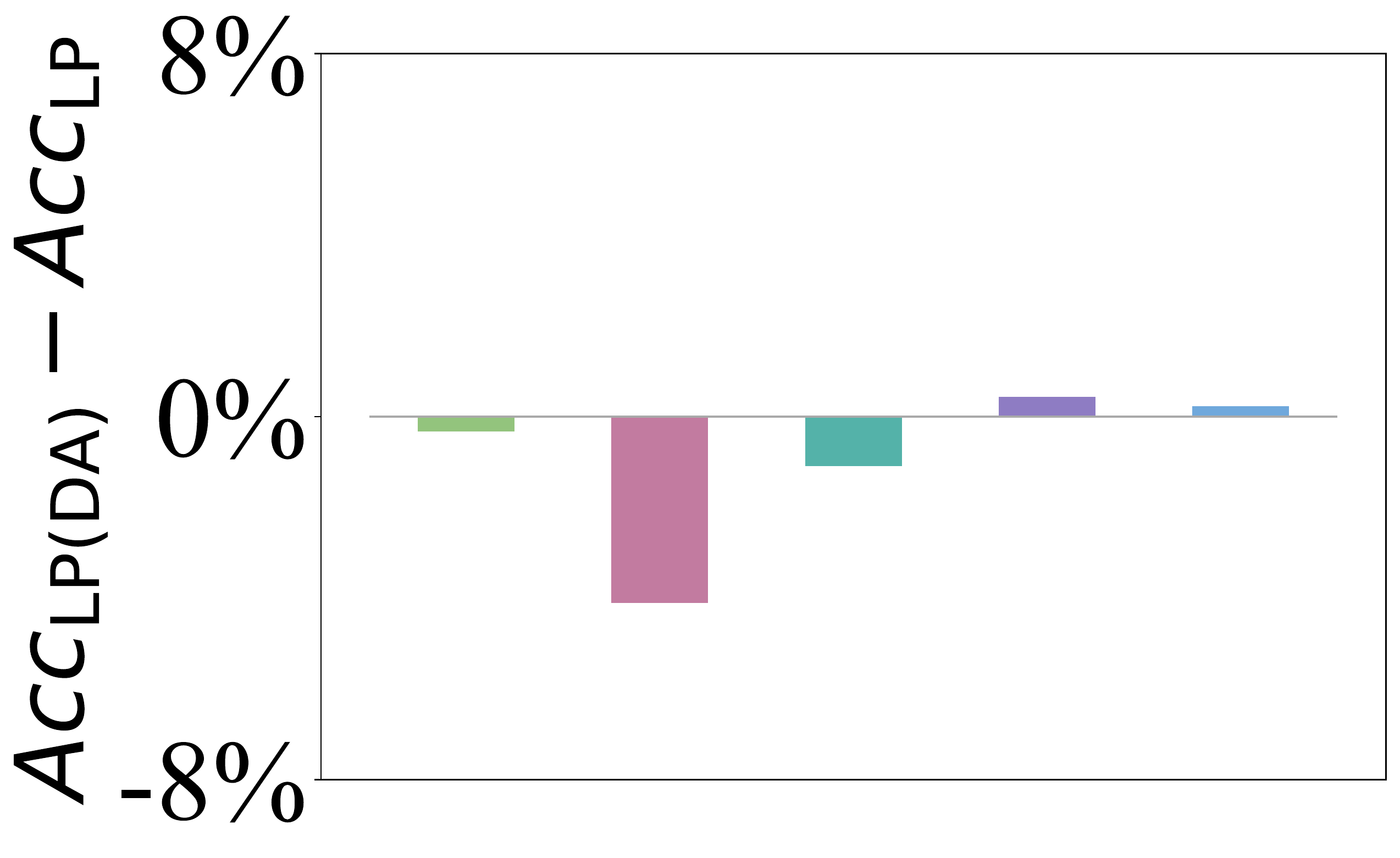}
        \caption{Base Aug}
    \end{subfigure} \hfill
    \begin{subfigure}[h]{0.312\linewidth}
    \includegraphics[width=\linewidth]{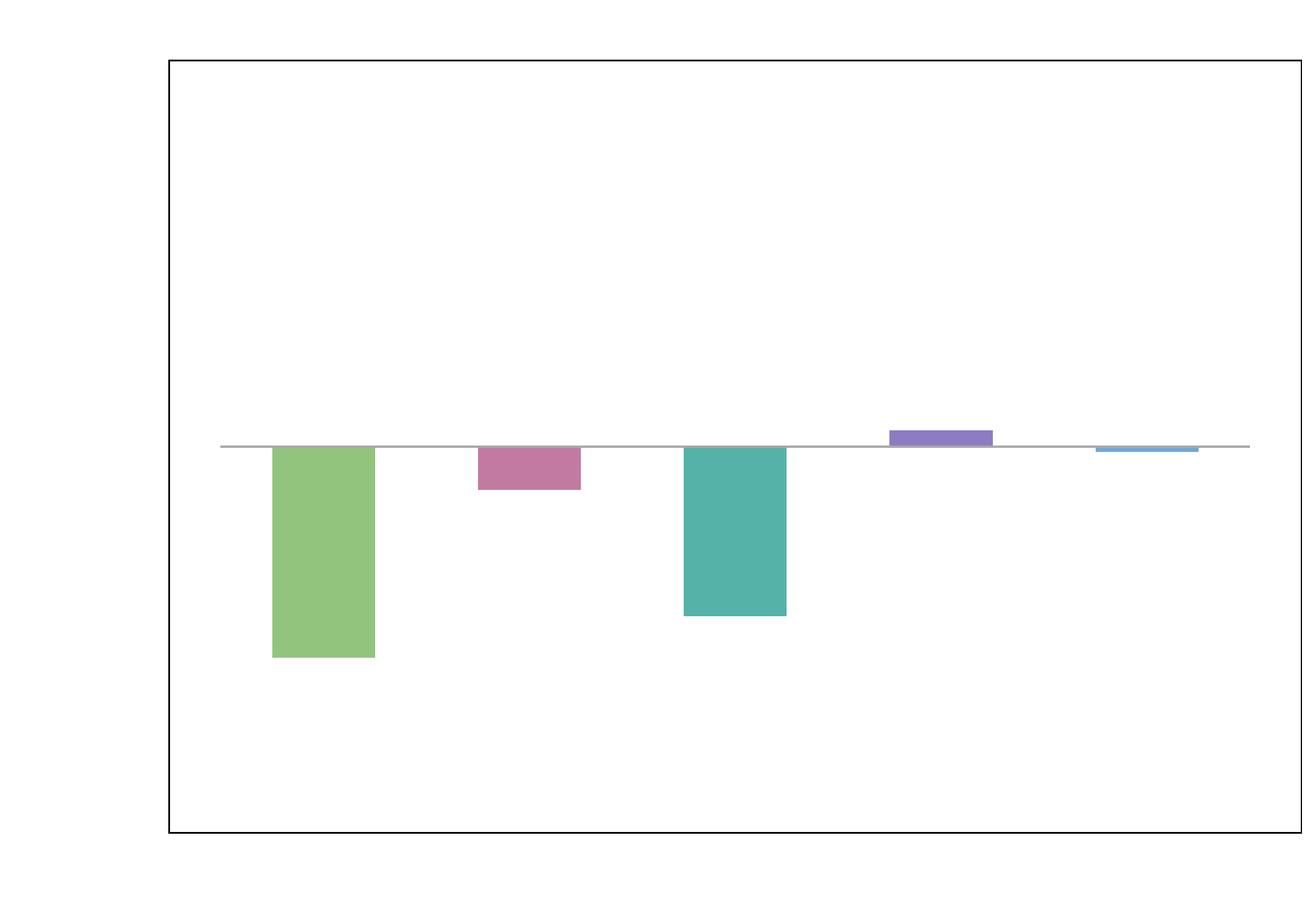}
        \caption{MixUp}
    \end{subfigure} \hfill
    \begin{subfigure}[h]{0.312\linewidth}
    \includegraphics[width=\linewidth]{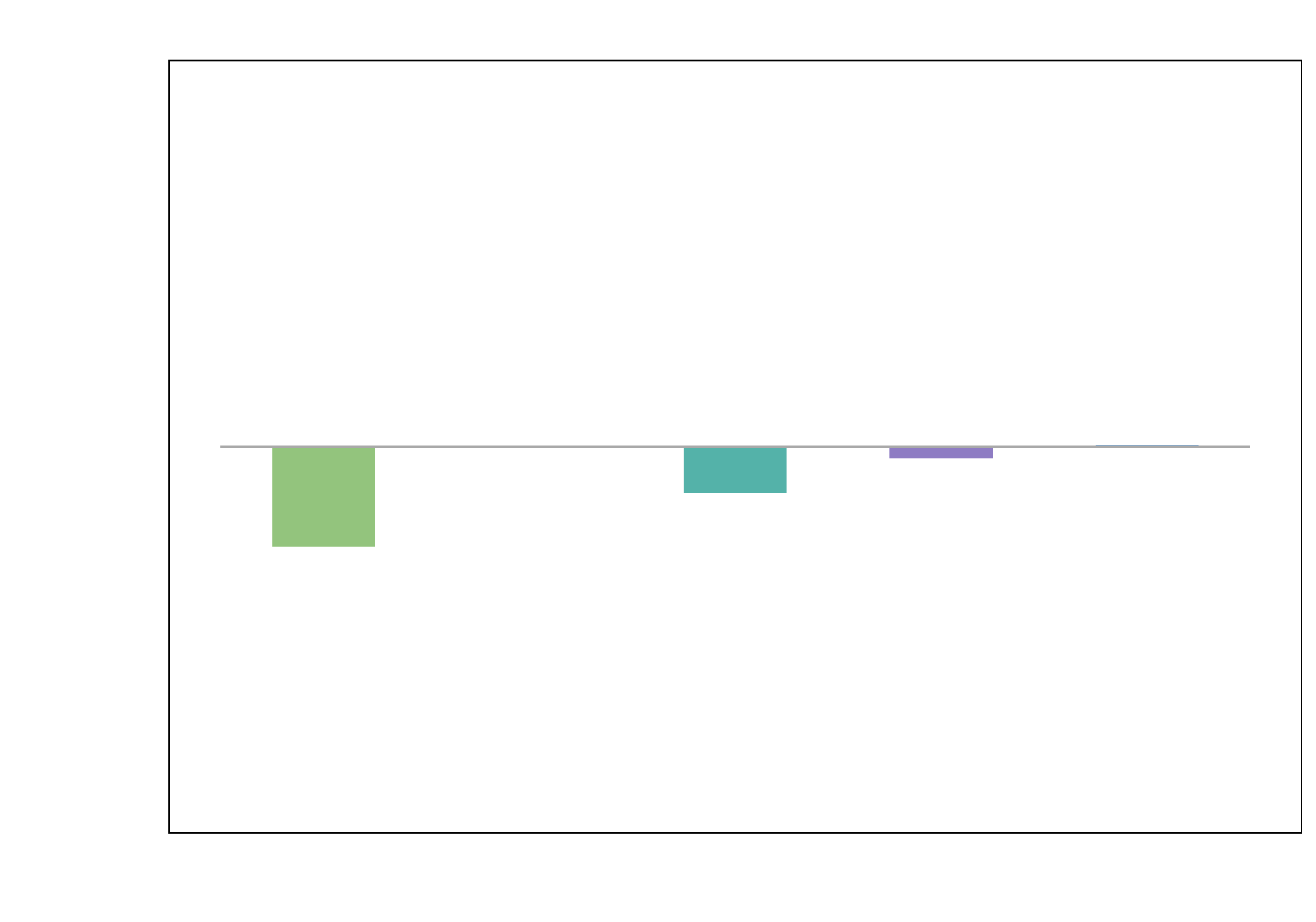}
        \caption{CutMix}
    \end{subfigure}
    \caption{5-way 5-shot performance difference of LP(Aug) with LP according to the augmentation techniques.}
    \label{fig:miniIN_da_lp_5shot}
\end{figure}
    
\begin{figure}[h!]
    \centering   
    \begin{subfigure}[h]{0.355\linewidth}
    \includegraphics[width=\linewidth]{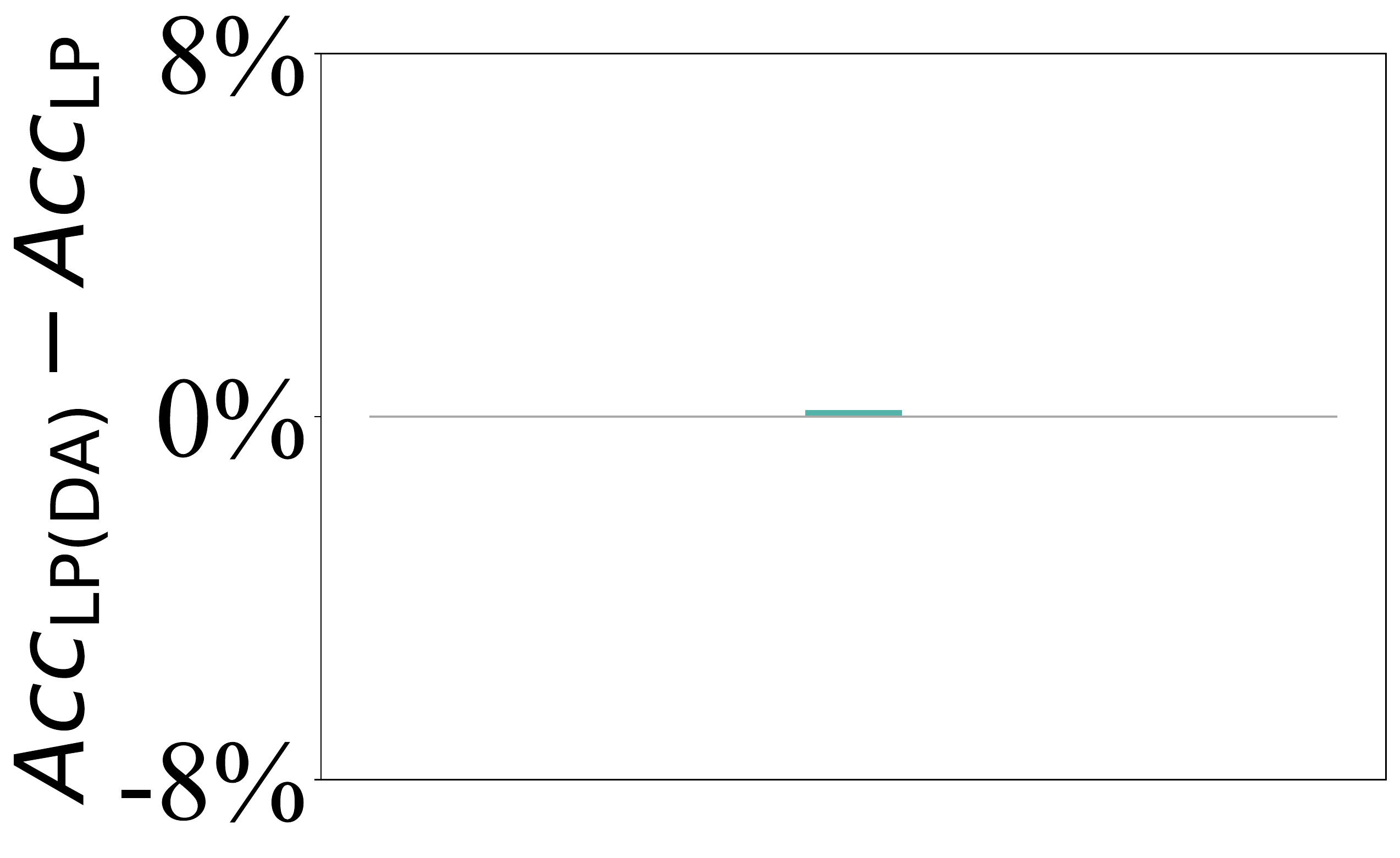}
        \caption{HFlip}
    \end{subfigure} \hfill
    \begin{subfigure}[h]{0.312\linewidth}
    \includegraphics[width=\linewidth]{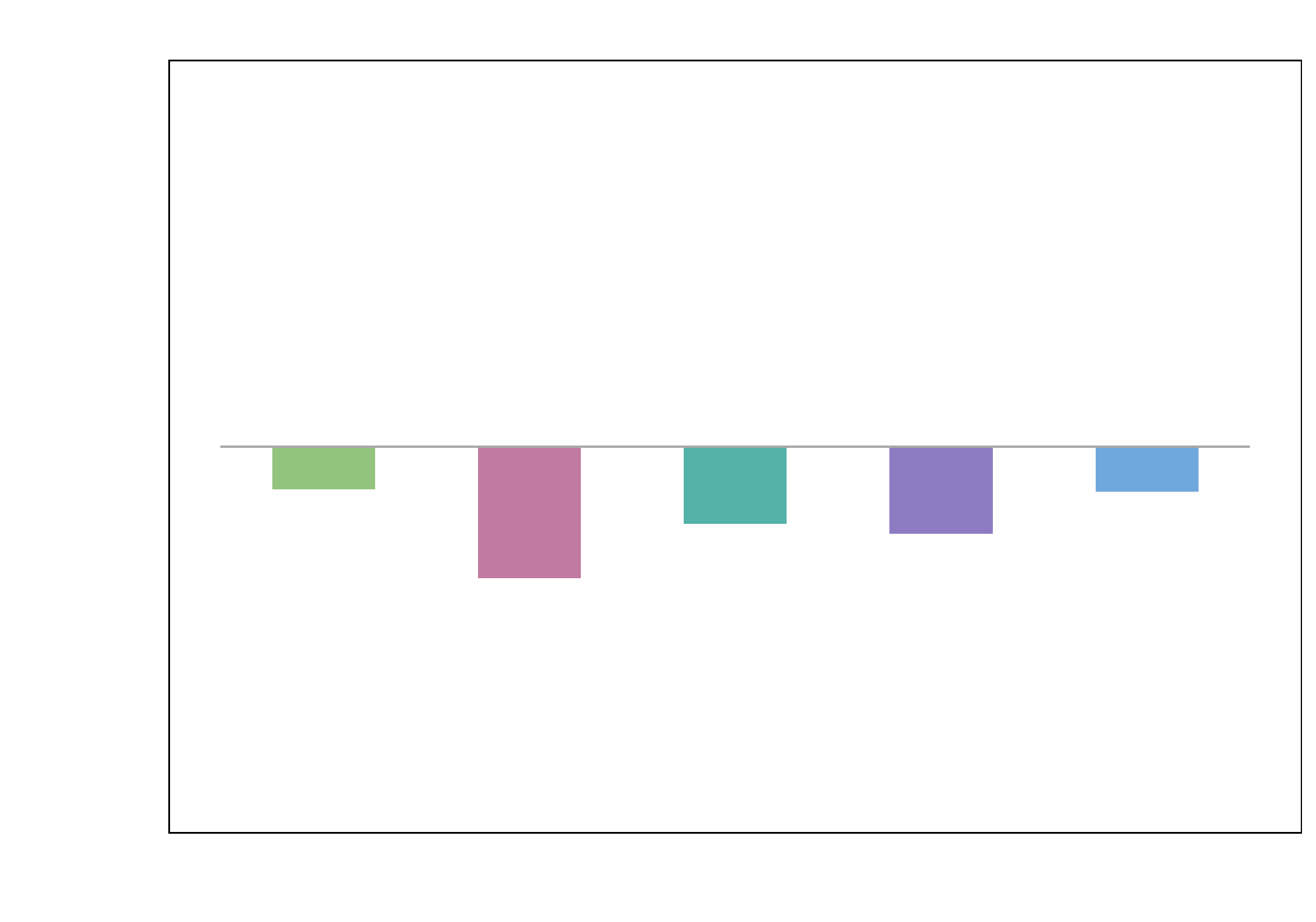}
        \caption{RCrop}
    \end{subfigure} \hfill
    \begin{subfigure}[h]{0.312\linewidth}
    \includegraphics[width=\linewidth]{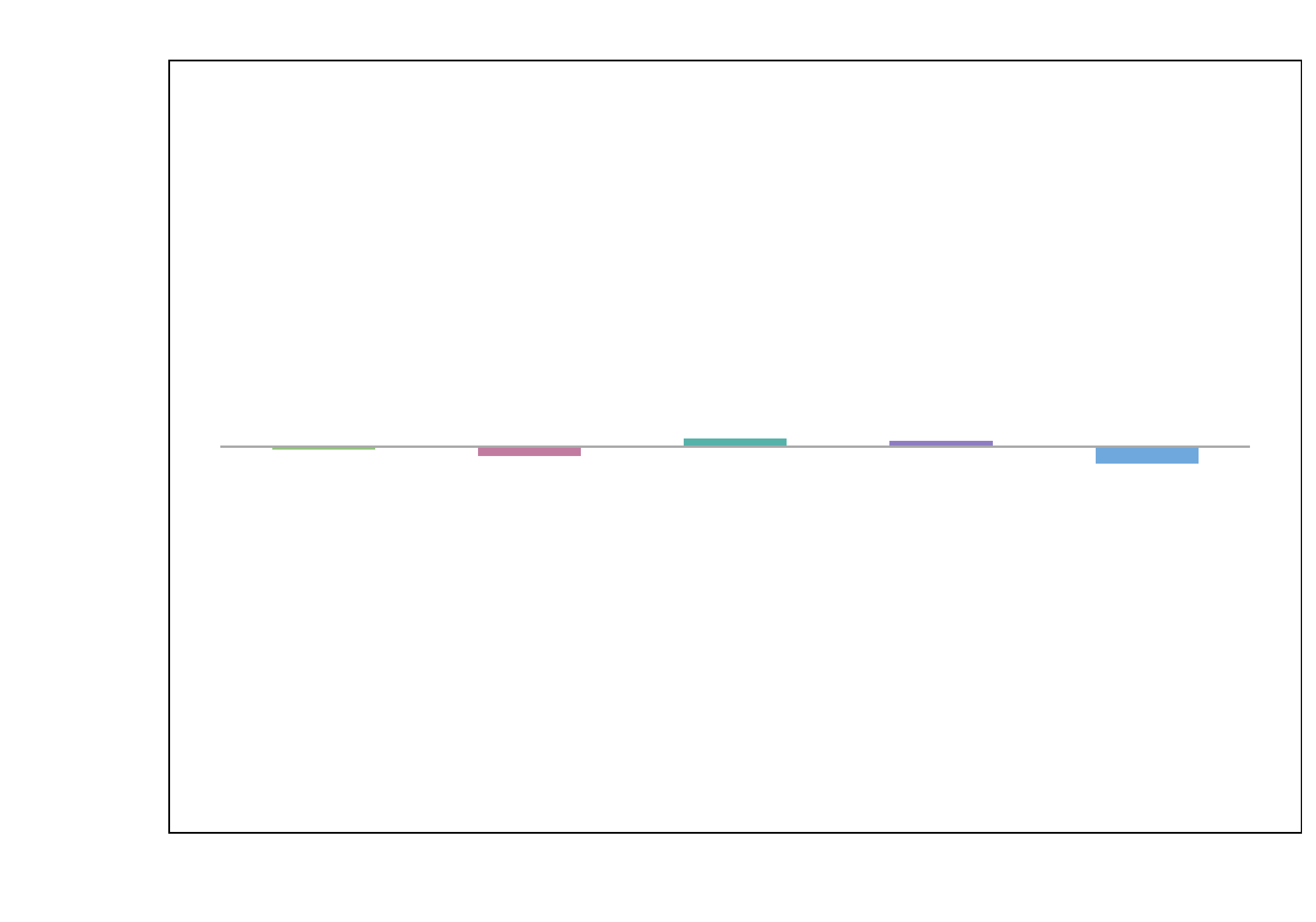}
        \caption{CJitter}
    \end{subfigure}
    
    \begin{subfigure}[h]{0.355\linewidth}
    \includegraphics[width=\linewidth]{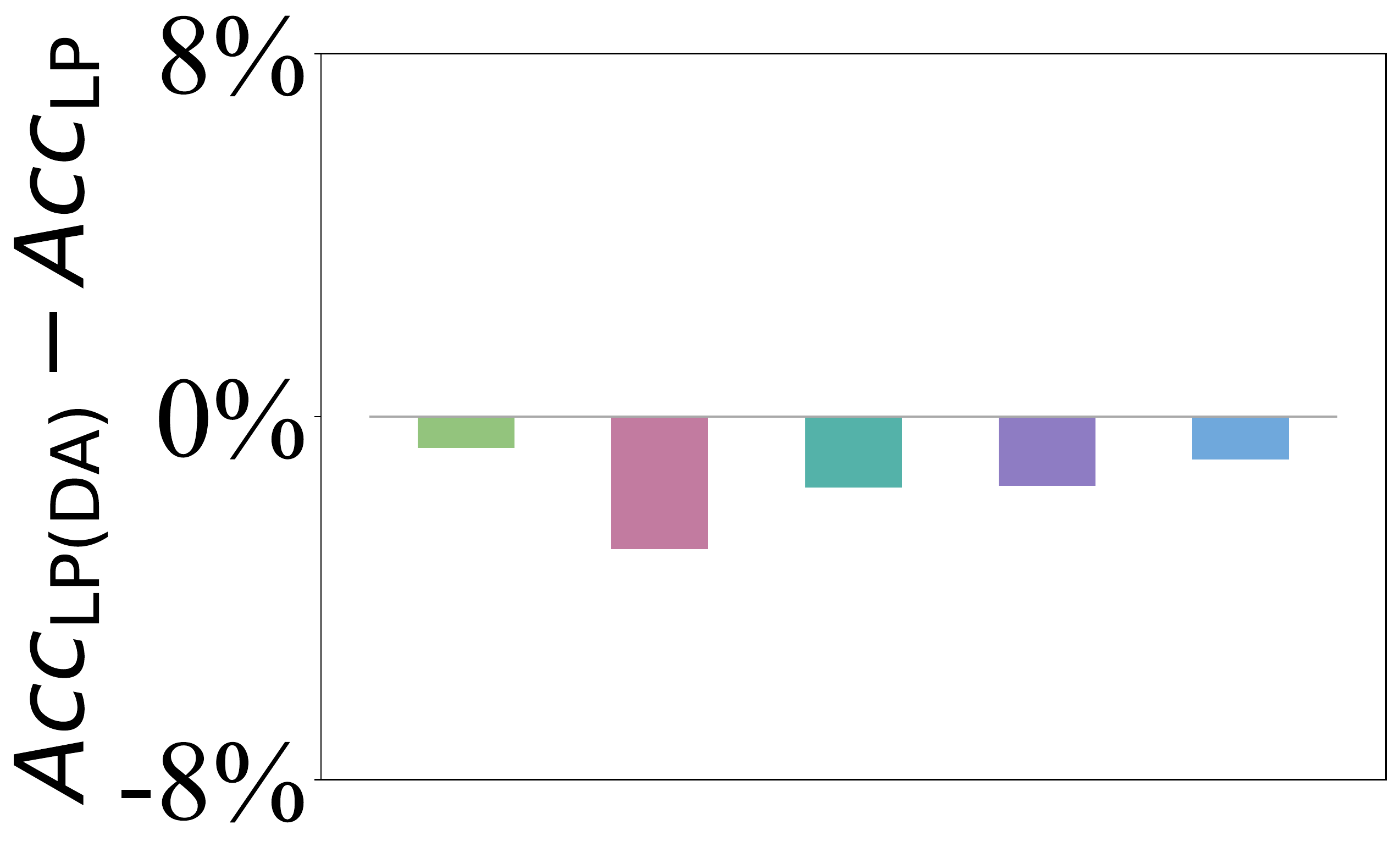}
        \caption{Base Aug}
    \end{subfigure} \hfill
    \begin{subfigure}[h]{0.312\linewidth}
    \includegraphics[width=\linewidth]{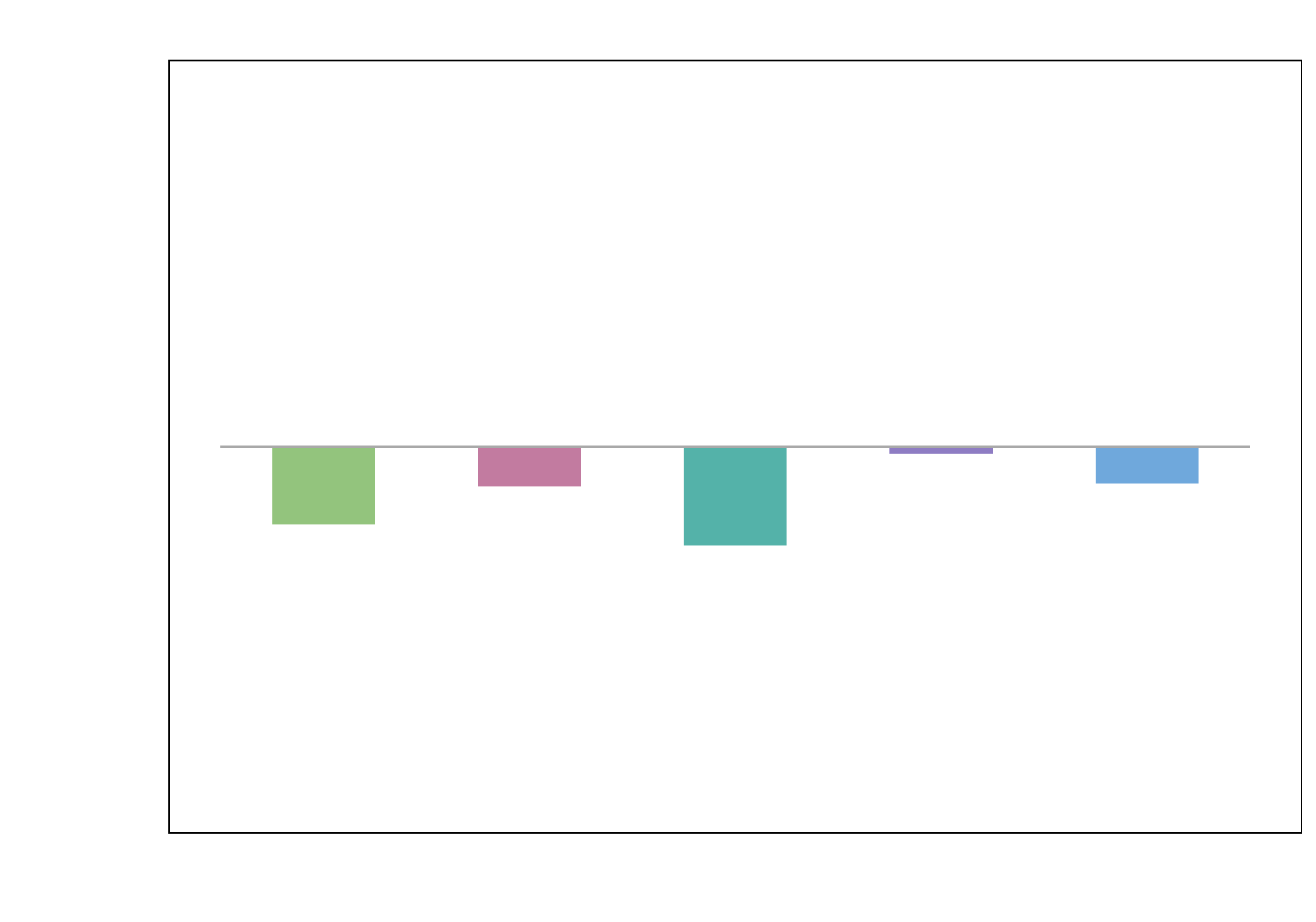}
        \caption{MixUp}
    \end{subfigure} \hfill
    \begin{subfigure}[h]{0.312\linewidth}
    \includegraphics[width=\linewidth]{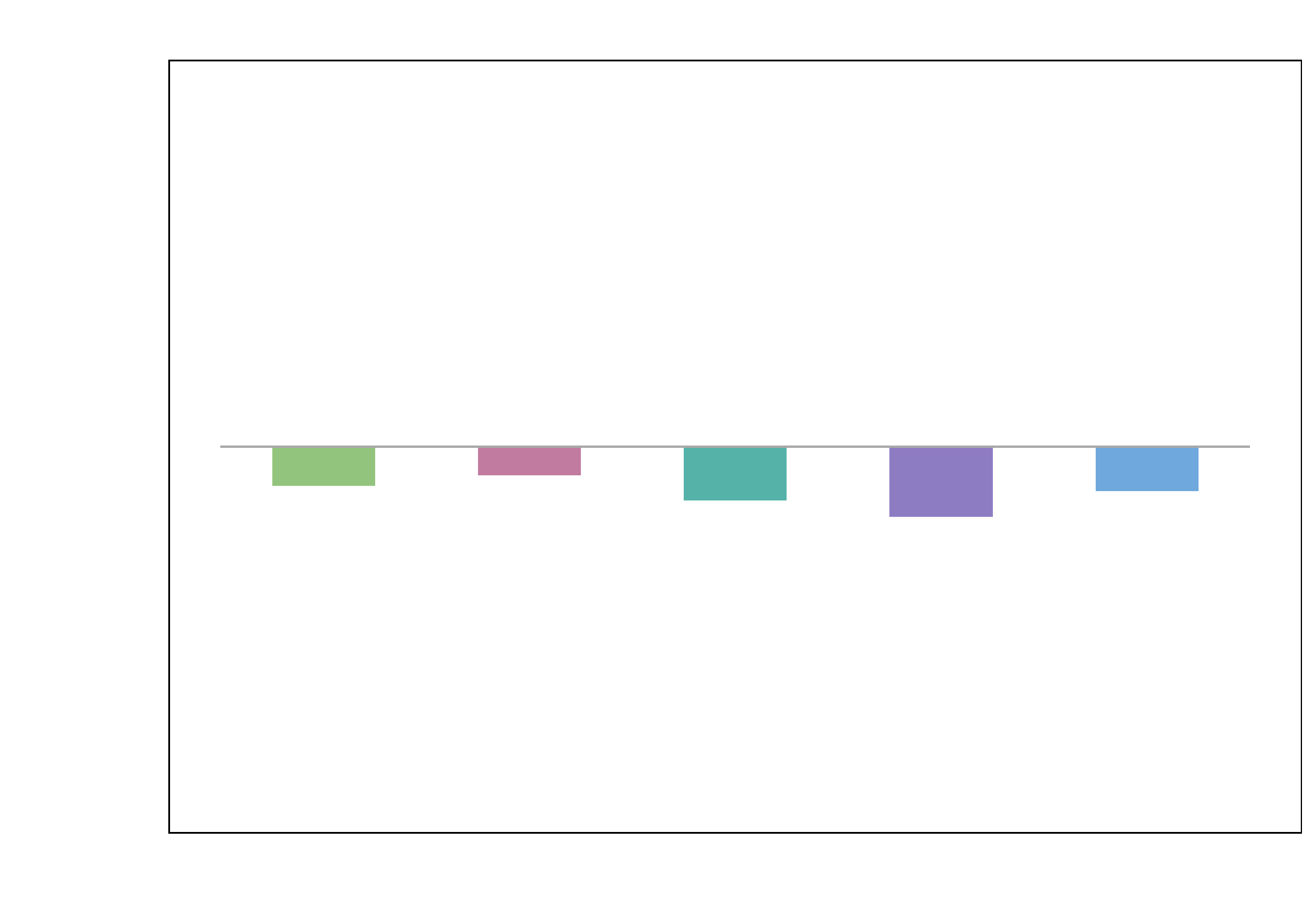}
        \caption{CutMix}
    \end{subfigure}
    
    \caption{5-way 20-shot performance difference of LP(Aug) with LP according to the augmentation techniques.}
    \label{fig:miniIN_da_lp_20shot}
\end{figure}

\subsection{Intensity of Single Augmentation}
\begin{figure}[h!]
    \centering
    \begin{subfigure}[h]{\linewidth}
    \includegraphics[width=\linewidth]{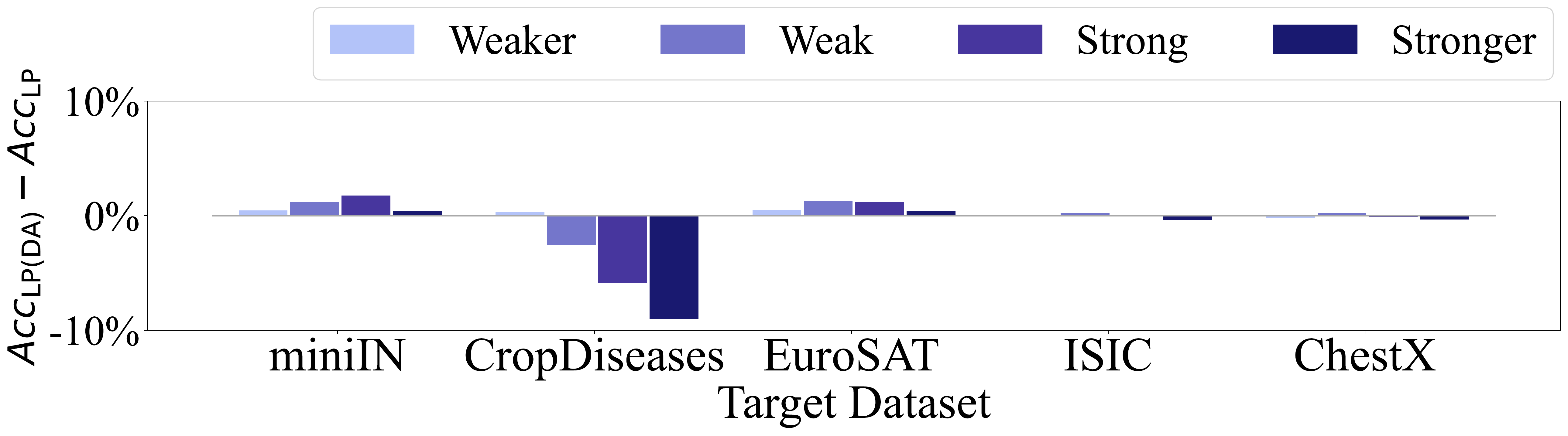}
    \caption{$k$ = 1}
    \end{subfigure}
    \begin{subfigure}[h]{\linewidth}
    \includegraphics[width=\linewidth]{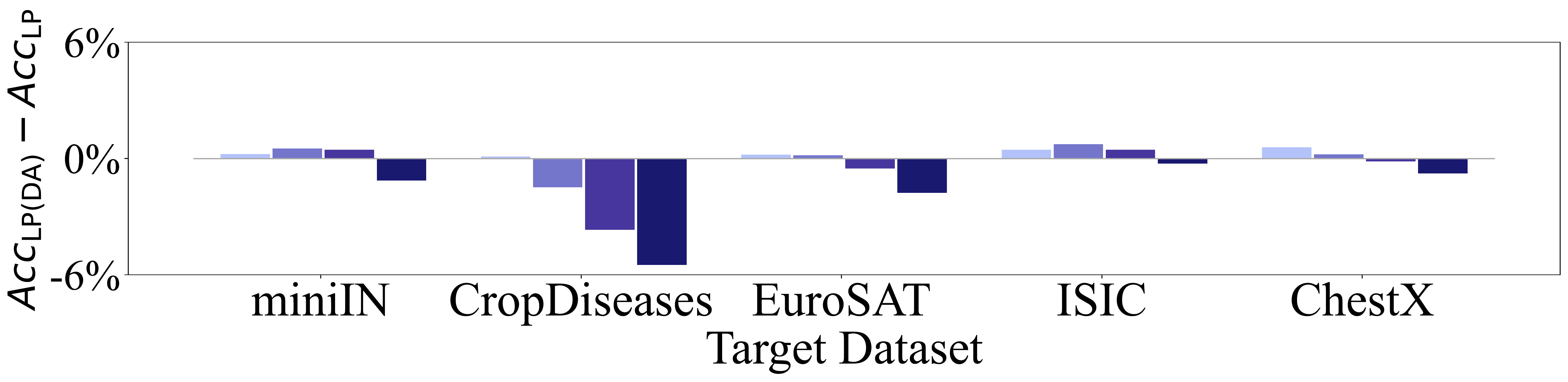}
    \caption{$k$ = 5}
    \end{subfigure}
    \begin{subfigure}[h]{\linewidth}
    \includegraphics[width=\linewidth]{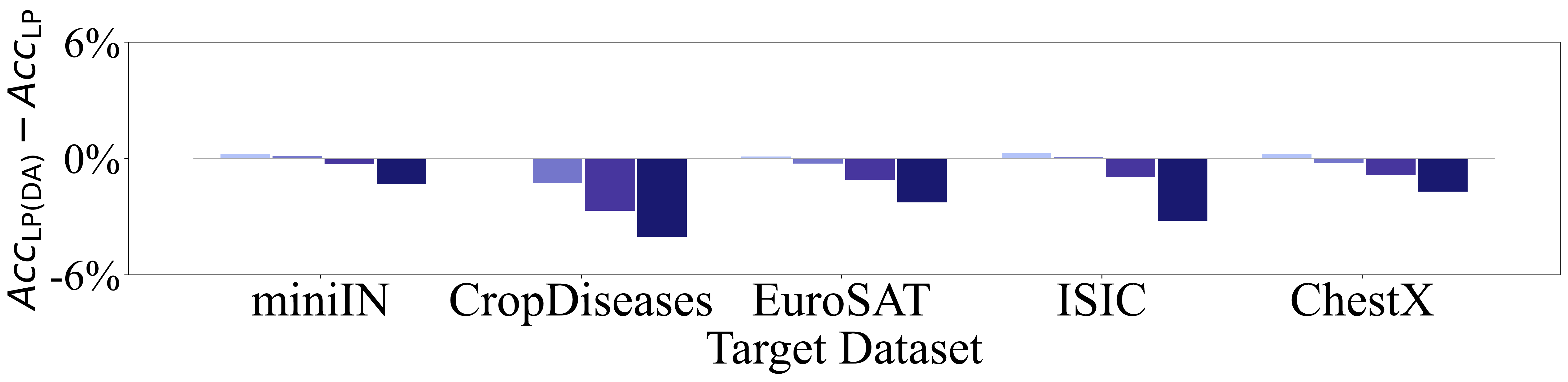}
    \caption{$k$ = 20}
    \end{subfigure}
    \caption{5-way $k$-shot performance difference of LP(DA) with LP according to the designed intensity of Base Aug.}
    \label{fig:base_magnitude_lp}
\end{figure}

Figure \ref{fig:base_magnitude_lp} describes the difference between LP(DA) and LP according to the designed intensity for $k \in \{1, 5, 20\}$. The observations are the same with Figure \ref{fig:base_magnitude_1shot}, \ref{fig:base_magnitude_5shot}, and \ref{fig:base_magnitude}. In particular, CropDiseases has the best intensity of \textsc{weaker} among four designed intensities, hence its performance degrades significantly as the intensity gets stronger. 

\subsection{Intensity of Mixing Augmentation}
Table \ref{tab:mixup_decomp_lp} shows the performance of LP according to the decomposition of MixUp and CutMix. The trend of the result is similar to Table \ref{tab:mixup_decomp}, in which the performance degradation of mixing augmentation is mainly due to mixing between different classes\,(i.e., ``B'' in this table).

\begin{table}[h!]
\centering
\resizebox{\linewidth}{!}{
\begin{tabular}{ccccccc}
    \toprule
    $k$ & MixUp & miniIN & \!\!\!\!CropDiseases\!\!\!\! & EuroSAT & ISIC & ChestX \\
    \midrule
    \multirow{2}{*}{1} & -  & 56.20{\scriptsize$\pm$.78} & 73.17{\scriptsize$\pm$.87} & 64.95{\scriptsize$\pm$.88} & 29.74{\scriptsize$\pm$.53} & 22.23{\scriptsize$\pm$.41} \\
    & B & 51.24{\scriptsize$\pm$.79} & 71.77{\scriptsize$\pm$.89} & 60.07{\scriptsize$\pm$.85} & 29.84{\scriptsize$\pm$.55} & 22.24{\scriptsize$\pm$.42} \\
    \midrule
    \multirow{4}{*}{5} & -   & 77.80{\scriptsize$\pm$.60} & 91.25{\scriptsize$\pm$.48} & 84.48{\scriptsize$\pm$.58} & 40.88{\scriptsize$\pm$.53} & 26.27{\scriptsize$\pm$.44} \\
    & {W+B} & 73.40{\scriptsize$\pm$.65} & 90.37{\scriptsize$\pm$.50} & 80.96{\scriptsize$\pm$.61} & 41.22{\scriptsize$\pm$.53} & 25.96{\scriptsize$\pm$.43} \\
    & W   & 76.40{\scriptsize$\pm$.62} & 90.86{\scriptsize$\pm$.49} & 83.03{\scriptsize$\pm$.63} & 40.29{\scriptsize$\pm$.54} & 26.07{\scriptsize$\pm$.44} \\
    & B   & 72.91{\scriptsize$\pm$.65} & 90.02{\scriptsize$\pm$.52} & 80.40{\scriptsize$\pm$.62} & 41.64{\scriptsize$\pm$.55} & 26.00{\scriptsize$\pm$.43} \\
    \midrule
    \multirow{4}{*}{20} & -   & 86.59{\scriptsize$\pm$.44} & 96.55{\scriptsize$\pm$.27} & 91.09{\scriptsize$\pm$.39} & 51.48{\scriptsize$\pm$.54} & 30.83{\scriptsize$\pm$.45} \\
    & {W+B} & 84.93{\scriptsize$\pm$.47} & 95.38{\scriptsize$\pm$.32} & 88.95{\scriptsize$\pm$.43} & 51.18{\scriptsize$\pm$.54} & 30.22{\scriptsize$\pm$.45} \\
    & W   & 85.23{\scriptsize$\pm$.47} & 95.45{\scriptsize$\pm$.31} & 89.12{\scriptsize$\pm$.45} & 50.01{\scriptsize$\pm$.53} & 30.17{\scriptsize$\pm$.45} \\
    & {B}   & 84.80{\scriptsize$\pm$.46} & 95.40{\scriptsize$\pm$.32} & 88.62{\scriptsize$\pm$.43} & 51.34{\scriptsize$\pm$.53} & 30.24{\scriptsize$\pm$.44} \\
    \midrule
    $k$ & CutMix & miniIN & \!\!\!\!CropDiseases\!\!\!\! & EuroSAT & ISIC & ChestX \\
    \midrule
    \multirow{2}{*}{1} & -  & 56.20{\scriptsize$\pm$.78} & 73.17{\scriptsize$\pm$.87} & 64.95{\scriptsize$\pm$.88} & 29.74{\scriptsize$\pm$.53} & 22.23{\scriptsize$\pm$.41} \\
    & B & 55.20{\scriptsize$\pm$.77} & 73.46{\scriptsize$\pm$.87} & 64.87{\scriptsize$\pm$.84} & 29.67{\scriptsize$\pm$.54} & 22.26{\scriptsize$\pm$.40} \\
    \midrule
    \multirow{4}{*}{5} & -   & 77.80{\scriptsize$\pm$.60} & 91.25{\scriptsize$\pm$.48} & 84.48{\scriptsize$\pm$.58} & 40.88{\scriptsize$\pm$.53} & 26.27{\scriptsize$\pm$.44} \\
    & W+B & 75.70{\scriptsize$\pm$.63} & 91.25{\scriptsize$\pm$.47} & 83.52{\scriptsize$\pm$.58} & 40.63{\scriptsize$\pm$.54} & 26.11{\scriptsize$\pm$.42} \\
    & W   & 77.01{\scriptsize$\pm$.62} & 91.29{\scriptsize$\pm$.47} & 83.59{\scriptsize$\pm$.57} & 40.60{\scriptsize$\pm$.54} & 26.29{\scriptsize$\pm$.44} \\
    & B   & 75.52{\scriptsize$\pm$.62} & 91.02{\scriptsize$\pm$.48} & 83.51{\scriptsize$\pm$.57} & 40.57{\scriptsize$\pm$.53} & 26.06{\scriptsize$\pm$.44} \\
    \midrule
    \multirow{4}{*}{20} & -   & 86.59{\scriptsize$\pm$.44} & 96.55{\scriptsize$\pm$.27} & 91.09{\scriptsize$\pm$.39} & 51.48{\scriptsize$\pm$.54} & 30.83{\scriptsize$\pm$.45} \\
    & W+B & 85.73{\scriptsize$\pm$.45} & 95.61{\scriptsize$\pm$.32} & 89.88{\scriptsize$\pm$.43} & 49.88{\scriptsize$\pm$.51} & 30.06{\scriptsize$\pm$.44} \\
    & W   & 85.95{\scriptsize$\pm$.45} & 95.60{\scriptsize$\pm$.31} & 89.40{\scriptsize$\pm$.42} & 49.97{\scriptsize$\pm$.55} & 30.50{\scriptsize$\pm$.44} \\
    & B   & 85.70{\scriptsize$\pm$.44} & 95.62{\scriptsize$\pm$.32} & 89.93{\scriptsize$\pm$.43} & 49.82{\scriptsize$\pm$.53} & 30.04{\scriptsize$\pm$.45} \\
    \bottomrule
\end{tabular}}
\caption{5-way $k$-shot performance of LP according to the decomposition of MixUp and CutMix. W and B indicate mixing within the same class (for $k>1$) and mixing between different classes, respectively.}\label{tab:mixup_decomp_lp}
\end{table}

\subsection{Scheduling DA}
Table \ref{tab:scheduling_lp} shows the few-shot performance according to the MixUp scheduler during LP. In line with Section \ref{subsec:scheduling}, not augmenting at later epochs alleviates the performance degradation by MixUp. 

\begin{table}[h!]
    \centering
    \resizebox{\linewidth}{!}{
    \begin{tabular}{ccccccc}
    \toprule
    $k$ & Aug. epochs & miniIN & \!\!\!CropDiseases\!\!\! & EuroSAT & ISIC & ChestX \\ \midrule
    \multirow{5}{*}{1} & \!\!- & 56.20{\scriptsize$\pm$.78} & 73.17{\scriptsize$\pm$.87} & 64.95{\scriptsize$\pm$.88} & 29.74{\scriptsize$\pm$.53} & 22.23{\scriptsize$\pm$.41} \\
    & \,\,\,\,1-100  & 51.24{\scriptsize$\pm$.79} & 71.77{\scriptsize$\pm$.89} & 60.07{\scriptsize$\pm$.85} & 29.84{\scriptsize$\pm$.55} & 22.24{\scriptsize$\pm$.42} \\
    & \,31-100 & 51.64{\scriptsize$\pm$.81} & 71.78{\scriptsize$\pm$.89} & 60.29{\scriptsize$\pm$.87} & 29.51{\scriptsize$\pm$.55} & 22.11{\scriptsize$\pm$.42} \\
    & \,1-70   & 55.75{\scriptsize$\pm$.78} & 73.02{\scriptsize$\pm$.88} & 64.24{\scriptsize$\pm$.85} & 29.80{\scriptsize$\pm$.53} & 22.22{\scriptsize$\pm$.43} \\
    & \,31-70\,\,\,  & 56.10{\scriptsize$\pm$.78} & 73.02{\scriptsize$\pm$.87} & 64.35{\scriptsize$\pm$.84} & 29.72{\scriptsize$\pm$.54} & 22.24{\scriptsize$\pm$.41} \\
    \midrule
    \multirow{5}{*}{5} & \!\!- & 77.80{\scriptsize$\pm$.60} & 91.25{\scriptsize$\pm$.48} & 84.48{\scriptsize$\pm$.58} & 40.88{\scriptsize$\pm$.53} & 26.27{\scriptsize$\pm$.44} \\
    & \,\,\,\,1-100  & 73.40{\scriptsize$\pm$.65} & 90.37{\scriptsize$\pm$.50} & 80.96{\scriptsize$\pm$.61} & 41.22{\scriptsize$\pm$.53} & 25.96{\scriptsize$\pm$.43} \\
    & \,31-100 & 73.86{\scriptsize$\pm$.65} & 90.40{\scriptsize$\pm$.49} & 81.42{\scriptsize$\pm$.62} & 41.47{\scriptsize$\pm$.54} & 26.13{\scriptsize$\pm$.45} \\
    & \,1-70   & 76.68{\scriptsize$\pm$.61} & 91.21{\scriptsize$\pm$.47} & 84.04{\scriptsize$\pm$.58} & 41.13{\scriptsize$\pm$.54} & 26.29{\scriptsize$\pm$.45} \\
    & \,31-70\,\,\,  & 77.24{\scriptsize$\pm$.61} & 91.42{\scriptsize$\pm$.47} & 84.44{\scriptsize$\pm$.58} & 41.29{\scriptsize$\pm$.55} & 26.30{\scriptsize$\pm$.44} \\
    \midrule
    \multirow{5}{*}{20} & \!\!- & 86.59{\scriptsize$\pm$.44} & 96.55{\scriptsize$\pm$.27} & 91.09{\scriptsize$\pm$.39} & 51.48{\scriptsize$\pm$.54} & 30.83{\scriptsize$\pm$.45} \\
    & \,\,\,\,1-100  & 84.93{\scriptsize$\pm$.47} & 95.38{\scriptsize$\pm$.32} & 88.95{\scriptsize$\pm$.43} & 51.18{\scriptsize$\pm$.54} & 30.22{\scriptsize$\pm$.45} \\
    & \,31-100 & 85.01{\scriptsize$\pm$.47} & 95.59{\scriptsize$\pm$.31} & 89.40{\scriptsize$\pm$.42} & 51.56{\scriptsize$\pm$.55} & 30.64{\scriptsize$\pm$.44} \\
    & \,1-70   & 86.29{\scriptsize$\pm$.44} & 95.98{\scriptsize$\pm$.29} & 90.71{\scriptsize$\pm$.41} & 51.13{\scriptsize$\pm$.53} & 30.70{\scriptsize$\pm$.45} \\
    & \,31-70\,\,\,  & 86.54{\scriptsize$\pm$.43} & 96.14{\scriptsize$\pm$.29} & 91.12{\scriptsize$\pm$.39} & 51.42{\scriptsize$\pm$.53} & 30.84{\scriptsize$\pm$.44} \\
    \bottomrule
    \end{tabular}}
    \caption{5-way $k$-shot performance of LP according to the data augmentation scheduler. ``-'' and ``1-100'' indicate no augmentation and augmentation during all epochs, respectively.}\label{tab:scheduling_lp}
\end{table}

\section{Comparison Using ResNet18}\label{appx:datta_tiered}
In this section, we provide the performance comparison using ResNet18 pre-trained on tieredIN (Table \ref{tab:comparison_TTA_tiered}) and ImageNet (Table \ref{tab:comparison_TTA_IN}). The best performance is from either LP+DA+TTA or FT+DA+TTA on most datasets. In particular, the effectiveness of LP increases as the size of the model or dataset used for pre-training increases.

\begin{table*}[t!]
\centering
\resizebox{\linewidth}{!}{
\begin{tabular}{cccccccccccccc}
    \toprule
    $k$ & Update & DA & TTA & miniIN & tieredIN & Places & Plantae & Cars & CUB & \!\!\!\!CropDiseases\!\!\!\! & EuroSAT & ISIC & ChestX \\
    \midrule
    \multirow{8.5}{*}{1} & \multirow{4}{*}{LP} & & & 63.17{\scriptsize$\pm$.78} & 60.36{\scriptsize$\pm$.89} & 52.91{\scriptsize$\pm$.82} & 38.27{\scriptsize$\pm$.77} & 31.80{\scriptsize$\pm$.57} & 58.93{\scriptsize$\pm$.90} & 66.04{\scriptsize$\pm$.90} & 62.07{\scriptsize$\pm$.91} & 30.75{\scriptsize$\pm$.56} & 21.87{\scriptsize$\pm$.39} \\ 
    & & \checkmark & & 63.74{\scriptsize$\pm$.81} & 62.03{\scriptsize$\pm$.91} & 52.81{\scriptsize$\pm$.82} & 39.36{\scriptsize$\pm$.74} & 32.69{\scriptsize$\pm$.61} & 59.67{\scriptsize$\pm$.90} & 62.65{\scriptsize$\pm$.87} & 62.15{\scriptsize$\pm$.89} & 31.20{\scriptsize$\pm$.55} & 21.70{\scriptsize$\pm$.37} \\ 
    & & & \checkmark & 63.57{\scriptsize$\pm$.82} & 60.02{\scriptsize$\pm$.94} & 52.49{\scriptsize$\pm$.83} & 37.96{\scriptsize$\pm$.77} & 30.49{\scriptsize$\pm$.57} & 59.94{\scriptsize$\pm$.91} & 61.27{\scriptsize$\pm$.94} & 60.19{\scriptsize$\pm$.89} & 30.88{\scriptsize$\pm$.63} & 21.54{\scriptsize$\pm$.36} \\ 
    & & \checkmark & \checkmark & \textbf{65.90}{\scriptsize$\pm$.81} & \textbf{62.60}{\scriptsize$\pm$.91} & \textbf{53.85}{\scriptsize$\pm$.83} & \textbf{40.67}{\scriptsize$\pm$.77} & \textbf{35.33}{\scriptsize$\pm$.64} & \textbf{62.04}{\scriptsize$\pm$.90} & \textbf{69.88}{\scriptsize$\pm$.84} & \textbf{63.75}{\scriptsize$\pm$.91} & 32.83{\scriptsize$\pm$.60} & \textbf{22.08}{\scriptsize$\pm$.38}\\ \cmidrule{2-14}
    & \multirow{4}{*}{FT} & & & 53.49{\scriptsize$\pm$.75} & 52.32{\scriptsize$\pm$.88} & 44.36{\scriptsize$\pm$.73} & 33.04{\scriptsize$\pm$.61} & 30.47{\scriptsize$\pm$.56} & 48.32{\scriptsize$\pm$.81} & 63.17{\scriptsize$\pm$.89} & 53.79{\scriptsize$\pm$.80} & 31.87{\scriptsize$\pm$.61} & 21.85{\scriptsize$\pm$.37} \\
    & & \checkmark & & 57.38{\scriptsize$\pm$.78} & 55.41{\scriptsize$\pm$.86} & 47.07{\scriptsize$\pm$.77} & 35.53{\scriptsize$\pm$.65} & 31.09{\scriptsize$\pm$.59} & 52.87{\scriptsize$\pm$.82} & 59.58{\scriptsize$\pm$.88} & 55.08{\scriptsize$\pm$.79} & \textbf{33.24}{\scriptsize$\pm$.59} & 21.72{\scriptsize$\pm$.37} \\
    & & & \checkmark & 53.19{\scriptsize$\pm$.79} & 51.78{\scriptsize$\pm$.90} & 43.83{\scriptsize$\pm$.78} & 32.90{\scriptsize$\pm$.62} & 29.45{\scriptsize$\pm$.56} & 47.20{\scriptsize$\pm$.84} & 59.91{\scriptsize$\pm$.94} & 45.12{\scriptsize$\pm$.86} & 29.59{\scriptsize$\pm$.60} & 21.22{\scriptsize$\pm$.33} \\
    & & \checkmark & \checkmark & 59.34{\scriptsize$\pm$.80} & 56.67{\scriptsize$\pm$.87} & 48.09{\scriptsize$\pm$.83} & 37.15{\scriptsize$\pm$.71} & 33.48{\scriptsize$\pm$.63} & 54.69{\scriptsize$\pm$.87} & 67.48{\scriptsize$\pm$.88} & 56.35{\scriptsize$\pm$.83} & 32.74{\scriptsize$\pm$.62} & 21.99{\scriptsize$\pm$.36} \\
    \midrule
    \multirow{8.5}{*}{5} & \multirow{4}{*}{LP} & & & 82.85{\scriptsize$\pm$.48} & 79.05{\scriptsize$\pm$.68} & 72.87{\scriptsize$\pm$.67} & 54.83{\scriptsize$\pm$.84} & 45.84{\scriptsize$\pm$.71} & 77.42{\scriptsize$\pm$.76} & 86.62{\scriptsize$\pm$.56} & 80.17{\scriptsize$\pm$.63} & 42.62{\scriptsize$\pm$.56} & 24.84{\scriptsize$\pm$.42} \\ 
    & & \checkmark & & 82.93{\scriptsize$\pm$.50} & 79.40{\scriptsize$\pm$.68} & 72.54{\scriptsize$\pm$.68} & 55.48{\scriptsize$\pm$.79} & 45.59{\scriptsize$\pm$.69} & 78.16{\scriptsize$\pm$.73} & 83.38{\scriptsize$\pm$.63} & 78.96{\scriptsize$\pm$.64} & 42.99{\scriptsize$\pm$.54} & 24.48{\scriptsize$\pm$.39} \\ 
    & & & \checkmark & 83.51{\scriptsize$\pm$.50} & 79.34{\scriptsize$\pm$.69} & 71.74{\scriptsize$\pm$.69} & 55.43{\scriptsize$\pm$.84} & 44.17{\scriptsize$\pm$.71} & 78.72{\scriptsize$\pm$.76} & 80.35{\scriptsize$\pm$.72} & 76.86{\scriptsize$\pm$.68} & 43.18{\scriptsize$\pm$.58} & 24.03{\scriptsize$\pm$.39} \\ 
    & & \checkmark & \checkmark & \textbf{85.47}{\scriptsize$\pm$.48} & \textbf{81.06}{\scriptsize$\pm$.67} & \textbf{73.66}{\scriptsize$\pm$.67} & \textbf{59.77}{\scriptsize$\pm$.81} & \textbf{54.82}{\scriptsize$\pm$.73} & \textbf{81.85}{\scriptsize$\pm$.70} & 89.67{\scriptsize$\pm$.51} & \textbf{81.20}{\scriptsize$\pm$.61} & 46.63{\scriptsize$\pm$.55} & 26.18{\scriptsize$\pm$.41} \\ \cmidrule{2-14}
    & \multirow{4}{*}{FT} & & & 79.75{\scriptsize$\pm$.51} & 77.16{\scriptsize$\pm$.69} & 68.47{\scriptsize$\pm$.70} & 52.66{\scriptsize$\pm$.79} & 47.72{\scriptsize$\pm$.73} & 73.70{\scriptsize$\pm$.79} & 89.44{\scriptsize$\pm$.52} & 79.02{\scriptsize$\pm$.60} & 48.47{\scriptsize$\pm$.62} & 25.23{\scriptsize$\pm$.44} \\
    & & \checkmark & & 80.67{\scriptsize$\pm$.50} & 77.90{\scriptsize$\pm$.70} & 69.88{\scriptsize$\pm$.69} & 54.72{\scriptsize$\pm$.80} & 46.55{\scriptsize$\pm$.71} & 76.84{\scriptsize$\pm$.74} & 87.08{\scriptsize$\pm$.55} & 77.67{\scriptsize$\pm$.57} & 48.98{\scriptsize$\pm$.61} & 25.32{\scriptsize$\pm$.40} \\
    & & & \checkmark & 80.32{\scriptsize$\pm$.53} & 77.34{\scriptsize$\pm$.70} & 67.88{\scriptsize$\pm$.72} & 53.35{\scriptsize$\pm$.79} & 45.66{\scriptsize$\pm$.74} & 74.11{\scriptsize$\pm$.79} & 87.03{\scriptsize$\pm$.57} & 68.84{\scriptsize$\pm$.70} & 45.74{\scriptsize$\pm$.65} & 24.23{\scriptsize$\pm$.36} \\
    & & \checkmark & \checkmark& 83.41{\scriptsize$\pm$.49} & 80.02{\scriptsize$\pm$.68} & 71.17{\scriptsize$\pm$.69} & 58.71{\scriptsize$\pm$.80} & 53.77{\scriptsize$\pm$.75} & 80.57{\scriptsize$\pm$.72} & \textbf{91.58}{\scriptsize$\pm$.45} & 80.48{\scriptsize$\pm$.57} & \textbf{51.23}{\scriptsize$\pm$.64} & \textbf{26.75}{\scriptsize$\pm$.42} \\
    \midrule
    \multirow{8.5}{*}{20} & \multirow{4}{*}{LP} & & & 89.07{\scriptsize$\pm$.34} & 85.85{\scriptsize$\pm$.55} & 80.58{\scriptsize$\pm$.54} & 65.17{\scriptsize$\pm$.76} & 58.68{\scriptsize$\pm$.66} & 85.11{\scriptsize$\pm$.56} & 92.59{\scriptsize$\pm$.41} & 86.31{\scriptsize$\pm$.51} & 50.42{\scriptsize$\pm$.51} & 28.72{\scriptsize$\pm$.42} \\ 
    & & \checkmark & & 88.77{\scriptsize$\pm$.36} & 85.64{\scriptsize$\pm$.54} & 80.30{\scriptsize$\pm$.54} & 64.81{\scriptsize$\pm$.76} & 55.22{\scriptsize$\pm$.62} & 84.83{\scriptsize$\pm$.58} & 88.94{\scriptsize$\pm$.51} & 84.02{\scriptsize$\pm$.54} & 48.63{\scriptsize$\pm$.51} & 27.78{\scriptsize$\pm$.39} \\ 
    & & & \checkmark & 89.83{\scriptsize$\pm$.35} & 86.38{\scriptsize$\pm$.54} & 79.56{\scriptsize$\pm$.55} & 66.21{\scriptsize$\pm$.75} & 56.31{\scriptsize$\pm$.69} & 86.87{\scriptsize$\pm$.54} & 85.92{\scriptsize$\pm$.60} & 82.59{\scriptsize$\pm$.54} & 50.96{\scriptsize$\pm$.56} & 27.04{\scriptsize$\pm$.40} \\ 
    & & \checkmark & \checkmark & \textbf{91.32}{\scriptsize$\pm$.34} & 87.16{\scriptsize$\pm$.50} & \textbf{81.23}{\scriptsize$\pm$.53} & 69.43{\scriptsize$\pm$.73} & 67.44{\scriptsize$\pm$.62} & 88.93{\scriptsize$\pm$.50} & 94.08{\scriptsize$\pm$.35} & 86.12{\scriptsize$\pm$.52} & 53.85{\scriptsize$\pm$.52} & 30.38{\scriptsize$\pm$.43} \\ \cmidrule{2-14}
    & \multirow{4}{*}{FT} & & & 88.51{\scriptsize$\pm$.38} & 86.08{\scriptsize$\pm$.56} & 79.42{\scriptsize$\pm$.57} & 66.46{\scriptsize$\pm$.76} & 63.15{\scriptsize$\pm$.66} & 84.14{\scriptsize$\pm$.60} & 96.35{\scriptsize$\pm$.27} & 89.79{\scriptsize$\pm$.45} & 60.87{\scriptsize$\pm$.57} & 29.72{\scriptsize$\pm$.44} \\
    & & \checkmark & & 88.46{\scriptsize$\pm$.38} & 86.18{\scriptsize$\pm$.55} & 79.58{\scriptsize$\pm$.56} & 67.37{\scriptsize$\pm$.73} & 58.48{\scriptsize$\pm$.68} & 85.03{\scriptsize$\pm$.56} & 95.08{\scriptsize$\pm$.31} & 87.63{\scriptsize$\pm$.46} & 59.80{\scriptsize$\pm$.55} & 29.63{\scriptsize$\pm$.41} \\
    & & & \checkmark & 88.94{\scriptsize$\pm$.39} & 86.13{\scriptsize$\pm$.56} & 78.50{\scriptsize$\pm$.58} & 67.01{\scriptsize$\pm$.77} & 60.20{\scriptsize$\pm$.70} & 84.91{\scriptsize$\pm$.58} & 94.43{\scriptsize$\pm$.35} & 84.34{\scriptsize$\pm$.49} & 58.89{\scriptsize$\pm$.62} & 27.95{\scriptsize$\pm$.40} \\
    & & \checkmark & \checkmark & 91.16{\scriptsize$\pm$.34} & \textbf{87.64}{\scriptsize$\pm$.51} & 80.78{\scriptsize$\pm$.52} & \textbf{72.32}{\scriptsize$\pm$.72} & \textbf{68.58}{\scriptsize$\pm$.65} & \textbf{88.96}{\scriptsize$\pm$.49} & \textbf{97.14}{\scriptsize$\pm$.22} & \textbf{90.02}{\scriptsize$\pm$.43} & \textbf{63.72}{\scriptsize$\pm$.58} & \textbf{32.15}{\scriptsize$\pm$.44} \\
    \bottomrule
    \end{tabular}}
\caption{5-way $k$-shot performance comparison. DA and TTA indicate augmentation during updates and evaluation, respectively. Base Aug is used for both DA and TTA. For TTA, 32 predictions are ensembled. ResNet18 is pre-trained on tieredIN. The best result are from either LP+DA+TTA or FT+DA+TTA on most datasets. Similar to the results of ResNet10, LP is better FT for $k=1$ and FT is better than LP for $k=20$. However, LP is better than FT for $k=5$, unlike the results of ResNet10.} \label{tab:comparison_TTA_tiered}
\end{table*}

\begin{table*}[t!]
\centering
\resizebox{\linewidth}{!}{
\begin{tabular}{cccccccccccccc}
    \toprule
    $k$ & Update & DA & TTA & miniIN & tieredIN & Places & Plantae & Cars & CUB & \!\!\!\!CropDiseases\!\!\!\! & EuroSAT & ISIC & ChestX \\
    \midrule
    \multirow{8.5}{*}{1} & \multirow{4}{*}{LP} & & & 80.25{\scriptsize$\pm$.78} & 71.09{\scriptsize$\pm$.88} & 59.09{\scriptsize$\pm$.89} & 43.93{\scriptsize$\pm$.82} & 45.44{\scriptsize$\pm$.80} & 67.94{\scriptsize$\pm$.90} & 74.38{\scriptsize$\pm$.85} & 66.38{\scriptsize$\pm$.88} & 29.79{\scriptsize$\pm$.52} & \textbf{22.45}{\scriptsize$\pm$.39} \\ 
    & & \checkmark & & 87.17{\scriptsize$\pm$.62} & \textbf{72.16}{\scriptsize$\pm$.90} & 61.70{\scriptsize$\pm$.84} & 45.79{\scriptsize$\pm$.82} & 46.73{\scriptsize$\pm$.79} & 70.43{\scriptsize$\pm$.87} & 72.51{\scriptsize$\pm$.91} & 68.75{\scriptsize$\pm$.89} & 30.57{\scriptsize$\pm$.51} & 21.96{\scriptsize$\pm$.36} \\ 
    & & & \checkmark & 79.30{\scriptsize$\pm$.87} & 67.68{\scriptsize$\pm$.93} & 57.94{\scriptsize$\pm$.89} & 44.07{\scriptsize$\pm$.82} & 45.52{\scriptsize$\pm$.84} & 66.92{\scriptsize$\pm$.94} & 72.71{\scriptsize$\pm$.88} & 58.54{\scriptsize$\pm$.94} & 30.09{\scriptsize$\pm$.57} & 21.48{\scriptsize$\pm$.35} \\ 
    & & \checkmark & \checkmark & \textbf{88.11}{\scriptsize$\pm$.63} & 71.99{\scriptsize$\pm$.89} & \textbf{61.80}{\scriptsize$\pm$.85} & \textbf{47.47}{\scriptsize$\pm$.86} & \textbf{49.62}{\scriptsize$\pm$.84} & \textbf{70.92}{\scriptsize$\pm$.90} & \textbf{77.98}{\scriptsize$\pm$.89} & \textbf{70.42}{\scriptsize$\pm$.91} & \textbf{32.67}{\scriptsize$\pm$.57} & 22.42{\scriptsize$\pm$.37} \\ \cmidrule{2-14}
    & \multirow{4}{*}{FT} & & & 71.71{\scriptsize$\pm$.81} & 58.04{\scriptsize$\pm$.89} & 51.70{\scriptsize$\pm$.81} & 36.63{\scriptsize$\pm$.66} & 38.27{\scriptsize$\pm$.70} & 54.24{\scriptsize$\pm$.85} & 69.08{\scriptsize$\pm$.89} & 49.68{\scriptsize$\pm$.85} & 30.69{\scriptsize$\pm$.54} & 21.93{\scriptsize$\pm$.38} \\
    & & \checkmark & & 65.72{\scriptsize$\pm$.94} & 50.39{\scriptsize$\pm$.90} & 47.98{\scriptsize$\pm$.85} & 37.19{\scriptsize$\pm$.70} & 35.90{\scriptsize$\pm$.68} & 52.50{\scriptsize$\pm$.89} & 66.87{\scriptsize$\pm$.95} & 52.77{\scriptsize$\pm$.85} & 32.61{\scriptsize$\pm$.61} & 22.06{\scriptsize$\pm$.38} \\
    & & & \checkmark & 70.22{\scriptsize$\pm$.91} & 52.87{\scriptsize$\pm$.95} & 49.23{\scriptsize$\pm$.84} & 36.35{\scriptsize$\pm$.73} & 35.64{\scriptsize$\pm$.72} & 51.63{\scriptsize$\pm$.90} & 60.87{\scriptsize$\pm$1.07} & 36.97{\scriptsize$\pm$.94} & 27.60{\scriptsize$\pm$.52} & 21.32{\scriptsize$\pm$.30} \\
    & & \checkmark & \checkmark & 64.82{\scriptsize$\pm$1.00} & 48.90{\scriptsize$\pm$.93} & 46.83{\scriptsize$\pm$.89} & 38.35{\scriptsize$\pm$.75} & 37.28{\scriptsize$\pm$.71} & 51.83{\scriptsize$\pm$.94} & 67.39{\scriptsize$\pm$.98} & 51.00{\scriptsize$\pm$.89} & 32.20{\scriptsize$\pm$.61} & 22.10{\scriptsize$\pm$.38} \\
    \midrule
    \multirow{8.5}{*}{5} & \multirow{4}{*}{LP} & & & 96.96{\scriptsize$\pm$.21} & \textbf{89.33}{\scriptsize$\pm$.50} & \textbf{83.28}{\scriptsize$\pm$.53} & 63.83{\scriptsize$\pm$.81} & 69.05{\scriptsize$\pm$.78} & 89.10{\scriptsize$\pm$.56} & 93.07{\scriptsize$\pm$.44} & 85.69{\scriptsize$\pm$.50} & 42.84{\scriptsize$\pm$.60} & 25.55{\scriptsize$\pm$.43} \\ 
    & & \checkmark & & 96.96{\scriptsize$\pm$.20} & 88.90{\scriptsize$\pm$.52} & 82.44{\scriptsize$\pm$.56} & 66.18{\scriptsize$\pm$.78 }& 70.11{\scriptsize$\pm$.79} & 89.02{\scriptsize$\pm$.55} & 91.76{\scriptsize$\pm$.48} & 85.57{\scriptsize$\pm$.50} & 43.03{\scriptsize$\pm$.58} & 25.74{\scriptsize$\pm$.42} \\ 
    & & & \checkmark & 97.40{\scriptsize$\pm$.19} & 88.59{\scriptsize$\pm$.50} & 83.11{\scriptsize$\pm$.54} & 65.34{\scriptsize$\pm$.82} & 69.58{\scriptsize$\pm$.80} & 89.79{\scriptsize$\pm$.56} & 91.13{\scriptsize$\pm$.50} & 79.89{\scriptsize$\pm$.61} & 43.52{\scriptsize$\pm$.60} & 24.21{\scriptsize$\pm$.38} \\ 
    & & \checkmark & \checkmark & \textbf{97.58}{\scriptsize$\pm$.19} & 88.77{\scriptsize$\pm$.52} & 82.77{\scriptsize$\pm$.57} & \textbf{70.35}{\scriptsize$\pm$.77} & \textbf{75.66}{\scriptsize$\pm$.76} & \textbf{90.43}{\scriptsize$\pm$.54} & \textbf{94.54}{\scriptsize$\pm$.40} & \textbf{87.75}{\scriptsize$\pm$.49} & 47.40{\scriptsize$\pm$.61} & 26.80{\scriptsize$\pm$.42} \\ \cmidrule{2-14}
    & \multirow{4}{*}{FT} & & & 90.72{\scriptsize$\pm$.42} & 80.55{\scriptsize$\pm$.66} & 73.79{\scriptsize$\pm$.71} & 58.33{\scriptsize$\pm$.81} & 61.84{\scriptsize$\pm$.77} & 80.99{\scriptsize$\pm$.70} & 92.22{\scriptsize$\pm$.47} & 75.41{\scriptsize$\pm$.63} & \textbf{48.97}{\scriptsize$\pm$.62} & 24.70{\scriptsize$\pm$.41} \\
    & & \checkmark & & 85.34{\scriptsize$\pm$.54} & 70.95{\scriptsize$\pm$.73} & 68.03{\scriptsize$\pm$.72} & 59.92{\scriptsize$\pm$.81} & 61.01{\scriptsize$\pm$.79} & 77.68{\scriptsize$\pm$.69} & 90.68{\scriptsize$\pm$.50} & 71.05{\scriptsize$\pm$.64} & 46.86{\scriptsize$\pm$.62} & 25.90{\scriptsize$\pm$.41} \\
    & & & \checkmark & 89.28{\scriptsize$\pm$.51} & 75.83{\scriptsize$\pm$.78} & 70.20{\scriptsize$\pm$.76} & 58.23{\scriptsize$\pm$.83} & 56.44{\scriptsize$\pm$.85} & 78.77{\scriptsize$\pm$.80} & 85.98{\scriptsize$\pm$.73} & 58.88{\scriptsize$\pm$.90} & 43.30{\scriptsize$\pm$.68} & 22.99{\scriptsize$\pm$.34} \\
    & & \checkmark & \checkmark& 86.88{\scriptsize$\pm$.53} & 71.91{\scriptsize$\pm$.75} & 68.60{\scriptsize$\pm$.71} & 64.59{\scriptsize$\pm$.82} & 68.28{\scriptsize$\pm$.78} & 81.55{\scriptsize$\pm$.66} & 92.76{\scriptsize$\pm$.44} & 74.99{\scriptsize$\pm$.63} & 48.94{\scriptsize$\pm$.64} & \textbf{27.20}{\scriptsize$\pm$.42} \\
    \midrule
    \multirow{8.5}{*}{20} & \multirow{4}{*}{LP} & & & 98.54{\scriptsize$\pm$.13} & 93.46{\scriptsize$\pm$.37} & 89.18{\scriptsize$\pm$.42} & 76.01{\scriptsize$\pm$.67} & 82.19{\scriptsize$\pm$.60} & 94.36{\scriptsize$\pm$.38} & 97.18{\scriptsize$\pm$.24} & 91.93{\scriptsize$\pm$.32} & 54.67{\scriptsize$\pm$.57} & 29.40{\scriptsize$\pm$.44} \\ 
    & & \checkmark & & 98.60{\scriptsize$\pm$.13} & 93.23{\scriptsize$\pm$.37} & 89.18{\scriptsize$\pm$.41} & 77.76{\scriptsize$\pm$.62} & 82.14{\scriptsize$\pm$.60} & 94.50{\scriptsize$\pm$.37} & 96.42{\scriptsize$\pm$.27} & 91.47{\scriptsize$\pm$.34} & 53.66{\scriptsize$\pm$.56} & 30.13{\scriptsize$\pm$.43} \\ 
    & & & \checkmark & 98.93{\scriptsize$\pm$.11} & 92.96{\scriptsize$\pm$.39} & 89.08{\scriptsize$\pm$.41} & 78.60{\scriptsize$\pm$.64} & 82.42{\scriptsize$\pm$.61} & 95.18{\scriptsize$\pm$.36} & 95.40{\scriptsize$\pm$.34} & 87.62{\scriptsize$\pm$.44} & 55.66{\scriptsize$\pm$.59} & 27.76{\scriptsize$\pm$.41} \\ 
    & & \checkmark & \checkmark & \textbf{98.96}{\scriptsize$\pm$.10} & \textbf{93.70}{\scriptsize$\pm$.37} & \textbf{89.75}{\scriptsize$\pm$.38} & 82.63{\scriptsize$\pm$.57} & 88.71{\scriptsize$\pm$.48} & \textbf{95.89}{\scriptsize$\pm$.32} & 98.08{\scriptsize$\pm$.19}& \textbf{93.07}{\scriptsize$\pm$.34} & 59.94{\scriptsize$\pm$.56} & 32.83{\scriptsize$\pm$.45} \\ \cmidrule{2-14}
    & \multirow{4}{*}{FT} & & & 96.79{\scriptsize$\pm$.19} & 91.57{\scriptsize$\pm$.43} & 85.77{\scriptsize$\pm$.49} & 75.38{\scriptsize$\pm$.68} & 83.97{\scriptsize$\pm$.54} & 92.84{\scriptsize$\pm$.42} & 97.99{\scriptsize$\pm$.21} & 90.24{\scriptsize$\pm$.38} & 61.30{\scriptsize$\pm$.60} & 29.47{\scriptsize$\pm$.44} \\
    & & \checkmark & & 95.96{\scriptsize$\pm$.22} & 89.34{\scriptsize$\pm$.47} & 85.44{\scriptsize$\pm$.47} & 79.06{\scriptsize$\pm$.61} & 85.32{\scriptsize$\pm$.48} & 92.55{\scriptsize$\pm$.40} & 98.14{\scriptsize$\pm$.20} & 89.81{\scriptsize$\pm$.37} & 60.16{\scriptsize$\pm$.63} & 32.64{\scriptsize$\pm$.47} \\
    & & & \checkmark & 96.62{\scriptsize$\pm$.24} & 89.35{\scriptsize$\pm$.50} & 83.76{\scriptsize$\pm$.53} & 76.67{\scriptsize$\pm$.67} & 78.24{\scriptsize$\pm$.72} & 92.13{\scriptsize$\pm$.46} & 95.14{\scriptsize$\pm$.40} & 78.24{\scriptsize$\pm$.70} & 56.52{\scriptsize$\pm$.67} & 26.34{\scriptsize$\pm$.38} \\
    & & \checkmark & \checkmark & 96.91{\scriptsize$\pm$.19} & 89.80{\scriptsize$\pm$.46} & 85.70{\scriptsize$\pm$.47} & \textbf{84.02}{\scriptsize$\pm$.54} & \textbf{91.50}{\scriptsize$\pm$.38} & 94.88{\scriptsize$\pm$.34} & \textbf{98.85}{\scriptsize$\pm$.14} & 91.99{\scriptsize$\pm$.34} & \textbf{64.95}{\scriptsize$\pm$.62} & \textbf{35.43}{\scriptsize$\pm$.50} \\
    \bottomrule
    \end{tabular}}
\caption{5-way $k$-shot performance comparison. DA and TTA indicate augmentation during updates and evaluation, respectively. Base Aug is used for both DA and TTA. For TTA, 32 predictions are ensembled. ResNet18 pre-trained on ImageNet from PyTorch \cite{paszke2019pytorch} is used. The best result are from either LP+DA+TTA or FT+DA+TTA on most datasets. However, when ImageNet is used for pre-training, there are cases where LP is still better than FT for $k=20$.}\label{tab:comparison_TTA_IN}
\end{table*}

\section{Combination of Intensities (LP)}\label{appx:abla_comb}
We analyze the combination of four designed intensities for DA and TTA, when only a linear classifier is updated.

\begin{figure}[h!]
    \centering
    \begin{subfigure}[h]{0.325\linewidth}
    \includegraphics[width=\linewidth]{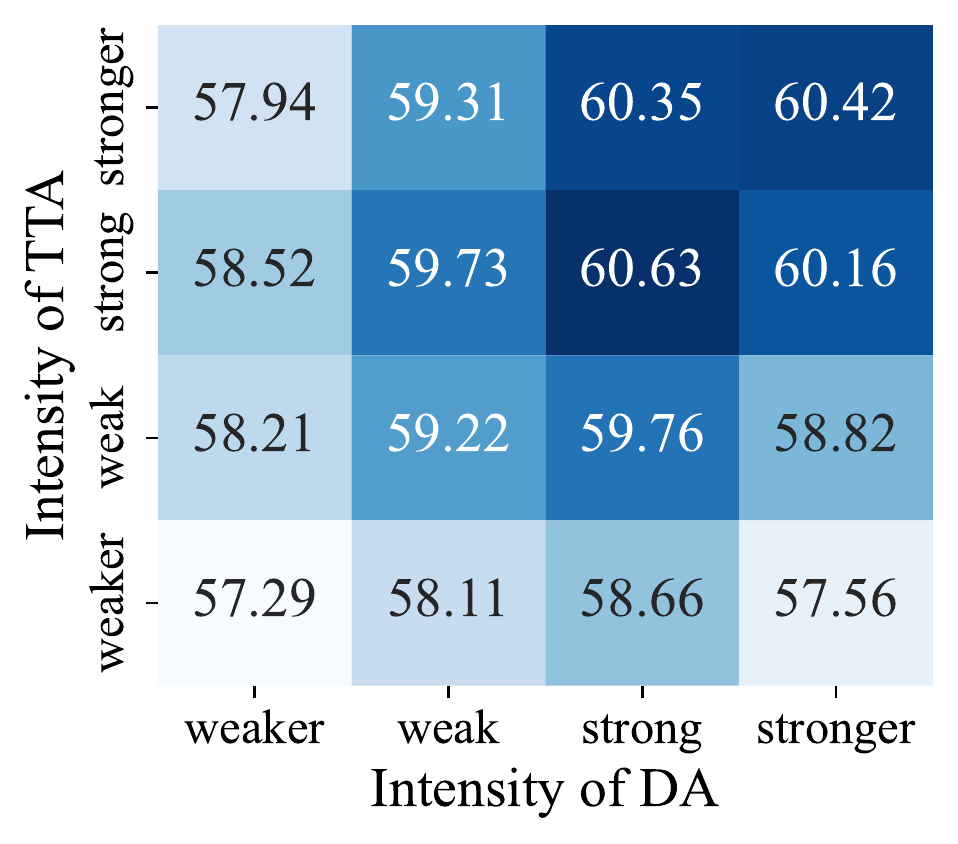}
        \caption{miniIN ($k=1$)}
    \end{subfigure}
    \begin{subfigure}[h]{0.325\linewidth}
    \includegraphics[width=\linewidth]{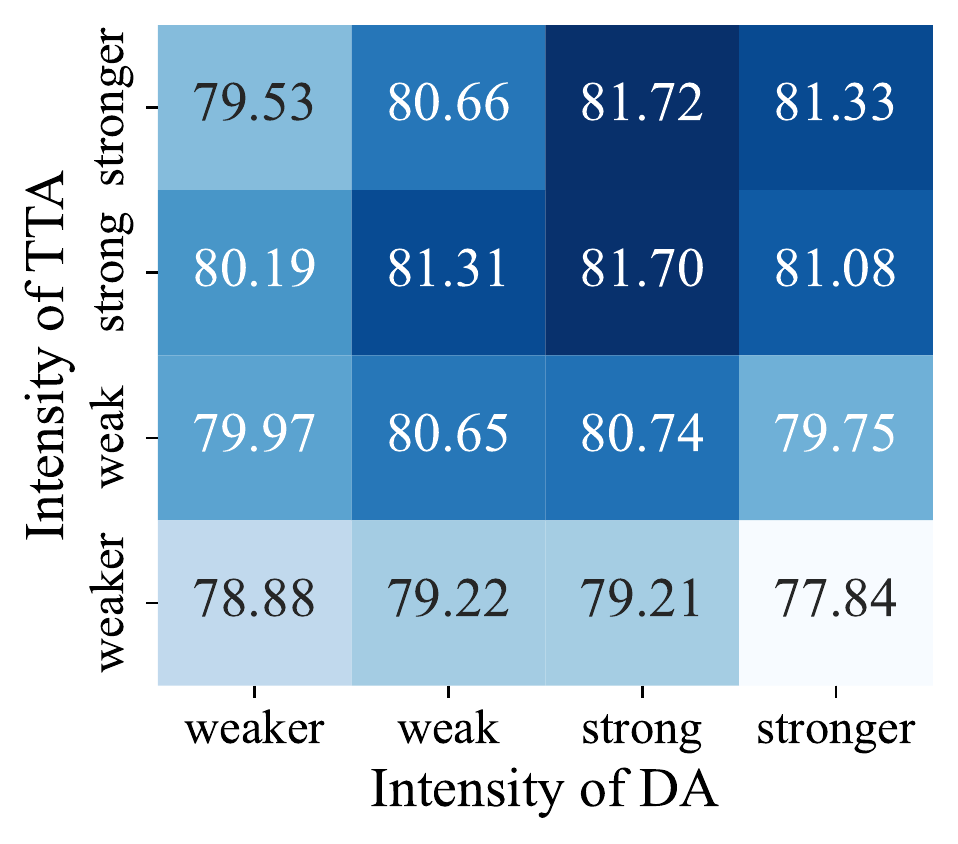}
        \caption{miniIN ($k=5$)}
    \end{subfigure}
    \begin{subfigure}[h]{0.325\linewidth}
    \includegraphics[width=\linewidth]{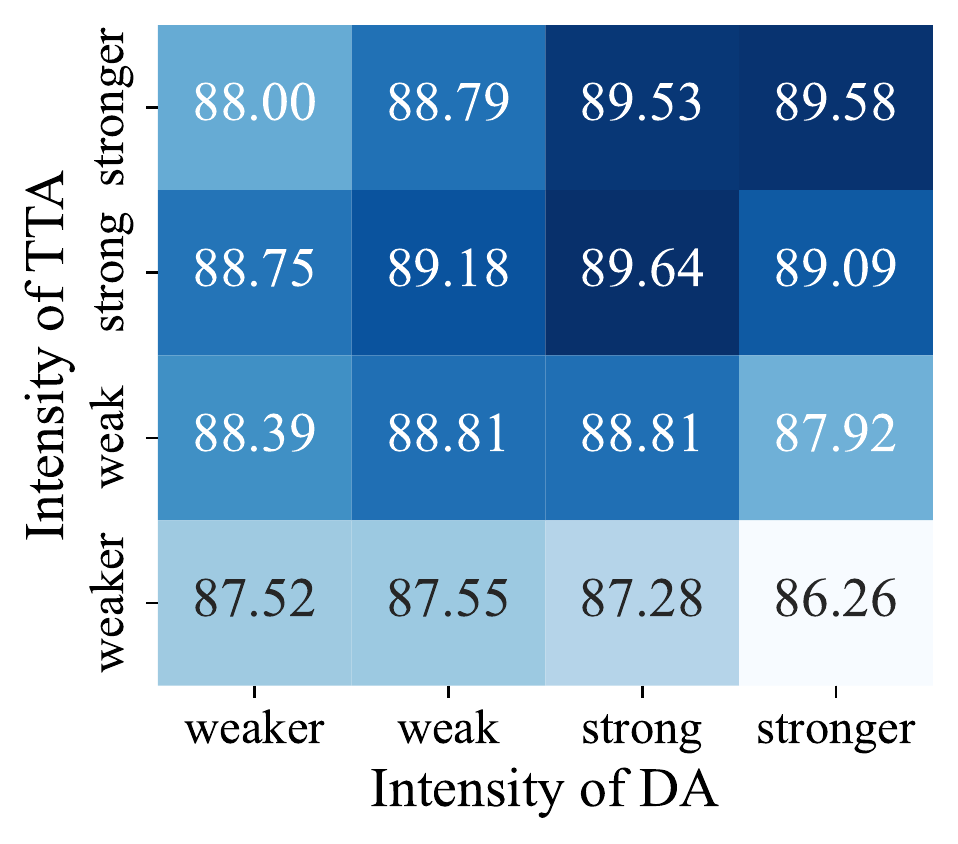}
        \caption{miniIN ($k=20$)}
    \end{subfigure}
    
    \begin{subfigure}[h]{0.325\linewidth}
    \includegraphics[width=\linewidth]{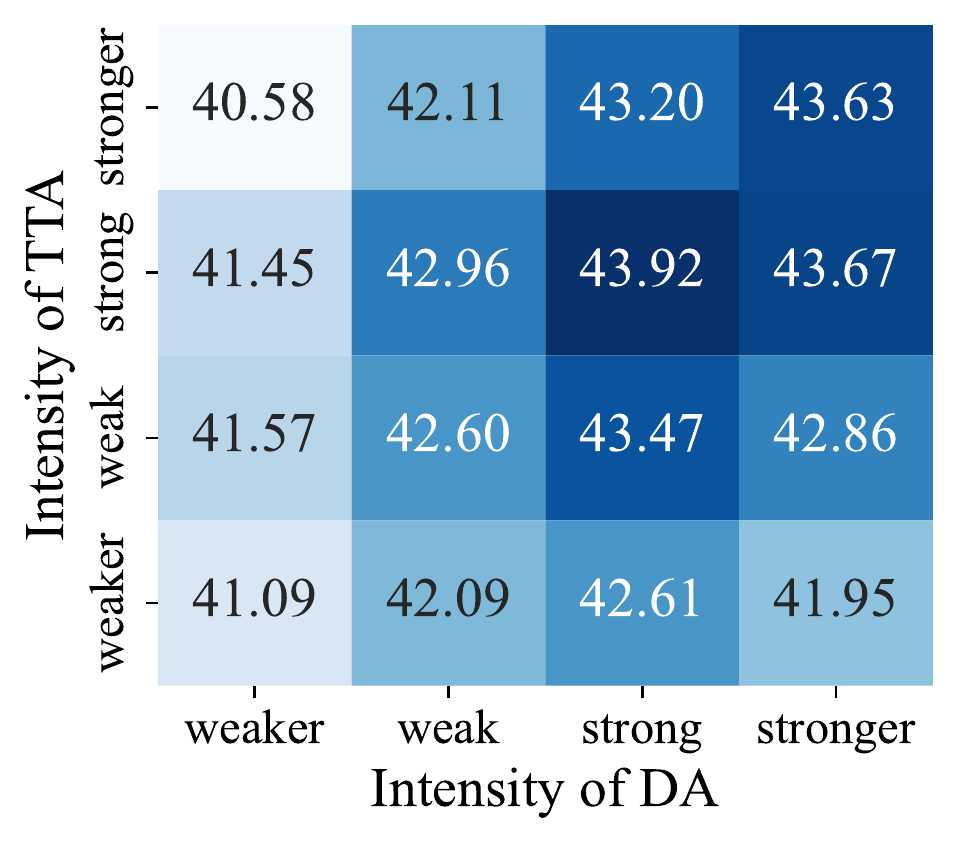}
        \caption{CUB ($k=1$)}
    \end{subfigure}
    \begin{subfigure}[h]{0.325\linewidth}
    \includegraphics[width=\linewidth]{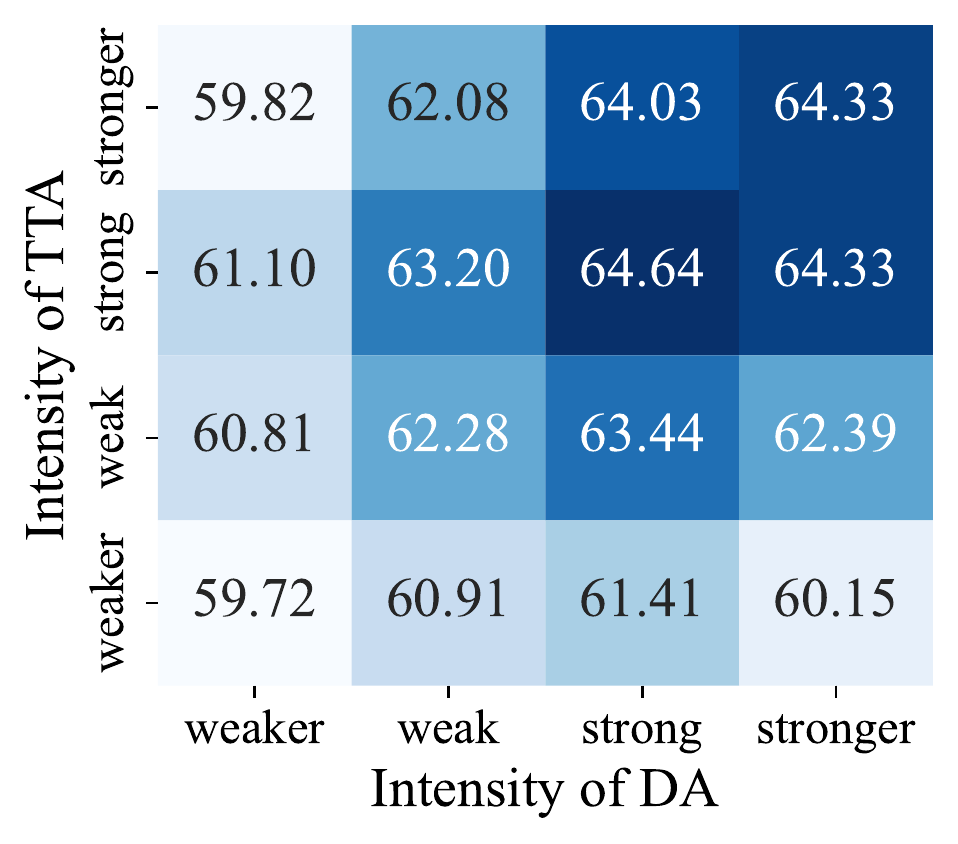}
        \caption{CUB ($k=5$)}
    \end{subfigure}
    \begin{subfigure}[h]{0.325\linewidth}
    \includegraphics[width=\linewidth]{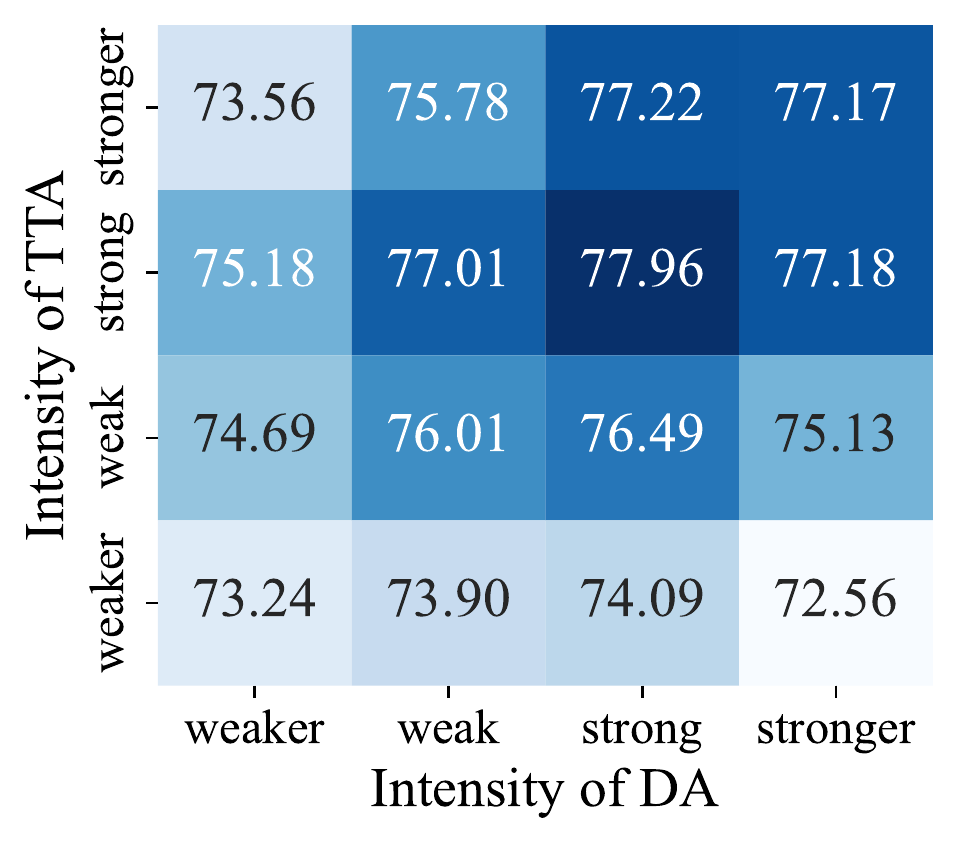}
        \caption{CUB ($k=20$)}
    \end{subfigure}
    \caption{5-way $k$-shot performance according to the designed intensity of DA and TTA. Base Aug is used for both DA and TTA. For TTA, 32 predictions are ensembled.}
    \label{fig:matching_appx}
\end{figure}

\section{The Number of TTA Samples (LP)}\label{appx:abla_num}
\begin{figure}[h!]
    \centering
    \includegraphics[width=\linewidth]{figure/tta/legend2.pdf}
    
    \begin{subfigure}[h]{0.34\linewidth}
    \includegraphics[width=\linewidth]{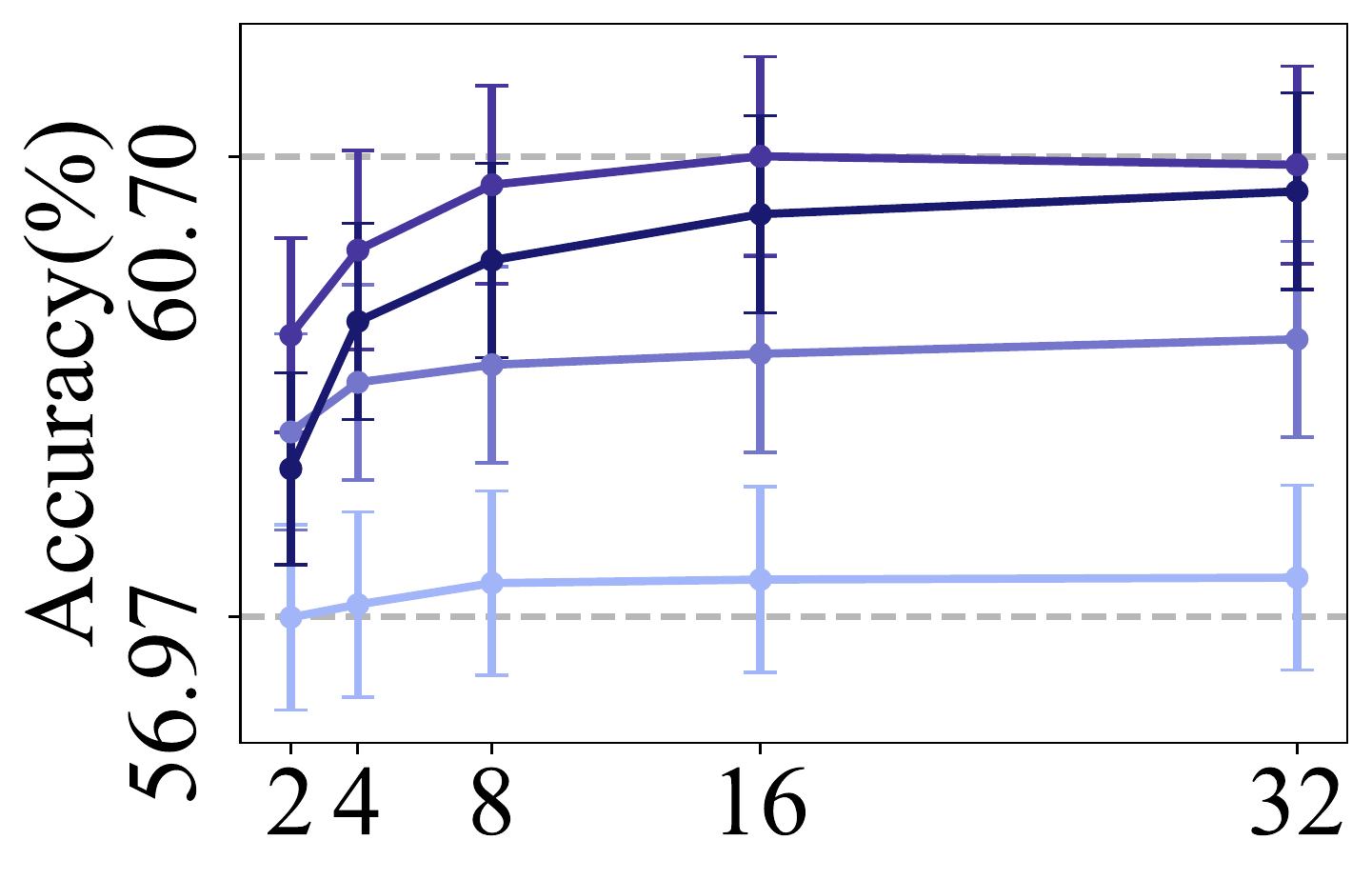}
        \caption{miniIN ($k=1$)}
    \end{subfigure}
    \begin{subfigure}[h]{0.315\linewidth}
    \includegraphics[width=\linewidth]{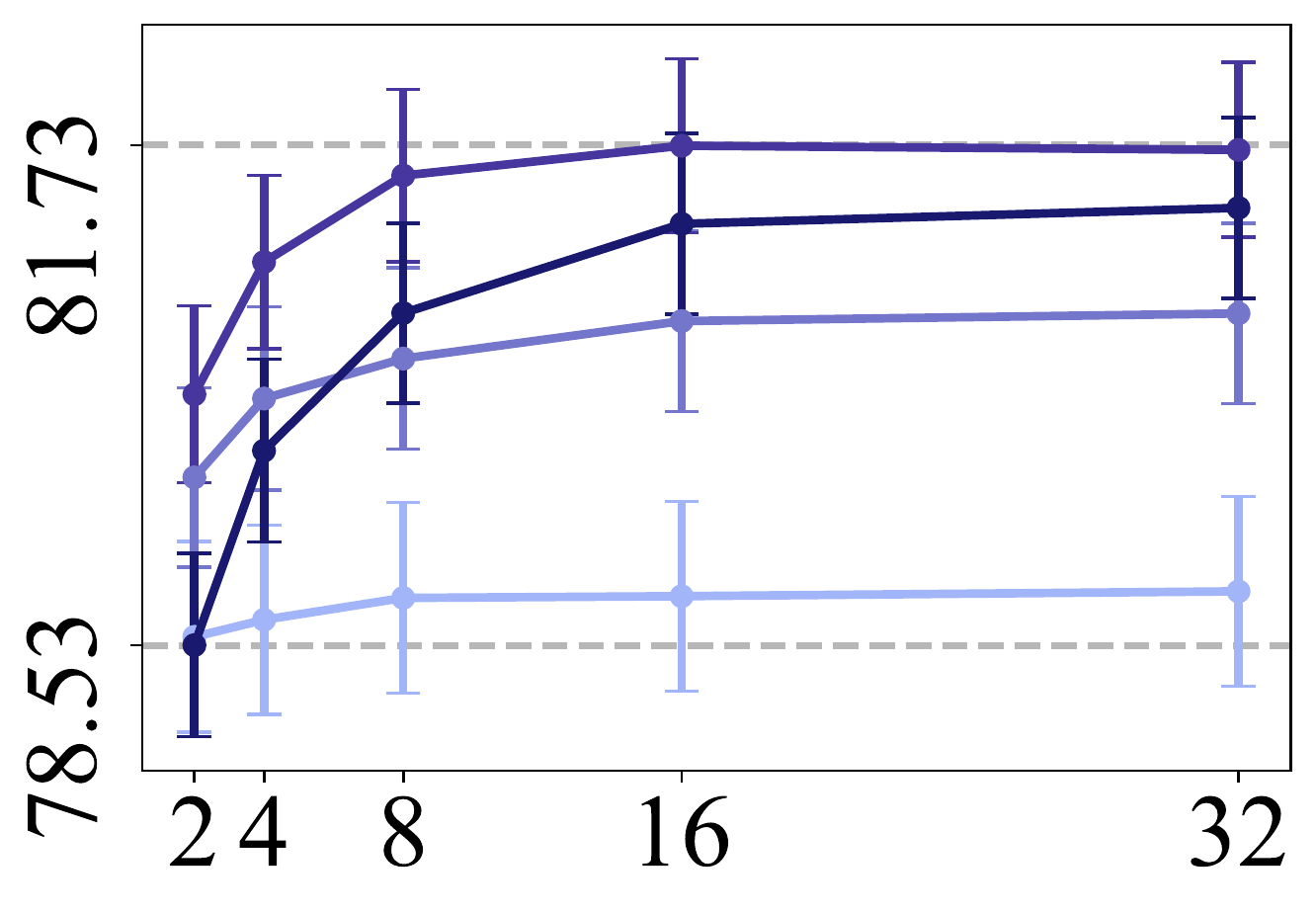}
        \caption{miniIN ($k=5$)}
    \end{subfigure}
    \begin{subfigure}[h]{0.315\linewidth}
    \includegraphics[width=\linewidth]{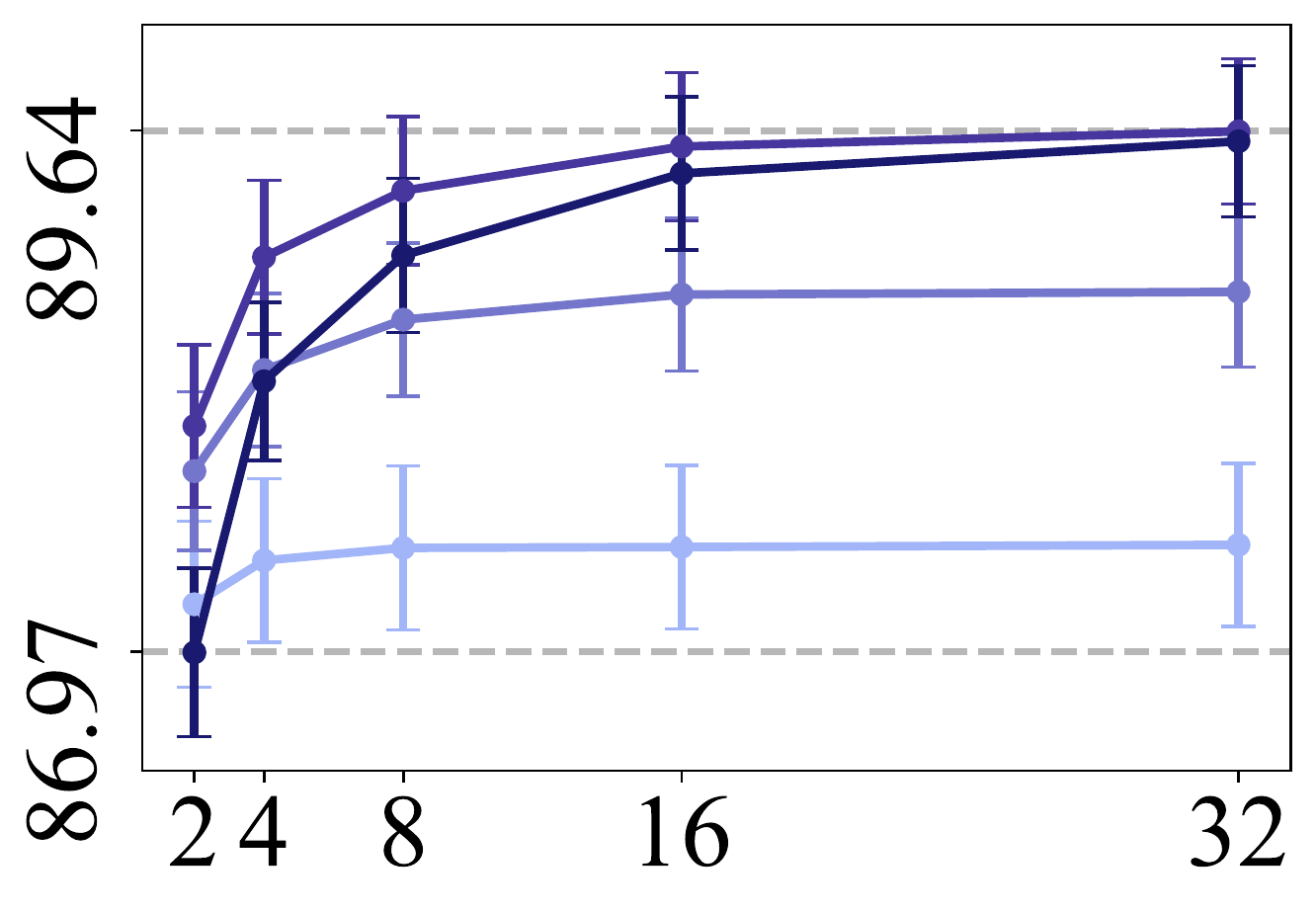}
        \caption{miniIN ($k=20$)}
    \end{subfigure}
    
    \begin{subfigure}[h]{0.34\linewidth}
    \includegraphics[width=\linewidth]{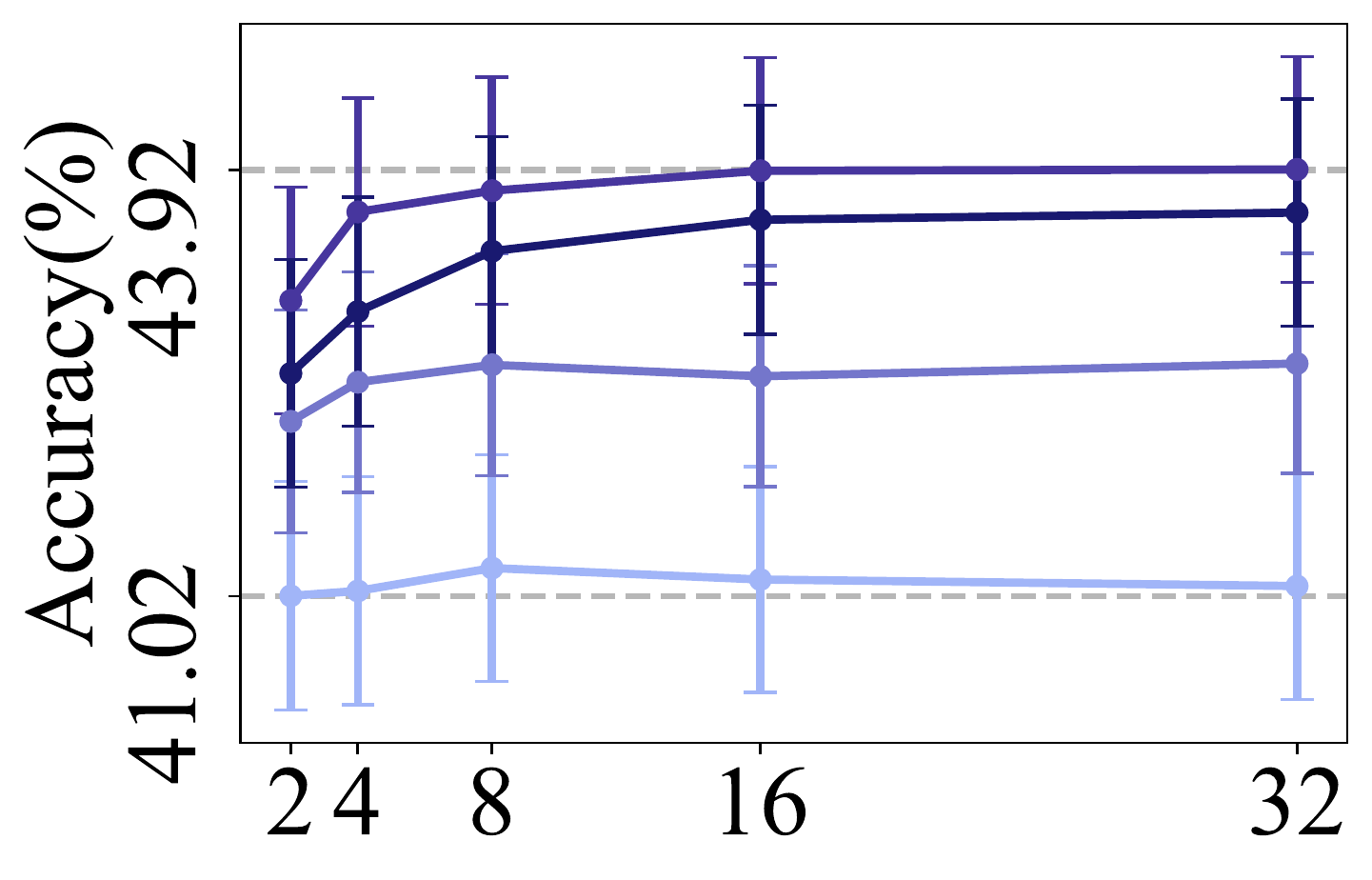}
        \caption{CUB ($k=1$)}
    \end{subfigure}
    \begin{subfigure}[h]{0.315\linewidth}
    \includegraphics[width=\linewidth]{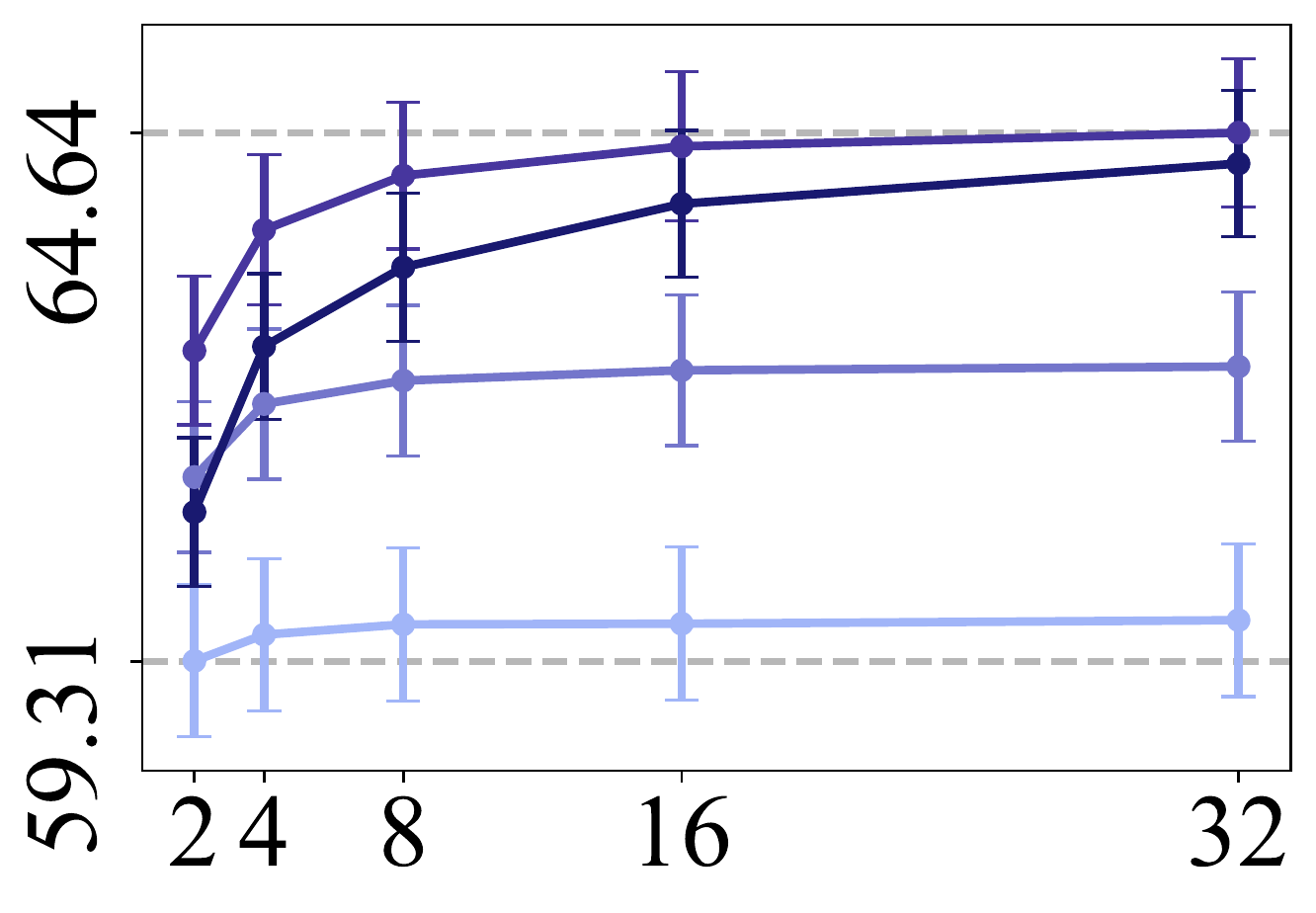}
        \caption{CUB ($k=5$)}
    \end{subfigure}
    \begin{subfigure}[h]{0.315\linewidth}
    \includegraphics[width=\linewidth]{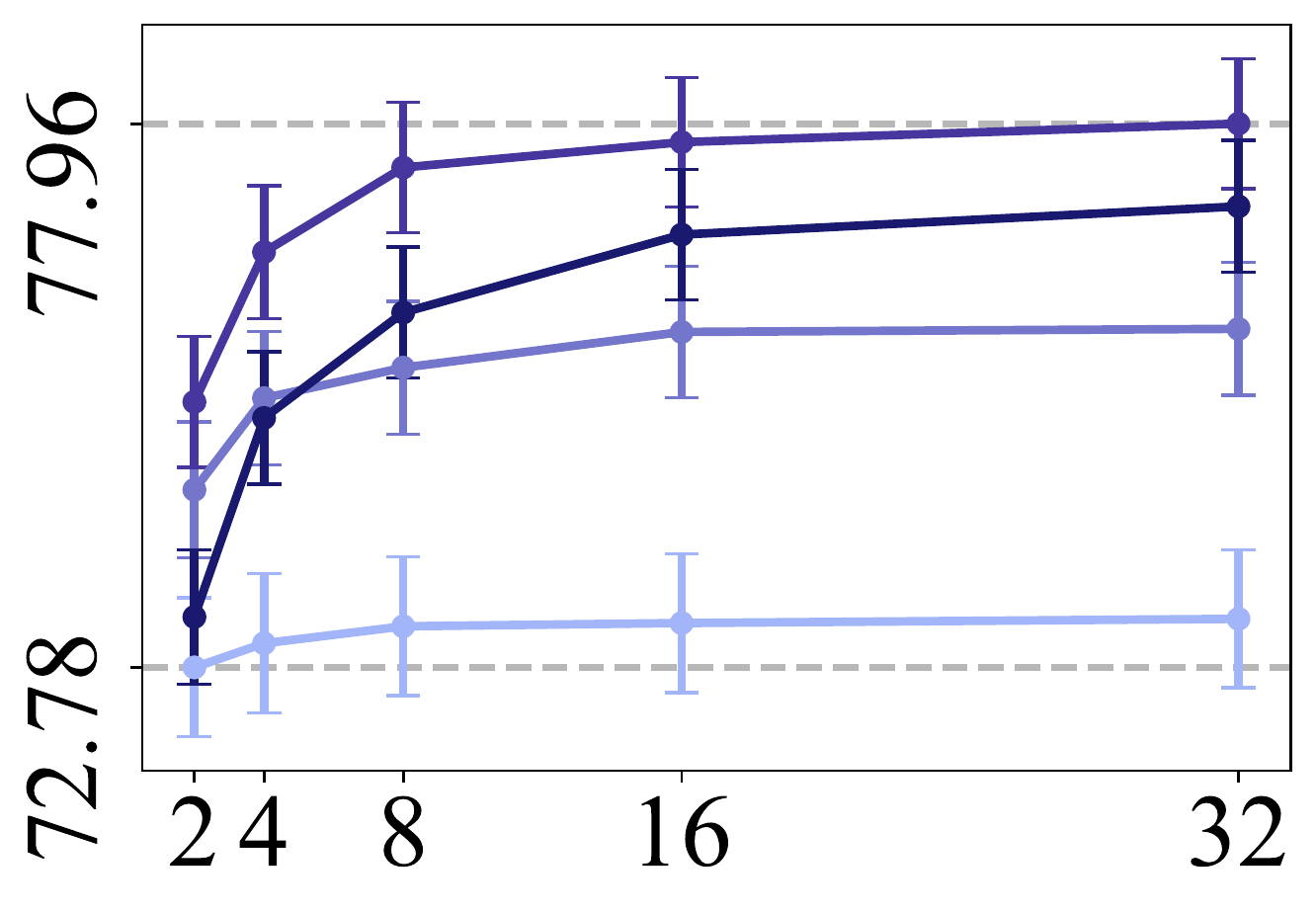}
        \caption{CUB ($k=20$)}
    \end{subfigure}
    \caption{5-way $k$-shot performance according to the number of augmented samples for TTA.}
    \label{fig:tta_abla_appx}
\end{figure}

We investigate the effectiveness of increasing the number of samples for TTA according to the intensity of augmentation, when only a linear classifier is updated. Figure \ref{fig:tta_abla_appx} describes the performance according to the number of augmented samples for TTA using the four matched intensities. 

\section{Related Works on (CD-)FSL}\label{appx:related}
FSL focuses on the underlying problem of deep learning where there are few training samples available. 
There are two dominant approaches for FSL, meta-learning and transfer learning. 
%
Meta-learning aims to design a generalized model that adapts to new tasks rapidly by learning a good initialization\,\cite{finn2017model, li2017meta, lee2018gradient, rajeswaran2019meta, oh2021boil}, learning the metrics\,\cite{koch2015siamese, vinyals2016matching, snell2017prototypical, sung2018learning}, or learning generative models to augment few-shot data\,\cite{antoniou2017data, hariharan2017low}.
In contrast, transfer learning aims to transfer knowledge from the source task to the target few-shot tasks.
Based on the capacity of deep networks trained on the source task, pre-trained knowledge is effectively transferred to different target tasks\,\cite{chen2019closer, dhillon2019baseline, rodriguez2020embedding, yazdanpanah2021shift, shen2021partial}.

Cross-domain few-shot learning (CD-FSL), where there are large distributional differences between the source and target data, has gained increased attention for more realistic FSL.
\citet{guo2020broader} provided CD-FSL benchmark datasets and demonstrated that transfer learning-based approaches achieved higher performances than existing meta-learning approaches. 
Therefore, to address a distribution shift challenge, prior works have focused on transfer learning-based approaches\,\cite{luo2017label,phoo2020self,islam2021dynamic,das2021confess}.

\section{Discussions}\label{appx:discussion}
In this section, we cover additional discussions on updating the model for few-shot learning.

\subsection{Performance Variation between Episodes}\label{appx:instab_with}

The evaluation in FSL is conducted by averaging the accuracies of 600 episodes. However, the variation of performance between these episodes is substantial. For example, the performance of LP fine-tuned with tieredIN varies from 18.67$\%$ to 93.33$\%$. 

\begin{figure}[h!]
    \centering
    
    \begin{subfigure}[h]{\linewidth}
        \includegraphics[width=\linewidth]{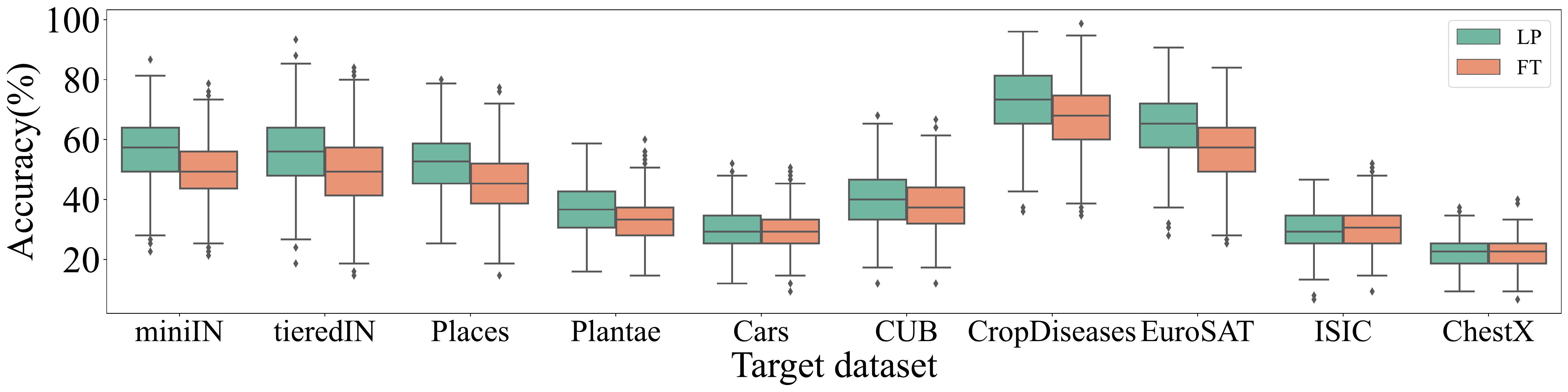}
    \caption{$k=1$}
    \label{fig:1shot_btw}
    \end{subfigure}
    
    \begin{subfigure}[h]{\linewidth}
        \includegraphics[width=\linewidth]{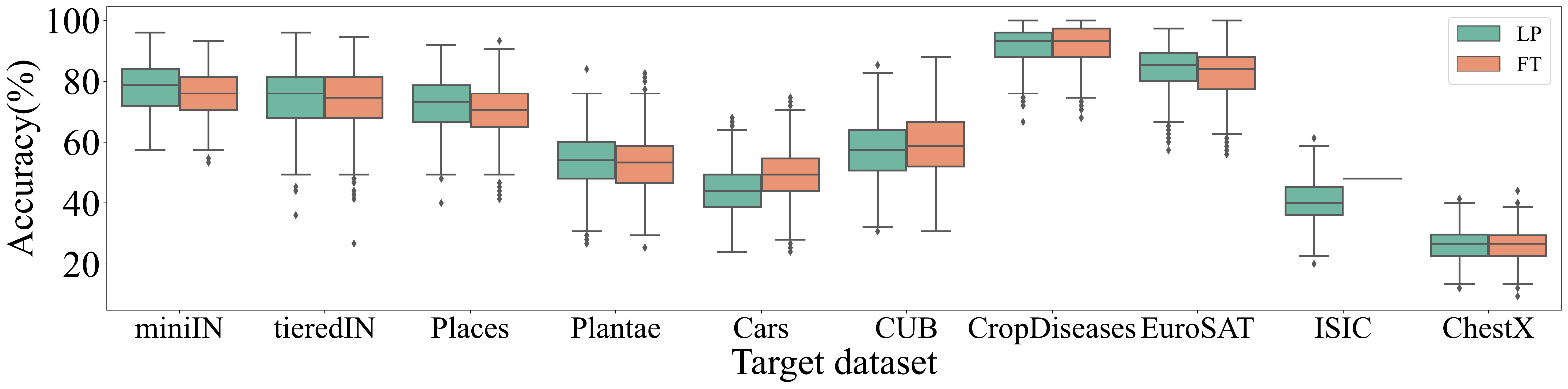}
    \caption{$k=5$}
    \label{fig:5shot_btw}
    \end{subfigure}

    \begin{subfigure}[h]{\linewidth}
        \includegraphics[width=\linewidth]{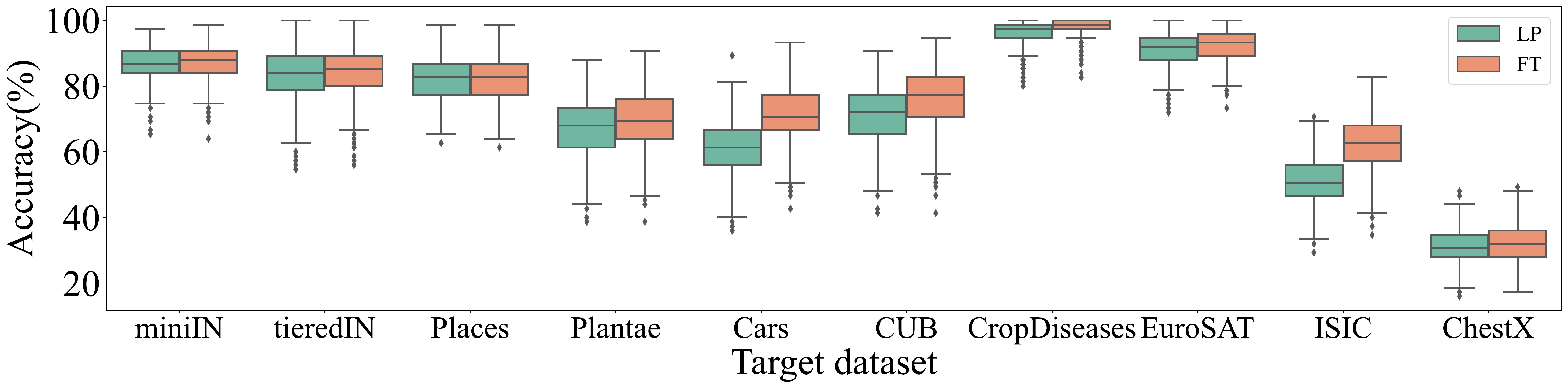}
    \caption{$k=20$}
    \label{fig:20shot_btw}
    \end{subfigure}
    \caption{5-way $k$-shot performance of LP and FT over 600 episodes according to different target dataset. The pre-trained ResNet10 on miniIN is used as the feature extractor.}
    \label{fig:between_instab}
\end{figure}

In Figure \ref{fig:between_instab}, we visualize this variation according to target dataset and updating method\,(i.e., FT or LP), where ResNet10 pre-trained on miniIN is used for $k\in \{1, 5, 20\}$. The green and orange colors indicate performance through LP and FT, respectively. The degrees of performance variation of LP and FT are similar, which means that the variation does not come from updating methods.  

We believe that this variation comes from many factors such as classifier initialization, data ordering for updating (similar to \citep{dodge2020fine}), and combinations of support and target sets.
%

    
    


\subsection{Instability within an Episode}\label{appx:instab_btw}
For transfer-based FSL, the model updated for 100 epochs is used for evaluation without early stopping using the validation set.
This is because \emph{it is difficult to construct the validation set} due to scarcity of the training set.

\begin{figure}[h!]
    \centering
    
    \begin{subfigure}[h]{\linewidth}
        \includegraphics[width=\linewidth]{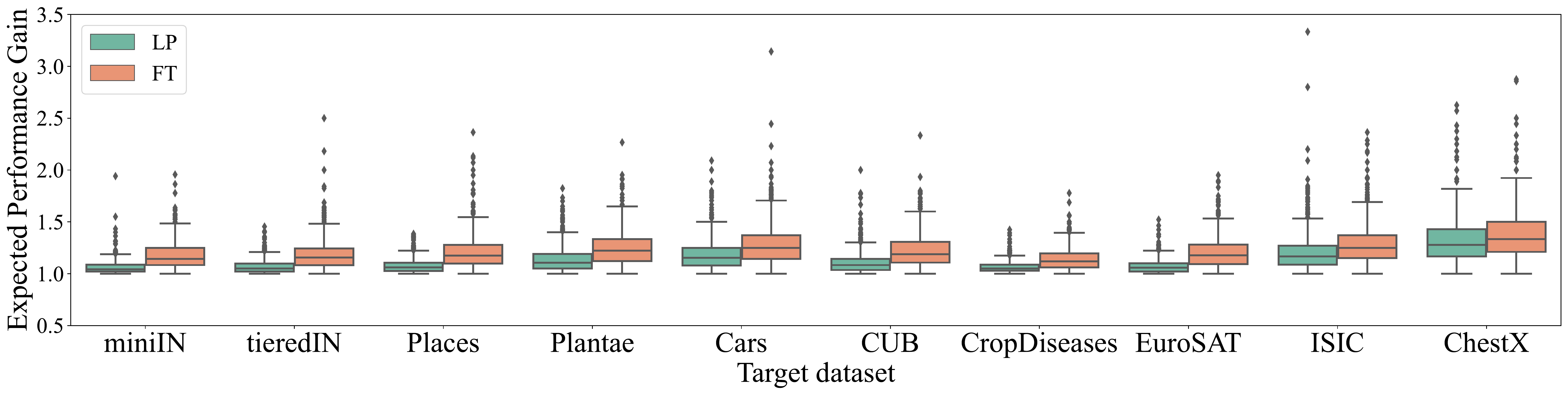}
    \caption{$k=1$}
    \end{subfigure}
    
    \begin{subfigure}[h]{\linewidth}
        \includegraphics[width=\linewidth]{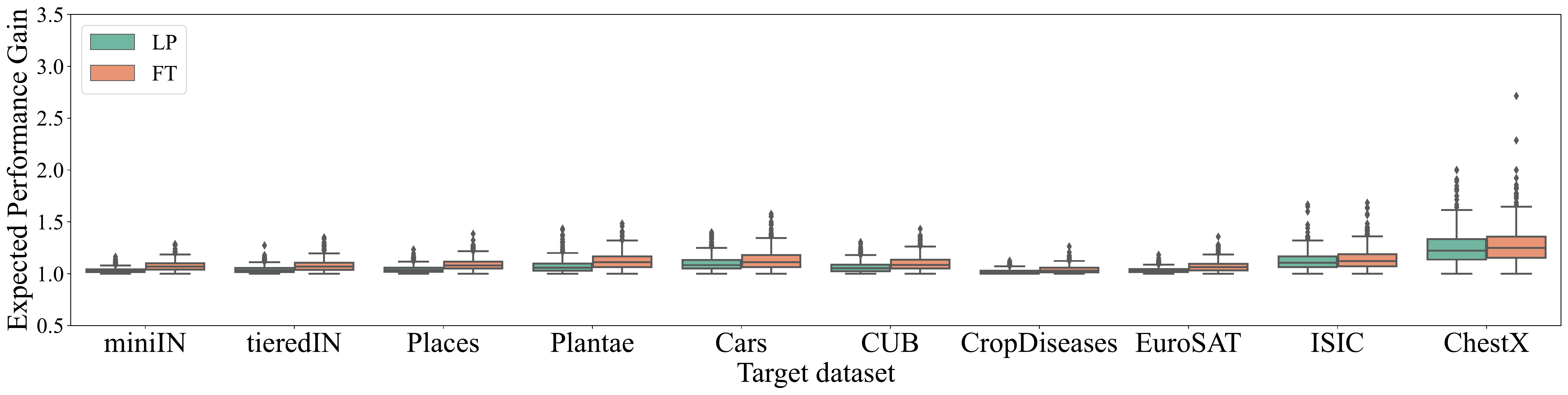}
    \caption{$k=5$}
    \end{subfigure}
    
    \begin{subfigure}[h]{\linewidth}
        \includegraphics[width=\linewidth]{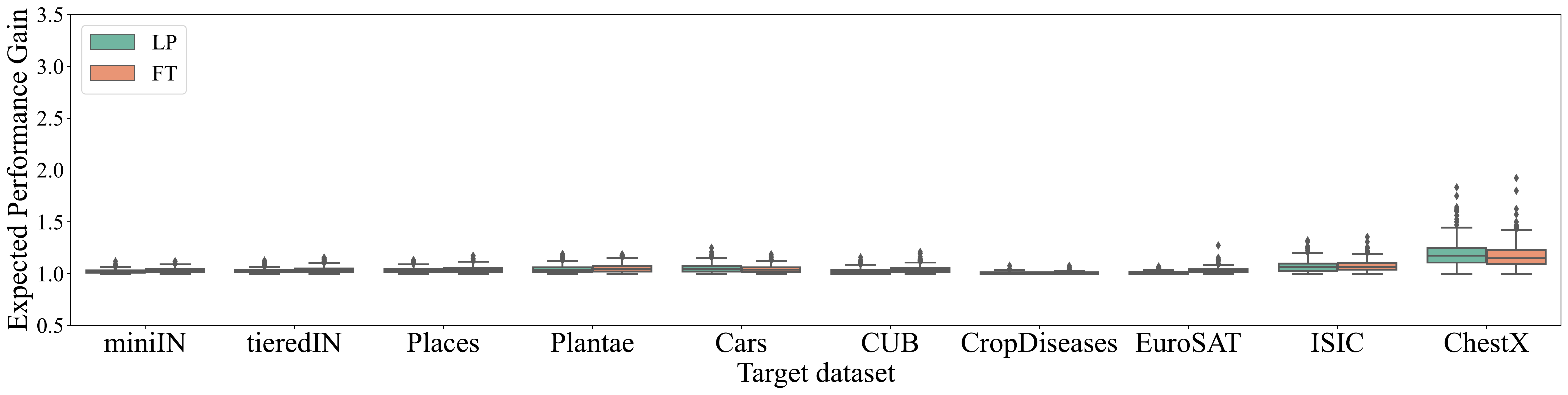}
    \caption{$k=20$}
    \end{subfigure}
    \caption{5-way $k$-shot expected performance gain ($ {\rm Acc}_{\sf best} / {\rm Acc}_{\sf last} $) over 600 episodes.}
    \label{fig:within_instab}
\end{figure}

Figure \ref{fig:within_instab} describes the expected performance gain over 600 episodes according to the target dataset.
The expected performance gain is defined by the ratio between accuracy at the best epoch and accuracy at the last epoch, i.e., $ {\rm Acc}_{\sf best} / {\rm Acc}_{\sf last} $.
%
%
The expected performance gain is amplified when the sample size is smaller (i.e., $k$=1) and when the similarity between the source and target data is bigger (i.e., ISIC and ChestX).

\begin{figure}[h!]
    \centering
    \begin{subfigure}[h]{0.19\linewidth}
             \includegraphics[width=\linewidth]{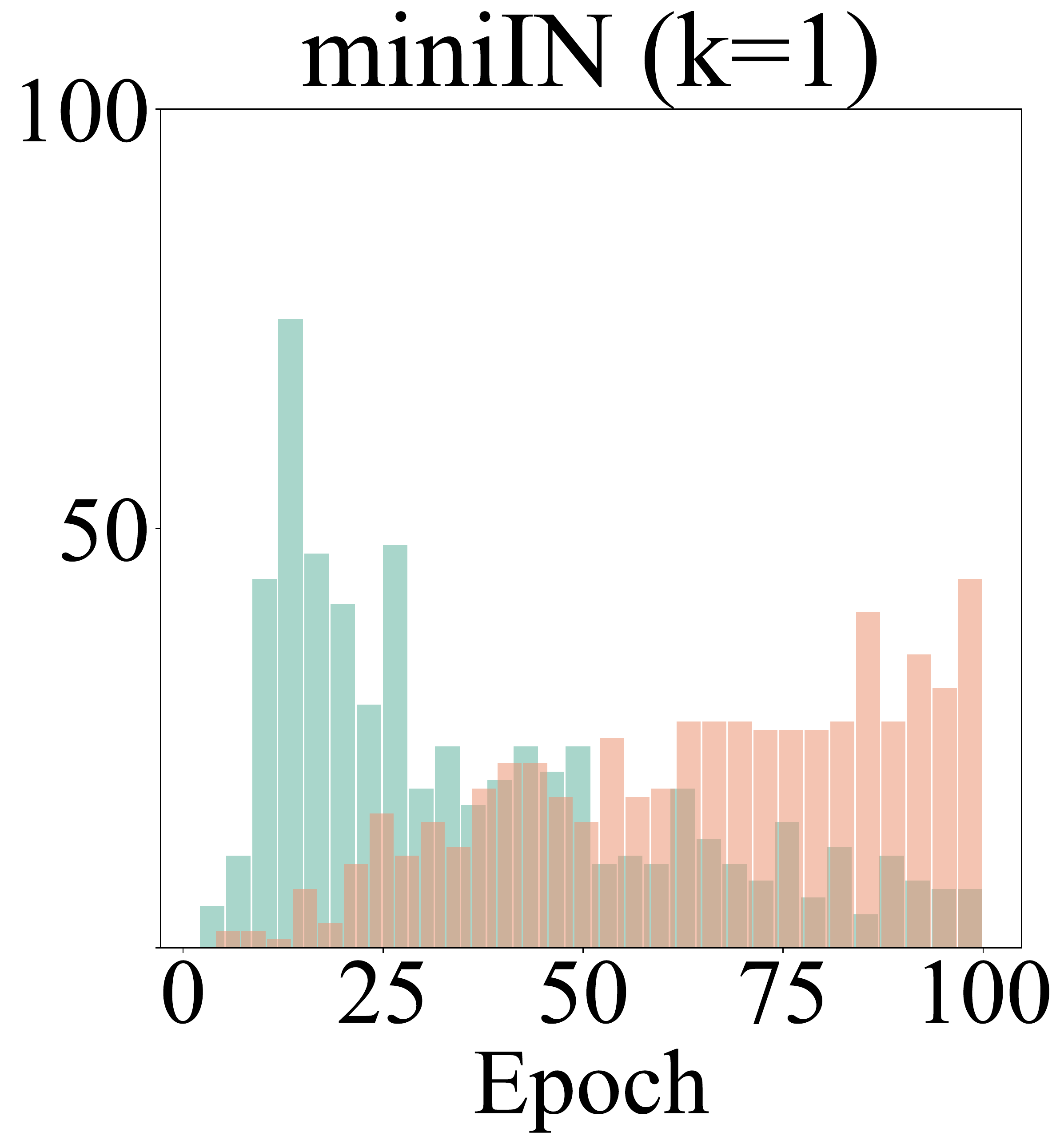}
         \end{subfigure}
     \begin{subfigure}[h]{0.19\linewidth}
         \includegraphics[width=\linewidth]{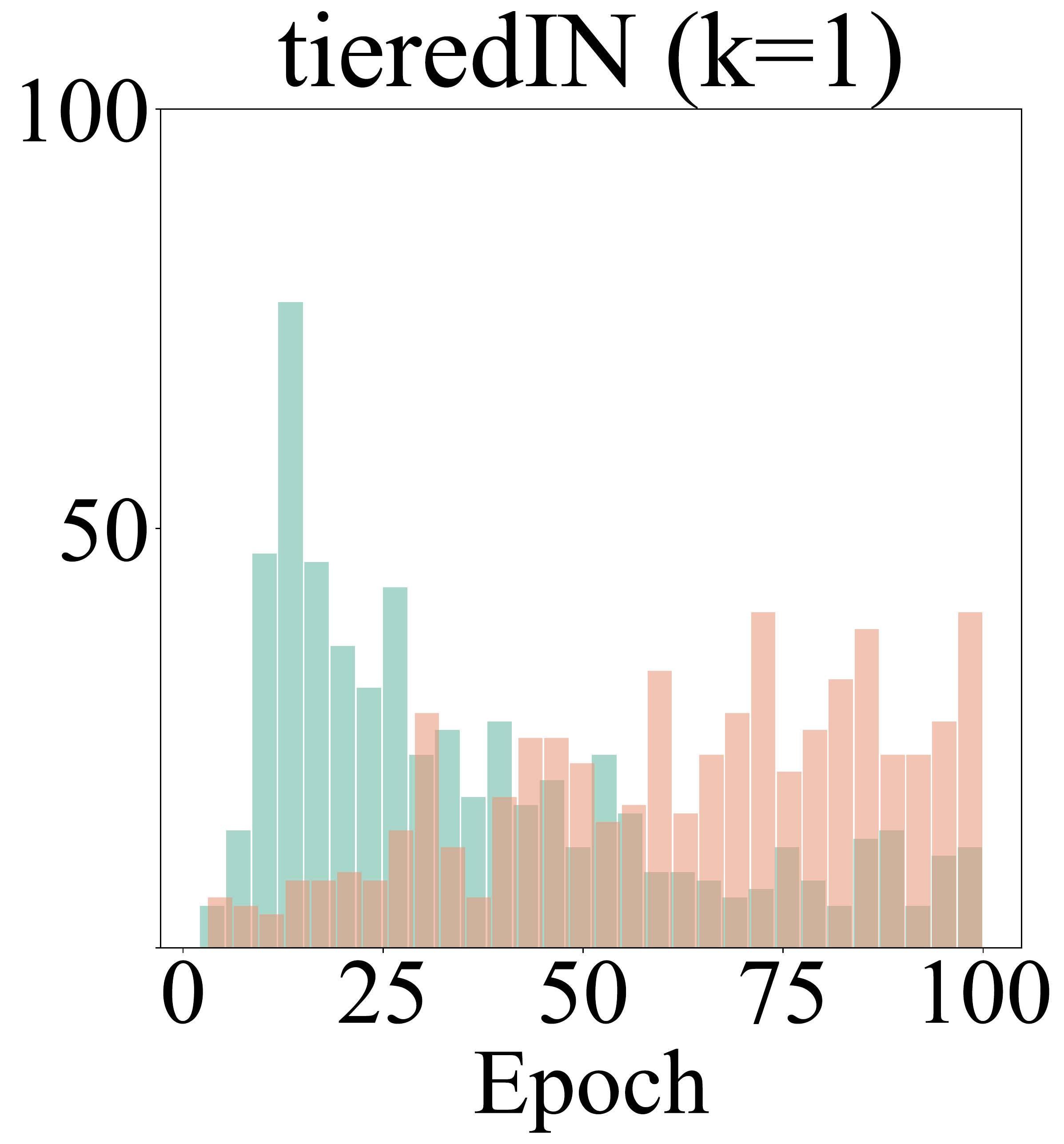}
     \end{subfigure}
     \begin{subfigure}[h]{0.19\linewidth}
         \includegraphics[width=\linewidth]{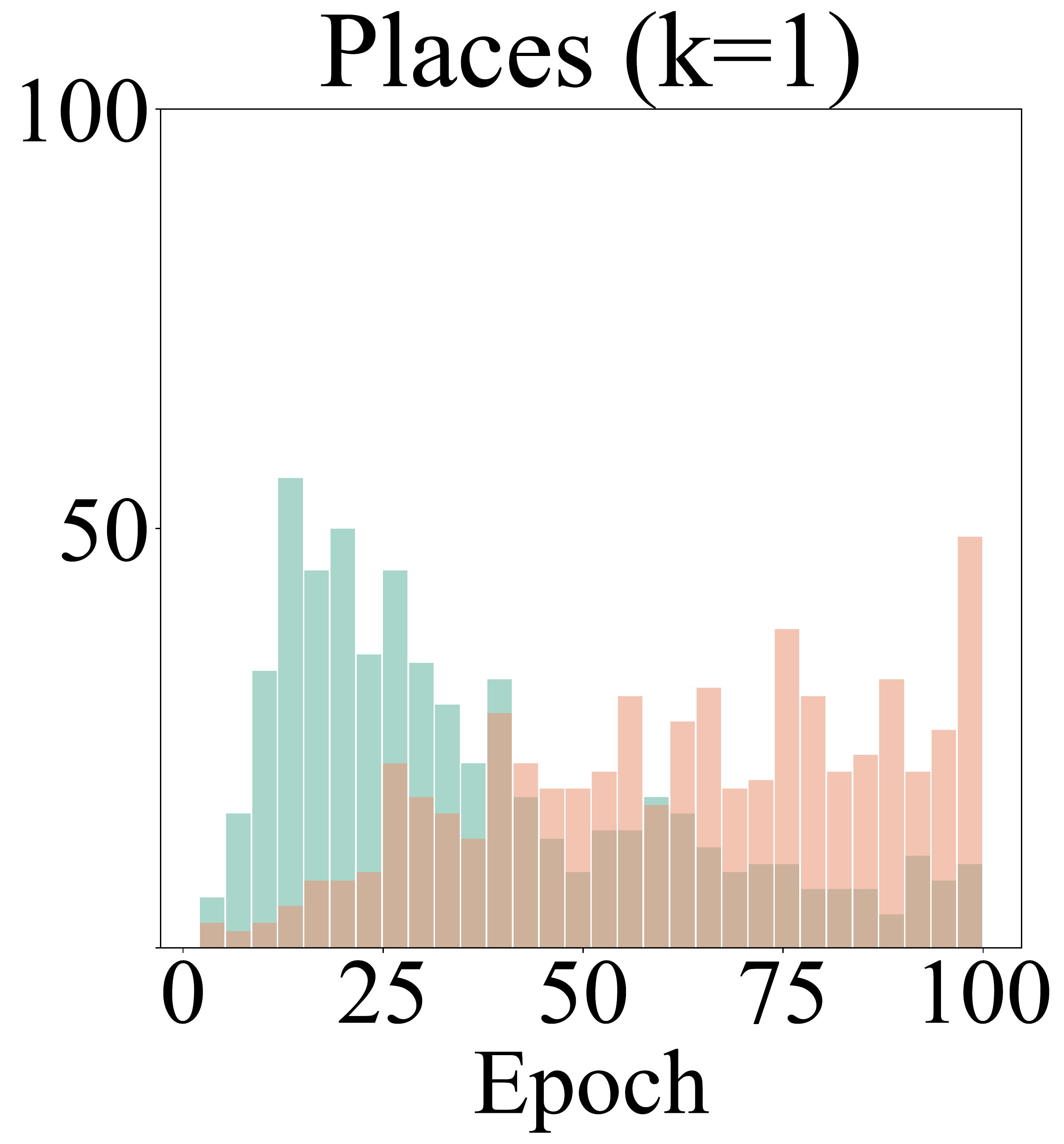}
     \end{subfigure}
     \begin{subfigure}[h]{0.19\linewidth}
         \includegraphics[width=\linewidth]{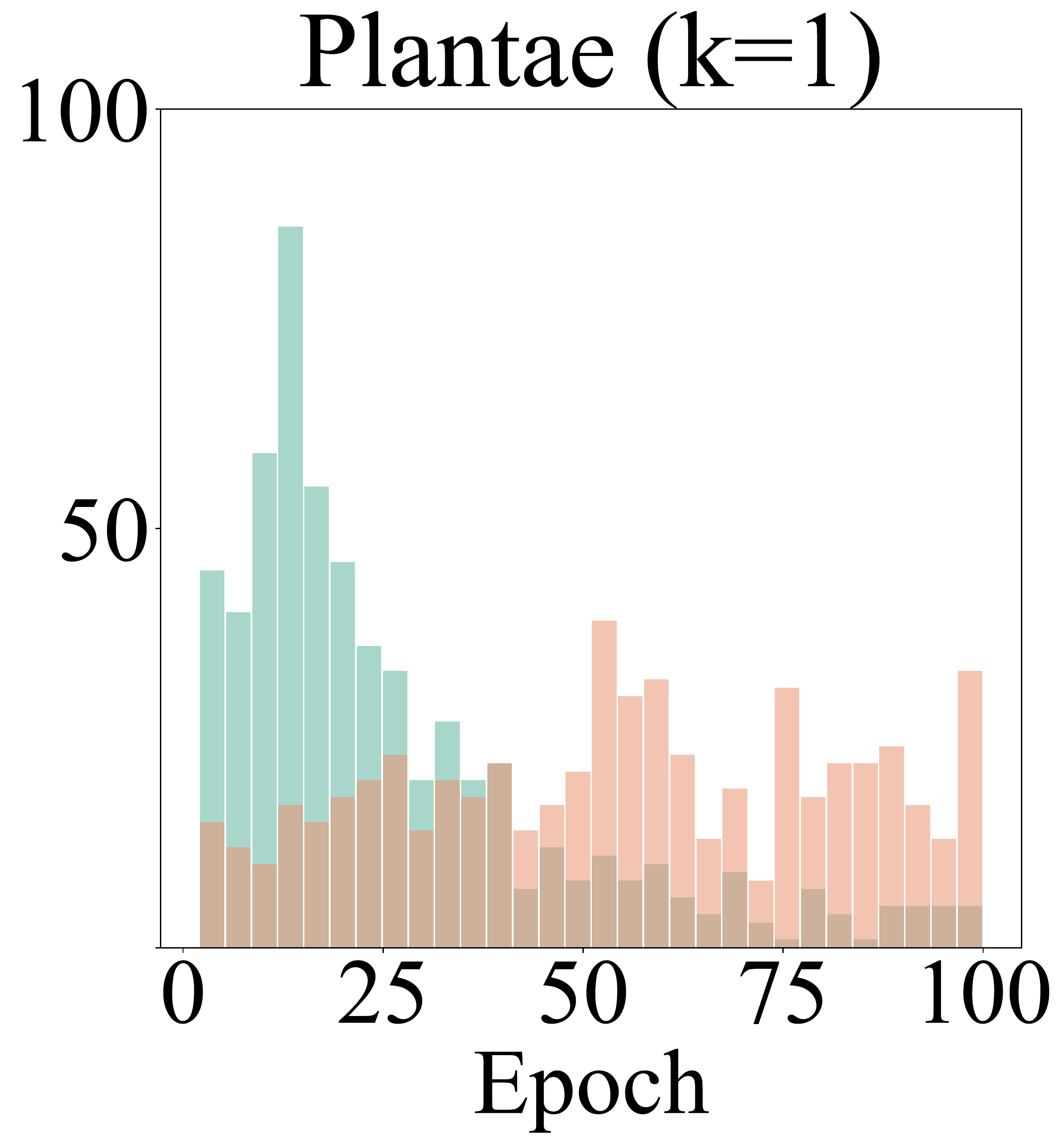}
     \end{subfigure}
     \begin{subfigure}[h]{0.19\linewidth}
         \includegraphics[width=\linewidth]{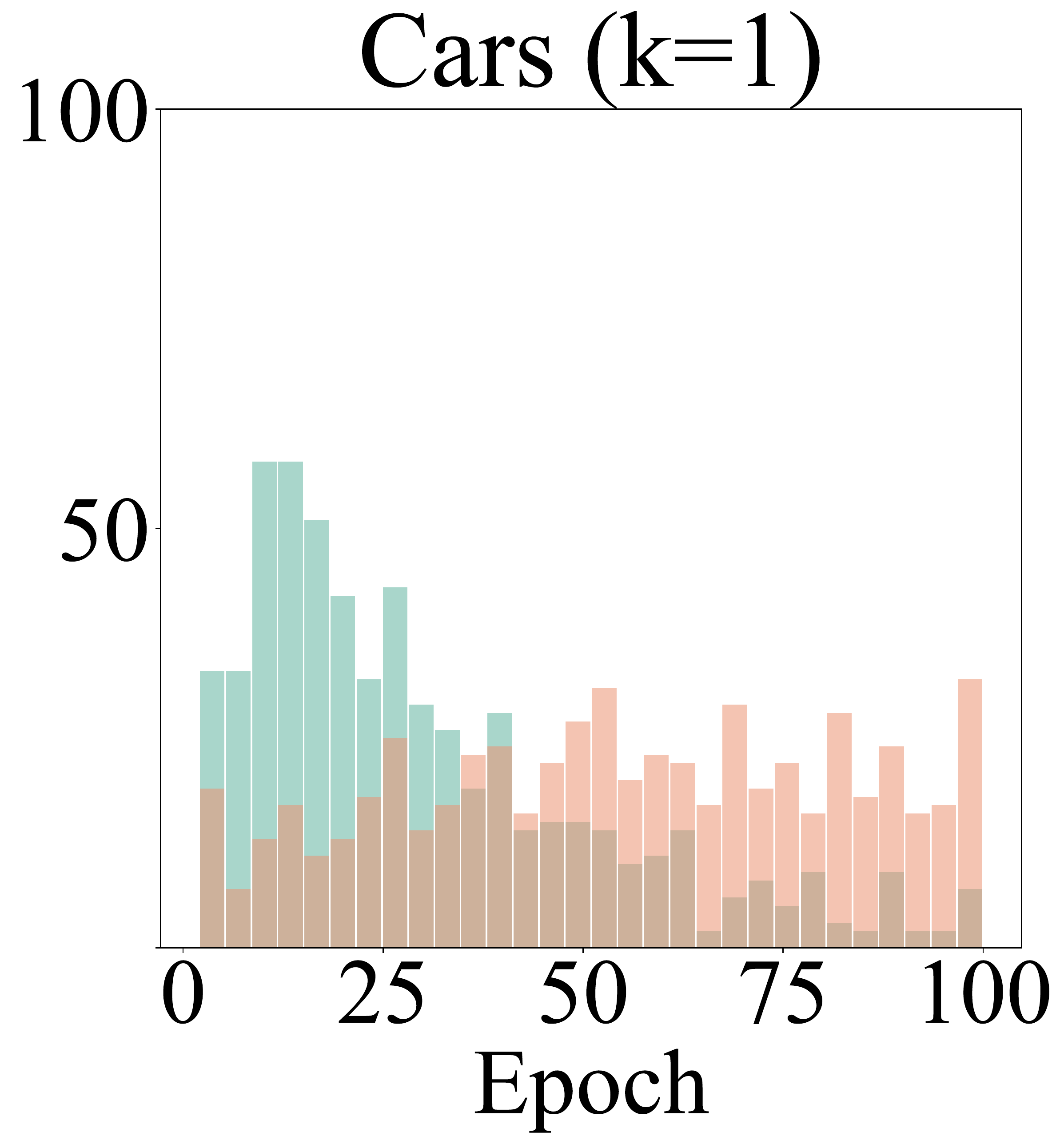}
     \end{subfigure}
     
     \begin{subfigure}[h]{0.19\linewidth}
             \includegraphics[width=\linewidth]{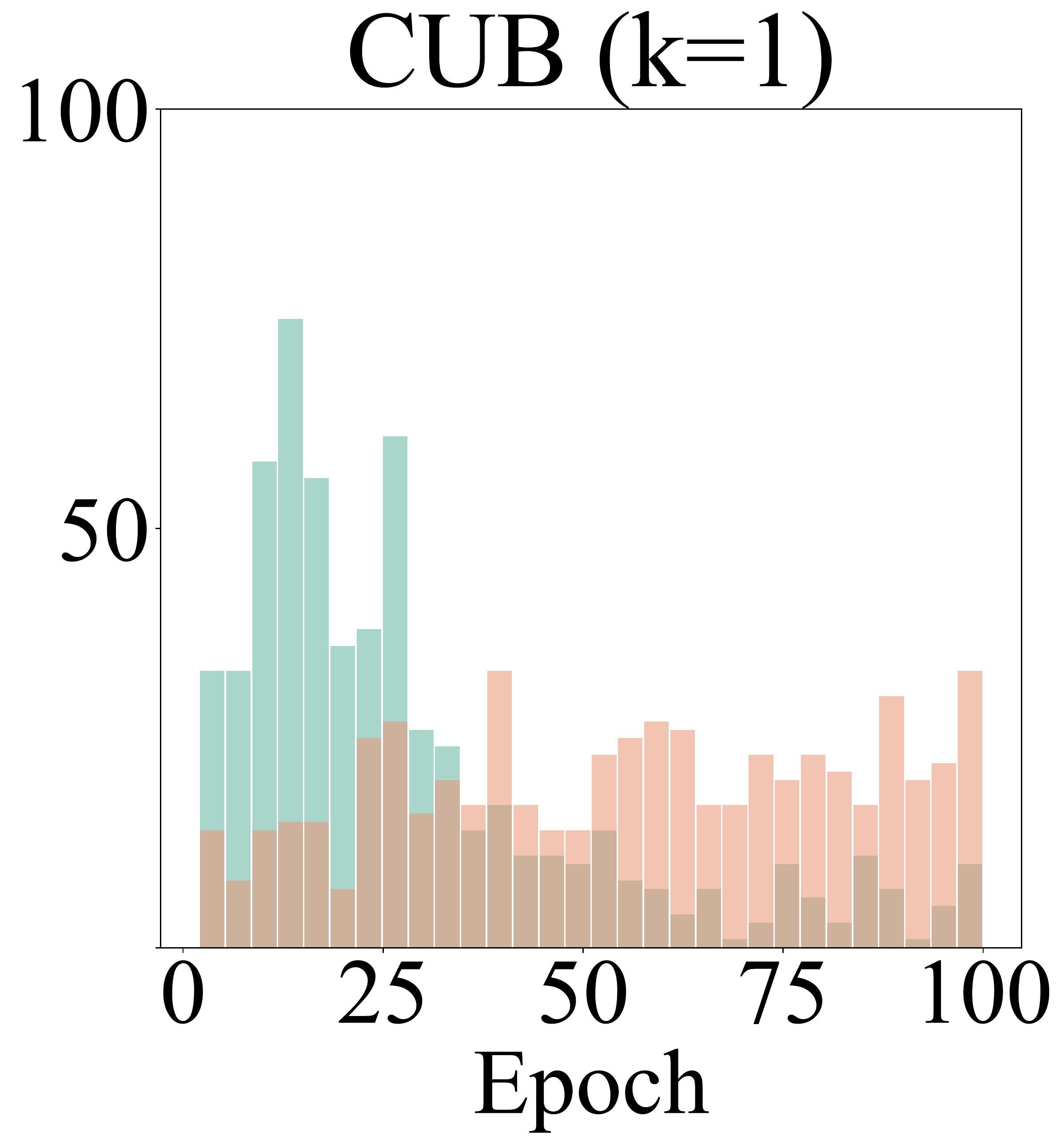}
         \end{subfigure}
     \begin{subfigure}[h]{0.19\linewidth}
         \includegraphics[width=\linewidth]{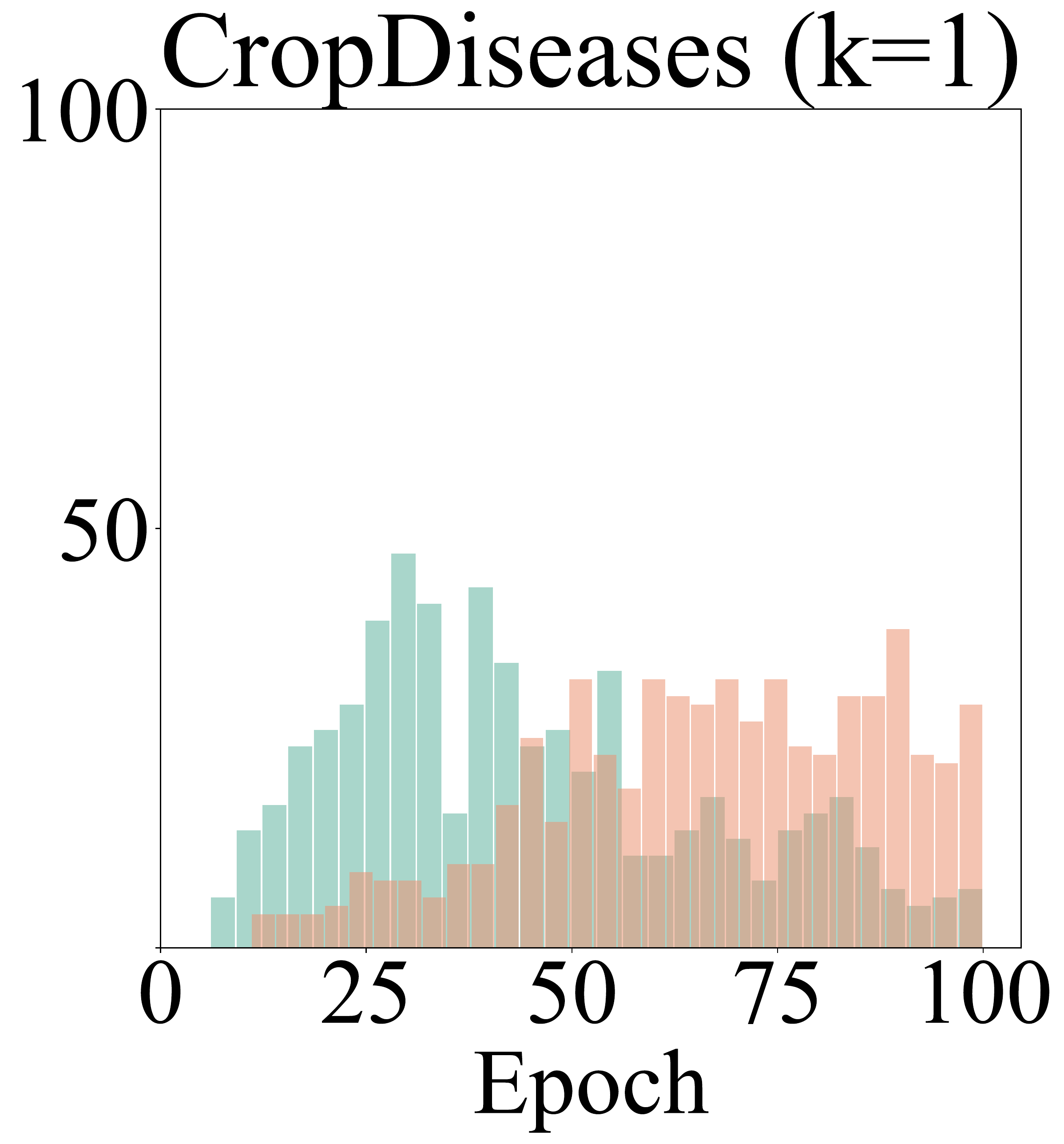}
     \end{subfigure}
     \begin{subfigure}[h]{0.19\linewidth}
         \includegraphics[width=\linewidth]{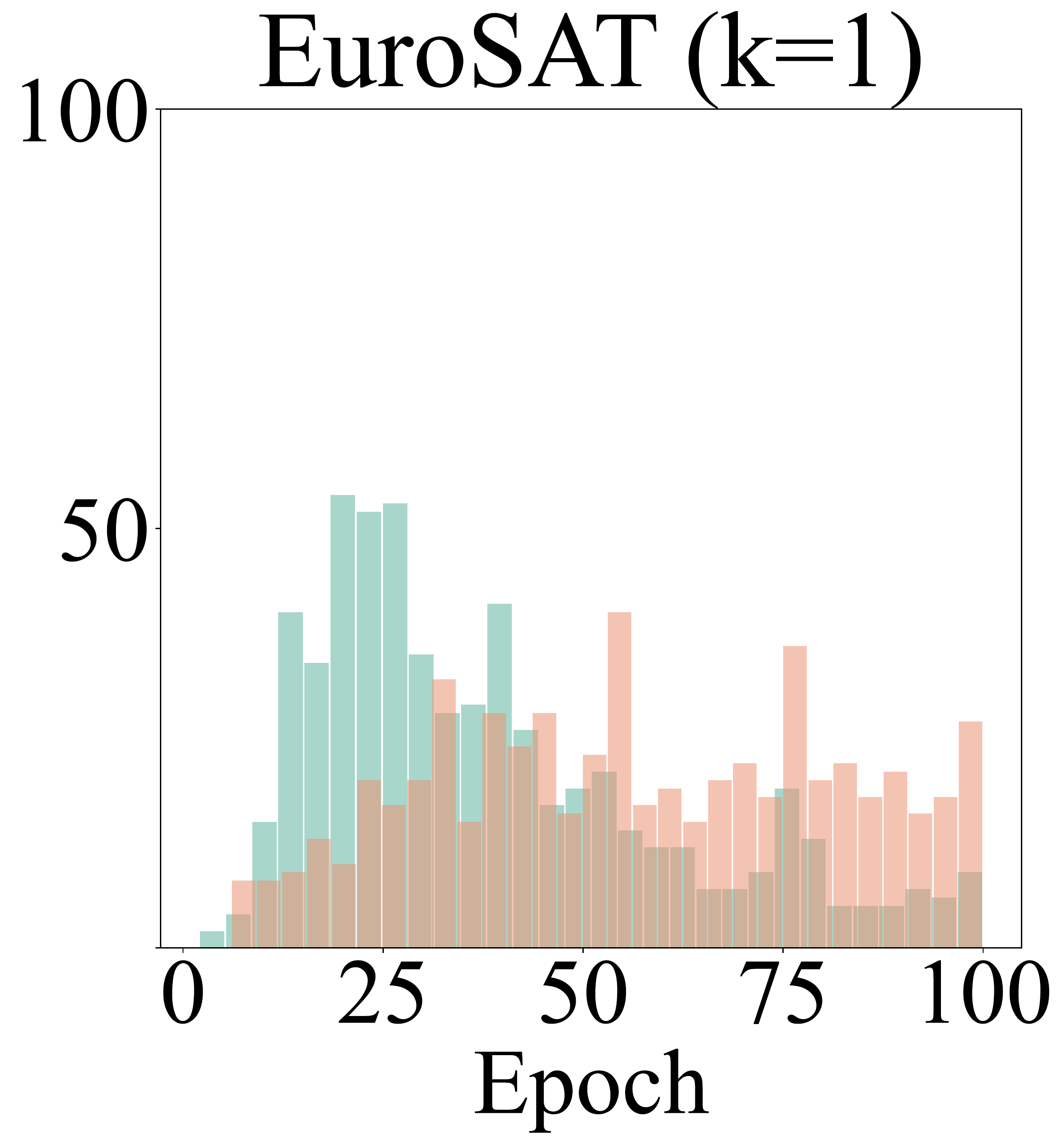}
     \end{subfigure}
     \begin{subfigure}[h]{0.19\linewidth}
         \includegraphics[width=\linewidth]{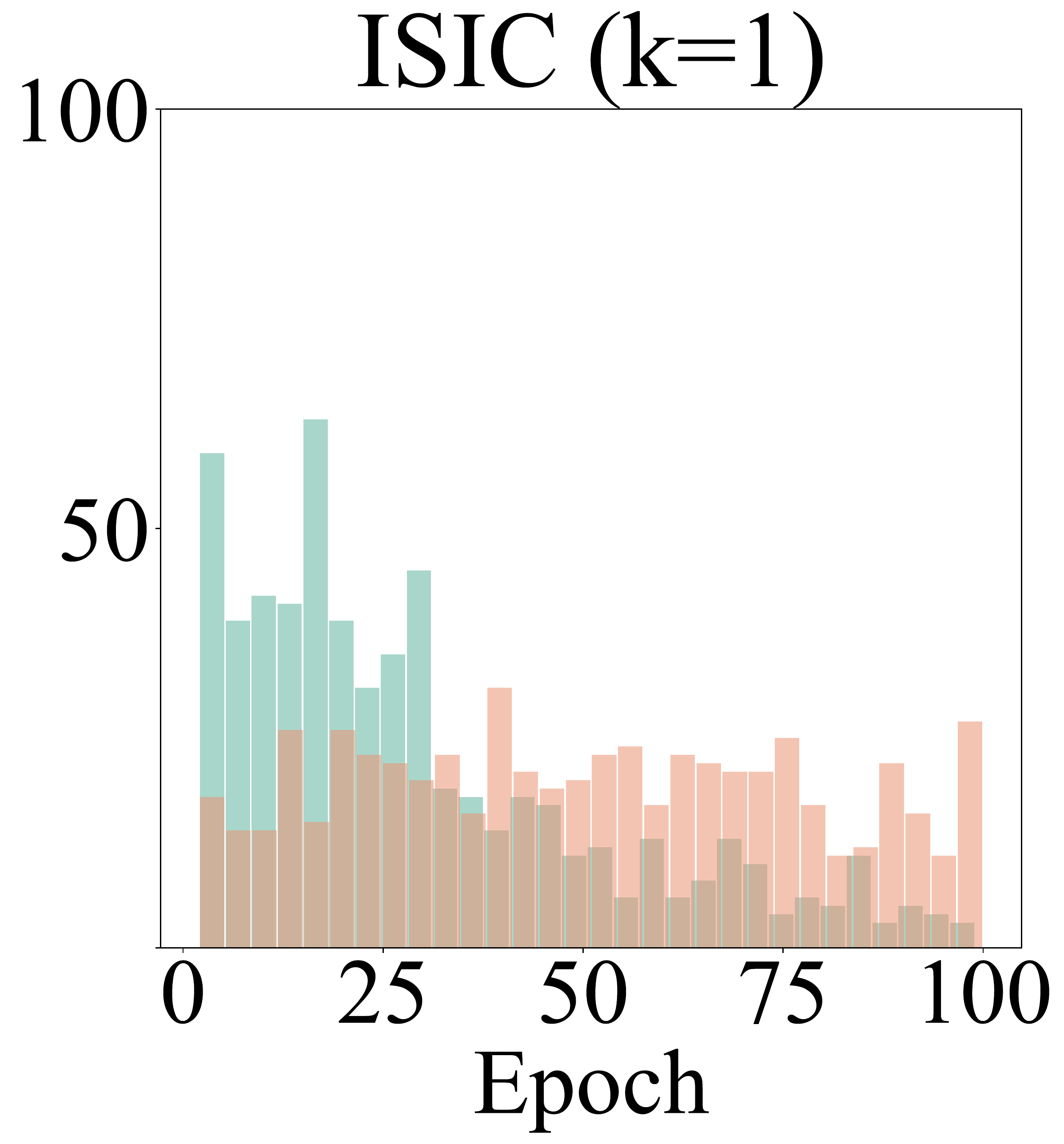}
     \end{subfigure}
     \begin{subfigure}[h]{0.19\linewidth}
         \includegraphics[width=\linewidth]{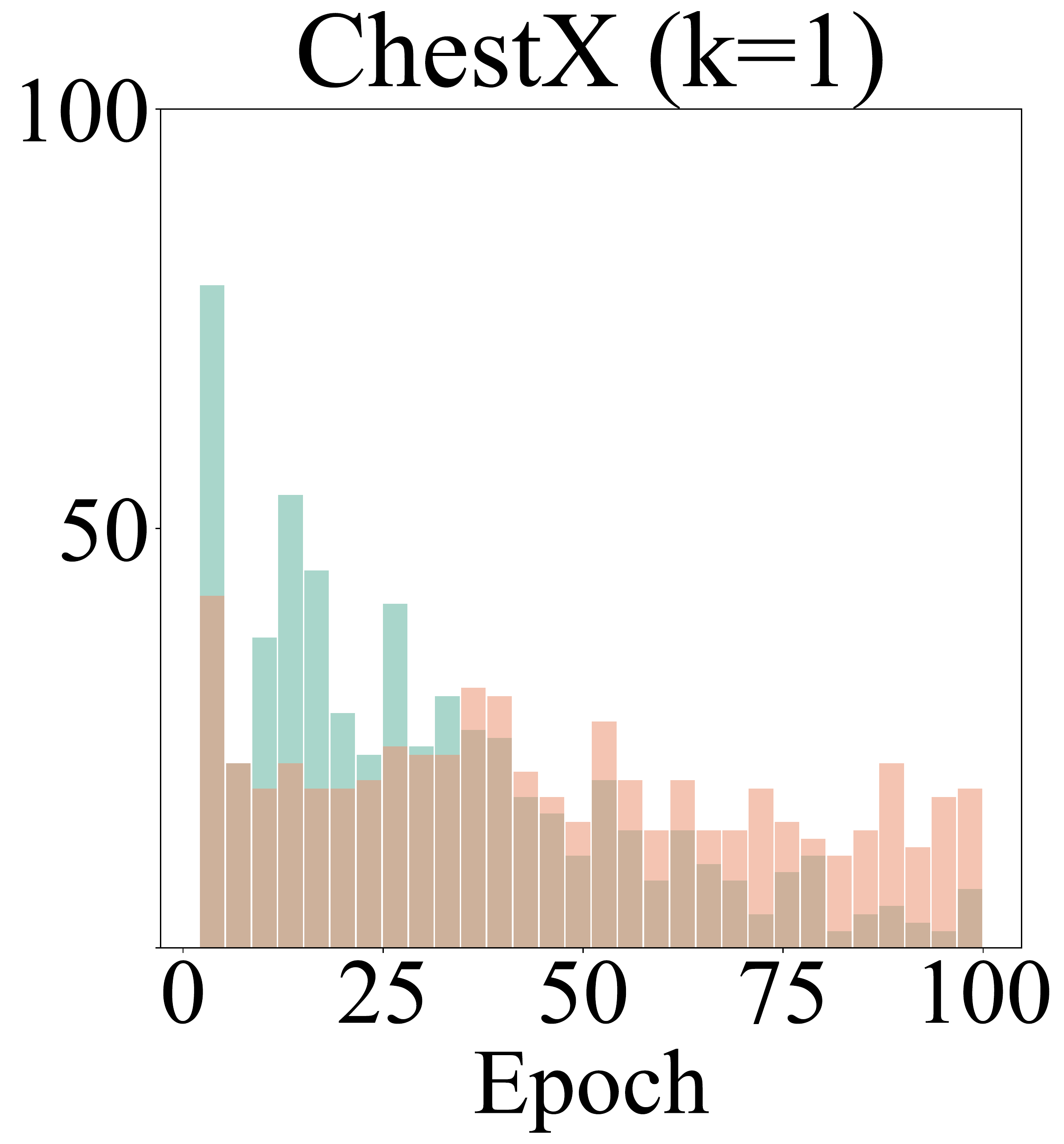}
     \end{subfigure}
    
     \hrulefill

     \begin{subfigure}[h]{0.19\linewidth}
             \includegraphics[width=\linewidth]{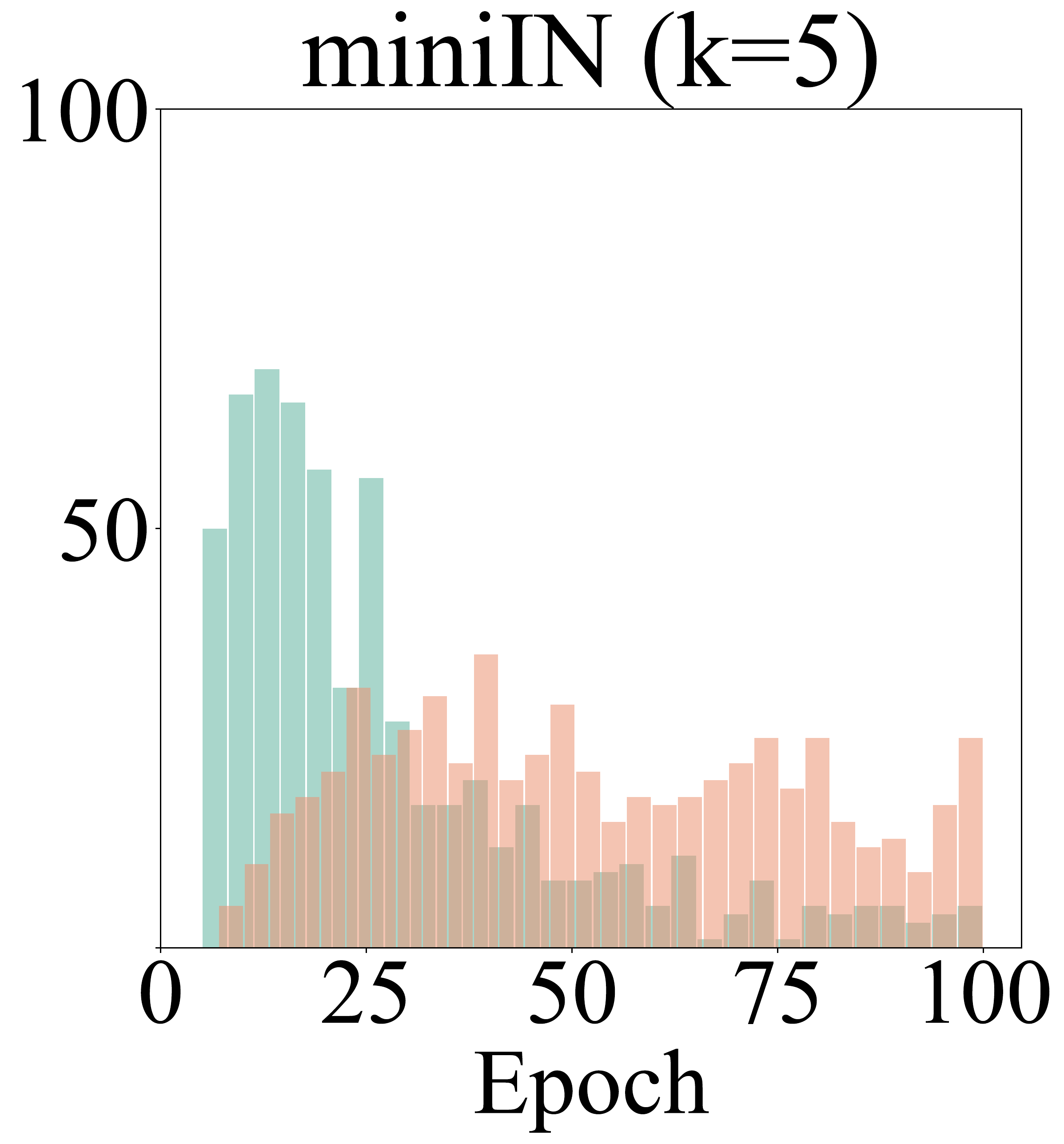}
         \end{subfigure}
     \begin{subfigure}[h]{0.19\linewidth}
         \includegraphics[width=\linewidth]{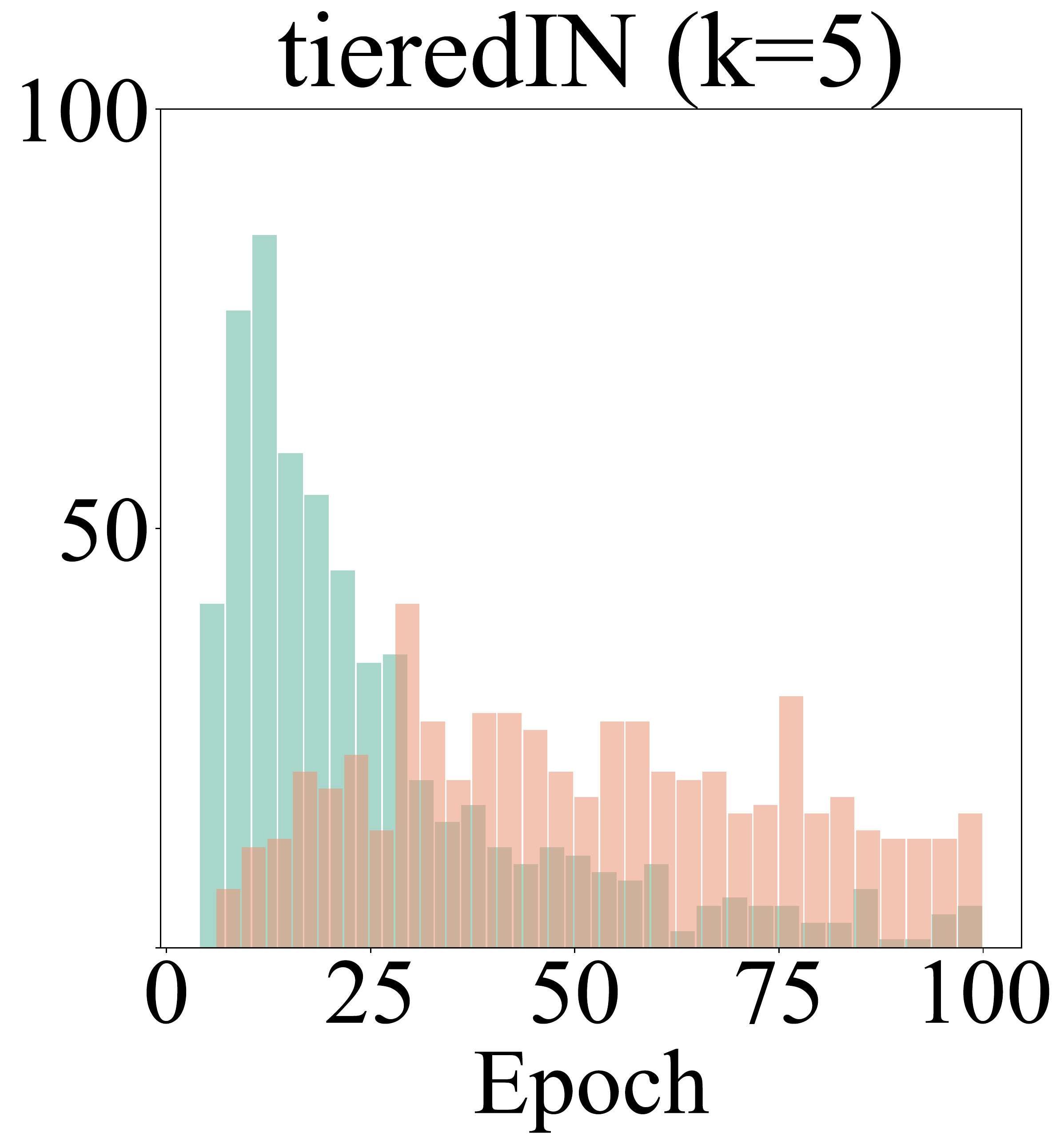}
     \end{subfigure}
     \begin{subfigure}[h]{0.19\linewidth}
         \includegraphics[width=\linewidth]{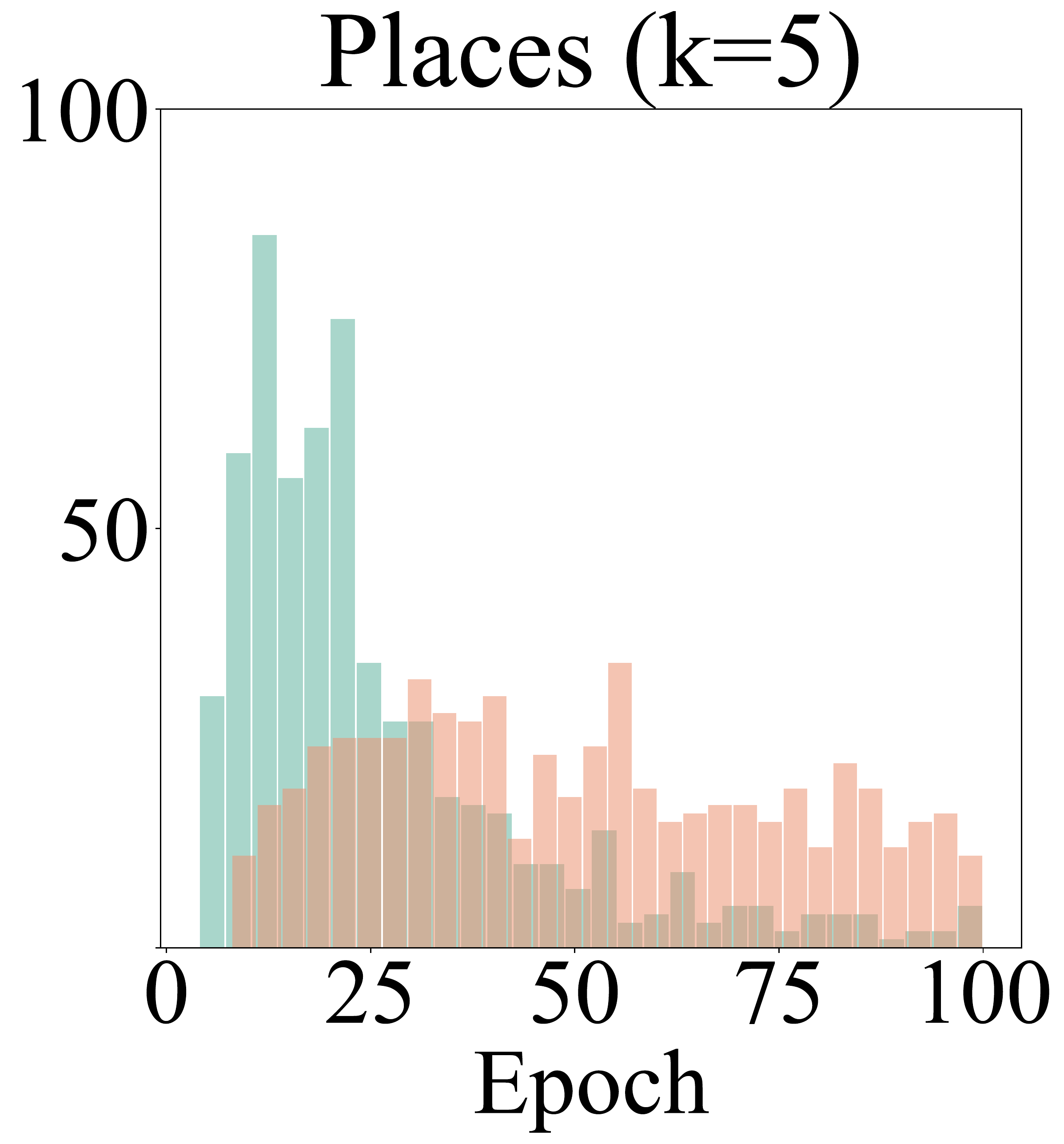}
     \end{subfigure}
     \begin{subfigure}[h]{0.19\linewidth}
         \includegraphics[width=\linewidth]{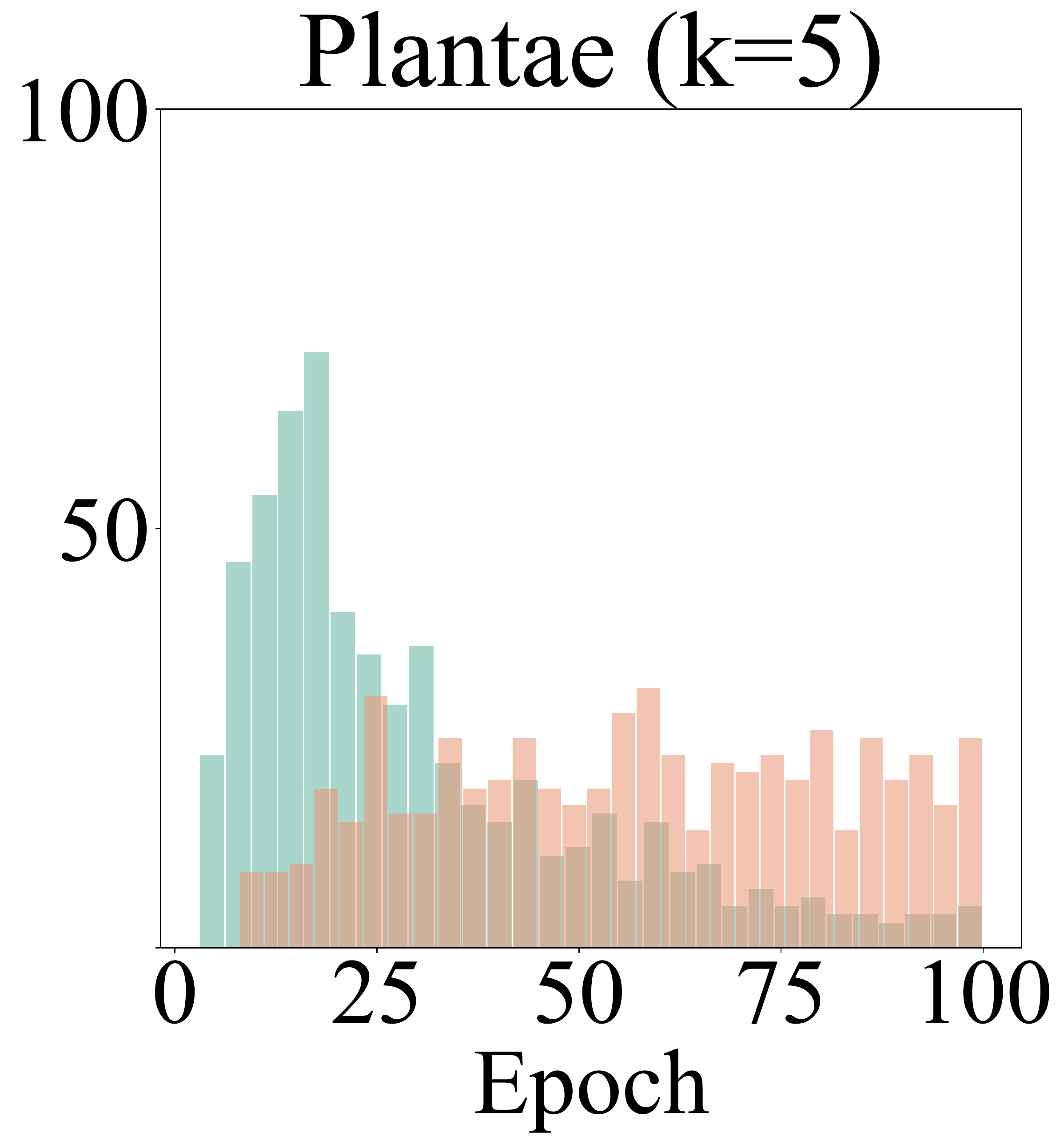}
     \end{subfigure}
     \begin{subfigure}[h]{0.19\linewidth}
         \includegraphics[width=\linewidth]{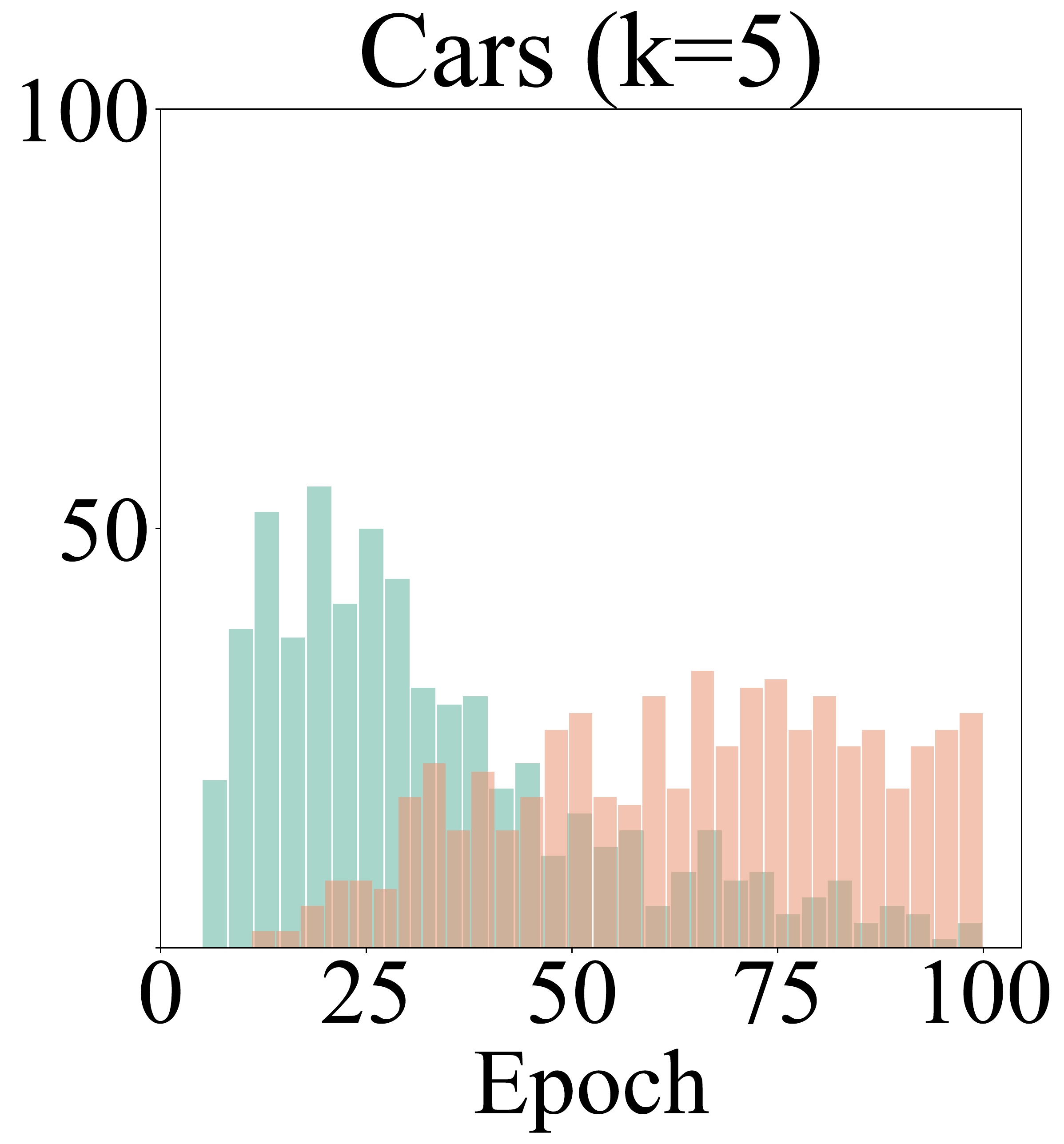}
     \end{subfigure}
     
     \begin{subfigure}[h]{0.19\linewidth}
             \includegraphics[width=\linewidth]{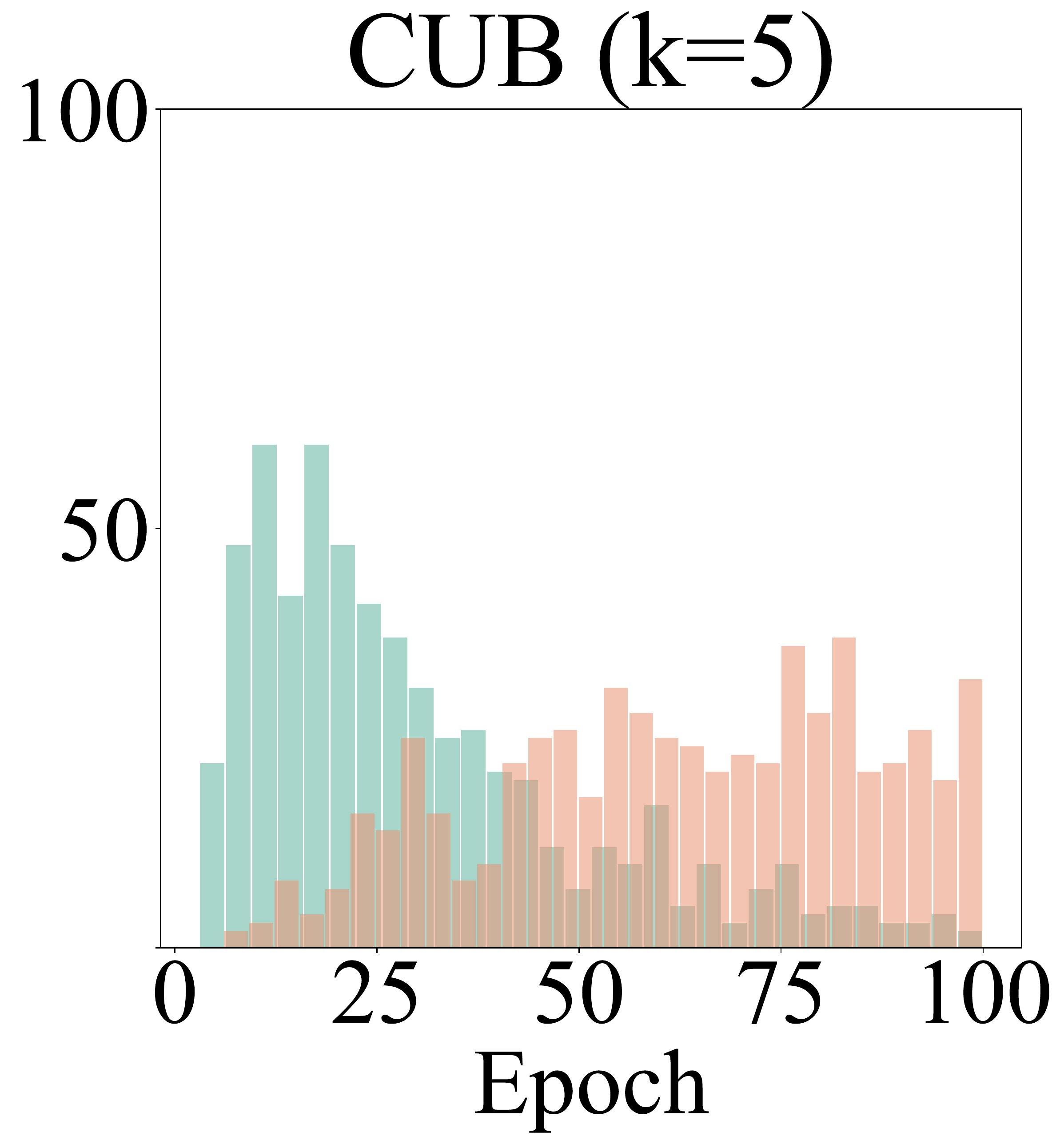}
         \end{subfigure}
     \begin{subfigure}[h]{0.19\linewidth}
         \includegraphics[width=\linewidth]{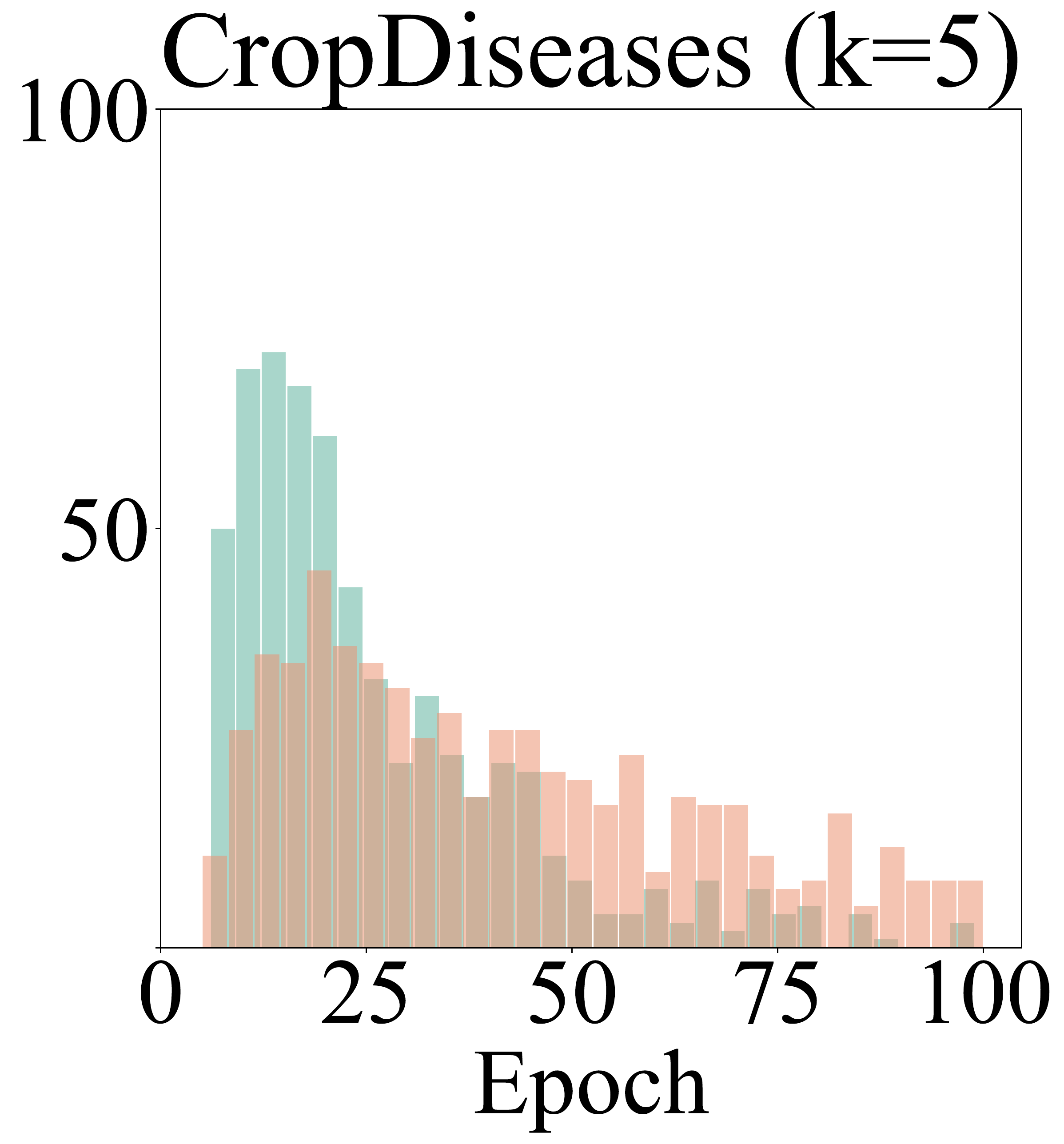}

     \end{subfigure}
     \begin{subfigure}[h]{0.19\linewidth}
         \includegraphics[width=\linewidth]{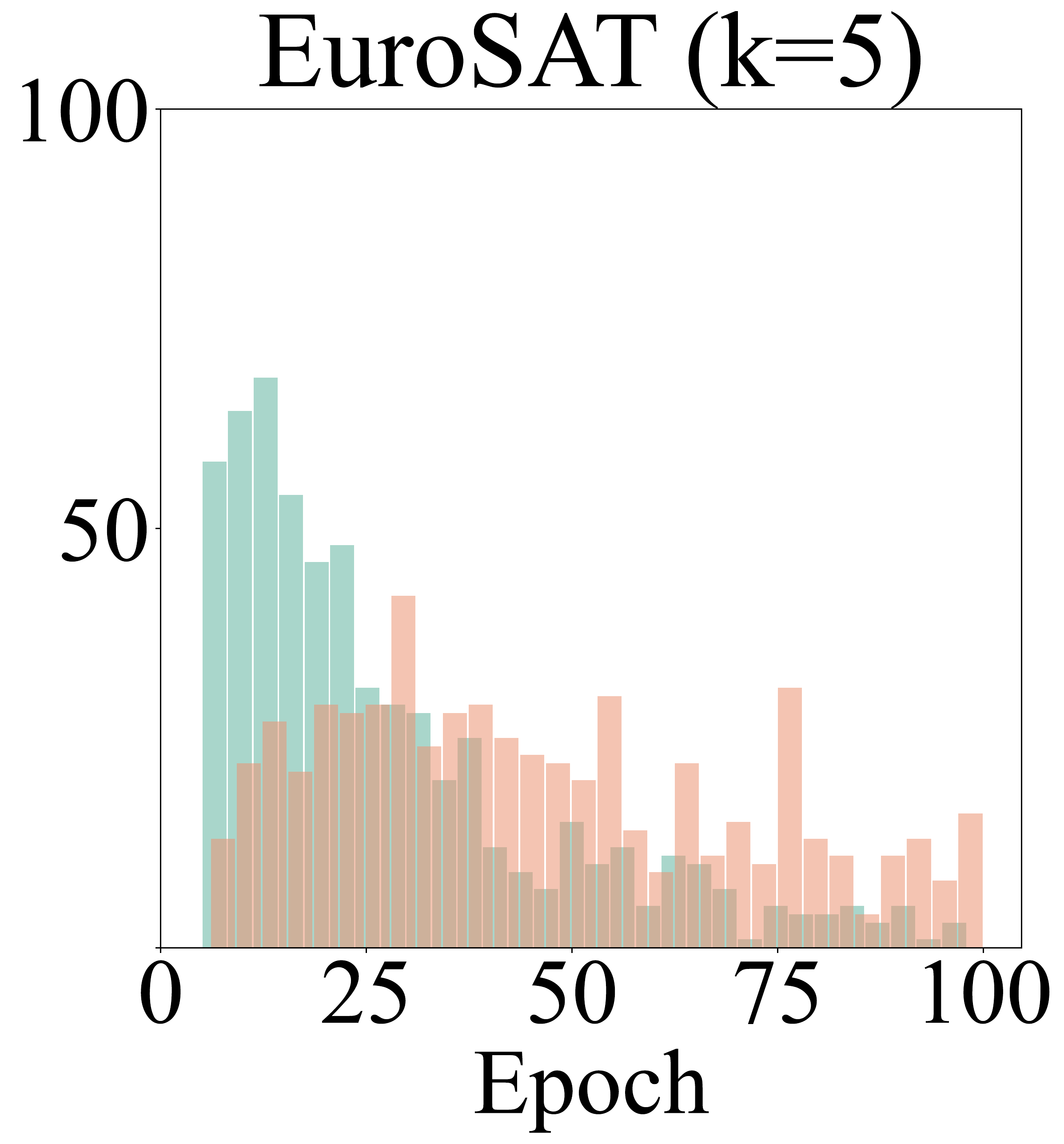}

     \end{subfigure}
     \begin{subfigure}[h]{0.19\linewidth}
         \includegraphics[width=\linewidth]{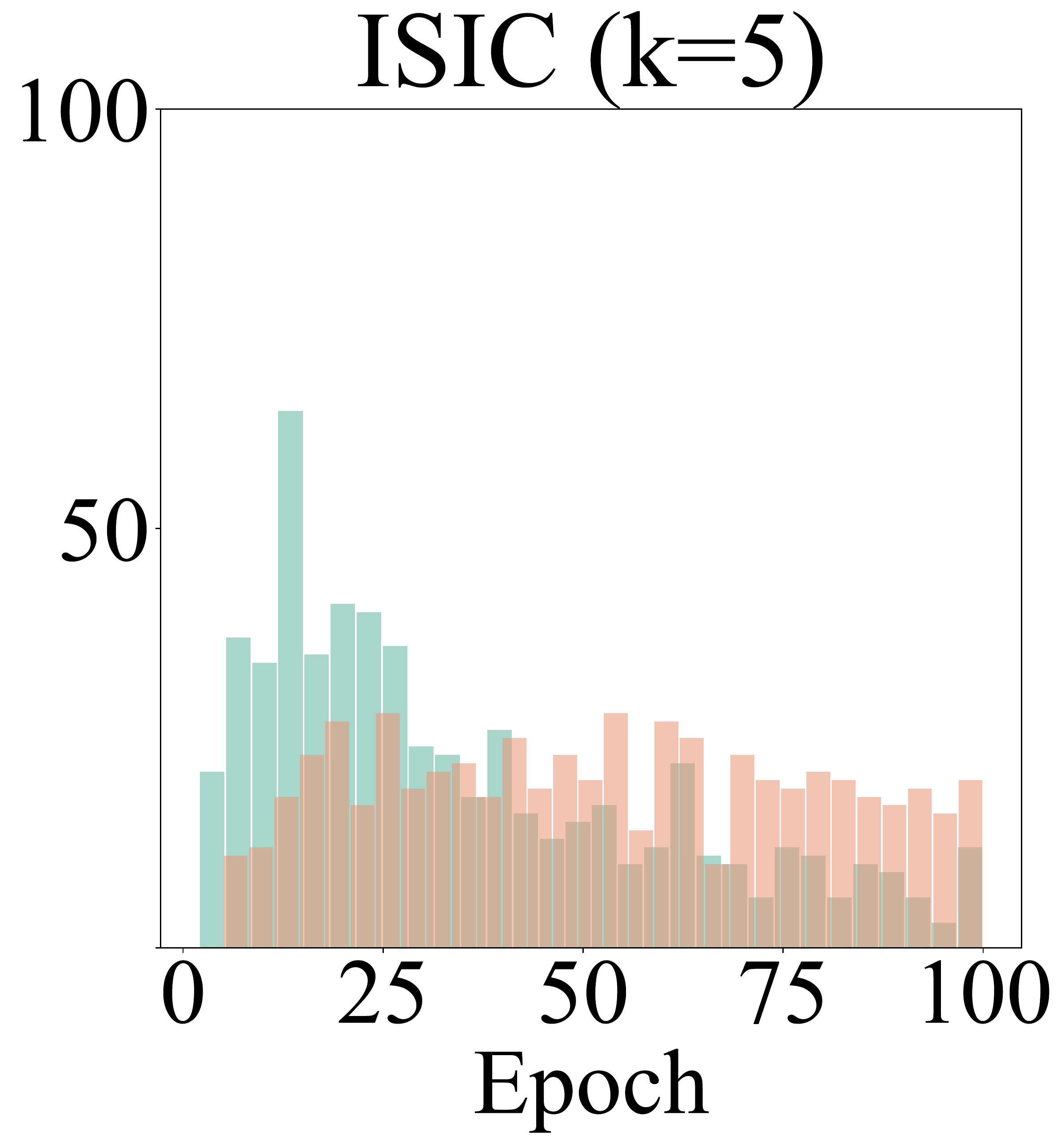}

     \end{subfigure}
     \begin{subfigure}[h]{0.19\linewidth}
         \includegraphics[width=\linewidth]{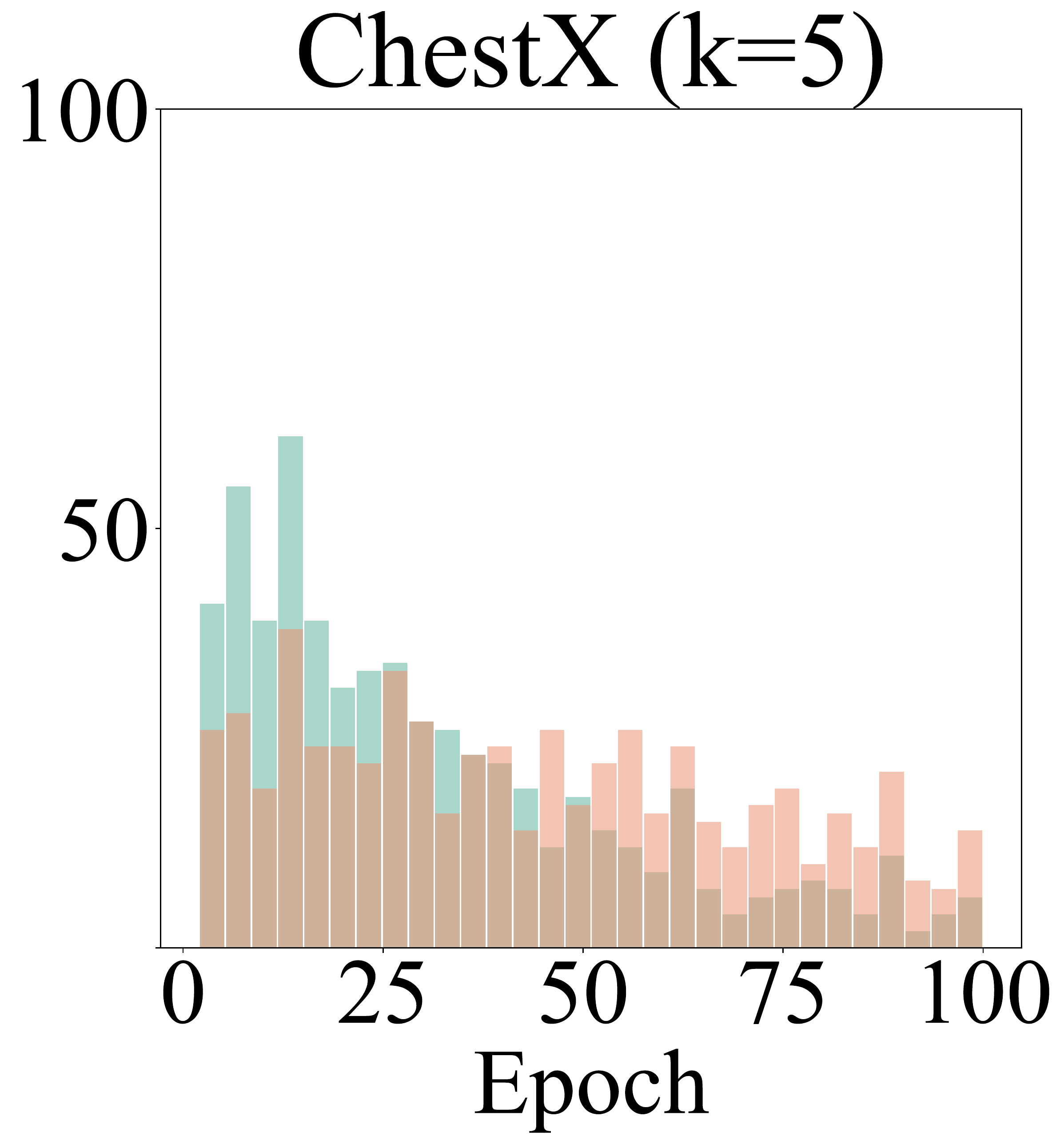}

     \end{subfigure}

     \hrulefill
     
    \begin{subfigure}[h]{0.19\linewidth}
             \includegraphics[width=\linewidth]{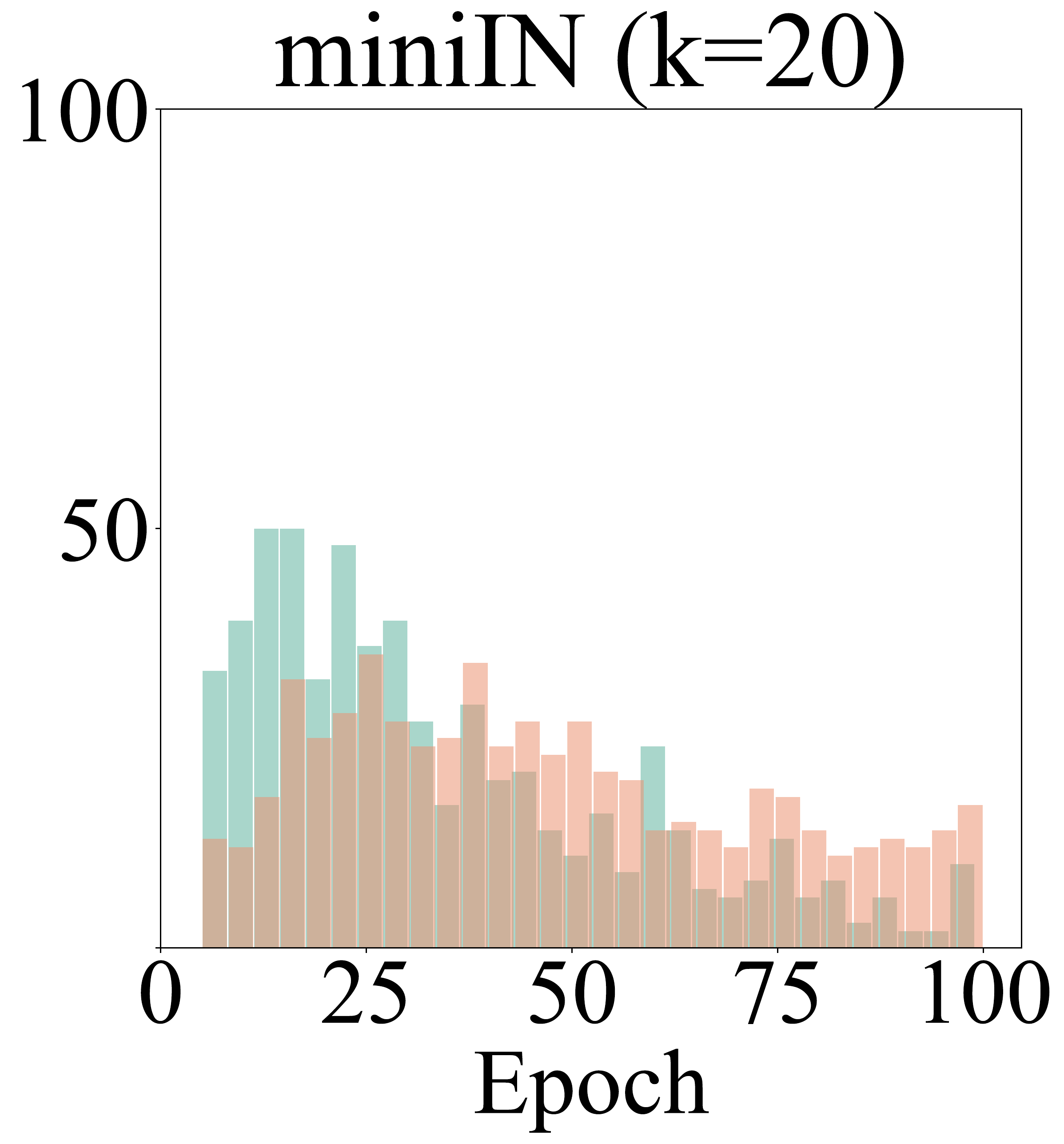}
         \end{subfigure}
     \begin{subfigure}[h]{0.19\linewidth}
         \includegraphics[width=\linewidth]{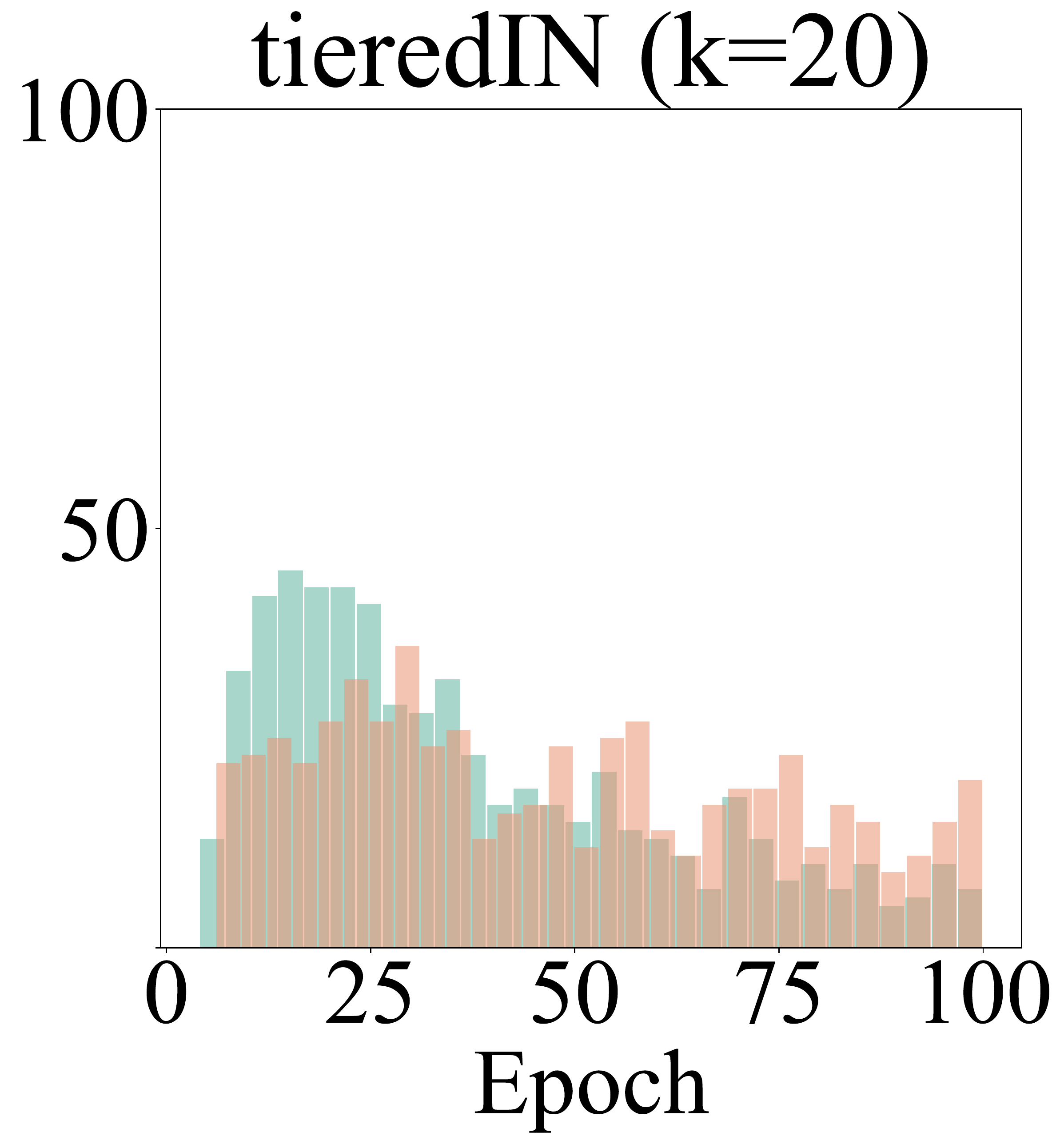}
     \end{subfigure}
     \begin{subfigure}[h]{0.19\linewidth}
         \includegraphics[width=\linewidth]{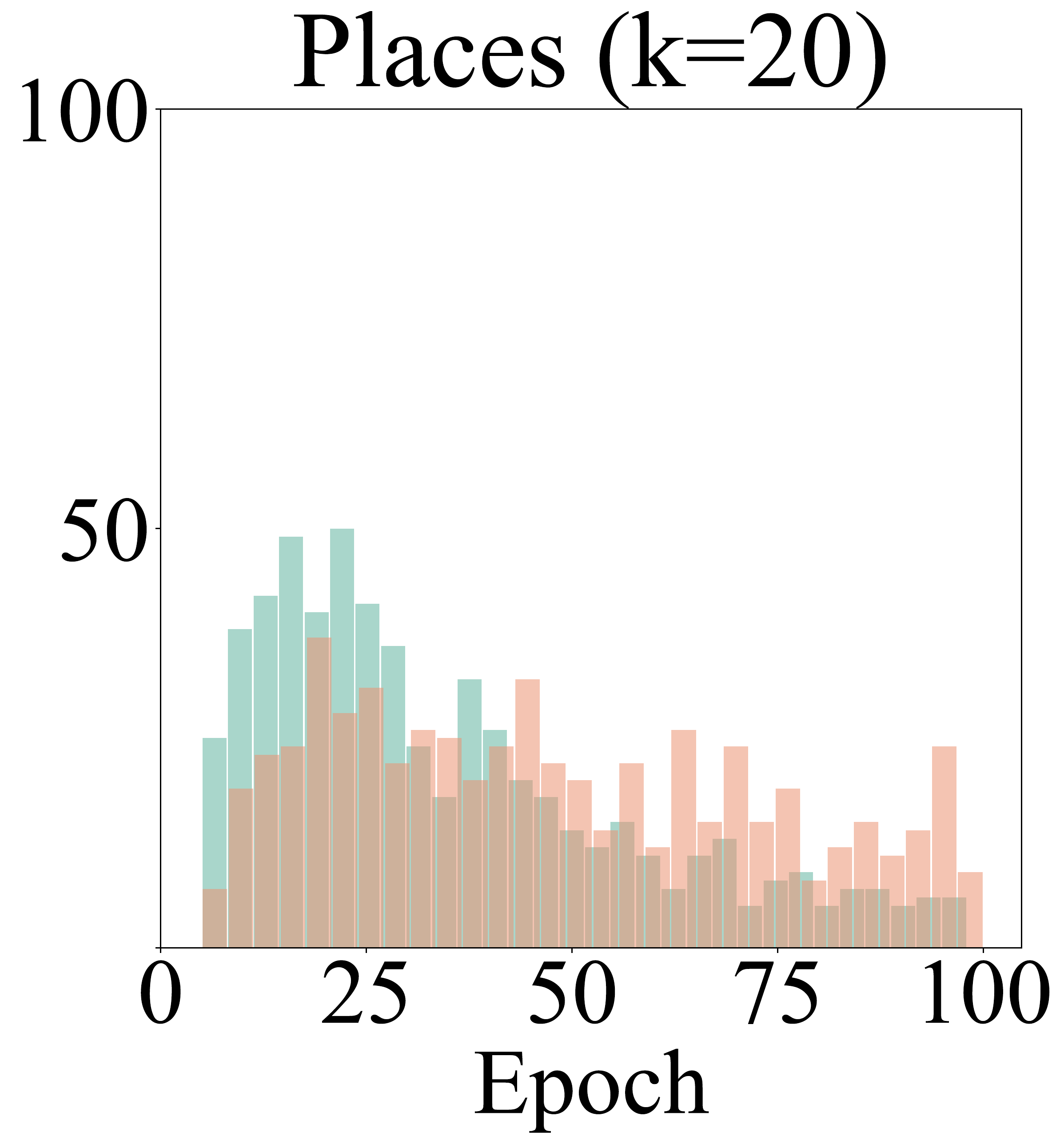}
     \end{subfigure}
     \begin{subfigure}[h]{0.19\linewidth}
         \includegraphics[width=\linewidth]{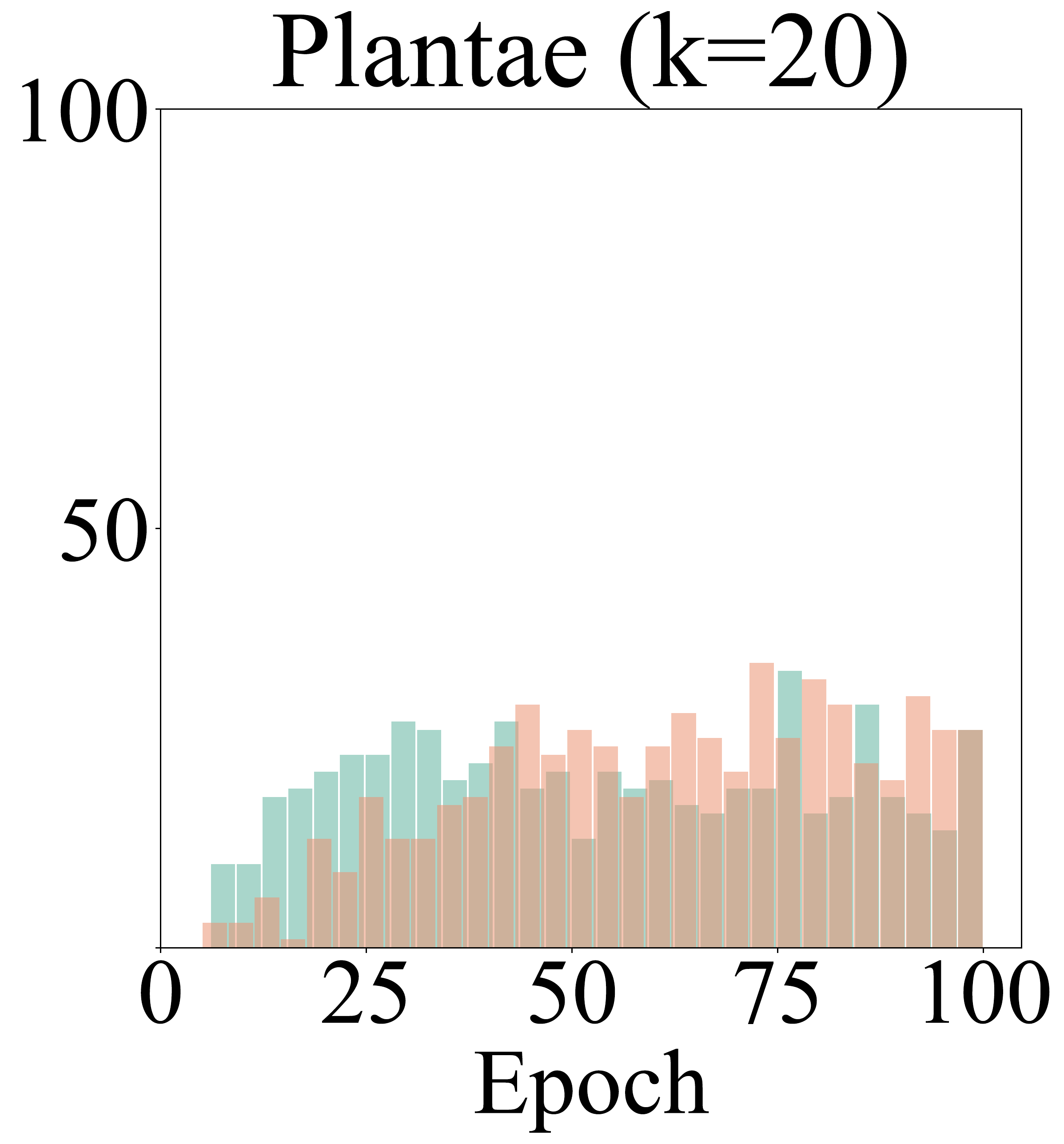}
     \end{subfigure}
     \begin{subfigure}[h]{0.19\linewidth}
         \includegraphics[width=\linewidth]{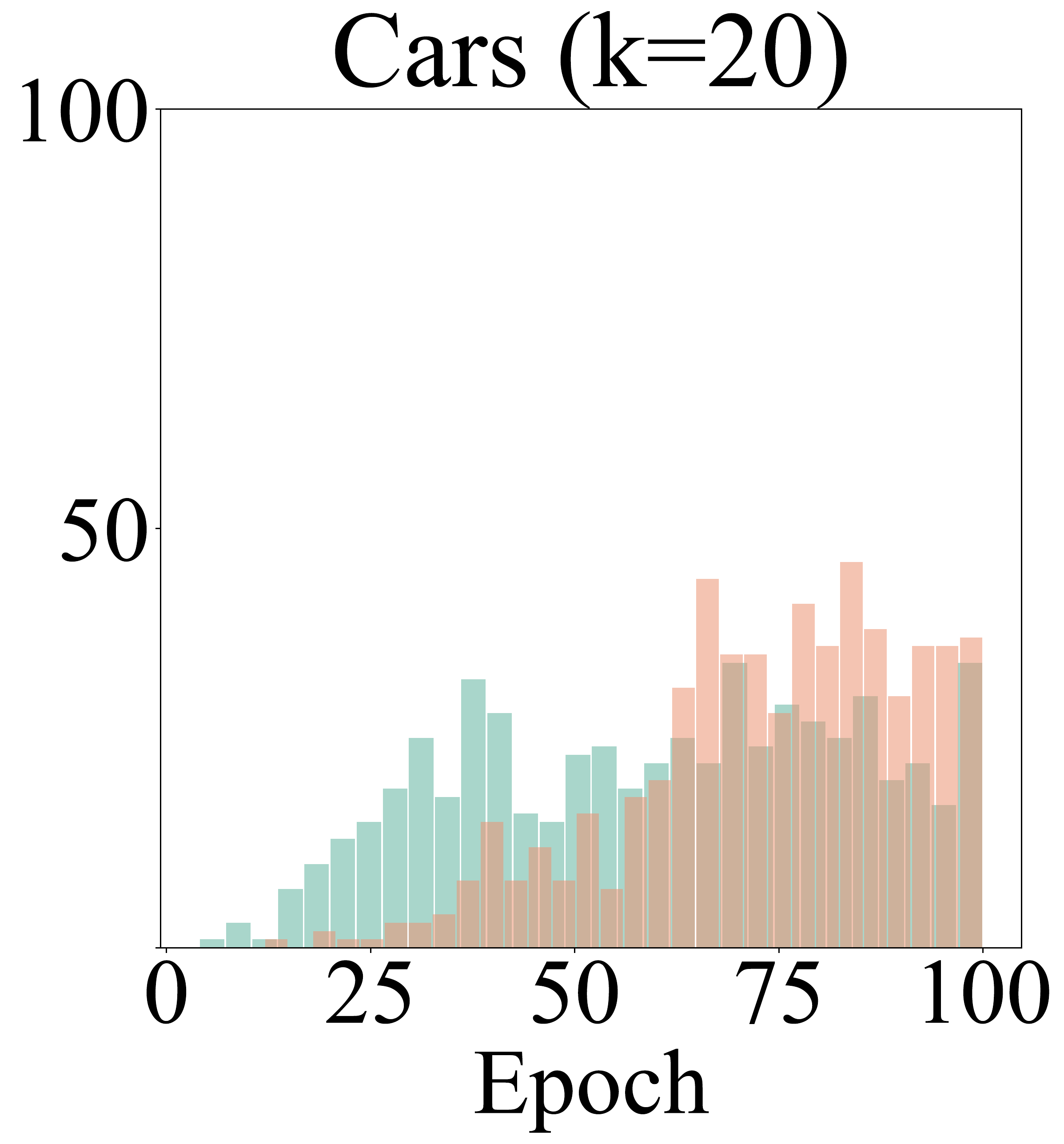}
     \end{subfigure}
     
     \begin{subfigure}[h]{0.19\linewidth}
             \includegraphics[width=\linewidth]{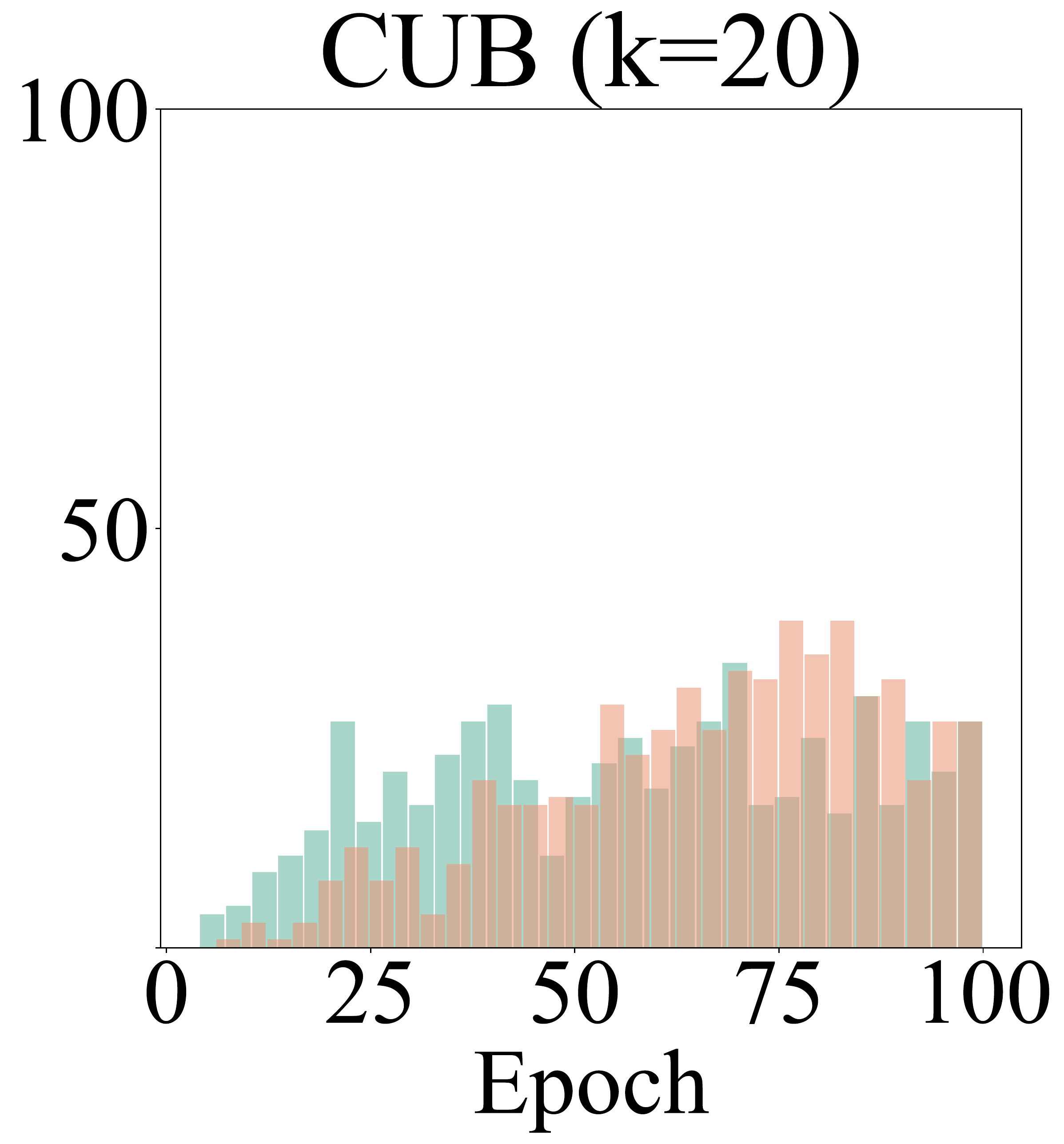}
         \end{subfigure}
     \begin{subfigure}[h]{0.19\linewidth}
         \includegraphics[width=\linewidth]{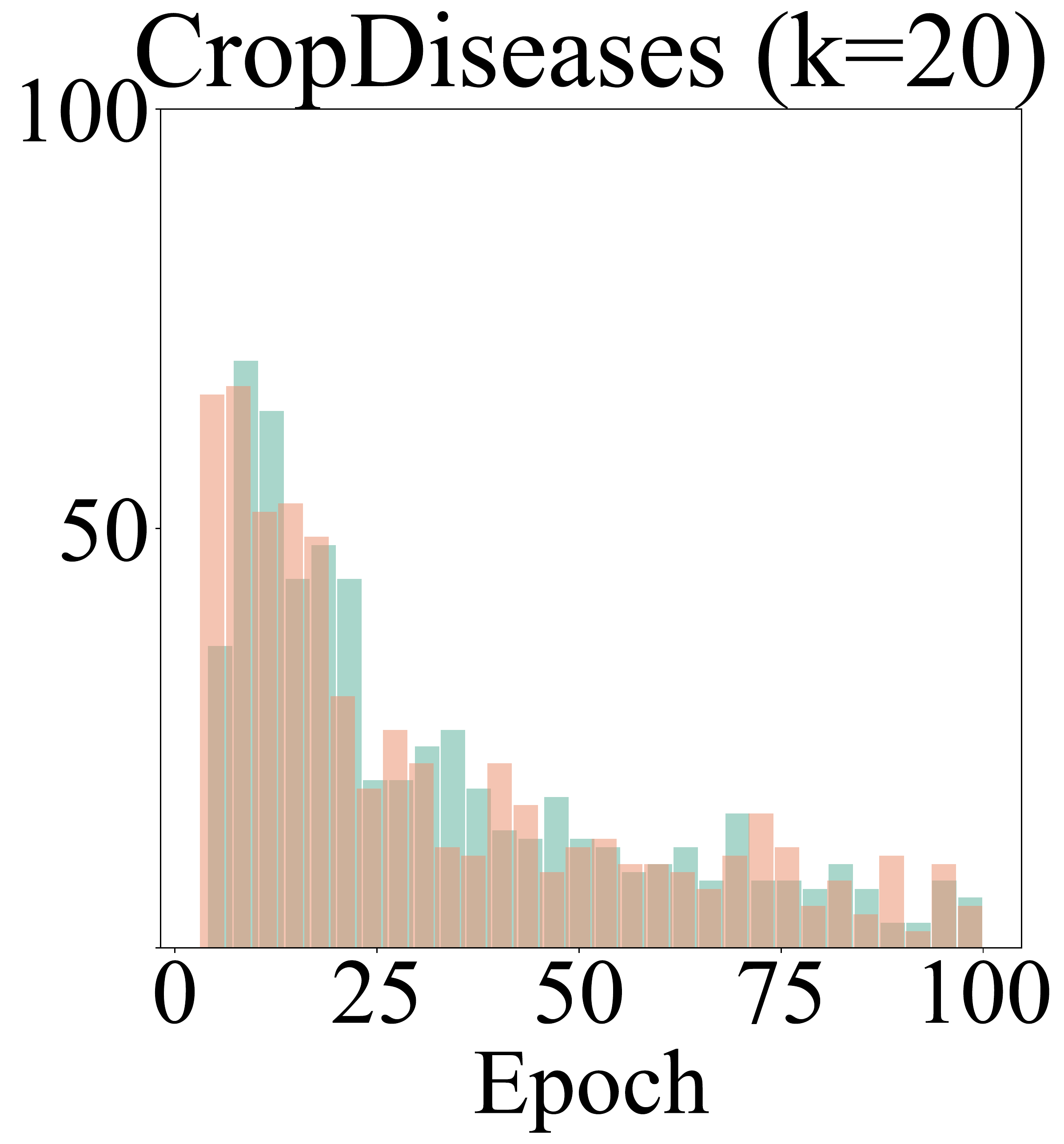}
     \end{subfigure}
     \begin{subfigure}[h]{0.19\linewidth}
         \includegraphics[width=\linewidth]{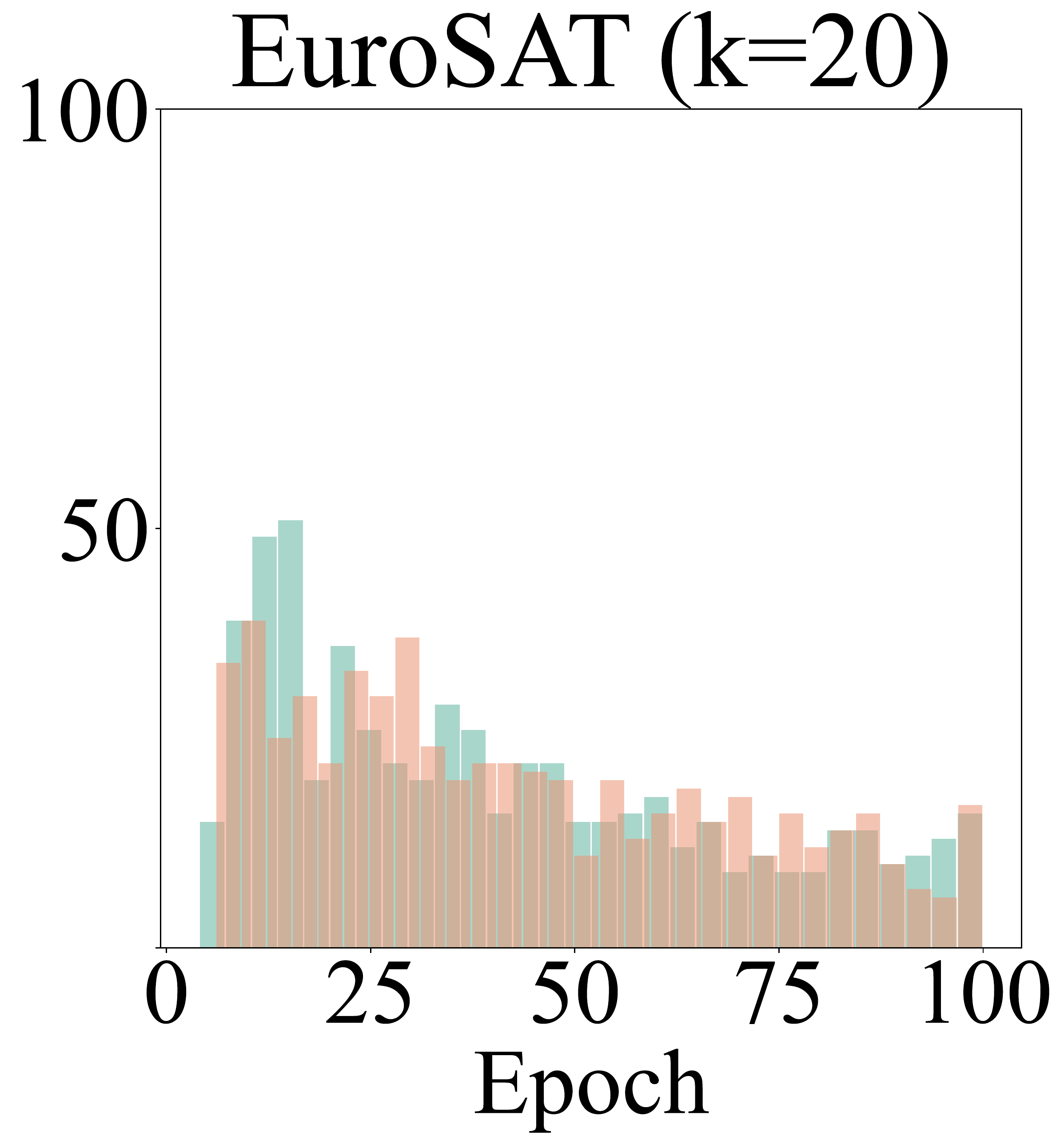}
     \end{subfigure}
     \begin{subfigure}[h]{0.19\linewidth}
         \includegraphics[width=\linewidth]{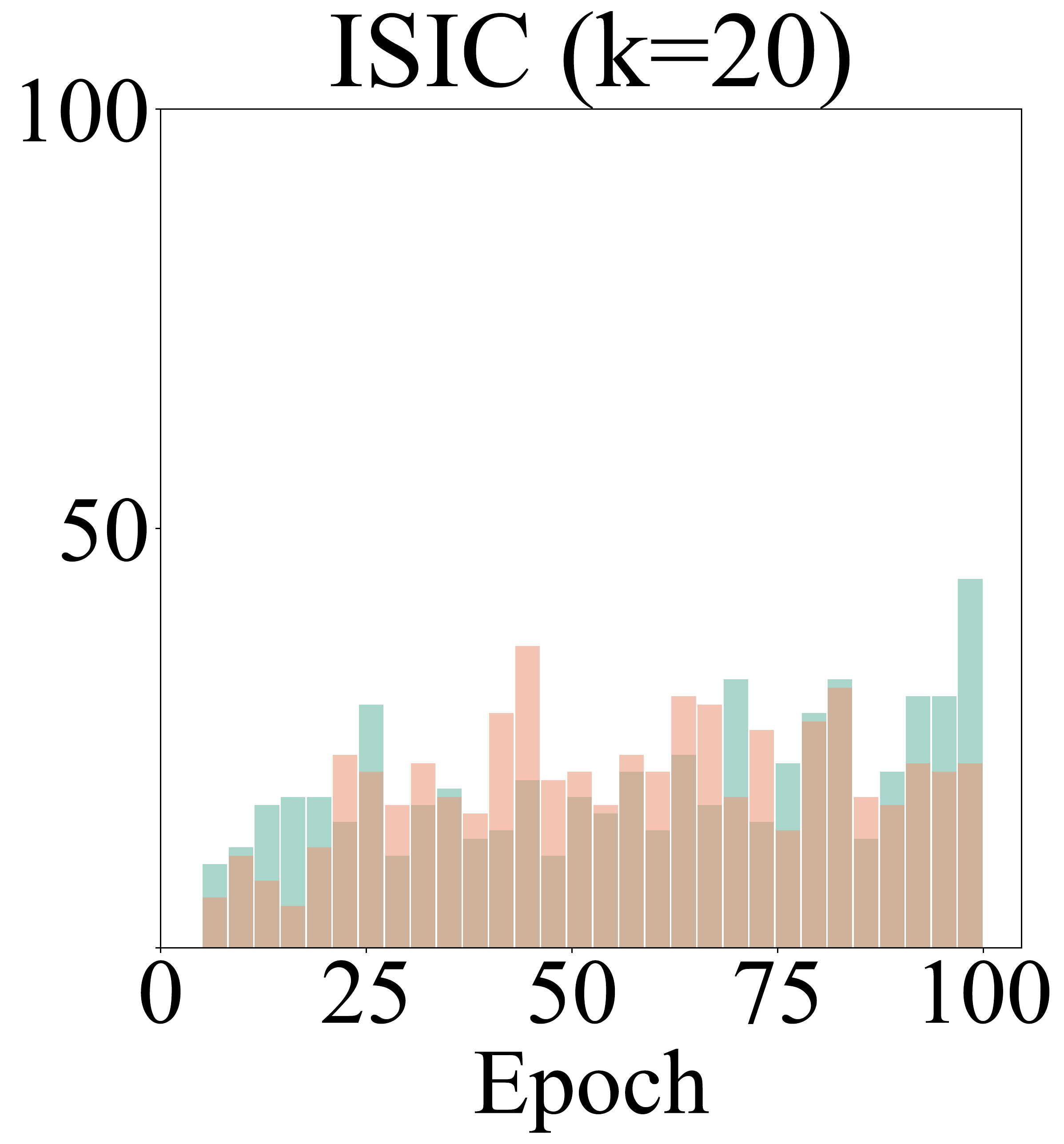}
     \end{subfigure}
     \begin{subfigure}[h]{0.19\linewidth}
         \includegraphics[width=\linewidth]{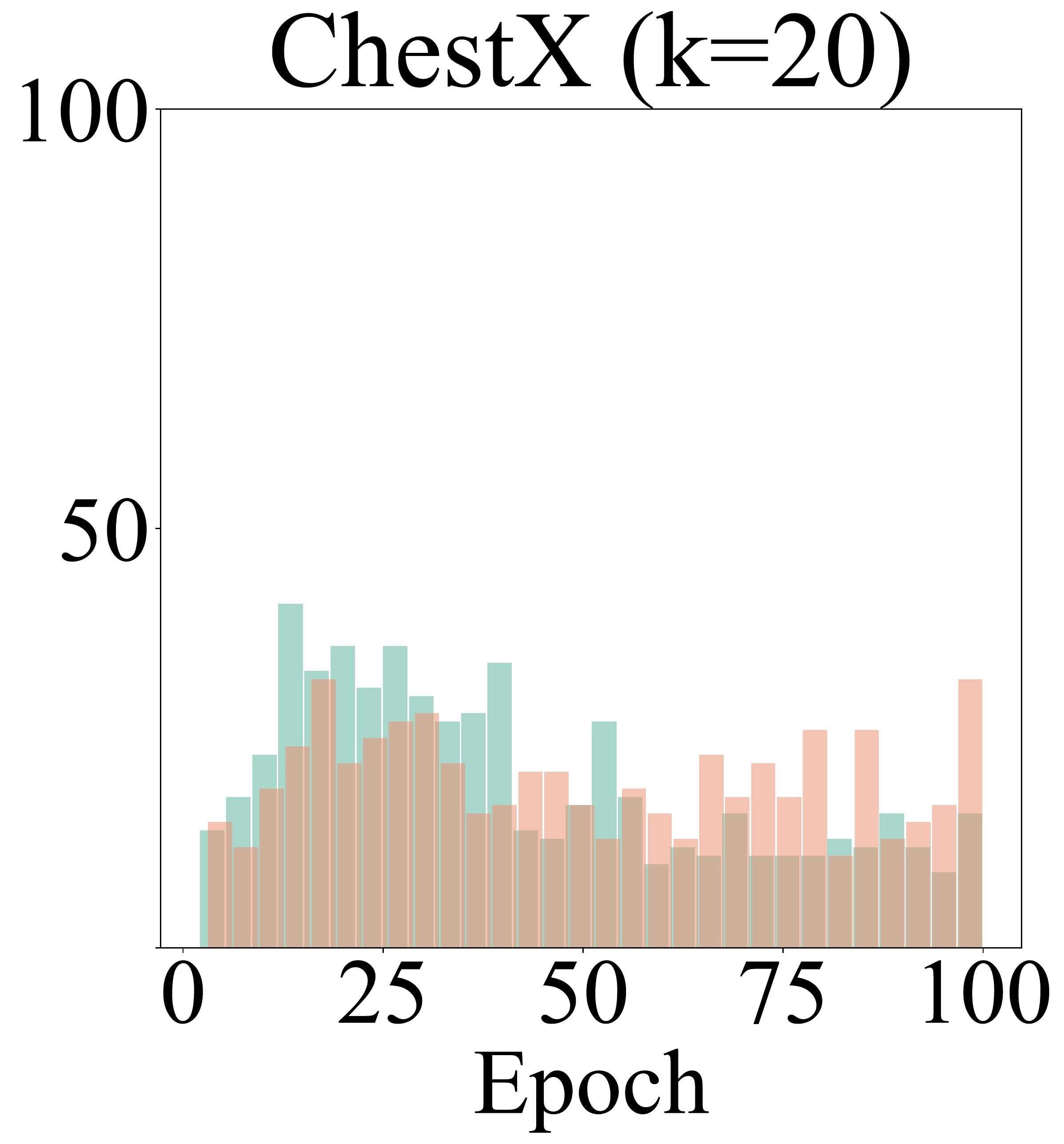}
     \end{subfigure}
     
     \caption{Histogram of best epoch during updating over 600 episodes. ResNet10 is pre-trained on miniIN. Green and orange colors indicate LP and FT, respectively.}
     \label{fig:best_hist}
\end{figure}

Figure \ref{fig:best_hist} illustrates histogram of best epochs over 600 episodes, considering different shot and target dataset. 
The green- and the orange-colored histograms indicate the result by LP and FT, respectively.
For LP, the best performance appears in the early- to mid-epochs, but for FT, the distribution of best epochs are generally skewed left or distributed uniformly.  
It seems that LP is prone to overfitting than FT.

\subsection{Layer Difference through FT}\label{appx:layer_diff}
\begin{figure}[h!] 
\centering
     \begin{subfigure}[h]{\linewidth}
         \includegraphics[width=\linewidth]{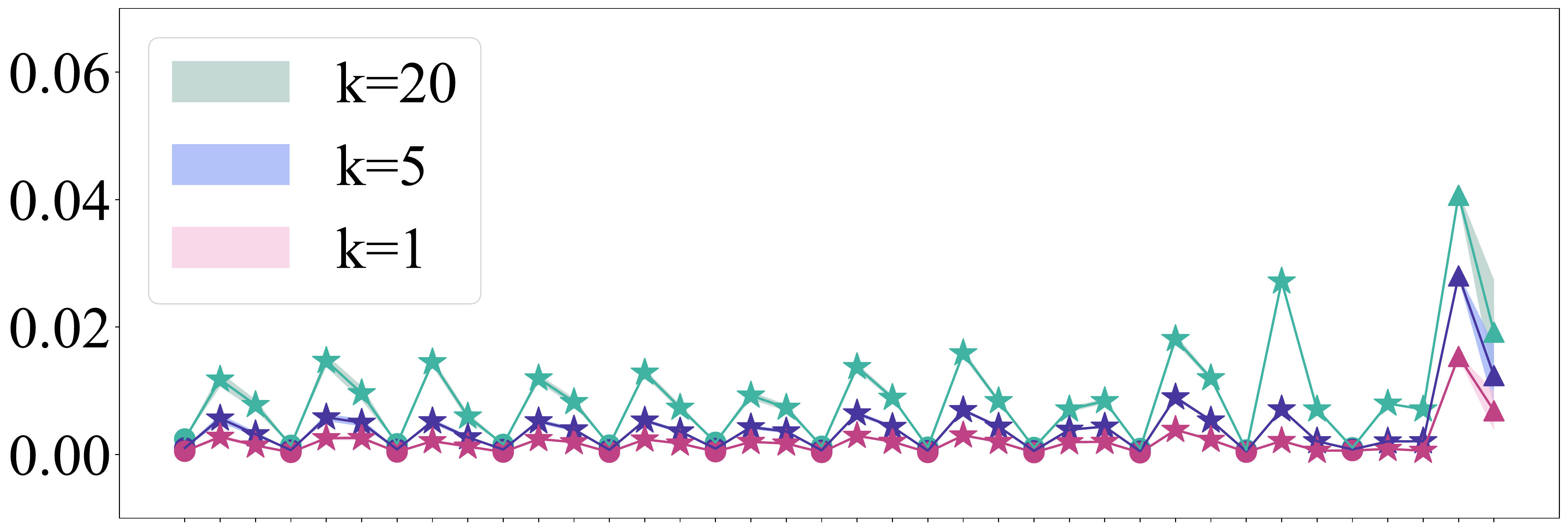}
         \caption{miniIN}
     \end{subfigure}
      \begin{subfigure}[h]{\linewidth}
         \includegraphics[width=\linewidth]{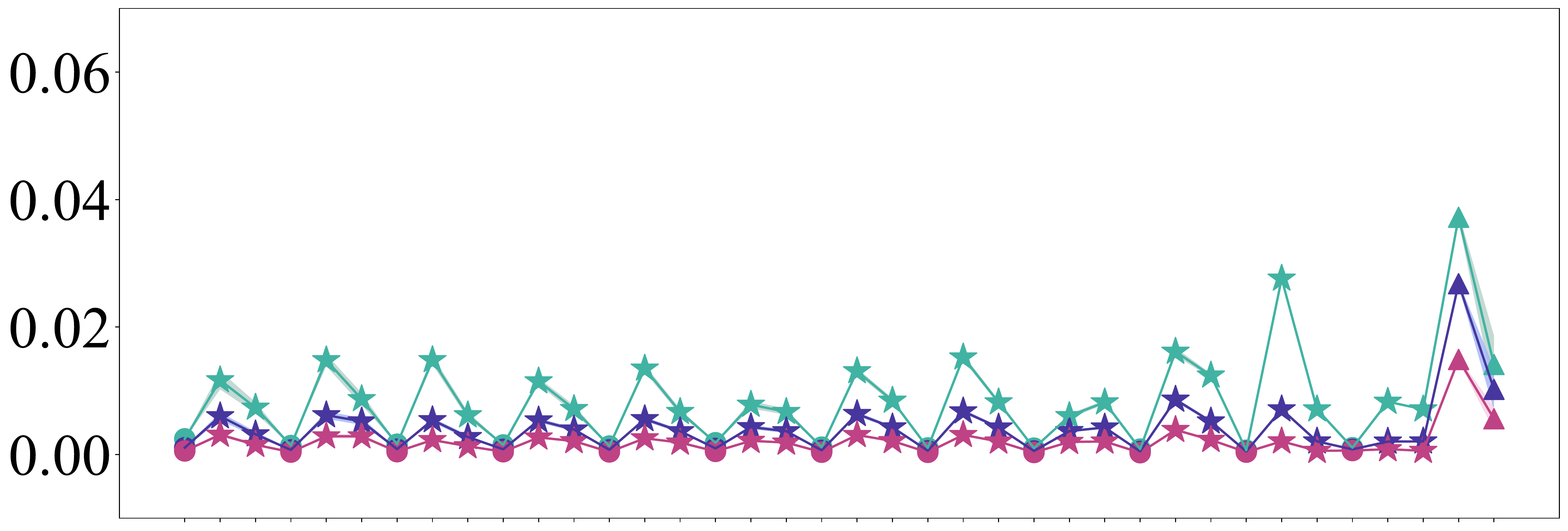}
         \caption{CropDiseases}
     \end{subfigure}
      \begin{subfigure}[h]{\linewidth}
         \includegraphics[width=\linewidth]{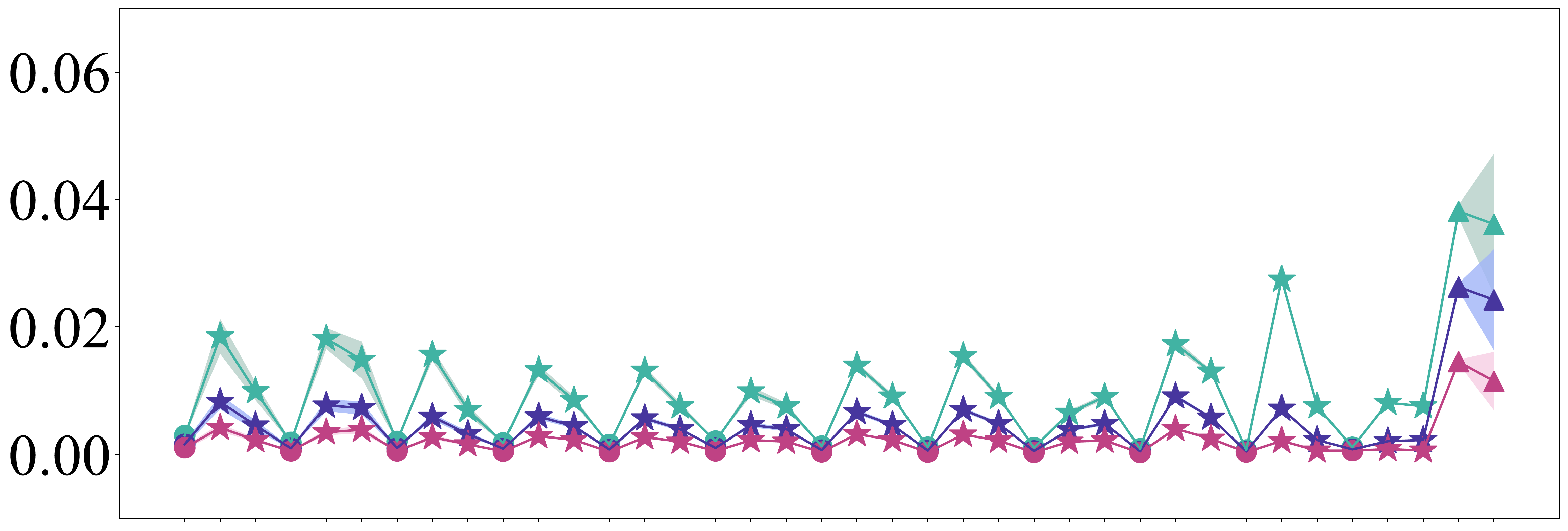}
         \caption{EuroSAT}
     \end{subfigure}
      \begin{subfigure}[h]{\linewidth}
         \includegraphics[width=\linewidth]{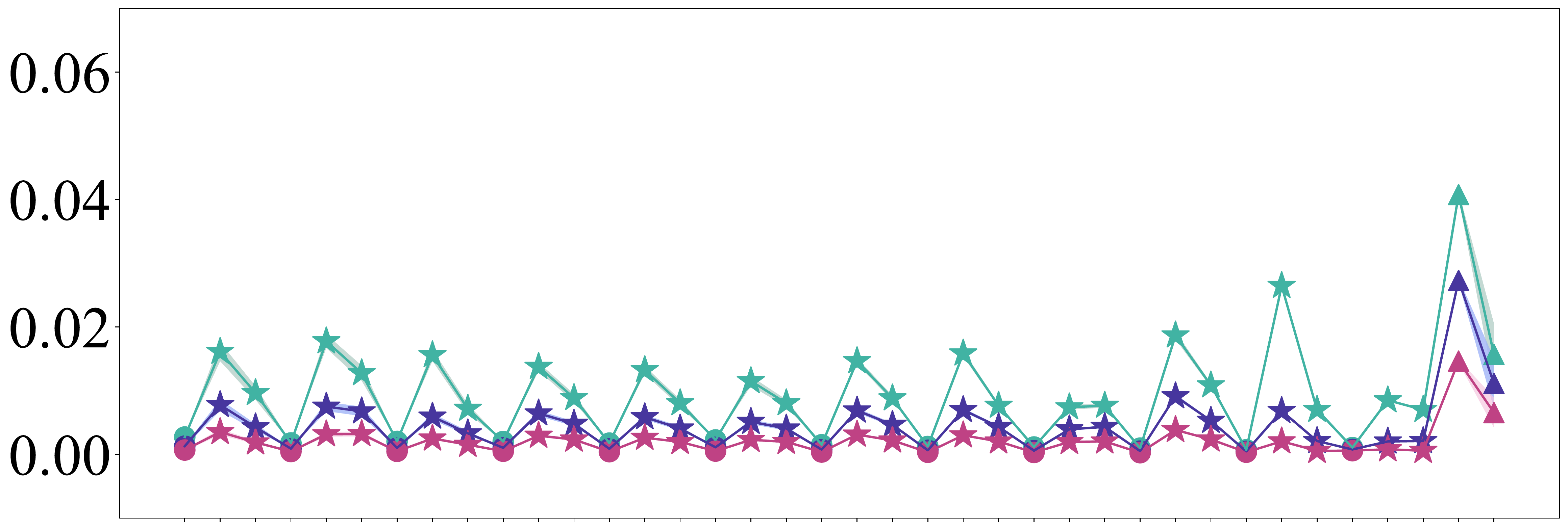}
         \caption{ISIC}
     \end{subfigure}
      \begin{subfigure}[h]{\linewidth}
         \includegraphics[width=\linewidth]{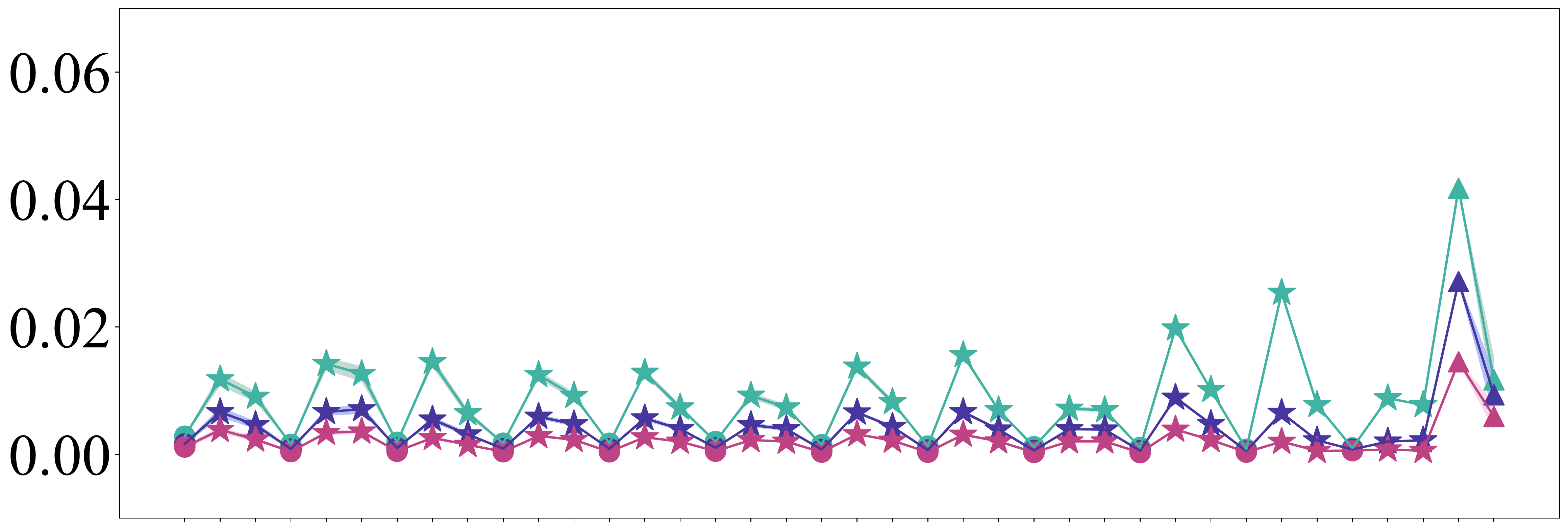}
         \caption{ChestX}
     \end{subfigure}
    \caption{Layer difference through FT according to target dataset using ResNet10 pre-trained on miniIN.}
     \label{fig:layer_diff_mini1}
\end{figure}

We analyze how layers of the model are updated by FT. For each layer, we evaluate \emph{layer difference}, which is L1 distance between parameters of LP and FT.
Figure \ref{fig:layer_diff_mini1} illustrates the layer differences of each layer, which is averaged over 100 episodes. X-ticks are as follows in order: Stem[Conv.weight, BN.scale, BN.shift] - Stage1[Conv1.weight, BN1.scale, BN1.shift, Conv2.weight, BN2.scale, BN2.shift] - Stage2[Conv1.weight, BN1.scale, BN1.shift, Conv2.weight, BN2.scale, BN2.shift, ShortCutConv.weight, ShortCutBN.scale, ShortCutBN.shift] - Stage3[Conv1.weight, BN1.scale, BN1.shift, Conv2.weight, BN2.scale, BN2.shift, ShortCutConv.weight, ShortCutBN.scale, ShortCutBN.shift] - Stage4[Conv1.weight, BN1.scale, BN1.shift, Conv2.weight, BN2.scale, BN2.shift, ShortCutConv.weight, ShortCutBN.scale, ShortCutBN.shift] - Classifier[weight, bias]. Conv-related, BN-related, and Classifier-related are marked as circle, star, and triangle, respectively.

It is observed that layers change a lot in the order of Classifier-related, BN-related, and Conv-related. Because a classifier is randomly initialized before FT, it is reasonable that the change in a classifier is large. Surprisingly, BN-related layers change more than Conv-related layers when fine-tuning a pre-trained extractor, which demonstrates that batch normalization is an important factor for FT. This can be explained that the fine-tuned model with FT utilizes pre-trained convolution layers, but considerably changes shift and scale of the distribution of intermediate features, which seems to be related to FWT\,\citep{tseng2020cross}. Furthermore, the changes of original path is larger than those of shortcut path.

\subsection{Two-staged Update}
We provide the results of two-staged updates, following \citep{kumar2022finetuning}. They showed that LP $\rightarrow$ FT improves the performance of OOD tasks. Similar to them, we update through one update method for the first 50 epochs, and then update through another update method for the following 50 epochs. Table \ref{tab:two_stage} describes few-shot performance of two-staged updates, FT $\rightarrow$ LP and LP $\rightarrow$ FT. Unlike OOD tasks, two-staged updates do not improve the performance of FSL. 

\begin{table}[h]
    \centering
    \resizebox{\linewidth}{!}{
    \begin{tabular}{ccccccc}
    \toprule
    $k$ & Update & miniIN & tieredIN & Places & Plantae & Cars\\ \midrule
    \multirow{4}{*}{1} & LP & 56.23{\scriptsize$\pm$.77} & 55.43{\scriptsize$\pm$.91} & 52.83{\scriptsize$\pm$.81} & 36.49{\scriptsize$\pm$.69} & 29.79{\scriptsize$\pm$.51} \\
     & FT & 49.34{\scriptsize$\pm$.77} & 48.96{\scriptsize$\pm$.86} & 46.24{\scriptsize$\pm$.78} & 33.08{\scriptsize$\pm$.59} & 29.55{\scriptsize$\pm$.52} \\
    & FT$\rightarrow$LP & 51.24{\scriptsize$\pm$.76} & 50.48{\scriptsize$\pm$.86} & 48.56{\scriptsize$\pm$.78} & 33.52{\scriptsize$\pm$.65} & 30.32{\scriptsize$\pm$.54} \\
     & LP$\rightarrow$FT & 54.55{\scriptsize$\pm$.77} & 54.20{\scriptsize$\pm$.87} & 51.45{\scriptsize$\pm$.81} & 35.14{\scriptsize$\pm$.63} & 30.37{\scriptsize$\pm$.55} \\
    \midrule
    \multirow{4}{*}{5} & LP & 77.77{\scriptsize$\pm$.61} & 75.02{\scriptsize$\pm$.71} & 73.26{\scriptsize$\pm$.66} & 54.21{\scriptsize$\pm$.74} & 44.92{\scriptsize$\pm$.67}\\
    & FT & 75.55{\scriptsize$\pm$.60} & 73.59{\scriptsize$\pm$.74} & 70.60{\scriptsize$\pm$.68} & 53.45{\scriptsize$\pm$.74} & 49.85{\scriptsize$\pm$.73}\\
   & FT$\rightarrow$LP & 75.12{\scriptsize$\pm$.61} & 73.07{\scriptsize$\pm$.74} & 70.44{\scriptsize$\pm$.79} & 53.22{\scriptsize$\pm$.74} & 49.29{\scriptsize$\pm$.70} \\
     & LP$\rightarrow$FT & 75.79{\scriptsize$\pm$.63} & 73.73{\scriptsize$\pm$.77} & 70.78{\scriptsize$\pm$.71} & 53.88{\scriptsize$\pm$.73} & 48.44{\scriptsize$\pm$.72} \\
    \midrule
    \multirow{4}{*}{20} & LP & 86.54{\scriptsize$\pm$.44} & 83.26{\scriptsize$\pm$.61} & 81.53{\scriptsize$\pm$.53} & 66.52{\scriptsize$\pm$.69} & 60.95{\scriptsize$\pm$.66}\\ 
     & FT & 86.94{\scriptsize$\pm$.44} & 84.48{\scriptsize$\pm$.59} & 81.30{\scriptsize$\pm$.53} & 69.08{\scriptsize$\pm$.68} & 70.62{\scriptsize$\pm$.64}\\
     & FT$\rightarrow$LP & 86.90{\scriptsize$\pm$.45} & 84.36{\scriptsize$\pm$.59} & 81.43{\scriptsize$\pm$.54} & 68.08{\scriptsize$\pm$.70} & 68.59{\scriptsize$\pm$.64} \\
     & LP$\rightarrow$FT & 87.19{\scriptsize$\pm$.44} & 84.73{\scriptsize$\pm$.58} & 82.05{\scriptsize$\pm$.52} & 68.91{\scriptsize$\pm$.70} & 69.42{\scriptsize$\pm$.64} \\ \midrule
     $k$ & Update & CUB & \!\!\!\!CropDiseases\!\!\!\! & EuroSAT & ISIC & ChestX \\ \midrule
    \multirow{4}{*}{1} & LP & 40.91{\scriptsize$\pm$.76} & 73.16{\scriptsize$\pm$.87} & 64.96{\scriptsize$\pm$.86} & 29.80{\scriptsize$\pm$.55} & 22.32{\scriptsize$\pm$.42} \\
     & FT & 37.57{\scriptsize$\pm$.70} & 66.82{\scriptsize$\pm$.89} & 57.25{\scriptsize$\pm$.83} & 31.20{\scriptsize$\pm$.55} & 22.06{\scriptsize$\pm$.38} \\
    & FT$\rightarrow$LP & 37.82{\scriptsize$\pm$.70} & 69.86{\scriptsize$\pm$.85} & 60.52{\scriptsize$\pm$.83} & 31.39{\scriptsize$\pm$.56} & 22.30{\scriptsize$\pm$.40} \\
     & LP$\rightarrow$FT & 39.23{\scriptsize$\pm$.74} & 71.77{\scriptsize$\pm$.87} & 62.96{\scriptsize$\pm$.85} & 31.60{\scriptsize$\pm$.58} & 22.23{\scriptsize$\pm$.39} \\
    \midrule
    \multirow{4}{*}{5} & LP & 58.38{\scriptsize$\pm$.76} & 91.27{\scriptsize$\pm$.48} & 84.48{\scriptsize$\pm$.57} & 40.88{\scriptsize$\pm$.54} & 26.07{\scriptsize$\pm$.44} \\
    & FT & 59.93{\scriptsize$\pm$.79} & 91.64{\scriptsize$\pm$.49} & 82.32{\scriptsize$\pm$.57} & 48.82{\scriptsize$\pm$.62} & 25.97{\scriptsize$\pm$.40} \\
    & FT$\rightarrow$LP & 59.18{\scriptsize$\pm$.78} & 91.35{\scriptsize$\pm$.50} & 81.31{\scriptsize$\pm$.58} & 48.70{\scriptsize$\pm$.61} & 26.36{\scriptsize$\pm$.44} \\
     & LP$\rightarrow$FT & 59.66{\scriptsize$\pm$.80} & 91.15{\scriptsize$\pm$.53} & 79.08{\scriptsize$\pm$.70} & 48.21{\scriptsize$\pm$.64} & 26.30{\scriptsize$\pm$.43} \\
    \midrule
    \multirow{4}{*}{20} & LP & 71.29{\scriptsize$\pm$.70} & 96.20{\scriptsize$\pm$.29} & 91.00{\scriptsize$\pm$.40} & 51.32{\scriptsize$\pm$.54} & 30.98{\scriptsize$\pm$.45} \\ 
     & FT & 75.91{\scriptsize$\pm$.67} & 97.72{\scriptsize$\pm$.22} & 92.54{\scriptsize$\pm$.36} & 62.36{\scriptsize$\pm$.59} & 32.65{\scriptsize$\pm$.47} \\
    & FT$\rightarrow$LP & 74.61{\scriptsize$\pm$.66} & 97.51{\scriptsize$\pm$.24} & 92.58{\scriptsize$\pm$.35} & 62.00{\scriptsize$\pm$.60} & 32.29{\scriptsize$\pm$.47} \\
     & LP$\rightarrow$FT & 75.91{\scriptsize$\pm$.65} & 97.44{\scriptsize$\pm$.24} & 92.54{\scriptsize$\pm$.37} & 61.86{\scriptsize$\pm$.58} & 32.54{\scriptsize$\pm$.46} \\
     \bottomrule
    \end{tabular}}
    \caption{5-way $k$-shot performance of two-staged updates. The backbone is ResNet10 pre-trained on miniIN.}\label{tab:two_stage}
\end{table}

\subsection{Partial Update between LP and FT}
Furthermore, we investigate the interpolation between LP and FT.
\citet{oh2021boil} showed that partial updates can improve few-shot performance based on meta-learning methods.
Following this work, we identify whether the partial updates still work for transfer learning-based FSL.

\begin{figure*}[!ht]
    \centering
    \begin{subfigure}[h]{0.18\linewidth}
    \includegraphics[width=\linewidth]{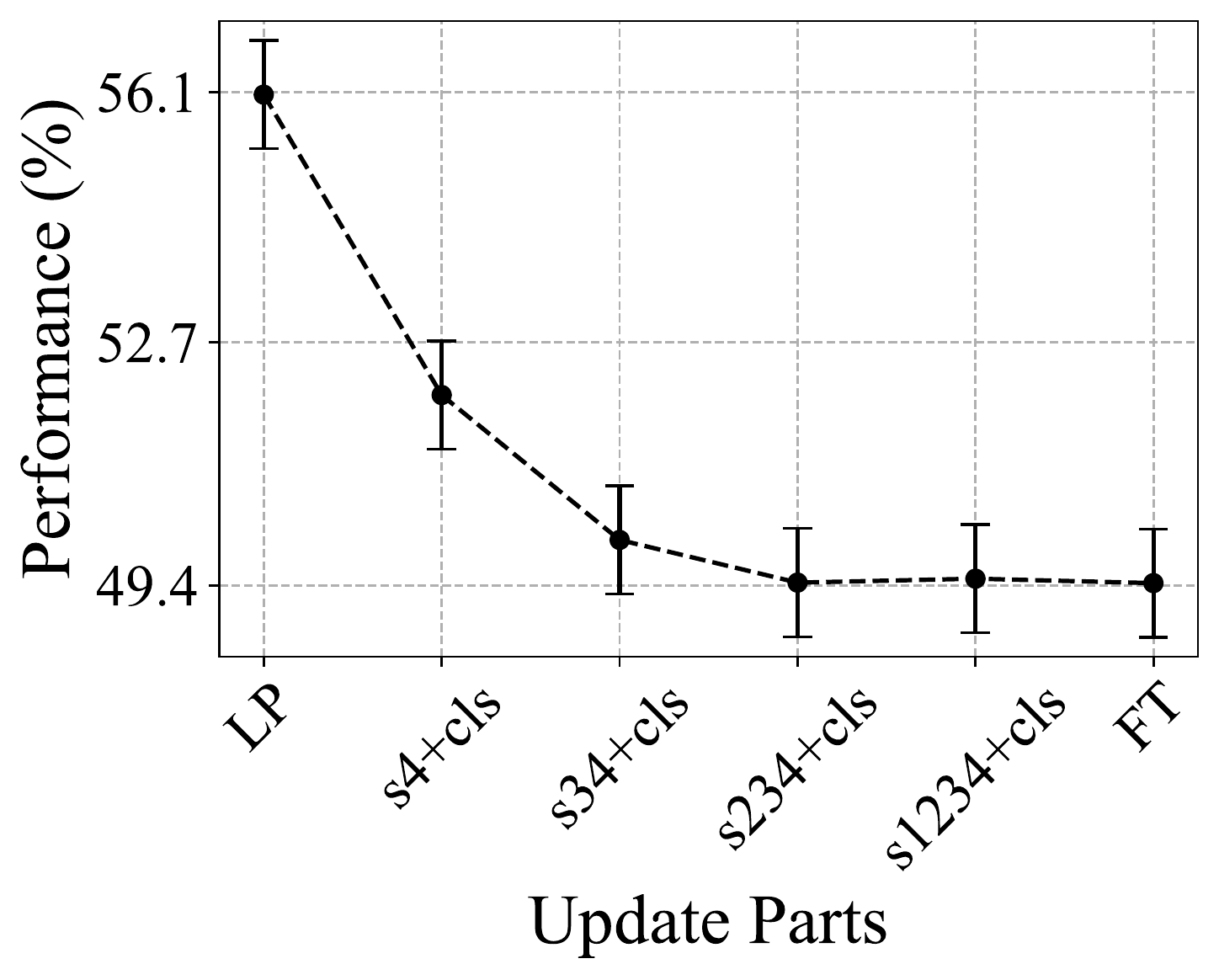}
    \caption{miniIN}
    \end{subfigure}
    \begin{subfigure}[h]{0.18\linewidth}
    \includegraphics[width=\linewidth]{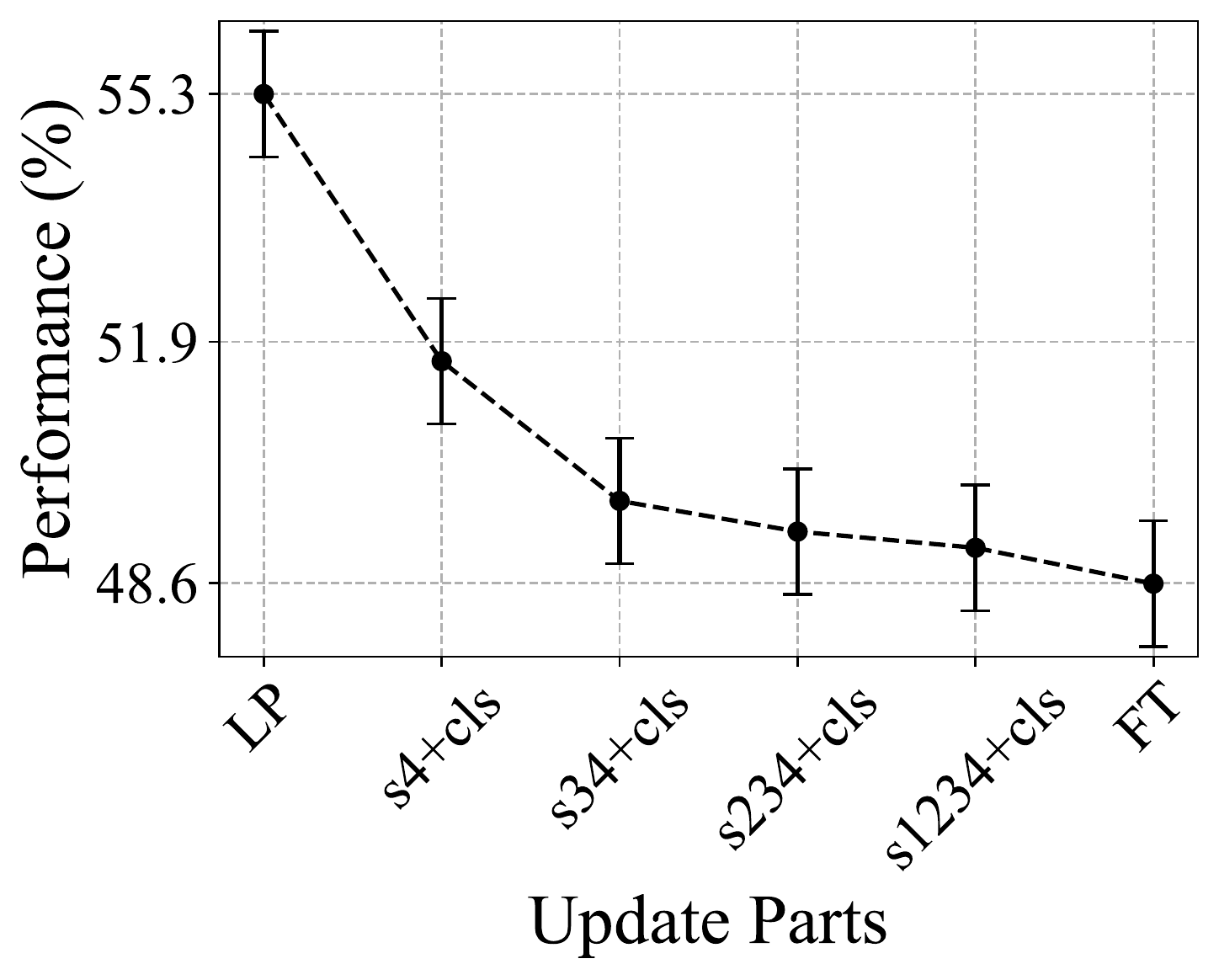}
    \caption{tieredIN}
    \end{subfigure}
    \centering
    \begin{subfigure}[h]{0.18\linewidth} 
    \includegraphics[width=\linewidth]{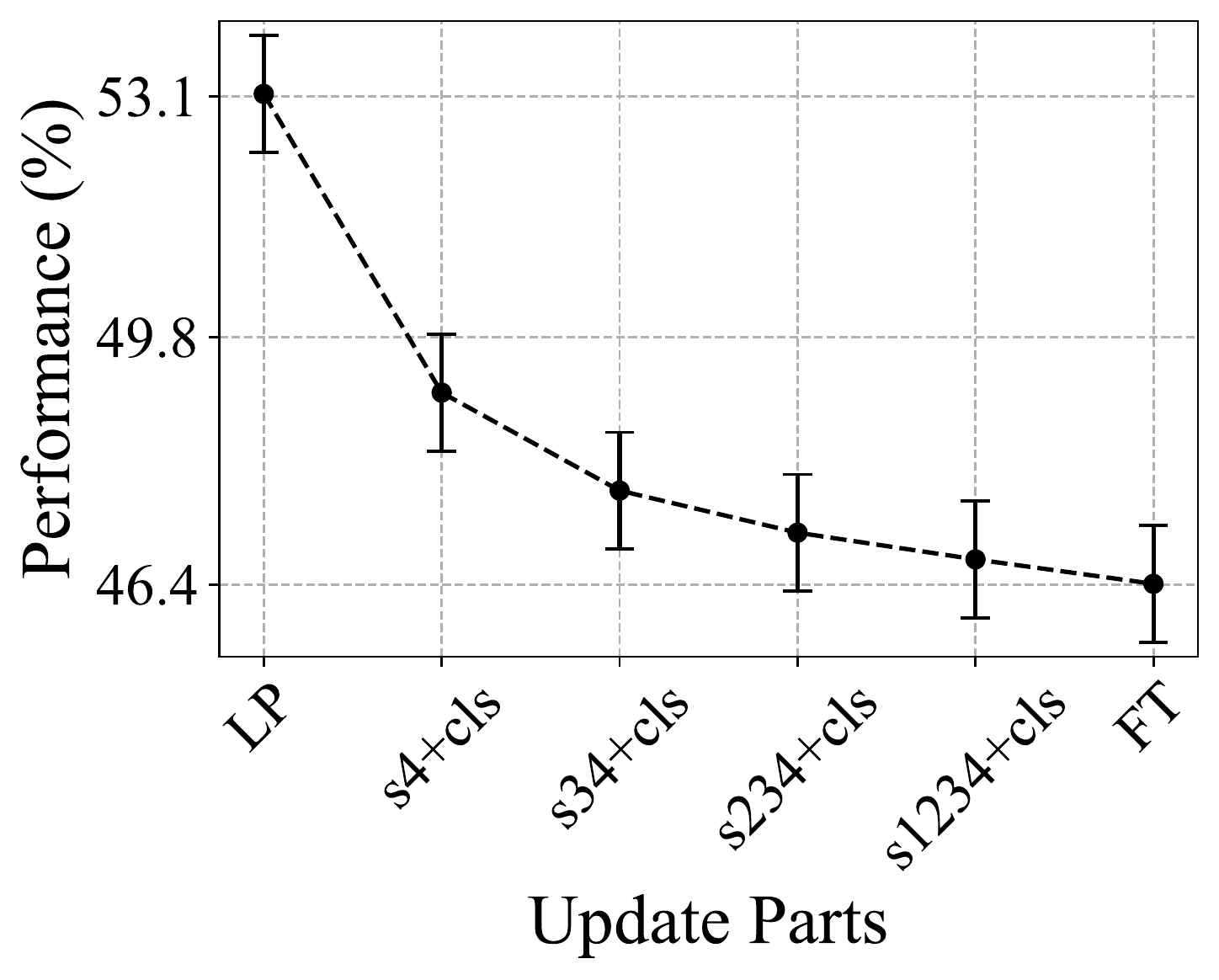}
    \caption{Places}
    \end{subfigure}
    \begin{subfigure}[h]{0.18\linewidth}
    \includegraphics[width=\linewidth]{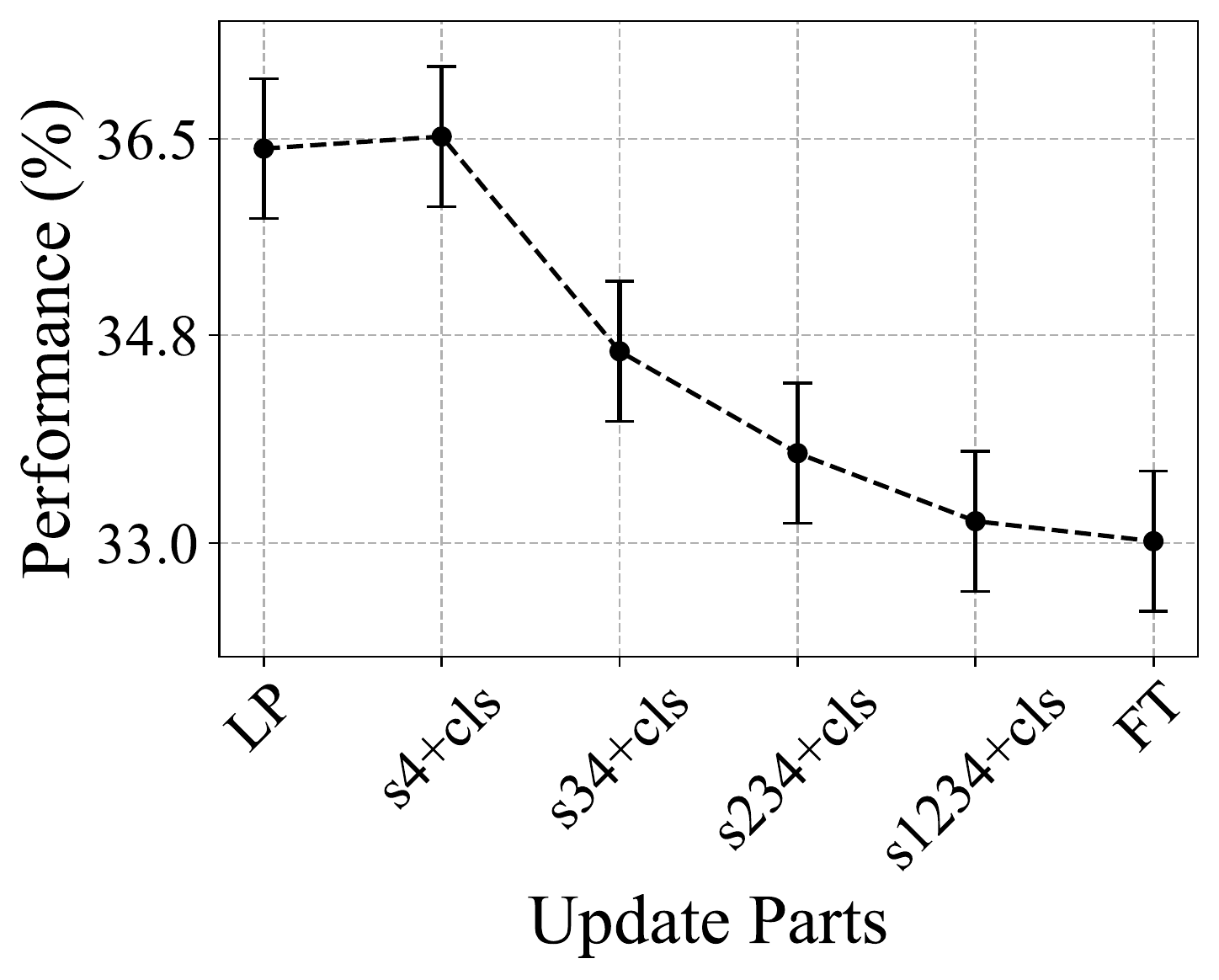}
    \caption{Plantae}
    \end{subfigure}
    \begin{subfigure}[h]{0.18\linewidth}
    \includegraphics[width=\linewidth]{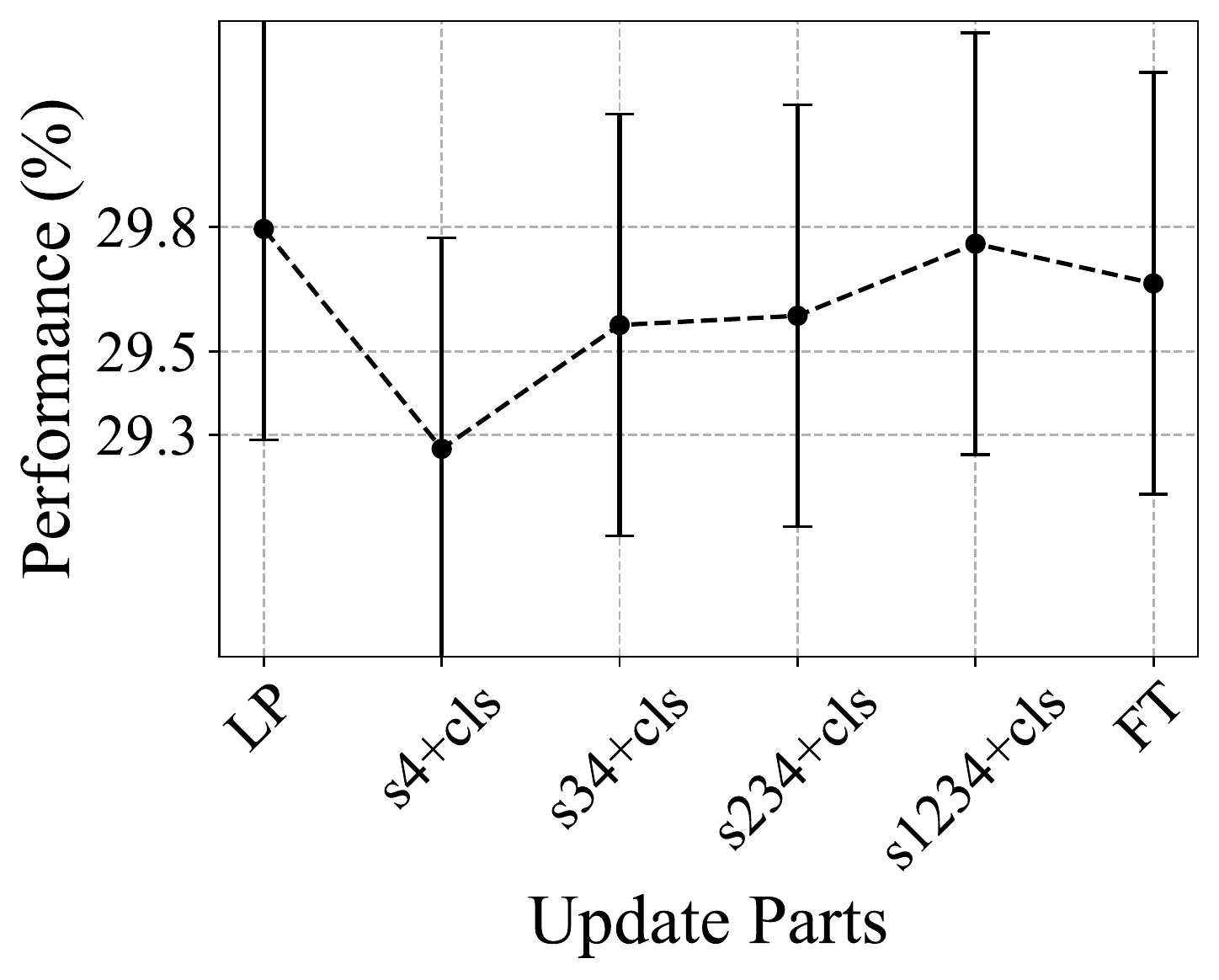}
    \caption{Cars}
    \end{subfigure}
    
    \centering
    \begin{subfigure}[h]{0.18\linewidth} 
    \includegraphics[width=\linewidth]{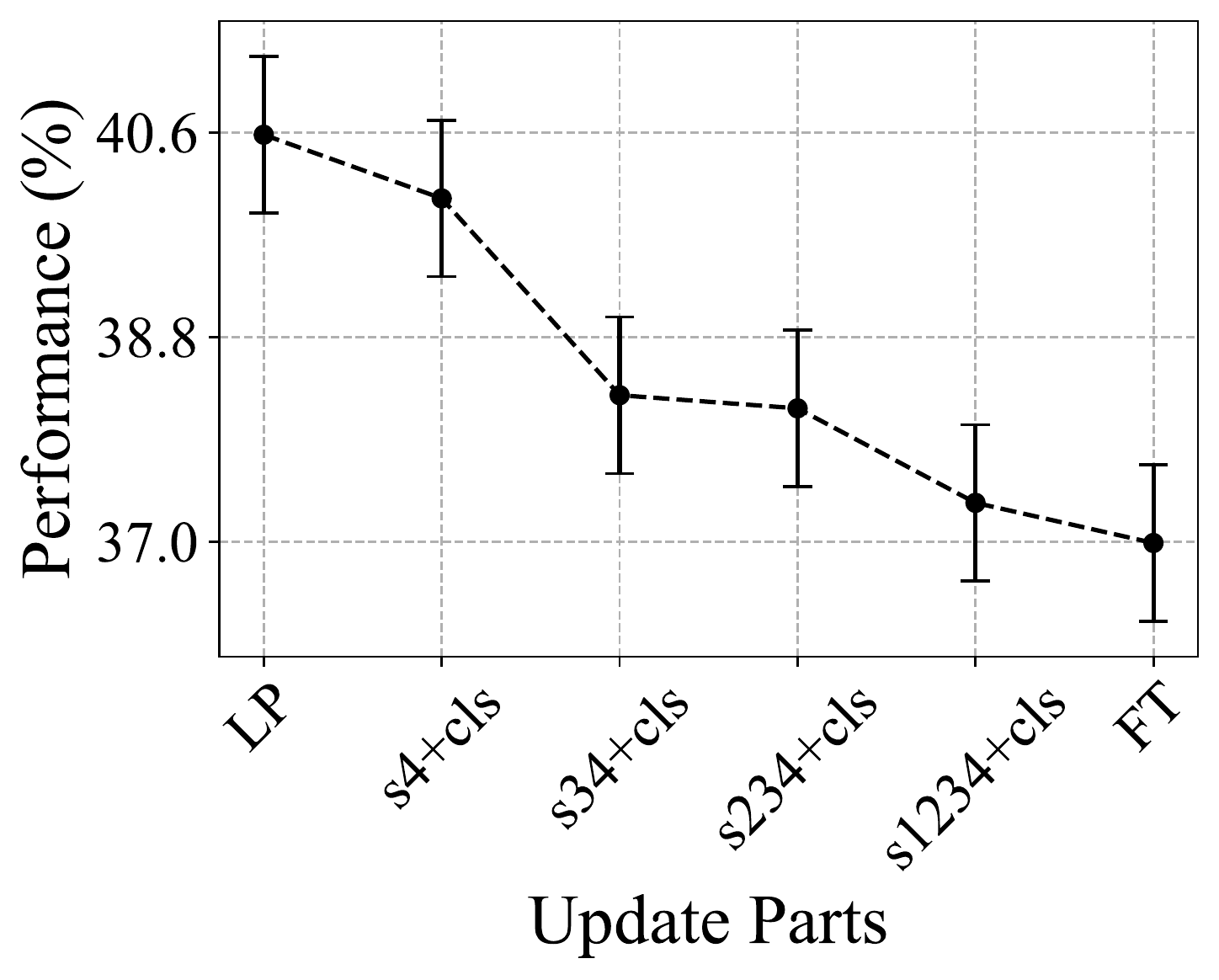}
    \caption{CUB}
    \end{subfigure}
    \begin{subfigure}[h]{0.18\linewidth}
    \includegraphics[width=\linewidth]{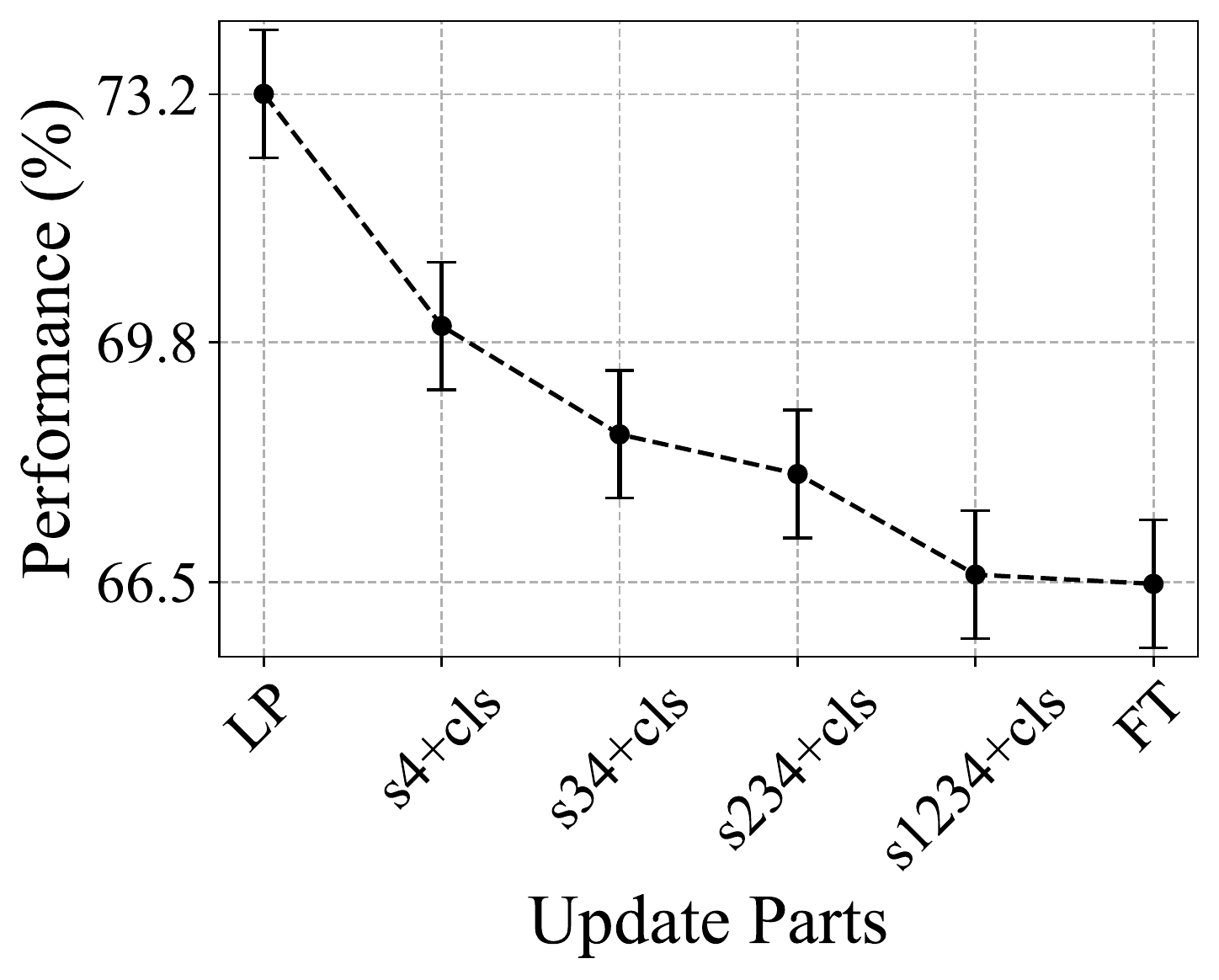}
    \caption{CropDiseases}
    \end{subfigure}
    \centering
    \begin{subfigure}[h]{0.18\linewidth} 
    \includegraphics[width=\linewidth]{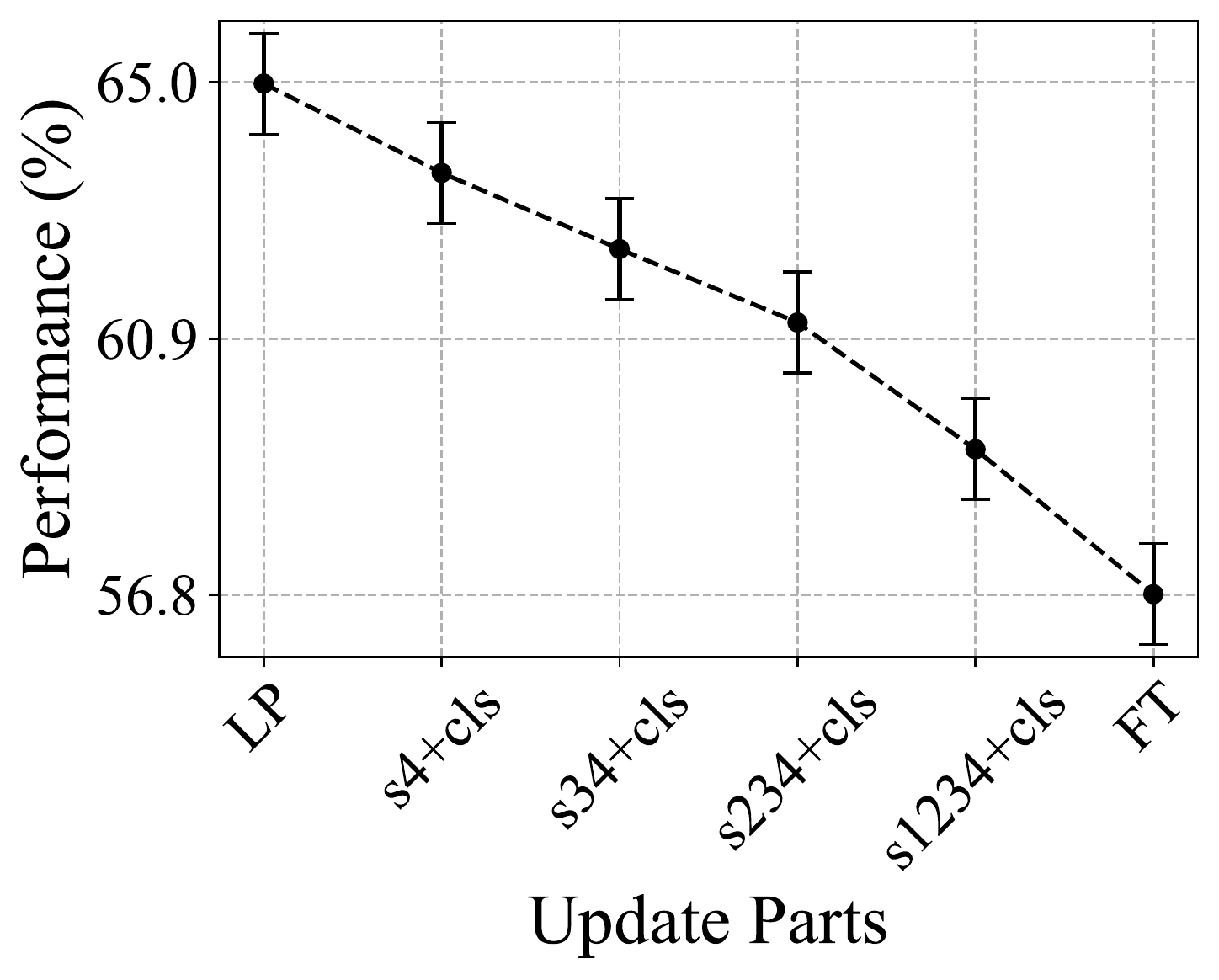}
    \caption{EuroSAT}
    \end{subfigure}
    \begin{subfigure}[h]{0.18\linewidth}
    \includegraphics[width=\linewidth]{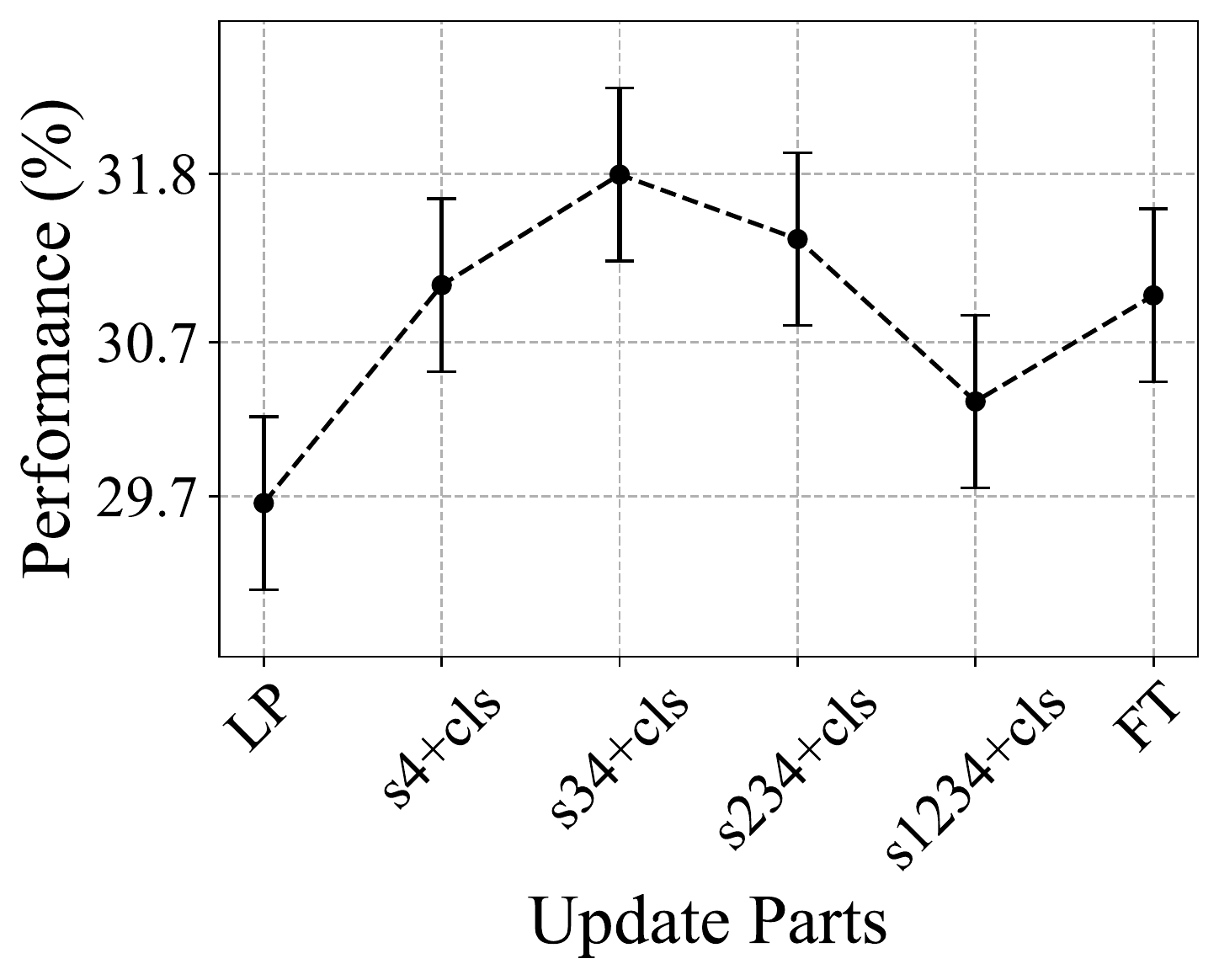}
    \caption{ISIC}
    \end{subfigure}
    \begin{subfigure}[h]{0.18\linewidth}
    \includegraphics[width=\linewidth]{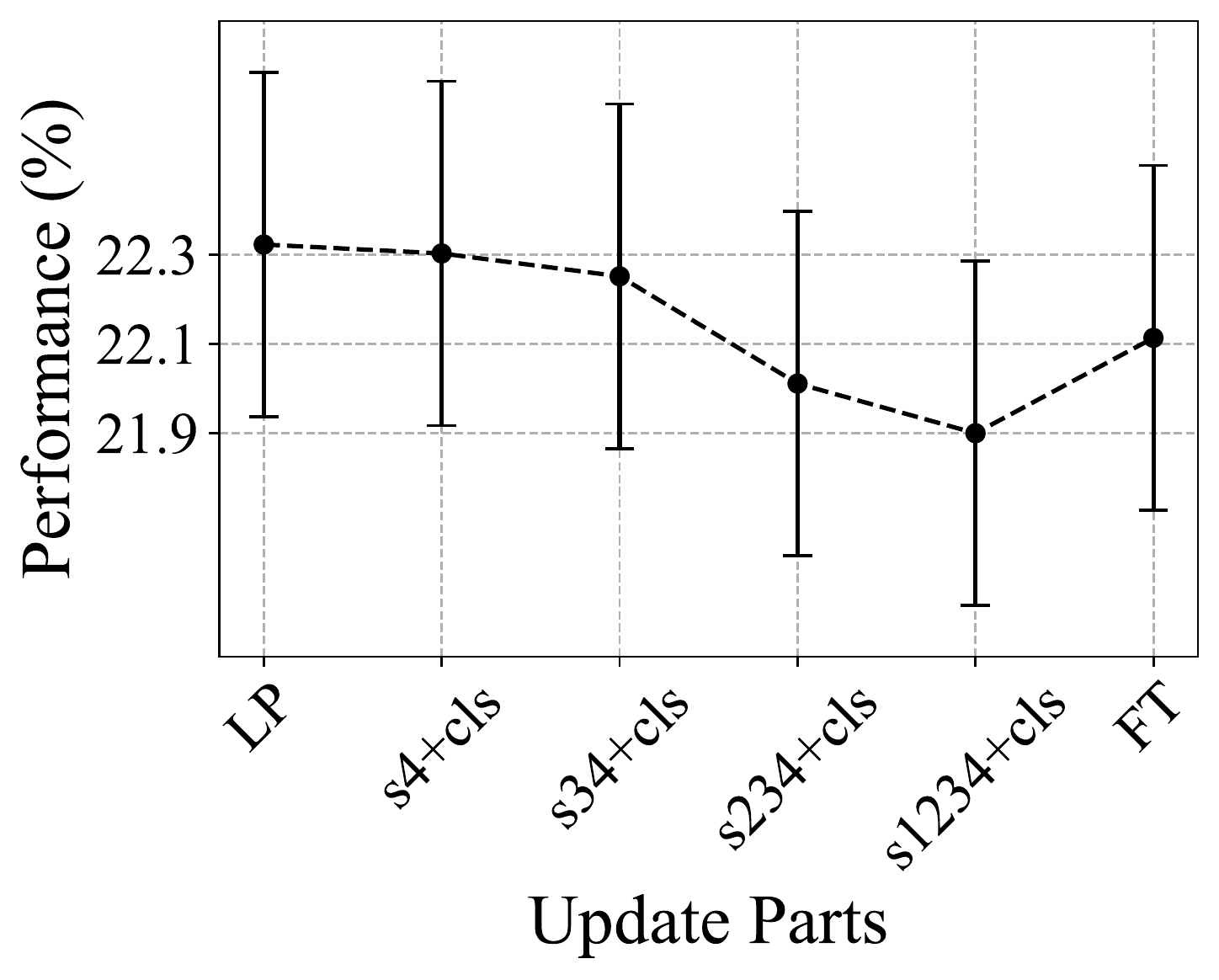}
    \caption{ChestX}
    \end{subfigure}
    \caption{$5$-way $1$-shot results of partial updates between LP and FT.}
    \label{fig:interpolation_1shot}
\end{figure*}

\begin{figure*}[!ht]
    \centering
    \begin{subfigure}[h]{0.18\linewidth}
    \includegraphics[width=\linewidth]{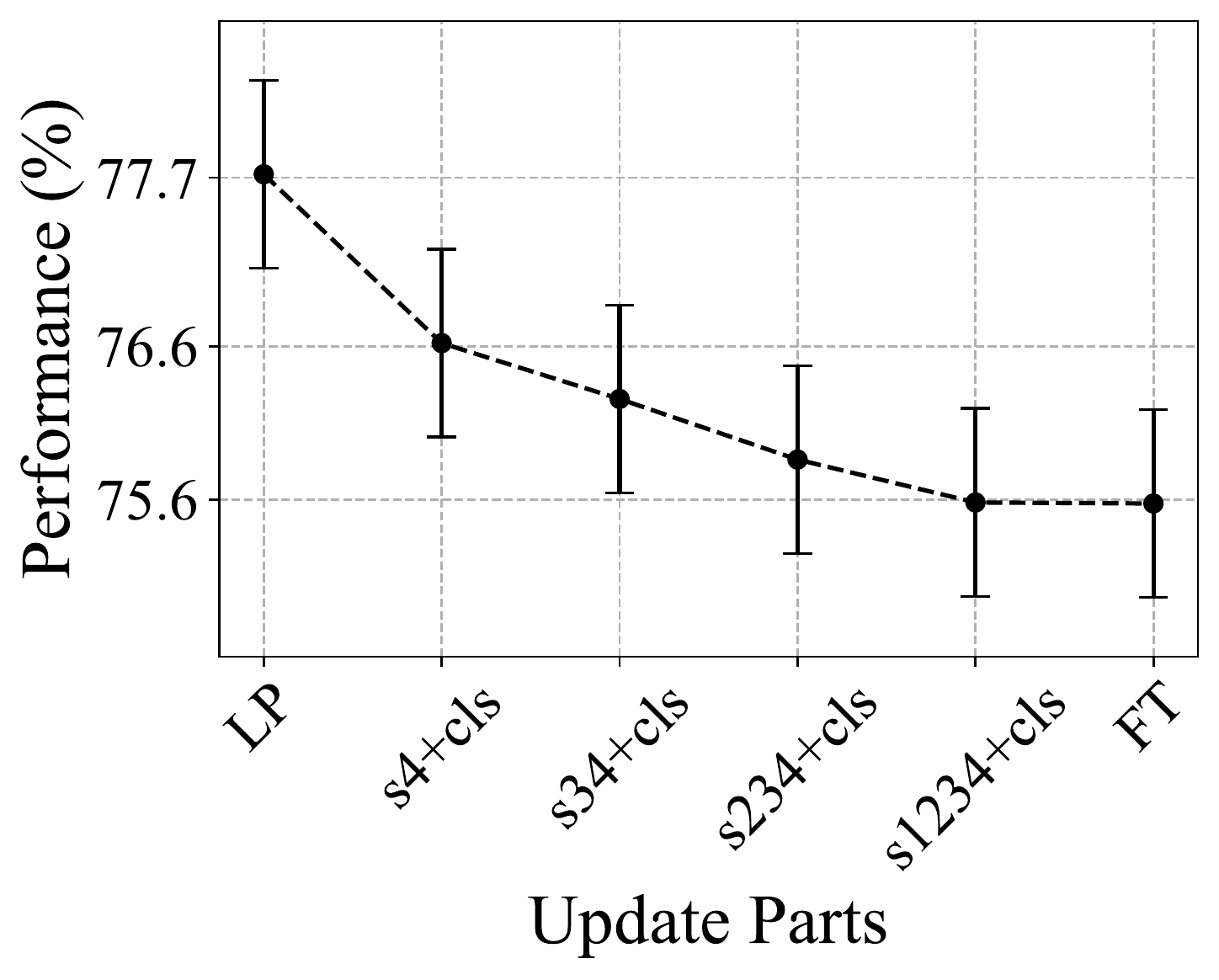}
    \caption{miniIN}
    \end{subfigure}
    \begin{subfigure}[h]{0.18\linewidth}
    \includegraphics[width=\linewidth]{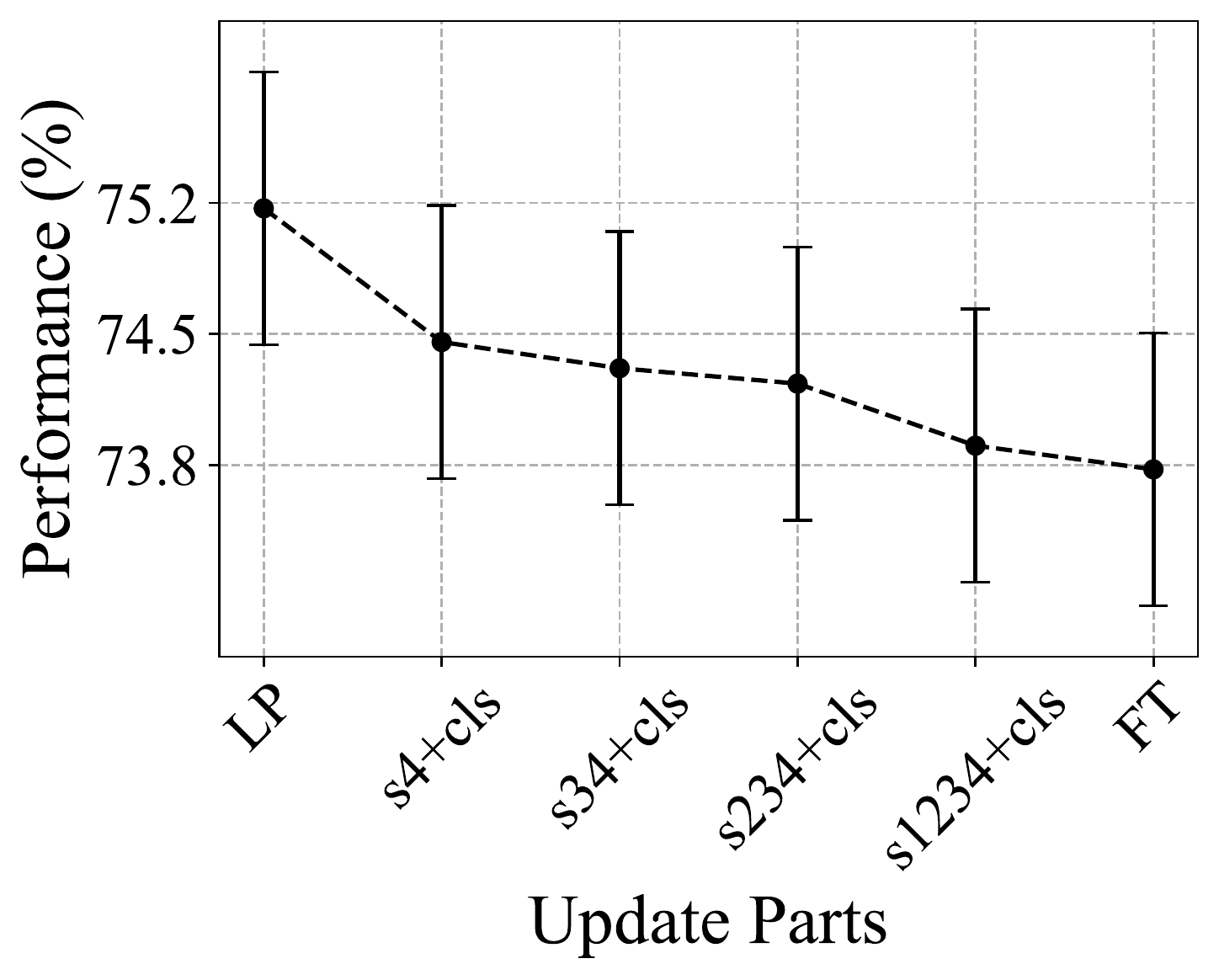}
    \caption{tieredIN}
    \end{subfigure}
    \centering
    \begin{subfigure}[h]{0.18\linewidth} 
    \includegraphics[width=\linewidth]{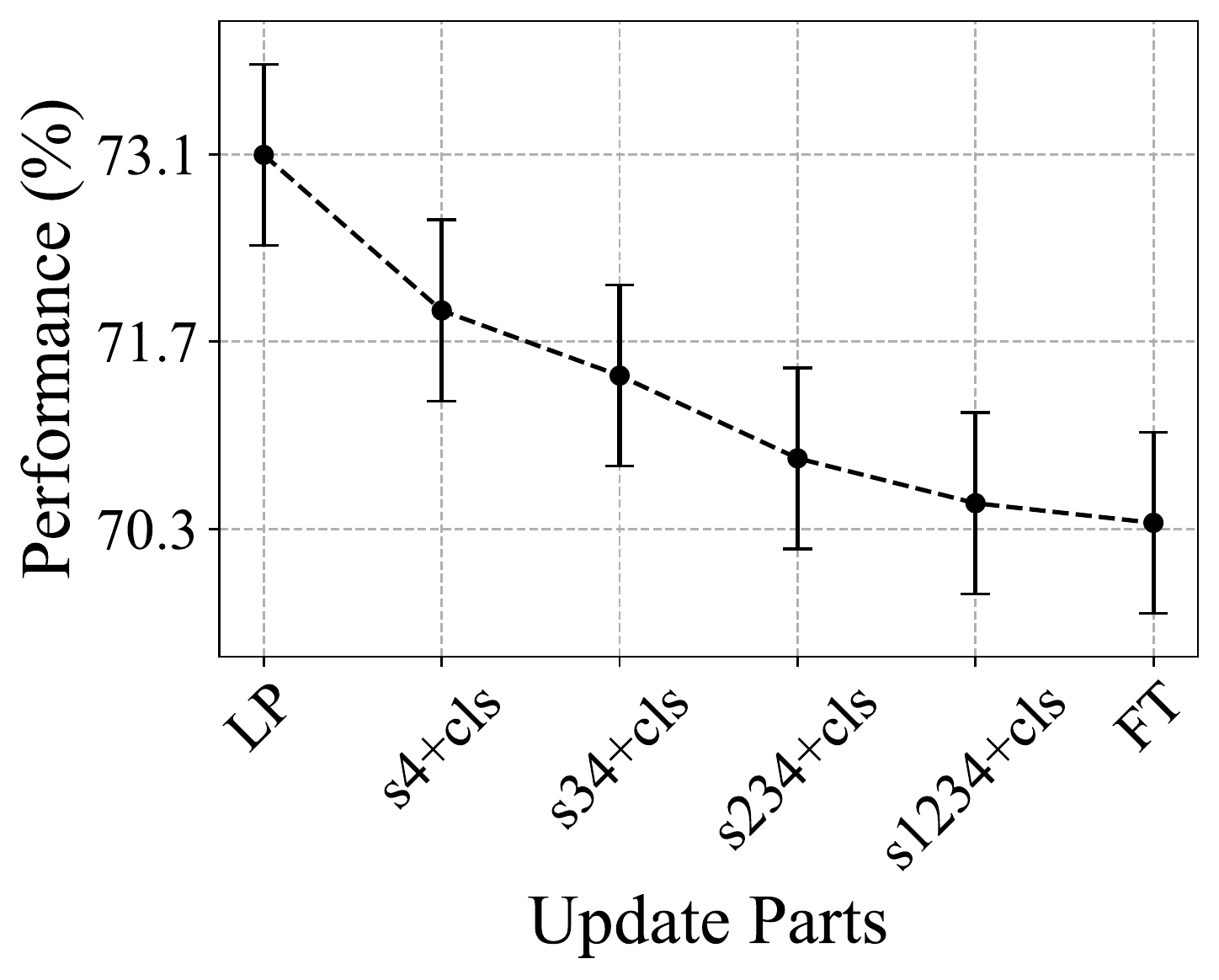}
    \caption{Places}
    \end{subfigure}
    \begin{subfigure}[h]{0.18\linewidth}
    \includegraphics[width=\linewidth]{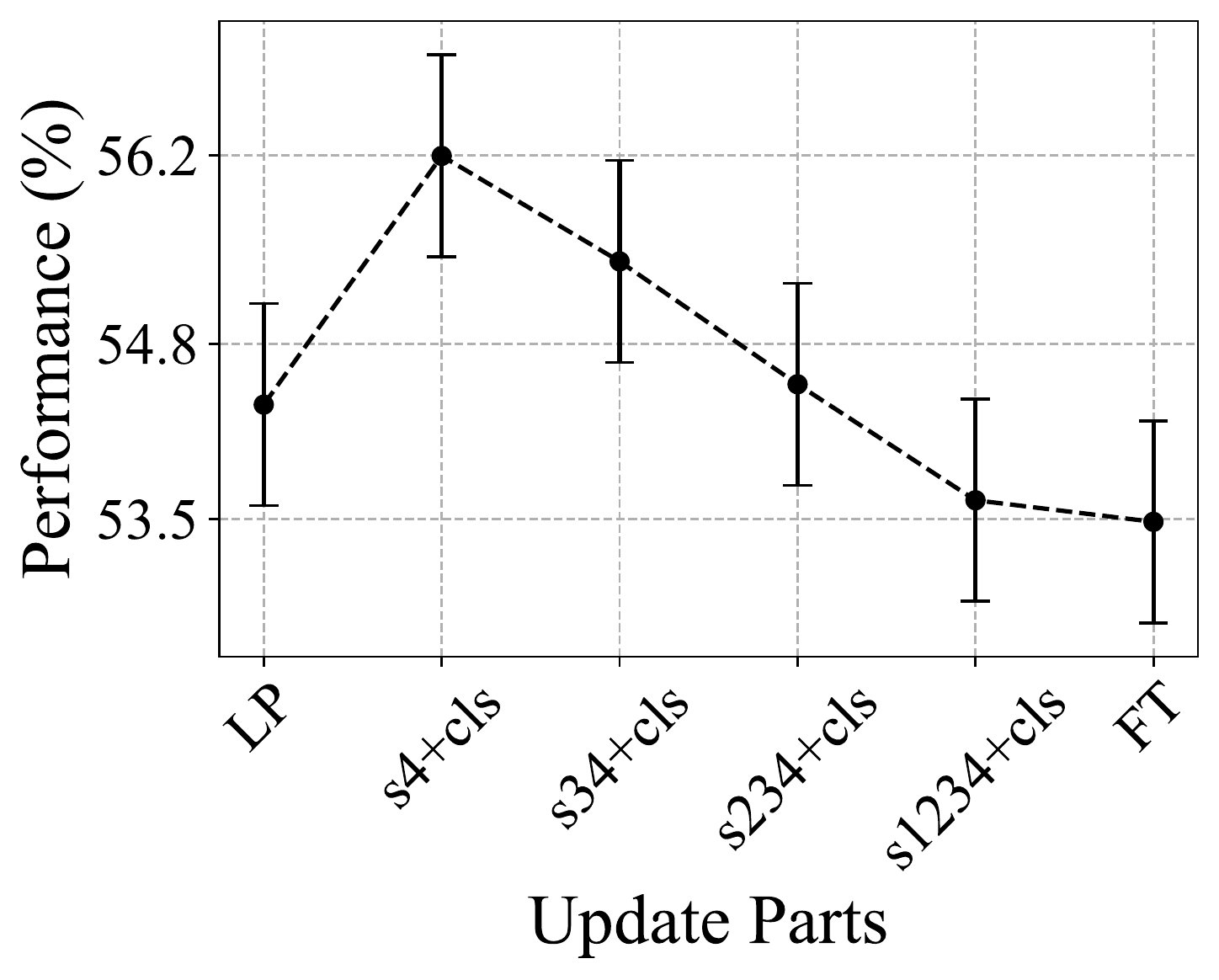}
    \caption{Plantae}
    \end{subfigure}
    \begin{subfigure}[h]{0.18\linewidth}
    \includegraphics[width=\linewidth]{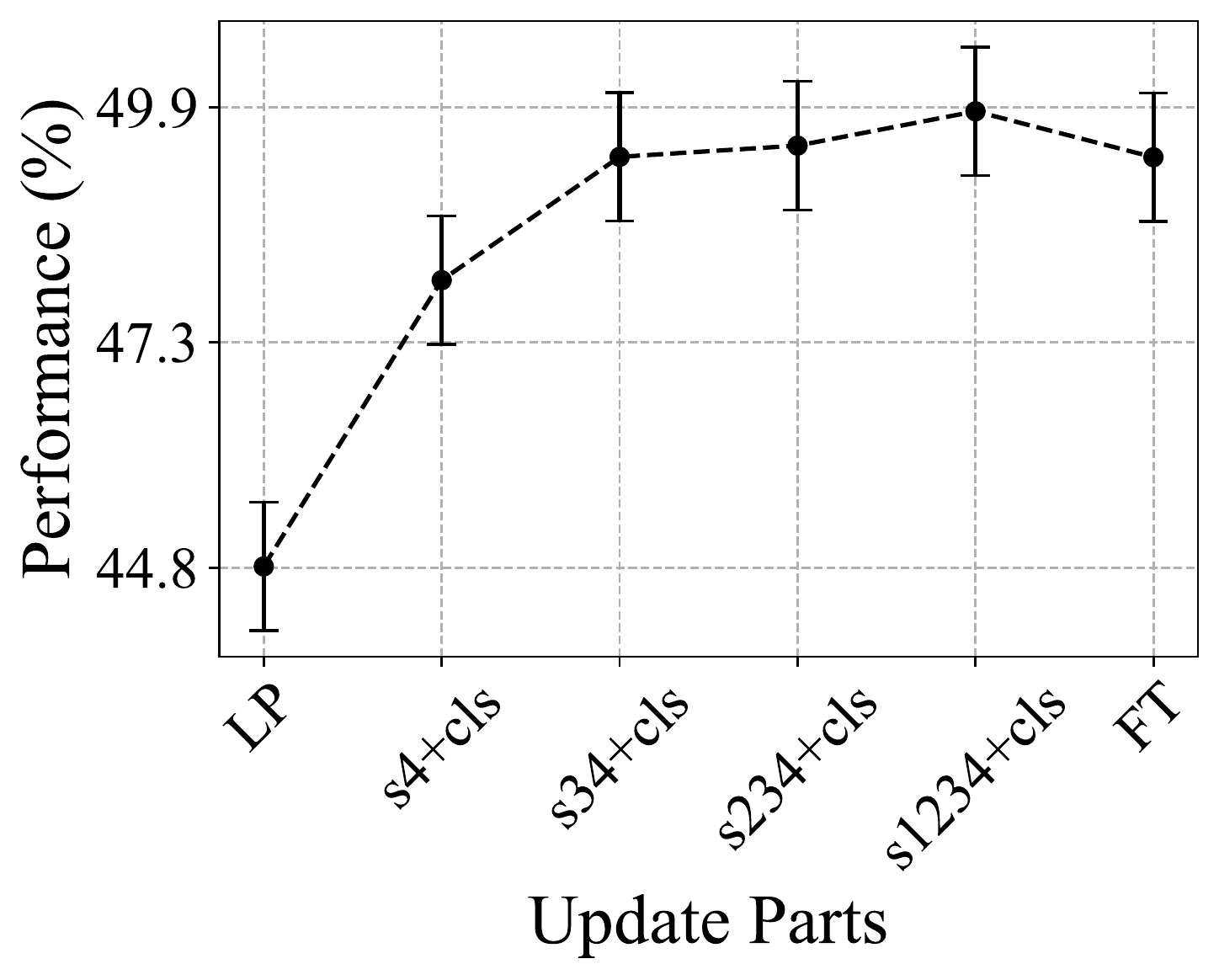}
    \caption{Cars}
    \end{subfigure}
    
    \centering
    \begin{subfigure}[h]{0.18\linewidth} 
    \includegraphics[width=\linewidth]{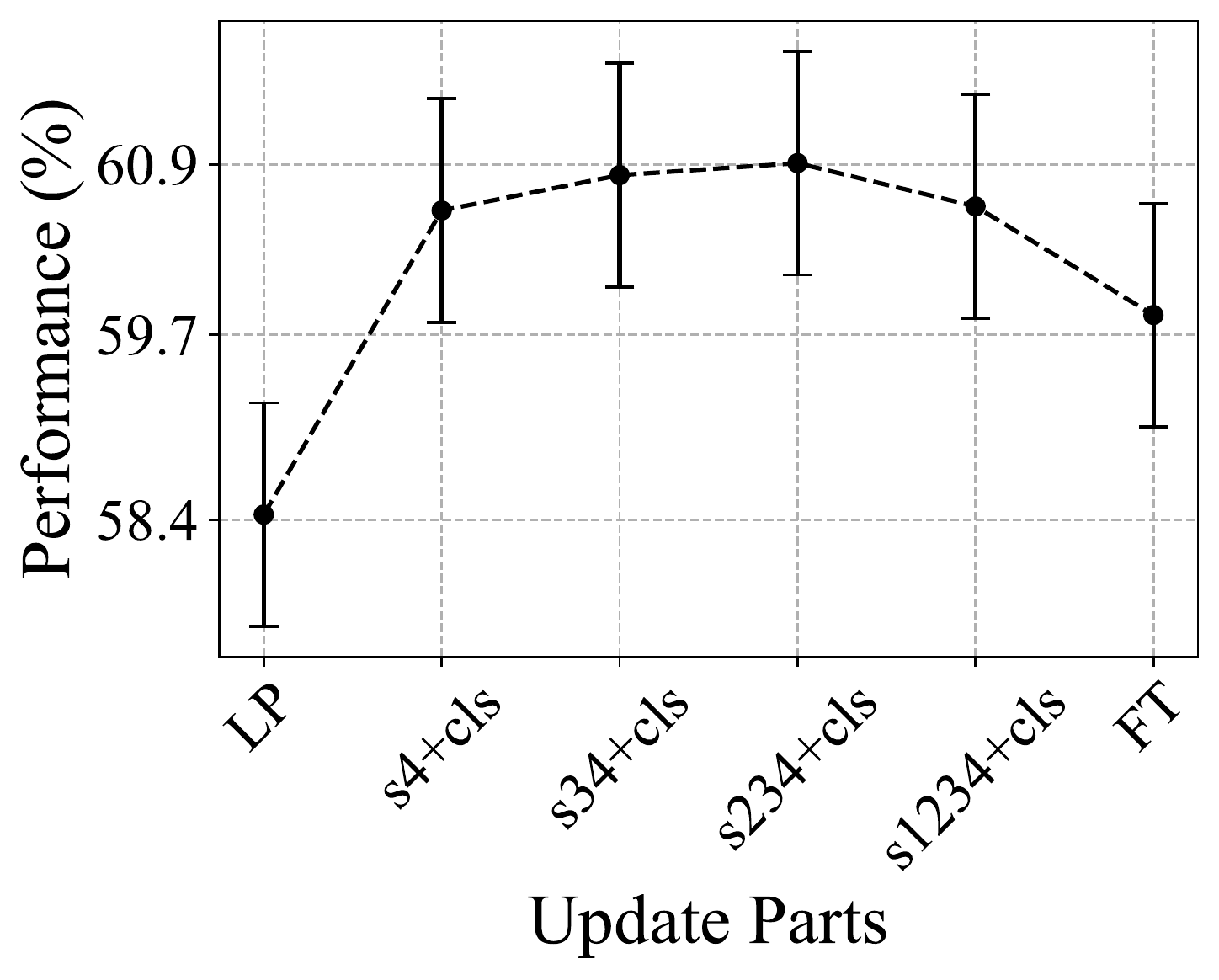}
    \caption{CUB}
    \end{subfigure}
    \begin{subfigure}[h]{0.18\linewidth}
    \includegraphics[width=\linewidth]{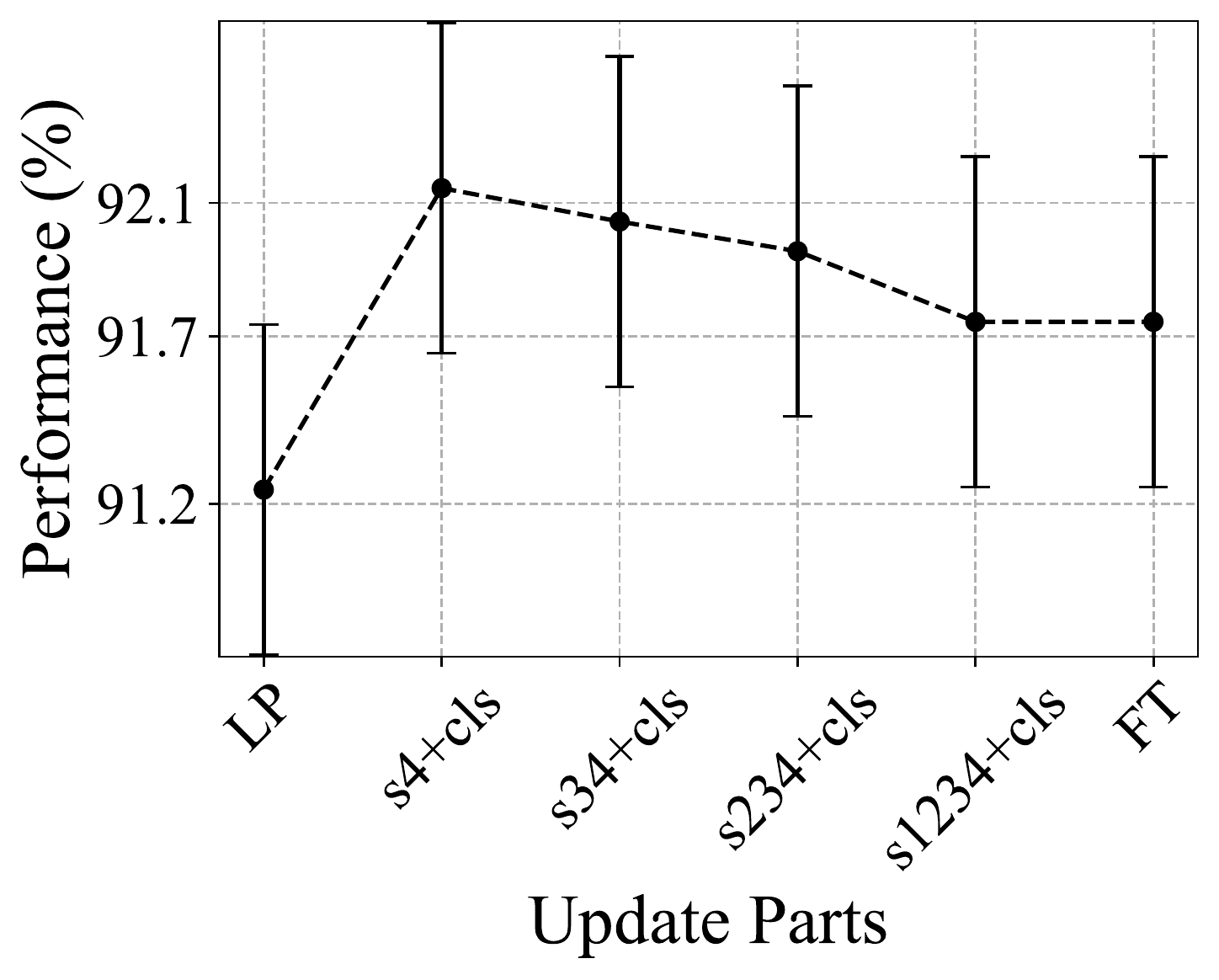}
    \caption{CropDiseases}
    \end{subfigure}
    \centering
    \begin{subfigure}[h]{0.18\linewidth} 
    \includegraphics[width=\linewidth]{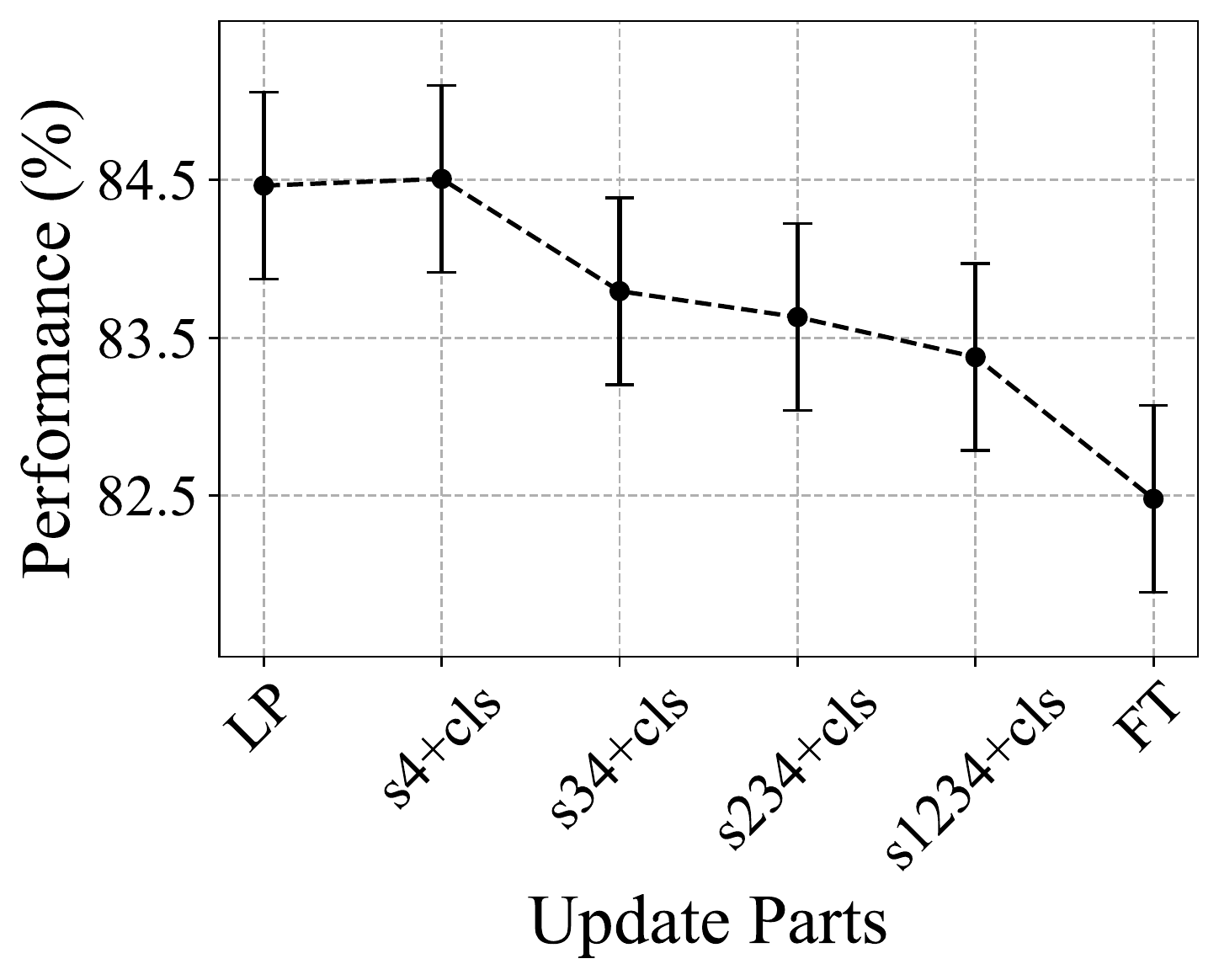}
    \caption{EuroSAT}
    \end{subfigure}
    \begin{subfigure}[h]{0.18\linewidth}
    \includegraphics[width=\linewidth]{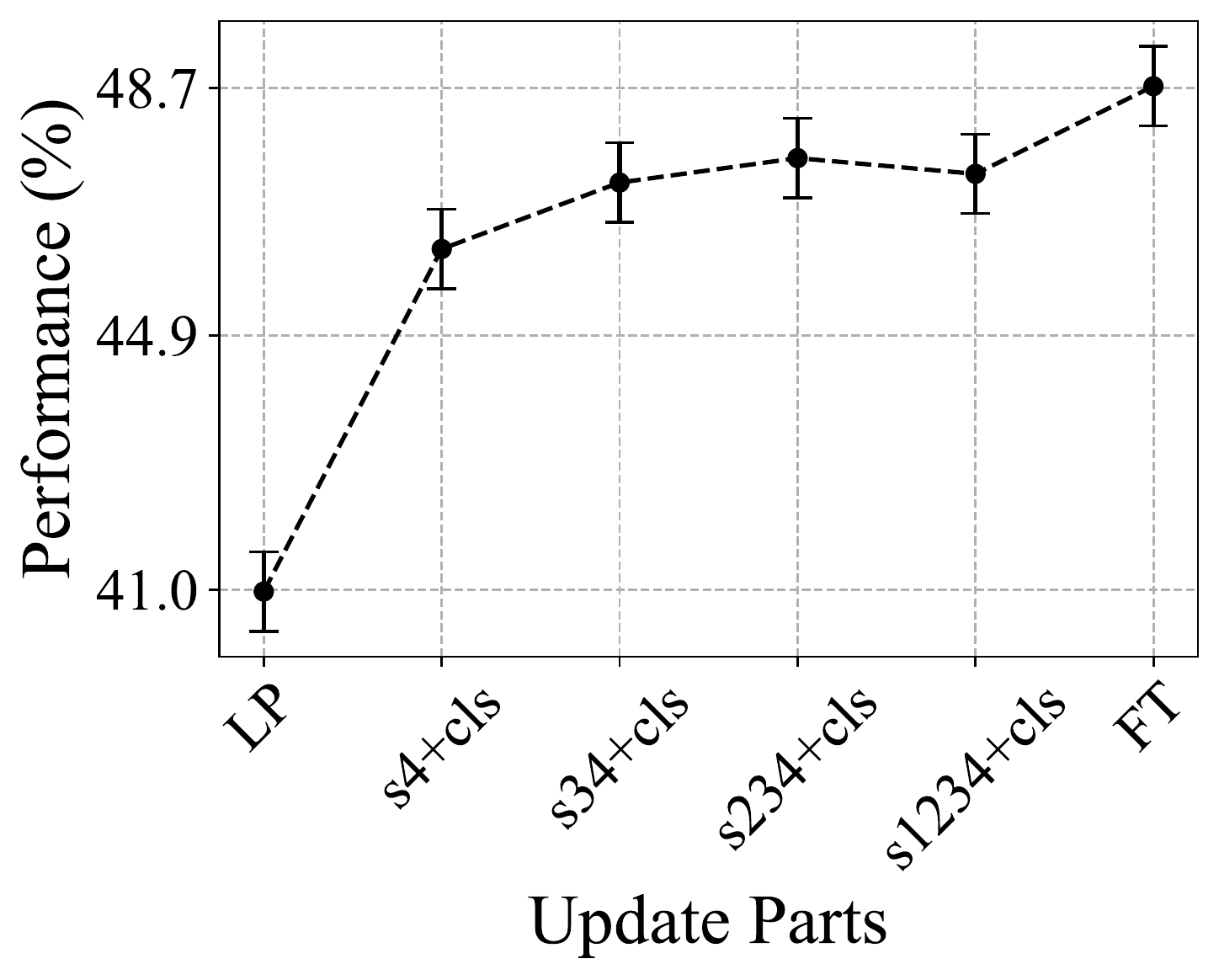}
    \caption{ISIC}
    \end{subfigure}
    \begin{subfigure}[h]{0.18\linewidth}
    \includegraphics[width=\linewidth]{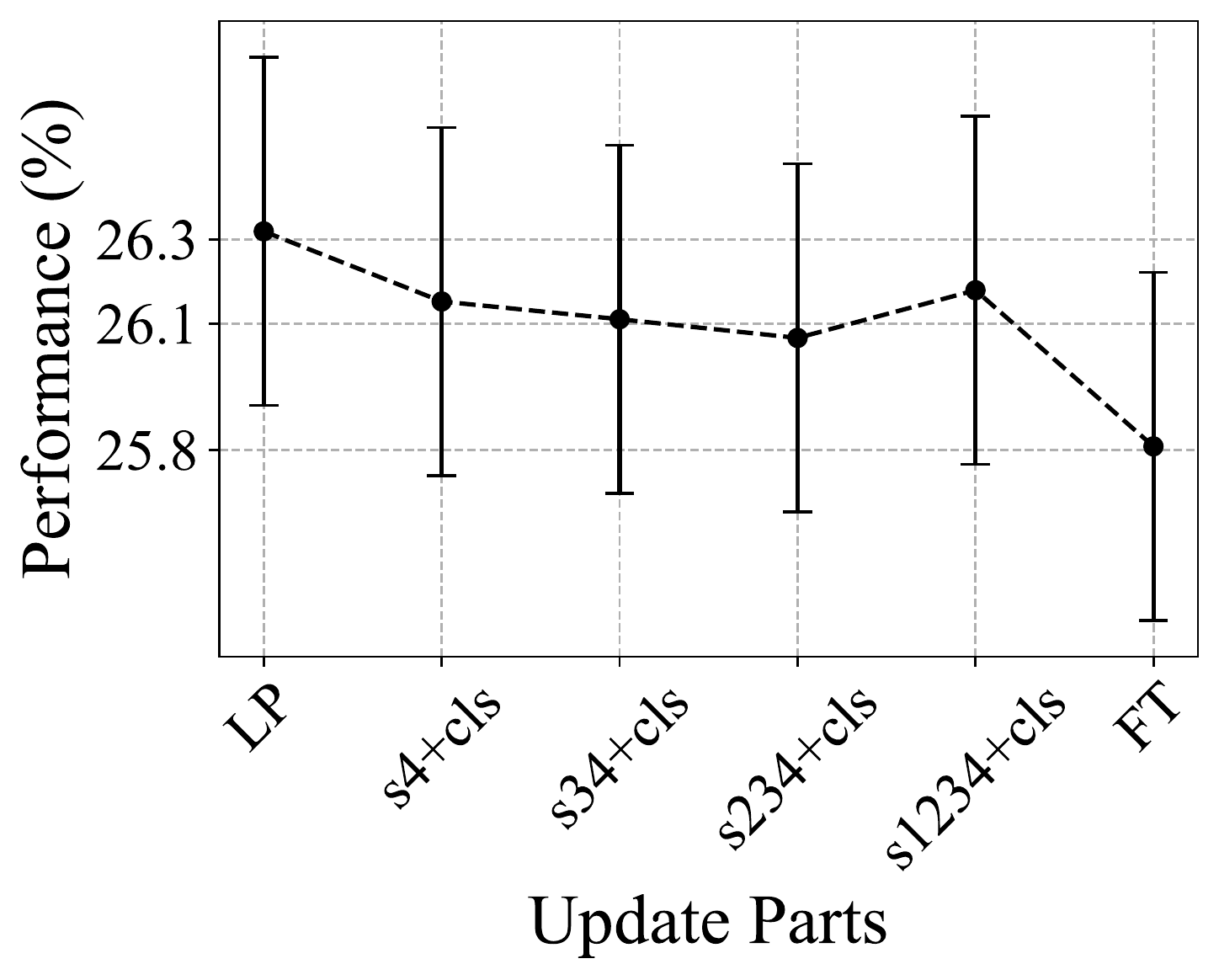}
    \caption{ChestX}
    \end{subfigure}
    \caption{$5$-way $5$-shot results of partial updates between LP and FT.}
    \label{fig:interpolation_5shot}
\end{figure*}

\begin{figure*}[!ht]
    \centering
    \begin{subfigure}[h]{0.18\linewidth} 
    \includegraphics[width=\linewidth]{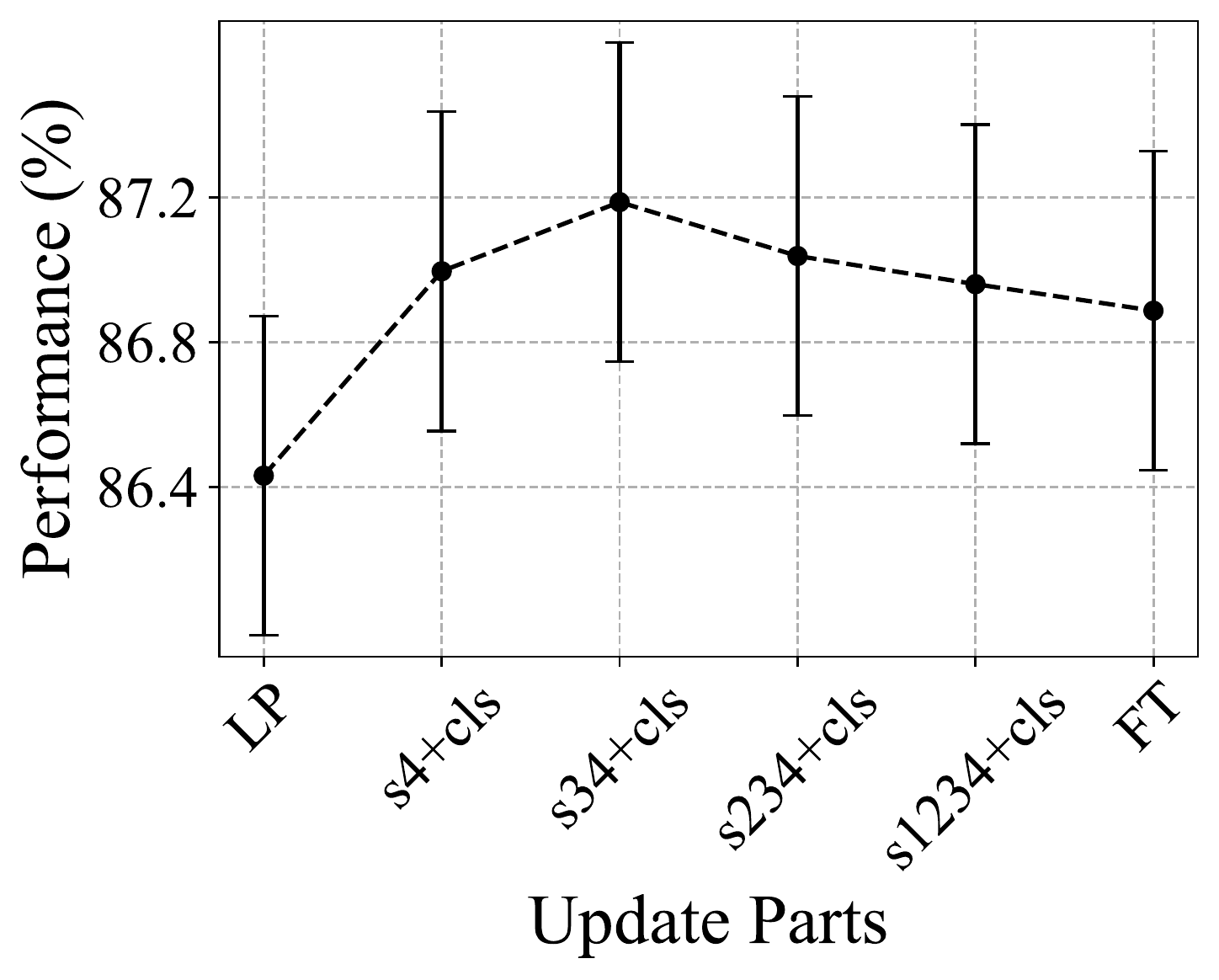}
    \caption{miniIN}
    \end{subfigure}
    \begin{subfigure}[h]{0.18\linewidth}
    \includegraphics[width=\linewidth]{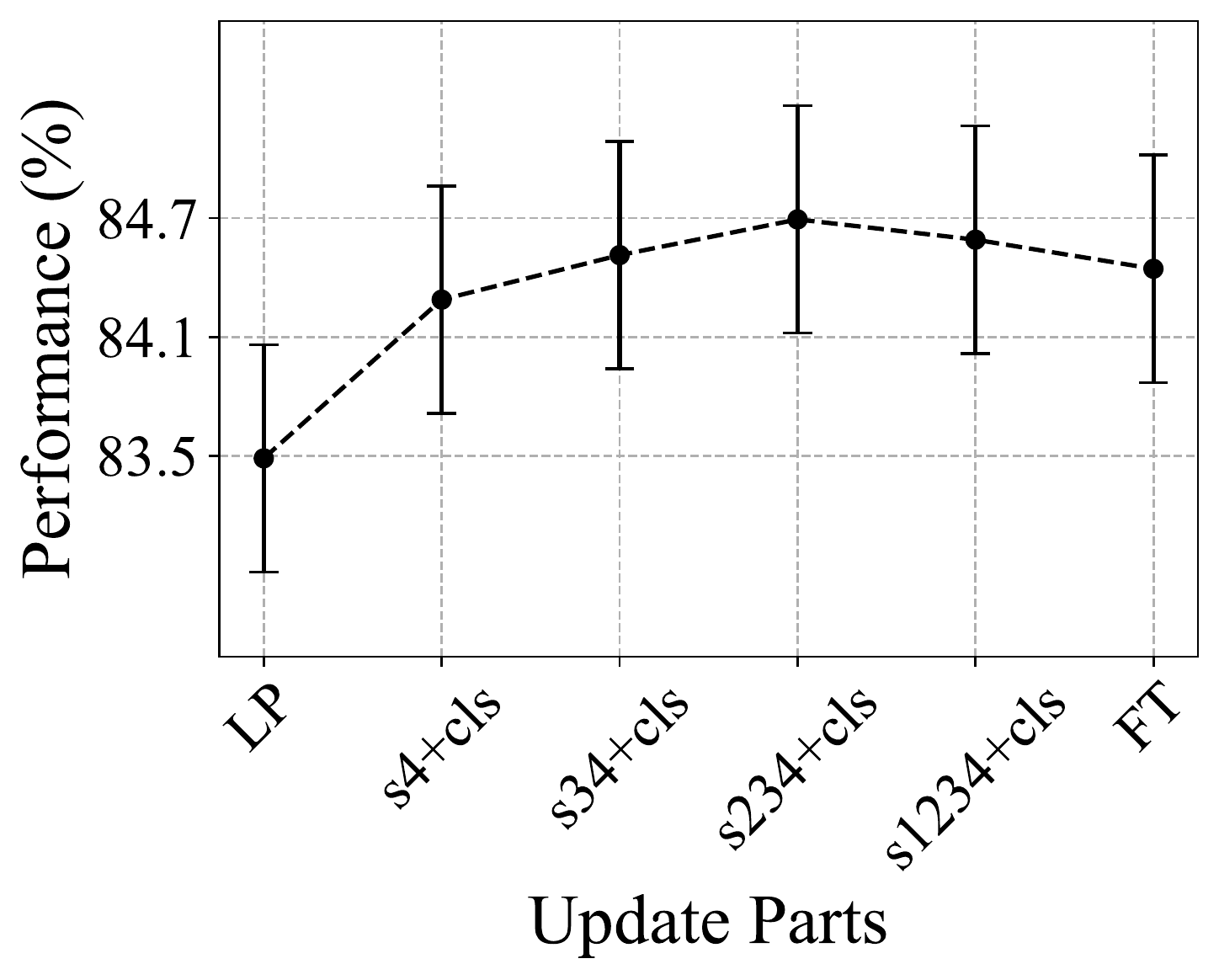}
    \caption{tieredIN}
    \end{subfigure}
    \centering
    \begin{subfigure}[h]{0.18\linewidth} 
    \includegraphics[width=\linewidth]{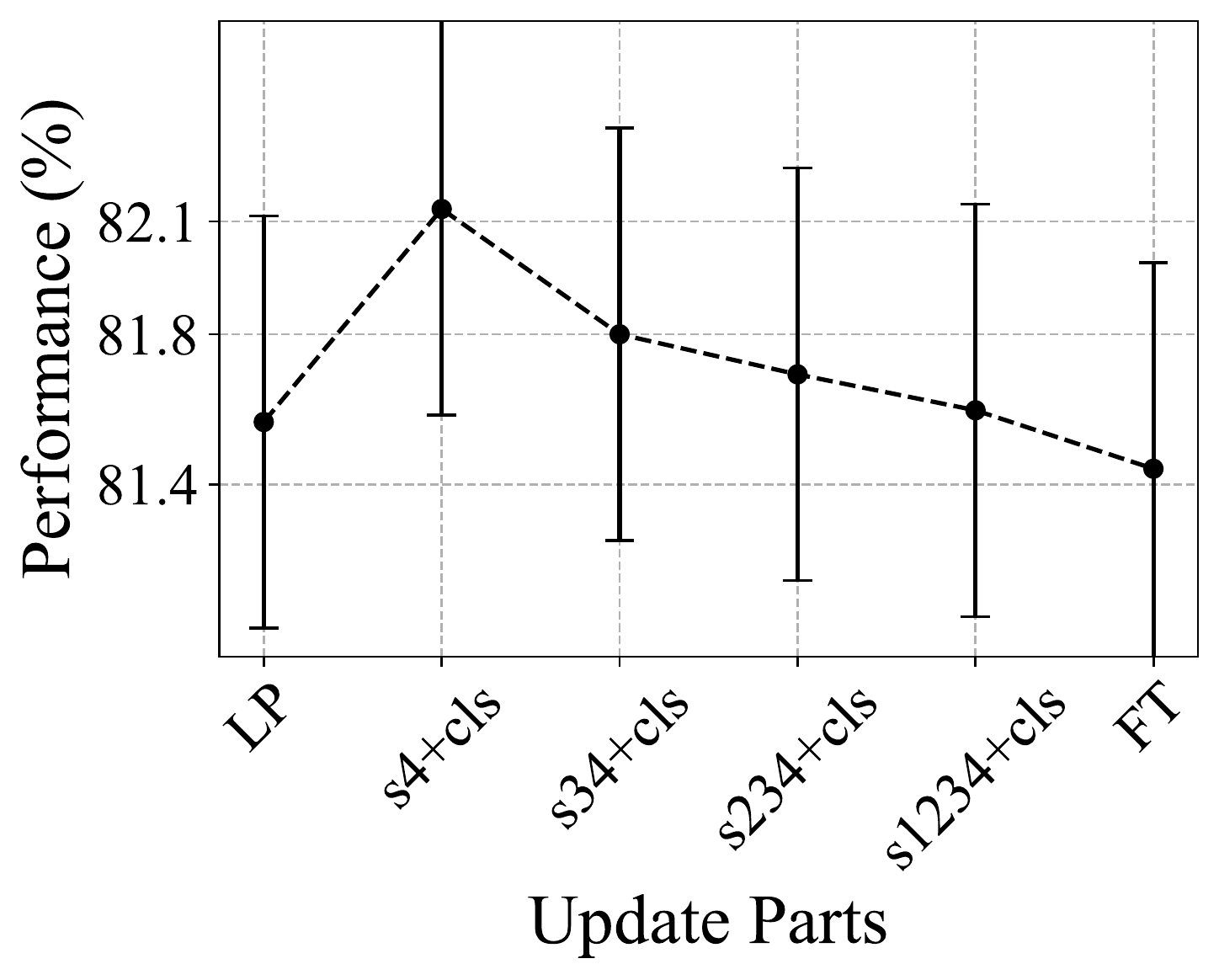}
    \caption{Places}
    \end{subfigure}
    \begin{subfigure}[h]{0.18\linewidth}
    \includegraphics[width=\linewidth]{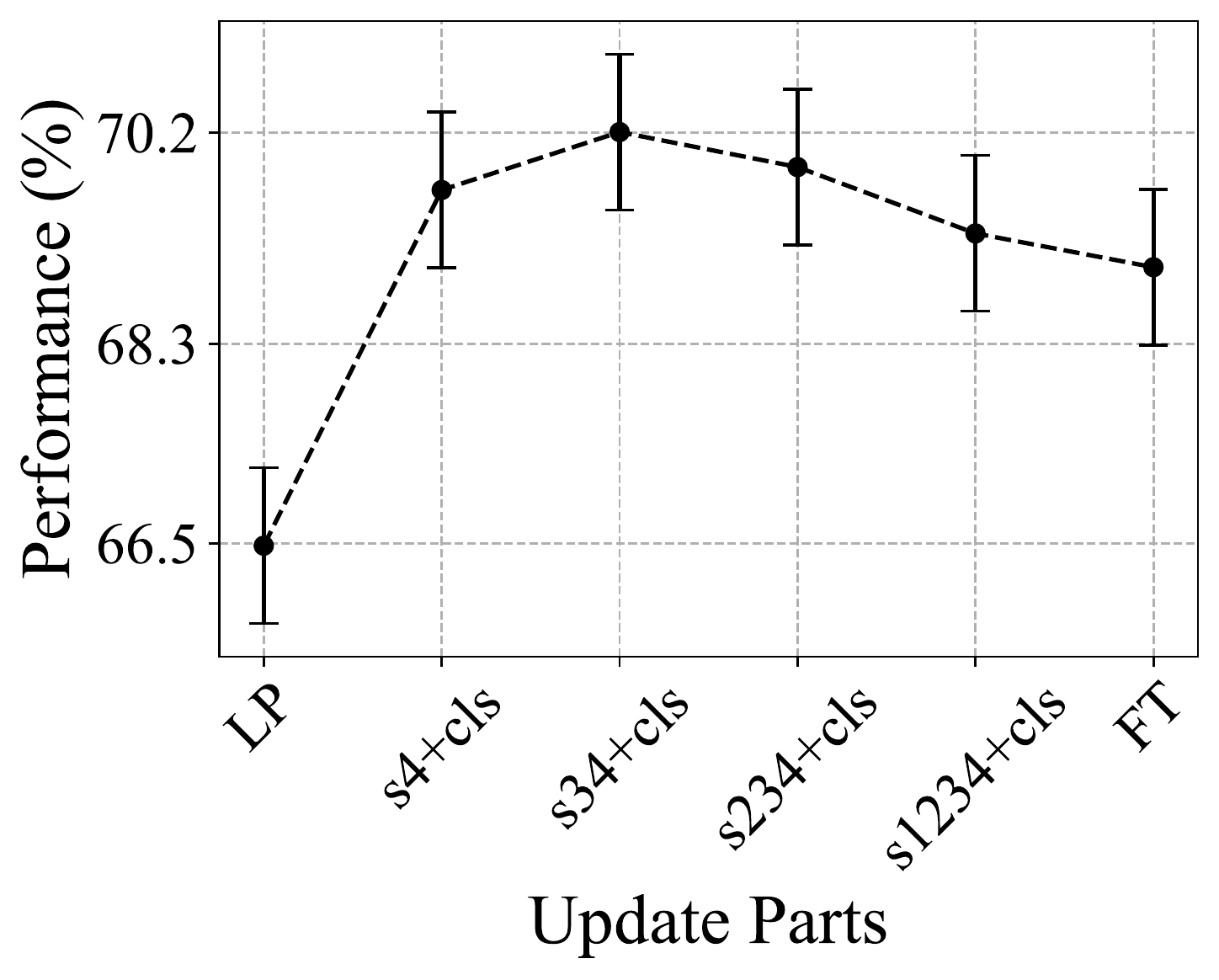}
    \caption{Plantae}
    \end{subfigure}
    \begin{subfigure}[h]{0.18\linewidth}
    \includegraphics[width=\linewidth]{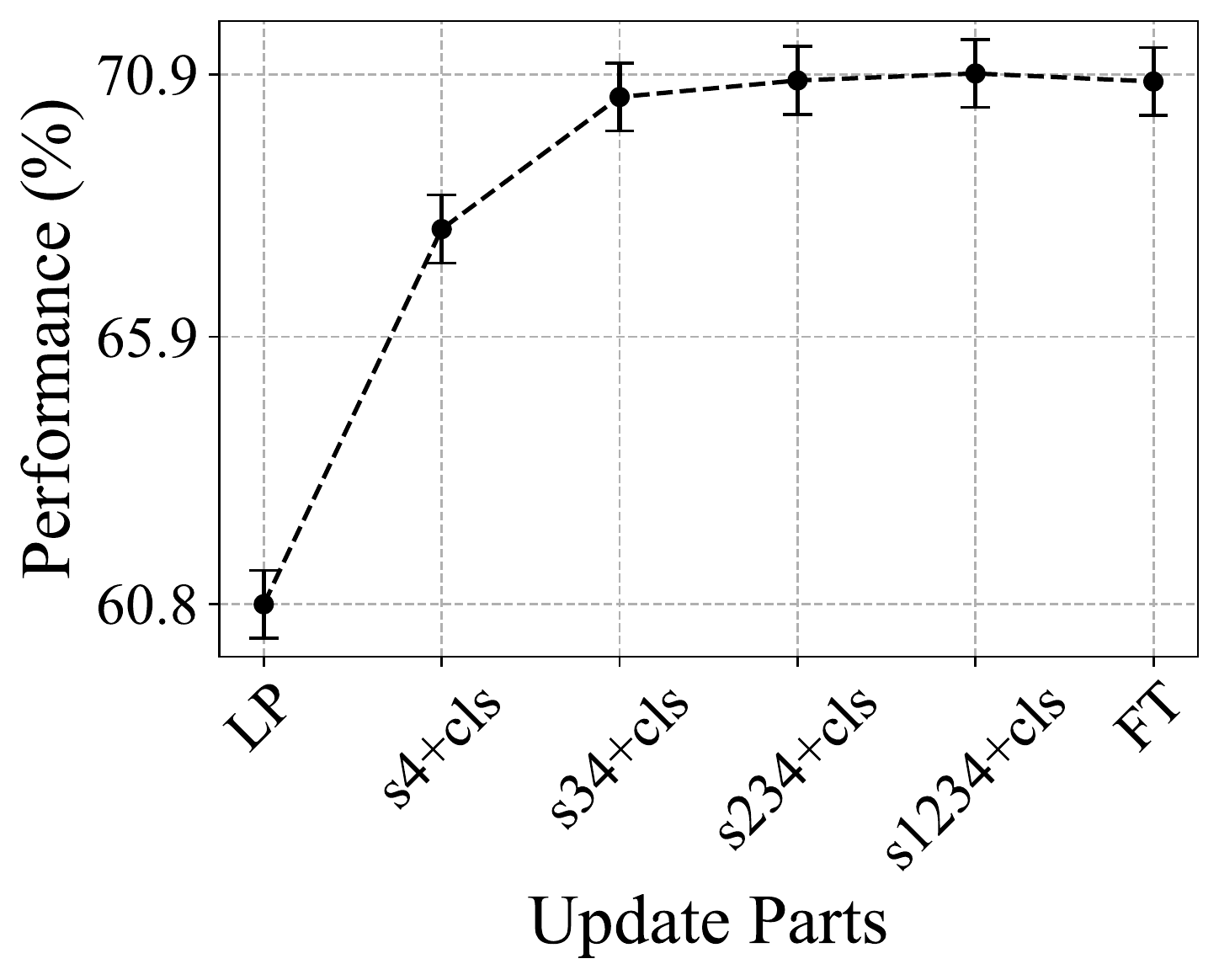}
    \caption{Cars}
    \end{subfigure}
    
    \centering
    \begin{subfigure}[h]{0.18\linewidth} 
    \includegraphics[width=\linewidth]{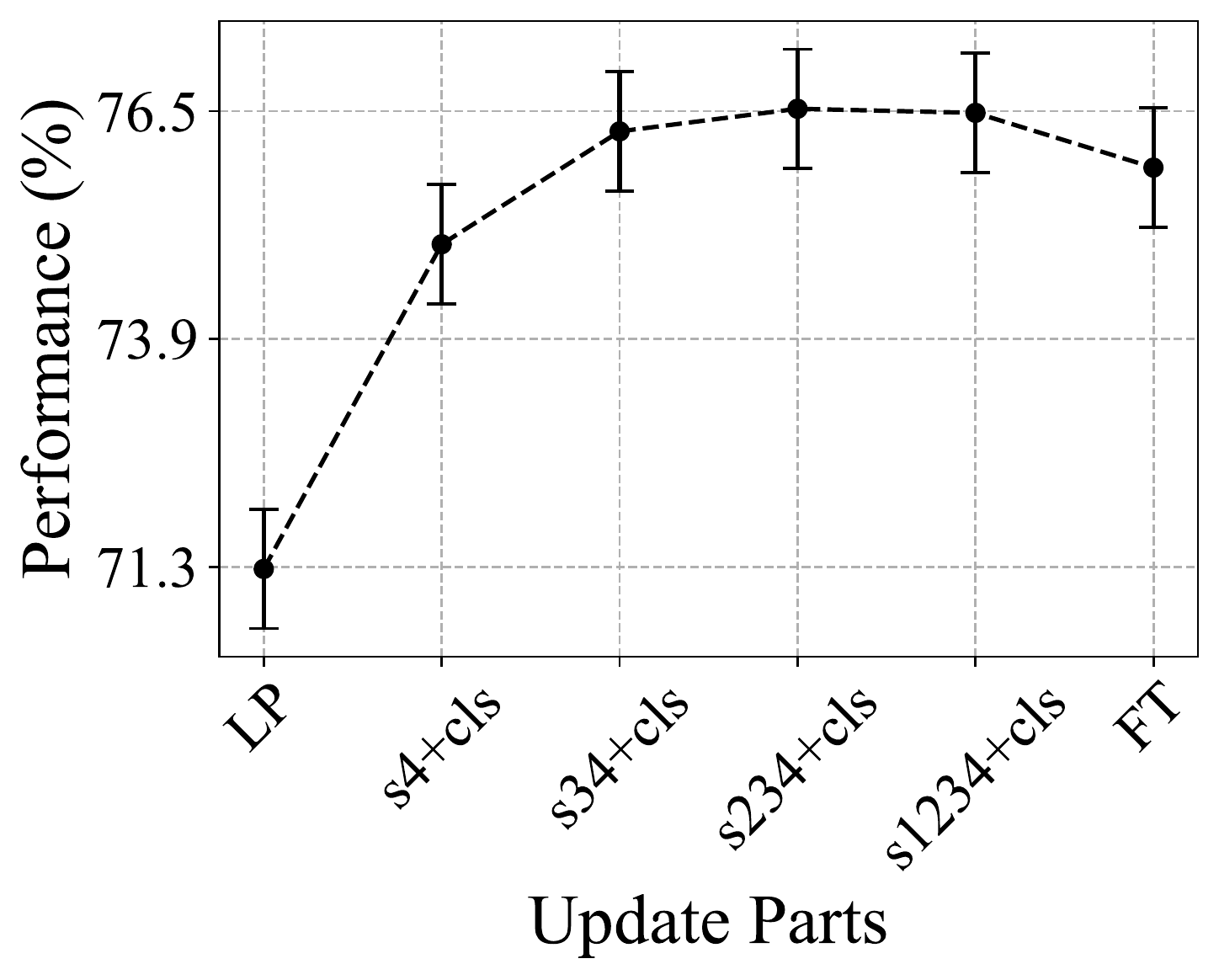}
    \caption{CUB}
    \end{subfigure}
    \begin{subfigure}[h]{0.18\linewidth}
    \includegraphics[width=\linewidth]{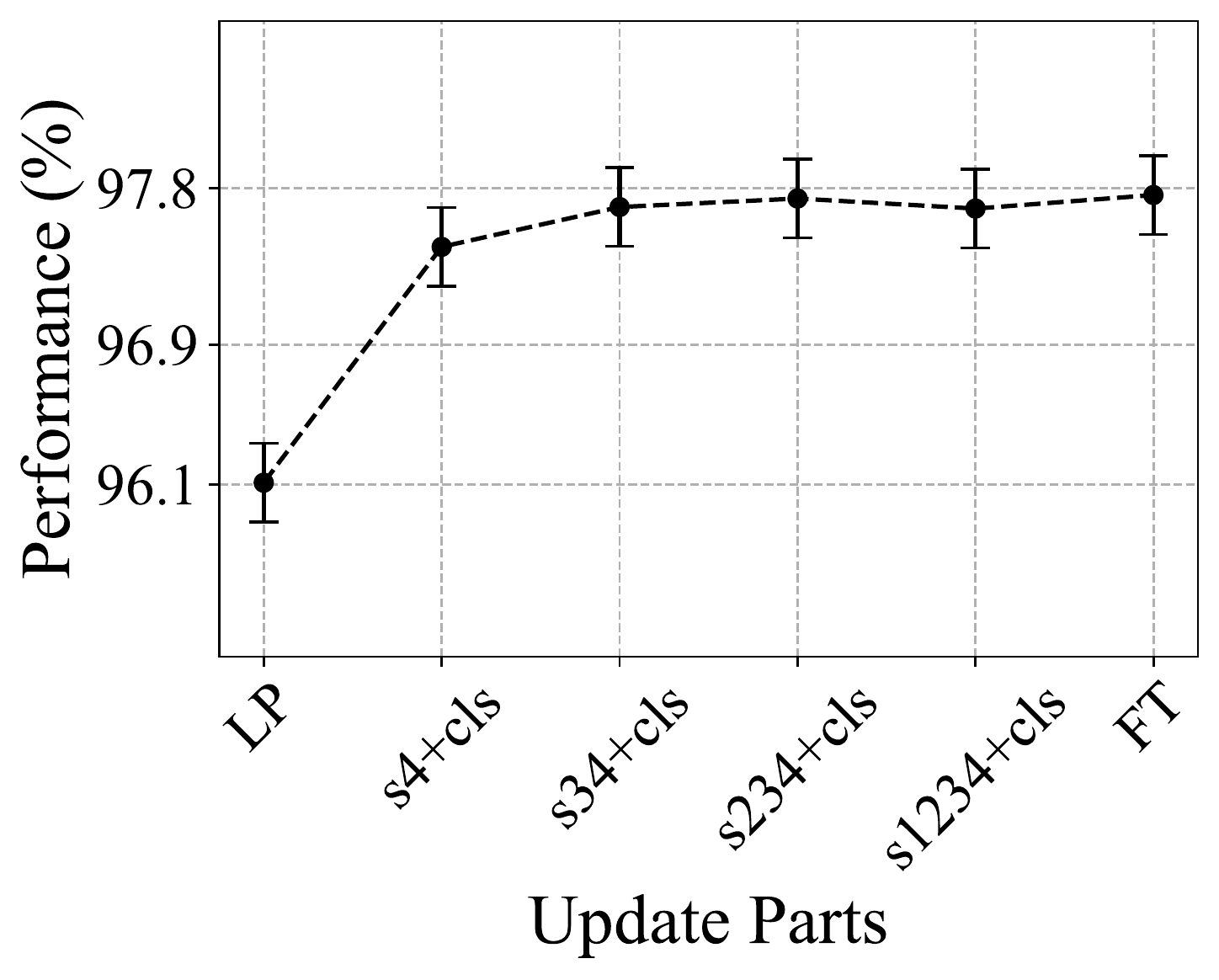}
    \caption{CropDiseases}
    \end{subfigure}
    \centering
    \begin{subfigure}[h]{0.18\linewidth} 
    \includegraphics[width=\linewidth]{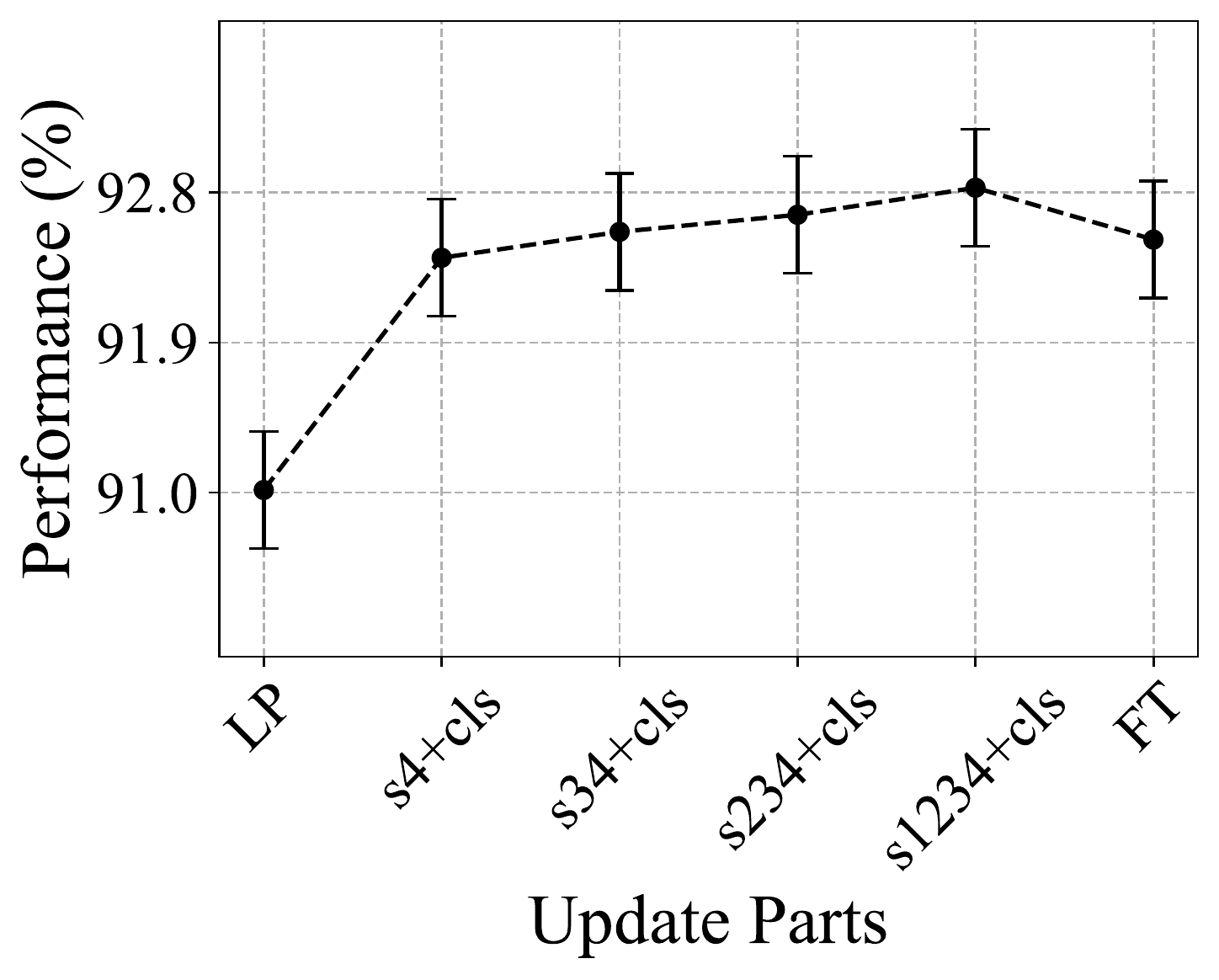}
    \caption{EuroSAT}
    \end{subfigure}
    \begin{subfigure}[h]{0.18\linewidth}
    \includegraphics[width=\linewidth]{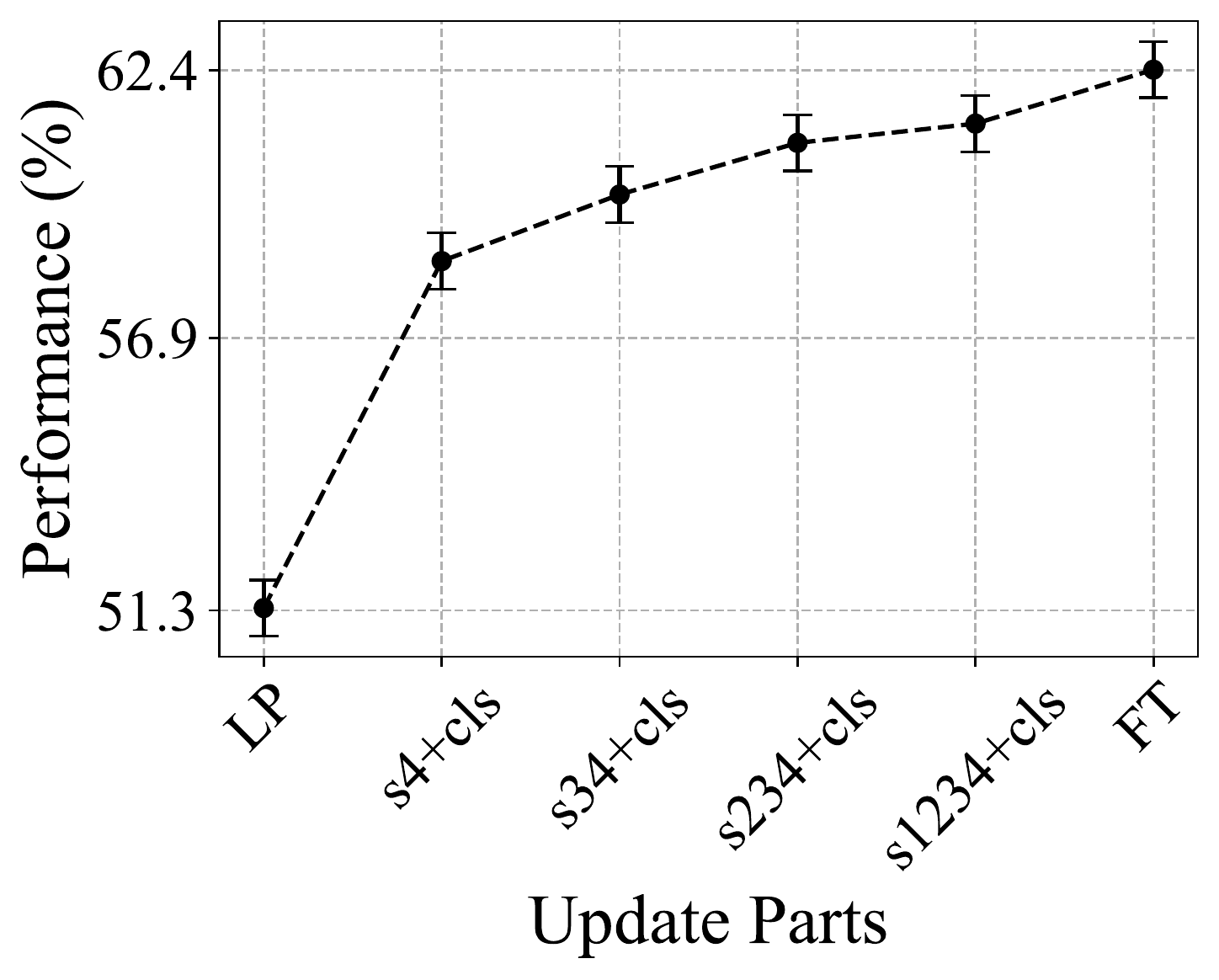}
    \caption{ISIC}
    \end{subfigure}
    \begin{subfigure}[h]{0.18\linewidth}
    \includegraphics[width=\linewidth]{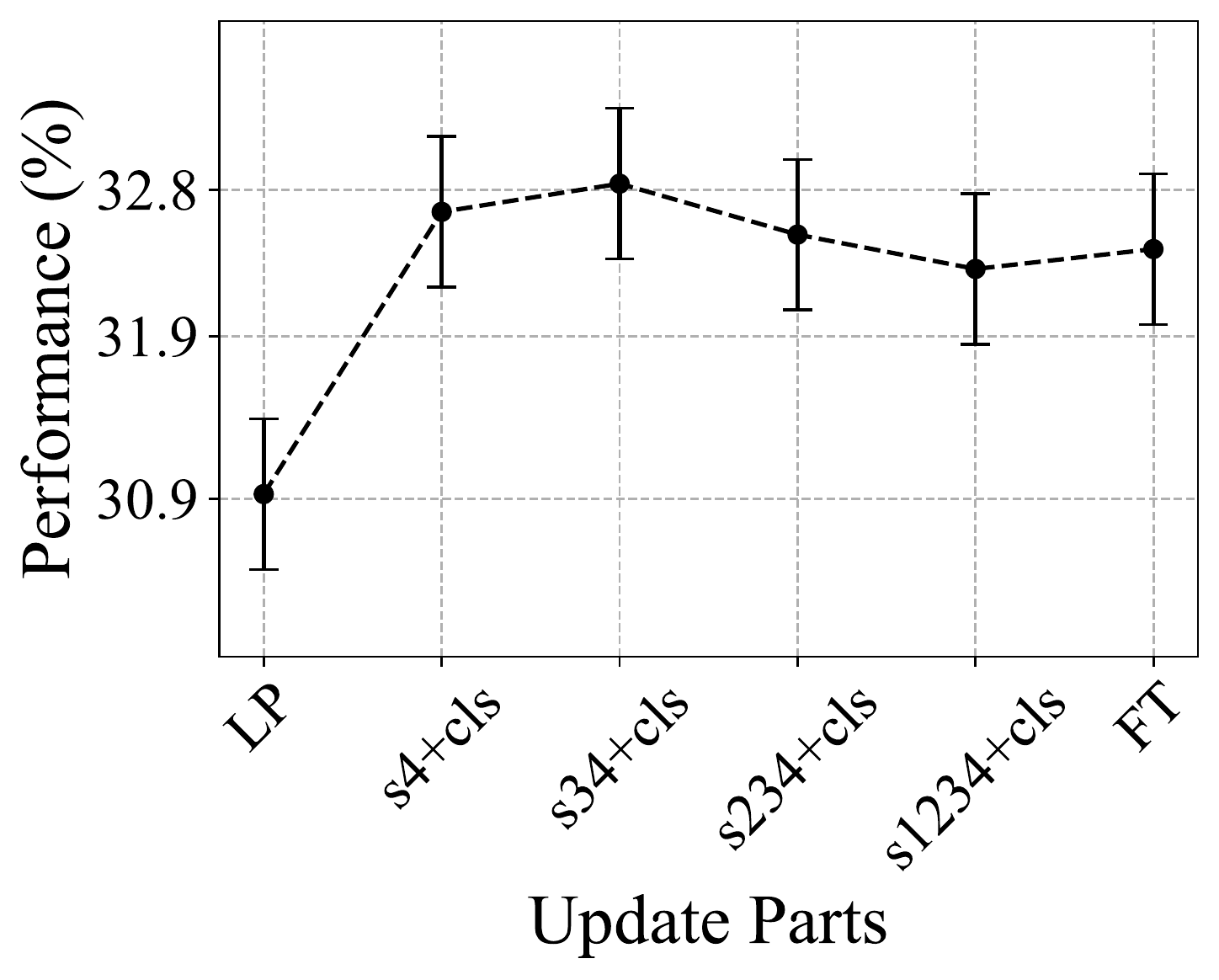}
    \caption{ChestX}
    \end{subfigure}
    \caption{$5$-way $20$-shot results of partial updates between LP and FT.}
    \label{fig:interpolation_20shot}
\end{figure*}

Figures \ref{fig:interpolation_1shot}, \ref{fig:interpolation_5shot}, and \ref{fig:interpolation_20shot} describe 1-, 5-, and 20-shot results of partial updates between LP and FT on various target datasets, respectively. In each subfigure, the leftmost and rightmost points indicate LP and FT, respectively.
When LP is better than FT, the performance decreases as update part gets larger. However, when FT is better than LP, it is observed that there can be a sweet-spot with smaller computation costs than FT, which is similar to observations provided in \citet{oh2021boil}.

\clearpage

\end{document}